\RequirePackage{fix-cm} % fix some latex issues see: http://texdoc.net/texmf-dist/doc/latex/base/fixltx2e.pdf
\documentclass[ oneside,openright,titlepage,numbers=noenddot,headinclude,%1headlines,% letterpaper a4paper
                footinclude=true,cleardoublepage=empty,abstractoff, % <--- obsolete, remove (todo)
                BCOR=5mm,paper=a4,fontsize=11pt,%11pt,a4paper,%
                ngerman,american,%
                ]{scrreprt}

\PassOptionsToPackage{utf8}{inputenc}
	\usepackage{inputenc}

\PassOptionsToPackage{
  drafting=false,    % print version information on the bottom of the pages
  tocaligned=false, % the left column of the toc will be aligned (no indentation)
  dottedtoc=true,  % page numbers in ToC flushed right
  eulerchapternumbers=true, % use AMS Euler for chapter font (otherwise Palatino)
  linedheaders=false,       % chaper headers will have line above and beneath
  floatperchapter=true,     % numbering per chapter for all floats (i.e., Figure 1.1)
  subfig=false,
  eulermath=false,  % use awesome Euler fonts for mathematical formulae (only with pdfLaTeX)
  beramono=true,    % toggle a nice monospaced font (w/ bold)
  palatino=true,    % deactivate standard font for loading another one, see the last section at the end of this file for suggestions
  parts=true,
  pdfspacing=true,
  listings=true,
  style=classicthesis % classicthesis, arsclassica
}{classicthesis}
\newcommand{\myTitle}{Hierarchical object-centric learning with Capsule Networks\xspace}
\newcommand{\mySubtitle}{Doctoral dissertation\xspace}

\newcommand{\myName}{Riccardo Renzulli\xspace}
\newcommand{\myProf}{Marco Grangetto\xspace}

\newcommand{\myFaculty}{School of Science of Nature\xspace}

\newcommand{\myUni}{University of Turin\xspace}

\newcommand{\myTime}{2023\xspace}
\newcommand{\myVersion}{version 4.2\xspace}

\newcounter{dummy} % necessary for correct hyperlinks (to index, bib, etc.)
\providecommand{\mLyX}{L\kern-.1667em\lower.25em\hbox{Y}\kern-.125emX\@}
\newcommand{\ie}{i.\,e.}

\newcommand{\eg}{e.\,g.}
 
	\usepackage{babel}                  

\usepackage{csquotes}
\PassOptionsToPackage{%
	backend=bibtex,bibencoding=ascii,%
	language=auto,%
	style=numeric-comp,%
    sorting=none, % name, year, title
    maxnames=1,
    maxbibnames=1, % default: 3, et al.
    natbib=true % natbib compatibility mode (\citep and \citet still work)
}{biblatex}
    \usepackage[maxbibnames=99]{biblatex}

\PassOptionsToPackage{T1}{fontenc} % T2A for cyrillics
    \usepackage{fontenc}     
\usepackage{textcomp} % fix warning with missing font shapes
\usepackage{scrhack} % fix warnings when using KOMA with listings package          
\usepackage{xspace} % to get the spacing after macros right  
\usepackage{mparhack} % get marginpar right
\PassOptionsToPackage{printonlyused,smaller}{acronym} 
    \usepackage{acronym} % nice macros for handling all acronyms in the thesis

\usepackage{amsmath}
\usepackage{gensymb}
\usepackage{algorithm}
\usepackage{algpseudocode}
\usepackage[export]{adjustbox}
\usepackage{bibentry}
\usepackage{tabularx} % better tables
\usepackage{caption}
\usepackage{subcaption}
\usepackage{makecell}
\usepackage{multirow}
\usepackage{lscape}

\usepackage{listings} 
\PassOptionsToPackage{pdftex,hyperfootnotes=false,pdfpagelabels}{hyperref}
    \usepackage{hyperref}  % backref linktocpage pagebackref
\PassOptionsToPackage{pdftex}{graphicx}
    \usepackage{graphicx}

\hypersetup{%
    colorlinks=true, linktocpage=true, pdfstartpage=3, pdfstartview=FitV,%
    breaklinks=true, pdfpagemode=UseNone, pageanchor=true, pdfpagemode=UseOutlines,%
    plainpages=false, bookmarksnumbered, bookmarksopen=true, bookmarksopenlevel=1,%
    hypertexnames=true, pdfhighlight=/O,%nesting=true,%frenchlinks,%
    urlcolor=webbrown, linkcolor=RoyalBlue, citecolor=webgreen, %pagecolor=RoyalBlue,%
    pdftitle={\myTitle},%
    pdfauthor={\textcopyright\ \myName, \myUni, \myFaculty},%
    pdfsubject={},%
    pdfkeywords={},%
    pdfcreator={pdfLaTeX},%
    pdfproducer={LaTeX with hyperref and classicthesis}%
}   

\makeatletter
\@ifpackageloaded{babel}%
    {%
       \addto\extrasamerican{%
                }%
       \addto\extrasngerman{% 
                }%  
    }{\relax}
\makeatother

\usepackage{classicthesis} 

\def\va{{\boldsymbol{a}}}
\def\vb{{\boldsymbol{b}}}
\def\vc{{\boldsymbol{c}}}

\def\vq{{\boldsymbol{q}}}
\def\vr{{\boldsymbol{r}}}
\def\vs{{\boldsymbol{s}}}

\def\vu{{\boldsymbol{u}}}
\def\vv{{\boldsymbol{v}}}

\def\vx{{\boldsymbol{x}}}

\def\mC{{\boldsymbol{C}}}

\def\mE{{\boldsymbol{E}}}

\def\mL{{\boldsymbol{L}}}

\def\mT{{\boldsymbol{T}}}
\def\mU{{\boldsymbol{U}}}

\def\mW{{\boldsymbol{W}}}

\DeclareMathAlphabet{\mathsfit}{\encodingdefault}{\sfdefault}{m}{sl}
\SetMathAlphabet{\mathsfit}{bold}{\encodingdefault}{\sfdefault}{bx}{n}

\newcommand{\Ls}{\mathcal{L}}

\newcommand{\normltwo}{L^2}

\DeclareMathOperator*{\argmin}{\arg\!\min}

\newcommand{\cond}[2]{p(#1\vert#2)}

\newcommand{\boldx}{\mathbf{X}}
\newcommand{\weightmatrix}{\mathbf{W}}
\newcommand{\descriptorvector}{\mathbf{w}}
\newcommand{\bigY}[1]{\mathbf{Y}_{#1}}
\newcommand{\continuousX}{\mathbf{\mathcal{X}}}
\newcommand{\kernel}{\mathbf{\kappa}}

\newcommand{\squarecoordinateframe}[1]{$\{C_{s,#1}\}$}
\newcommand{\localcoordinateframe}[1]{$\{C_{l,#1}\}$}

\def\xy{$(x, y)$}
\def\xycoords{\xy-coordinates}

\def\ground{ground-truth}
\def\Ground{Ground-truth}

\def\sota{state-of-the-art}
\def\ie{\textit{i.e.},}
\def\eg{\textit{e.g.},}

\def\etal{\textit{et al.}}

\def\km2{km$^2$}

\def\precomputedmap{\mathbf{M}}

\def\maparea{100 km$^2$}
\def\translationbetweenupdates{50}
\def\localcoordtosquarecoord{T_{l,k}^{s,k}}

\def\squaresizem{100m by 100m}
\def\squaresizepx{100px by 100px}
\def\likelihoodconversiontableheader{Likelihood conversion}
\def\modeltypetableheader{Model type}
\def\translerrorafterconvergence{12.6--18.7 m}
\def\numberofstepstoconvergence{23.2--44.4}

\def\kauhavamapdimensions{1.62 km by 3.82 km}

\def\figvspace{\vspace{0em}}

\def\eqref#1{equation~\ref{#1}}
\def\Eqref#1{Equation~\ref{#1}}
\defbibheading{renzulli-bib}{}
\newcommand{\makeauthorbold}[1]{%
  \DeclareNameFormat{author}{%
    \ifthenelse{\value{listcount}=1}
    {%
      {\expandafter\ifstrequal\expandafter{\namepartfamily}{#1}{\mkbibbold{\namepartfamily\addcomma\addspace \namepartgiveni}}{\namepartfamily\addcomma\addspace \namepartgiveni}}
    }{\ifnumless{\value{listcount}}{\value{liststop}}
        {\expandafter\ifstrequal\expandafter{\namepartfamily}{#1}{\mkbibbold{\addcomma\addspace \namepartfamily\addcomma\addspace \namepartgiveni}}{\addcomma\addspace \namepartfamily\addcomma\addspace \namepartgiveni}}
        {\expandafter\ifstrequal\expandafter{\namepartfamily}{#1}{\mkbibbold{\addcomma\addspace \namepartfamily\addcomma\addspace \namepartgiveni\addcomma\isdot}}{\addcomma\addspace \namepartfamily\addcomma\addspace \namepartgiveni\addcomma\isdot}}%
      }
    \ifthenelse{\value{listcount}<\value{liststop}}
    {\addcomma\space}{}
  }
}
\makeauthorbold{Renzulli}

\begin{document}
\frenchspacing
\raggedbottom
\selectlanguage{american} % american ngerman
%\renewcommand*{\bibname}{new name}
%\setbibpreamble{}
\pagenumbering{roman}
\pagestyle{plain}
%********************************************************************
% Frontmatter
%*******************************************************
%\include{FrontBackmatter/DirtyTitlepage}
% %*******************************************************
% % Titlepage
% %*******************************************************
% \begin{titlepage}
%     % if you want the titlepage to be centered, uncomment and fine-tune the line below (KOMA classes environment)
%     \begin{addmargin}[-1cm]{-3cm}
%     \begin{center}
%         \large  

%         \hfill

%         \vfill

%         \begingroup
%             \color{Maroon}\spacedallcaps{\myTitle} \\ \bigskip
%         \endgroup

%         \spacedlowsmallcaps{\myName}

%         \vfill

%         \includegraphics[width=6cm]{gfx/TFZsuperellipse_bw} \\ \medskip

%         \mySubtitle \\ \medskip   
%         %\myDegree \\
%         %\myDepartment \\                            
%         %\myFaculty \\
%         %\myUni \\ \bigskip

%         \myTime\ -- \myVersion

%         \vfill                      

%     \end{center}  
%   \end{addmargin}       
% \end{titlepage}
%*******************************************************
% Titlepage
%*******************************************************
\begin{titlepage}
    %\pdfbookmark[1]{\myTitle}{titlepage}
    % if you want the titlepage to be centered, uncomment and fine-tune the line below (KOMA classes environment)
    \begin{addmargin}[-3.5cm]{-3cm}
    %\begin{addmargin}[-1cm]{-3cm}
    \begin{center}
        \large
        \vfill
		\spacedallcaps{\myUni} \\ \medskip
		\hrulefill \\ \medskip \medskip
		\spacedlowsmallcaps{Doctoral Program in Computer Science} \\
		\spacedlowsmallcaps{Cycle XXXV} \\ \medskip 
        \vfill
                \begingroup
            \Large
            \color{CTtitle}\spacedallcaps{\myTitle} \\ \medskip
        \endgroup
        \spacedlowsmallcaps{Doctoral Dissertation} \\ \bigskip
        \vfill
        \includegraphics[width=7cm]{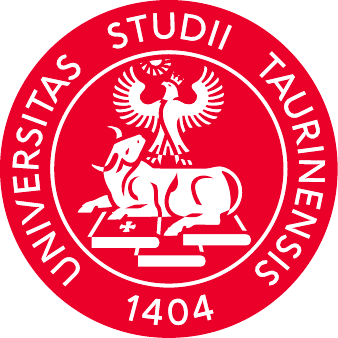} \\ \bigskip \bigskip
        \vfill

    \end{center}
    \begin{minipage}{2in}
		\textsc{Candidate:} \\
		\spacedlowsmallcaps{\myName} \\\\\\\\\\
		%\par\par
	\end{minipage}
	\hfill
	\begin{minipage}{2in}
		\textsc{Supervisor:} \\
		\spacedlowsmallcaps{Prof. \myProf}\\
		\par\par
		\textsc{Reviewers:} \\
		\spacedlowsmallcaps{Prof. Georgios Leontidis}\\
		\spacedlowsmallcaps{Prof. Peter Sincak}
	\end{minipage}	 \\ \medskip

	\hrulefill \medskip \medskip
	\center
	\spacedlowsmallcaps{November 2023}
  	\end{addmargin}
\end{titlepage}

\thispagestyle{empty}

\hfill

\vfill

\noindent\myName: \textit{\myTitle,} \mySubtitle, %\myDegree, 
\textcopyright\ \myTime

%\bigskip
%
%\noindent\spacedlowsmallcaps{Supervisors}: \\
%\myProf \\
%\myOtherProf \\ 
%\mySupervisor
%
%\medskip
%
%\noindent\spacedlowsmallcaps{Location}: \\
%\myLocation
%
%\medskip
%
%\noindent\spacedlowsmallcaps{Time Frame}: \\
%\myTime

\cleardoublepage%*******************************************************
% Dedication
%*******************************************************
\thispagestyle{empty}
%\phantomsection 
\refstepcounter{dummy}
\pdfbookmark[1]{Dedication}{Dedication}

\vspace*{3cm}

\begin{center}
    \emph{I hope that people will finally come to realize\\ that there is only one `race' - the human race - and\\ that we are all members of it}. \\ \medskip
    --- Margaret Atwood    
\end{center}

\medskip

\begin{center}
    Dedicated to all the people whose resilience, passion, and unwavering commitment define the collective spirit of pursuing knowledge. \\ \smallskip
\end{center}
%\cleardoublepage\include{FrontBackmatter/Foreword}
\cleardoublepage%*******************************************************
% Abstract
%*******************************************************
%\renewcommand{\abstractname}{Abstract}
\pdfbookmark[1]{Abstract}{Abstract}
\begingroup
\let\clearpage\relax
\let\cleardoublepage\relax
\let\cleardoublepage\relax

\chapter*{Abstract}
 Convolutional neural networks have been widely successful in various computer vision tasks. However, they lack an explicit representation of entities, and the loss of spatial information hinders their robustness to distributional shifts, such as object pose variations.
 Capsule networks were introduced to address these limitations, learning object-centric representations that are more robust, pose-aware, and interpretable. They organize neurons into groups called capsules, where each capsule encodes the instantiation parameters of an object or one of its parts. Moreover, a routing algorithm connects capsules in different layers, thereby capturing hierarchical part-whole relationships in the data. 

This thesis investigates the intriguing aspects of capsule networks and focuses on three key questions to unlock their full potential. First, we explore the effectiveness of the routing algorithm, particularly in small-sized networks. We propose a novel method that anneals the number of routing iterations during training, enhancing performance in architectures with fewer parameters.

Secondly, we investigate methods to extract more effective first-layer capsules, also known as primary capsules. By exploiting pruned backbones, we aim to improve computational efficiency by reducing the number of capsules while achieving high generalization. This approach reduces capsule networks 
memory requirements and computational effort.

Third, we delve into the exploration of part-relationship learning in capsule networks. Through extensive research, we demonstrate that capsules with low entropy can extract more concise and discriminative part-whole relationships compared to traditional capsule networks, even with reasonable network sizes. 

Lastly, we showcase how capsule networks can be utilized in real-world applications, including autonomous localization of unmanned aerial vehicles, quaternion-based rotations prediction in synthetic datasets, and lung nodule segmentation in biomedical imaging. 

The findings presented in this thesis contribute to a deeper understanding of capsule networks and highlight their potential to address complex computer vision challenges.
% \begin{center}
% \url{https://plg.uwaterloo.ca/~migod/research/beckOOPSLA.html}
% \end{center}

\vfill

% \begin{otherlanguage}{ngerman}
% \pdfbookmark[1]{Zusammenfassung}{Zusammenfassung}
% \chapter*{Sommario}
% Serve in italiano?\dots 
% \end{otherlanguage}

\endgroup			

\vfill
\cleardoublepage%*******************************************************
% Publications
%*******************************************************
\pdfbookmark[1]{Publications}{publications}
\chapter*{Publications}
% \graffito{This is just an early --~and currently ugly~-- test!}
% This might come in handy for PhD theses: some ideas and figures have appeared previously in the following publications:

%\noindent Put your publications from the thesis here. The packages \texttt{multibib} or \texttt{bibtopic} etc. can be used to handle multiple different bibliographies in your document.

    \begin{refsection}[renzulli-bib]
        \nocite{*}
        % \printbibliography[heading=renzulli-caps-bib, sorting=ydn]
        \printbibliography[heading=renzulli-bib]
    \end{refsection}

% \emph{Attention}: This requires a separate run of \texttt{bibtex} for your \texttt{refsection}, \eg, \texttt{ClassicThesis1-blx} for this file. You might also use \texttt{biber} as the backend for \texttt{biblatex}. See also \url{http://tex.stackexchange.com/questions/128196/problem-with-refsection}.
\cleardoublepage%*******************************************************
% Acknowledgments
%*******************************************************
\pdfbookmark[1]{Acknowledgments}{acknowledgments}

% \begin{flushright}{\slshape    
%     We have seen that computer programming is an art, \\ 
%     because it applies accumulated knowledge to the world, \\ 
%     because it requires skill and ingenuity, and especially \\
%     because it produces objects of beauty.} \\ \medskip
%     --- \defcitealias{knuth:1974}{Donald E. Knuth}\citetalias{knuth:1974} \citep{knuth:1974}
% \end{flushright}

\bigskip

\begingroup
\let\clearpage\relax
\let\cleardoublepage\relax
\let\cleardoublepage\relax
\chapter*{Acknowledgments}

I want to thank all those who have supported me in the past few years during the journey that led to this dissertation.

I extend my deepest gratitude to my advisor, Marco Grangetto, whose unwavering support, guidance, and expertise have been instrumental throughout this PhD journey. I would like to give a special thanks to Enzo Tartaglione, your mentorship has been invaluable, and I am truly grateful for the countless hours you dedicated to shaping and refining my research.

I would like to thank my defence committee members, Georgios Leontidis, Peter Sincak, Attilio Fiandrotti and Lia Morra, for their insightful feedback and constructive criticism. Your expertise has enriched the quality of this work, and I am honoured to have had your guidance.

I am grateful for the research visiting period at the Intelligent Robotics group at Aalto University, especially for the support of Ville Kyrki, Francesco Verdoja and Jouko Kinnari, who enabled me to enrich my skills and research activities essential for this thesis.

Special thanks to all the EIDOS group members and colleagues of the Department of Computer Science at the University of Turin, in particular to Andrea Bragagnolo, Carlo Alberto Barbano, Giovanni Bonetta, Gabriele Sartor, Mirko Zaffaroni, Daniele Perlo, Alberto Presta, Gabriele Spadaro, Francesco Di Sario and Gianluca Dalmasso for providing a conducive and stimulating academic environment. The exchange of ideas and the camaraderie made the research process both enriching and enjoyable. 

Heartfelt thanks to my friends Alvise, Alessio, Clarissa, Dorian, Elena, Joana, Marco and Viola for their unwavering encouragement and understanding during the challenging phases of this academic pursuit. 

I extend my deepest thanks to my family, who believed in me from the beginning. Your belief has been a constant source of motivation.

Thanks to Dario, my biggest supporter, my compass, always there to help me even when I was sinking into the deepest part of the ocean. 

\endgroup

\pagestyle{scrheadings}
\cleardoublepage%*******************************************************
% Table of Contents
%*******************************************************
%\phantomsection
\refstepcounter{dummy}
\pdfbookmark[1]{\contentsname}{tableofcontents}
\setcounter{tocdepth}{1} % <-- 2 includes up to subsections in the ToC
\setcounter{secnumdepth}{3} % <-- 3 numbers up to subsubsections
\manualmark
\markboth{\spacedlowsmallcaps{\contentsname}}{\spacedlowsmallcaps{\contentsname}}
\tableofcontents 
\automark[section]{chapter}
\renewcommand{\chaptermark}[1]{\markboth{\spacedlowsmallcaps{#1}}{\spacedlowsmallcaps{#1}}}
\renewcommand{\sectionmark}[1]{\markright{\thesection\enspace\spacedlowsmallcaps{#1}}}
%*******************************************************
% List of Figures and of the Tables
%*******************************************************
\clearpage

\begingroup 
    \let\clearpage\relax
    \let\cleardoublepage\relax
    \let\cleardoublepage\relax
    %*******************************************************
    % List of Figures
    %*******************************************************    
    %\phantomsection 
    \refstepcounter{dummy}
    %\addcontentsline{toc}{chapter}{\listfigurename}
    \pdfbookmark[1]{\listfigurename}{lof}
    \listoffigures

    \vspace{8ex}

    %*******************************************************
    % List of Tables
    %*******************************************************
    %\phantomsection 
    \refstepcounter{dummy}
    %\addcontentsline{toc}{chapter}{\listtablename}
    \pdfbookmark[1]{\listtablename}{lot}
    \listoftables
        
    \vspace{8ex}
%   \page
    
    % %*******************************************************
    % % List of Listings
    % %*******************************************************      
    %   %\phantomsection 
    % \refstepcounter{dummy}
    % %\addcontentsline{toc}{chapter}{\lstlistlistingname}
    % \pdfbookmark[1]{\lstlistlistingname}{lol}
    % \lstlistoflistings 

    % \vspace{8ex}
       
    %*******************************************************
    % Acronyms
    %*******************************************************
    %\phantomsection 
    % \refstepcounter{dummy}
    % \pdfbookmark[1]{Acronyms}{acronyms}
    % \markboth{\spacedlowsmallcaps{Acronyms}}{\spacedlowsmallcaps{Acronyms}}
    % \chapter*{Acronyms}
    % \begin{acronym}[UMLX]
    %     \acro{dof}[DOF]{degree-of-freedom}
    %     \acrodefplural{dof}[DOFs]{degrees-of-freedom}
    %     \acro{ransac}[RANSAC]{random sample consensus}
    %     \acro{map}[MAP]{maximum a posteriori}
    %     \acro{mle}[MLE]{maximum likelihood estimation}
    %     \acro{rms}[RMS]{root-mean-square}
    %     \acro{dem}[DEM]{digital elevation model}
    %     \acro{pdf}[pdf]{probability density function}
    %     \acro{ahrs}[AHRS]{attitude and heading reference system}
    %     \acro{lidar}[LIDAR]{light detection and ranging}
    %     \acro{relu}[ReLU]{rectified linear unit}
    %     \acro{rtk}[RTK]{real-time kinematic}
    %     \acro{gps}[GPS]{global positioning system}
    %     \acro{fcn}[FCN]{fully-connected network}
    %     \acro{brm}[BRM]{building ratio map}
    %     \acro{sfm}[SfM]{Structure-from-Motion}
    %     \acro{vpr}[VPR]{visual place recognition}
    % \end{acronym}
\endgroup

%********************************************************************
% Mainmatter
%*******************************************************
\cleardoublepage\pagenumbering{arabic}
% use \cleardoublepage here to avoid problems with pdfbookmark
\cleardoublepage
\chapter{Introduction}\label{ch:introduction}

Convolutional networks lose the spatial relationships between their parts because of max pooling layers, which progressively drop spatial information~\citep{hinton-dr}. Furthermore, convolutional networks do not have an explicit representation of an entity.
Capsule networks were introduced by~\citet{hinton-dr} to overcome the shortcomings of convolutional networks. 
Capsule networks group neurons into \textit{capsules} that encode objects and objects-parts of the entities in the input images. Therefore, they learn hierarchical structures from data. Thanks to a routing algorithm that connects capsules in different layers, we can carve out a parse tree composed of part-object relationships from the network. They are also more robust to viewpoint changes and affine transformations than traditional convolutional networks. 

Despite their potential advantages, capsule networks have not yet seen widespread adoption in the industry or research community compared to traditional convolutional networks or recently introduced vision transformers ~\cite{dosovitskiy2021an, liu2021Swin}. The lack of standardized architectures and pre-trained models might be a contributing factor. However, continued research and advancements in capsule networks are essential to harness their potential and address their limitations fully. 

In this thesis, we delve into the intriguing aspects of capsule networks, focusing on three key questions that aim to unlock their full potential:

\begin{itemize}
    \item \textit{Does the routing algorithm really improve the performance of capsule networks?}
    \item \textit{How can capsule networks learn effective low-level part descriptions?} 
    \item \textit{Can capsules achieve high generalization ability even when not accounting for the clutter?}
\end{itemize}

The first question addressed in this thesis revolves around the effectiveness of the routing algorithm. While previous research has demonstrated the robustness of capsule networks in various tasks, questions remain regarding the necessity of the routing-by-agreement mechanism, especially in small-sized networks. Through rigorous experimentation, we affirm the importance of the routing algorithm in small-sized networks. We propose a novel method that anneals the number of routing iterations during training, striking better performance, especially for architectures with fewer
parameters.

The second question explores methods to extract better first-layer capsules, often referred to as primary capsules. Convolutional layers without pooling have been a common choice in capsule networks, but they might lead to information loss and limited feature representation. To address this concern, we investigate the exploitation of pruned backbones. Promoting sparse topologies, for example, allows the deployment of deep neural networks models on embedded, resource-constrained devices. We explore sparsity besides capsule representations to improve their computational efficiency by
reducing the number of capsules. By leveraging this approach, we show how pruning with capsule network achieves high generalization with less
memory requirements, computational effort, and inference
and training time.

To answer the third question, we delve into the exploration of part-relationship learning in capsule networks. Understanding how capsules capture part-whole relationships is crucial in advancing our comprehension of capsule network dynamics. In this research, we demonstrate that capsules with low entropy hold the potential to extract more succinct and more discriminative part-relationships than traditional capsule networks, even with reasonable network sizes. These findings shed light on the underlying mechanisms of part-aware representations and pave the way for more interpretable and effective capsule networks.

This thesis will focus on answering the previous questions, also showing how we can exploit capsule networks ability to capture spatial relationships between features in applications such as autonomous localization and biomedical imaging. 

This thesis is organized as follows:
\begin{itemize}
    \item in Part~\ref{part:capsnet-fundamentals} we deepen the limitations of convolutional networks, providing an overview of the fundamentals on capsule networks.
    \item in Part~\ref{part:capsnet-components} we tackle the above open questions proposing three methods: routing annealing, efficient capsule networks with pruned backbones, and routing entropy minimization.
    \item in Part~\ref{part:capsnets-applications} we show how capsule networks can be applied to more difficult tasks: localization of autonomous unmanned aerial vehicles, quaternion-based rotations prediction in synthetic datasets, and segmentation of lung nodules on medical images.
\end{itemize}
\ctparttext{
This part discusses the drawbacks of convolutional neural networks in image and object recognition, such as the use of pooling layers, which can lead to the loss of spatial relationships between objects that reduce the robustness of a model to novel viewpoints and affine transformations of the input. Then we introduce capsule networks, a novel deep learning architecture proposed by Hinton, which attempts to solve the problems with these state-of-the-art methods grouping neurons into vectors called capsules.
}
\part{Learning with Capsule Networks}\label{part:capsnet-fundamentals}
\chapter{The limitations of Convolutional Networks}\label{ch:limitconvnets}
CNNs have been remarkably successful in the last few years and they outperform other methods on object recognition tasks. 
Their main characteristics are:
\begin{enumerate}
	\item They use many learned feature detectors with different receptive fields.
	\item Different types of feature detectors are replicated across space.
	\item Higher-level feature detector fields have bigger spatial dimensions.
	\item Convolutional layers are interleaved with pooling layers that downsample the input. They combine the outputs of nearby feature detectors neurons of the same type.
\end{enumerate}
Replicated feature detectors, inspired by the human vision system, allow recognition of similar textures, shapes and objects in different locations in the image. Pooling layers give a small amount of translation invariance at each level and reduce the number of neurons to the next feature extraction layer. Therefore, CNNs learn multiple types of features with bigger spatial domains.

However, \citeauthor{hinton-trans-autoencoders} \citep{hinton-trans-autoencoders} argue that pooling is a very inefficient way of learning. The four arguments against pooling are:
\begin{enumerate}
	\item It does not fit the psychological reality of part-whole hierarchies and coordinate frames in human vision.
	\item It is better to ask for invariance in the weights and equivariance in the activities.
	\item It fails to use the underlying linear manifold that handles natural image changes.
	\item Pooling is a primitive way to route information to the next layers.
\end{enumerate}
We will deepen these arguments in the following sections.

\section{Coordinate frames}
\label{sec:perception}
\begin{figure}[h]
	\centering
	\includegraphics[width=0.65\linewidth]{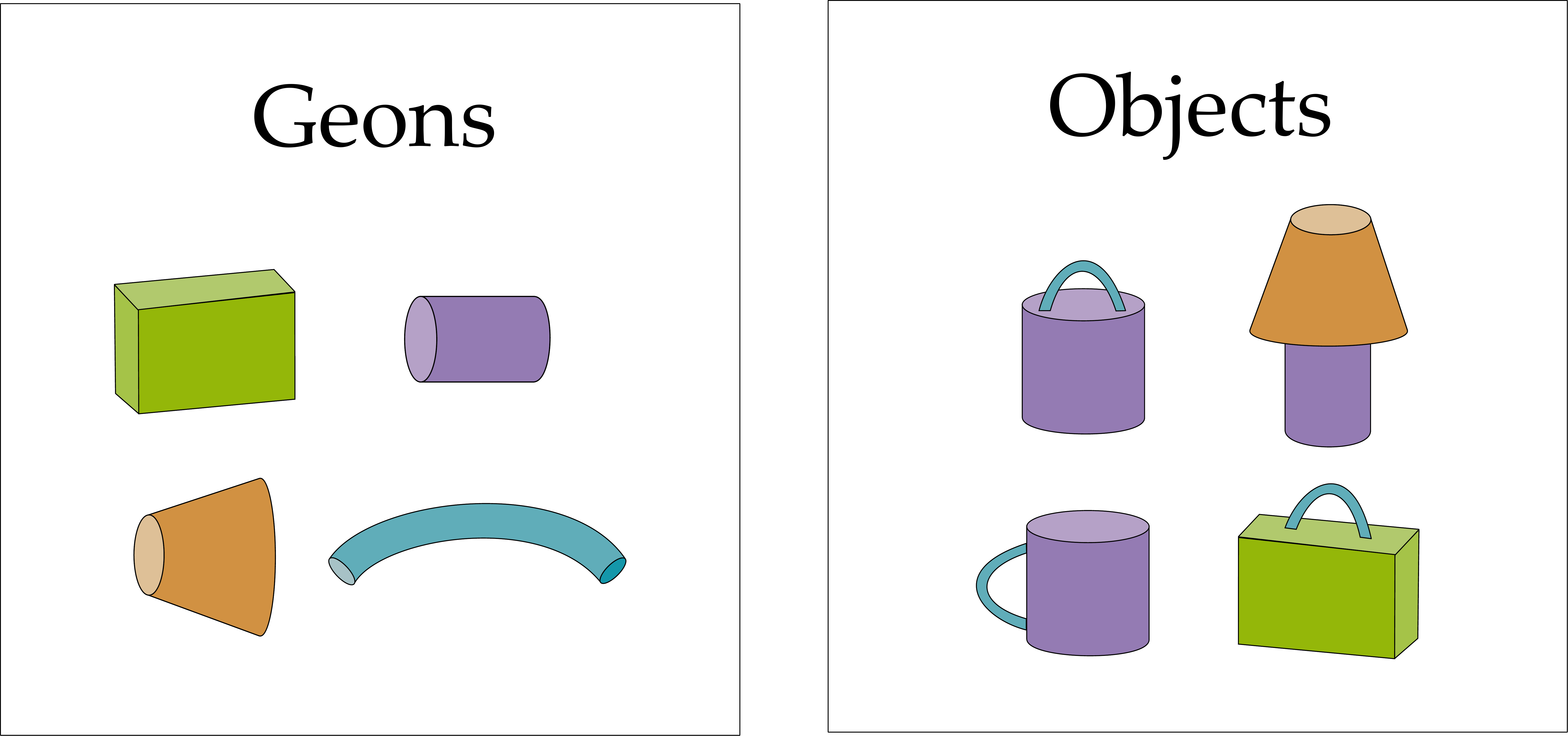}
	\caption{Different arrangements of the same components can produce different objects.}
	\label{fig:biederman}
\end{figure}

According to the recognition-by-components theory illustrated by~\citeauthor{biederman87}~\cite{biederman87}, the ``preferred'' mode of human objects recognition is to separate them into simple 3D shapes, called geons (blocks, cylinders, wedges, and cones), which are the parts that compose an object. ~\citeauthor{biederman87}~\cite{biederman87} suggested that geons can be assembled in various arrangements to form a virtually unlimited number of objects, as shown in Figure~\ref{fig:biederman}.
Additionally, \citeauthor{hinton-mental-imagery} \citep{hinton-mental-imagery} presented mental imagery tasks and perceptual demonstrations to illustrate the involvement of structural descriptions in the imagination and perception of 3D spatial structures. These descriptions are hierarchically organized: complex objects are assigned hierarchical structural descriptions by parsing into parts, each with a local system of significant directions.

Psychological evidence shows that our visual
systems impose coordinate frames for shape recognition~\citep{rock1973orientation,rock1982perception,hinton-mental-imagery, hinton1981frames}. 
\begin{figure}
	\centering
	\includegraphics[width=0.8\linewidth]{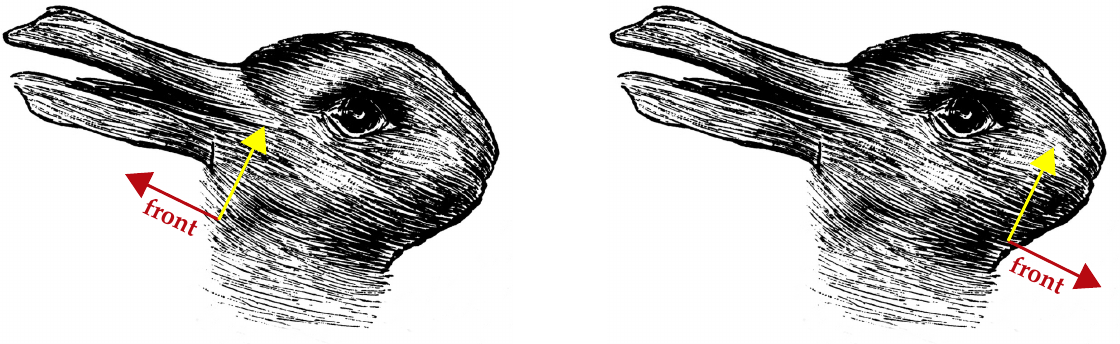}
    	\caption{The same object will look different depending upon the coordinate frame imposed (a duck if the front points to the left, a rabbit if the front points to the right).  }
	\label{fig:rabbit_duck}
\end{figure}
Figure \ref{fig:rabbit_duck} shows how coordinate frames are a fundamental aspect of human perception. We use as an example the rabbit–duck illusion, the famous ambiguous drawing of the German humor magazine Fliegende Blätter (1892), in which a rabbit or a duck can be seen. The percept of them is utterly different depending on which way they are seen. We can explain how it is possible that people can see both animals using coordinate frames. So, with the additional information that the front of the animal points from the center of the page to the left, the reader can recognize a duck. On the contrary, if the additional information is that the front points from the center of the page to the right, the reader can now see the resemblance with a rabbit instead.
Therefore, we get a completely different internal percept depending on what coordinate frame we impose. However, CNNs recognize objects in a very different way from people~\citep{hinton-trans-autoencoders}. Given an input image, they have only one percept for it, and this percept does not depend on imposing different coordinate frames. So CNNs can not explain how the same pixels can be perceived differently.

% As shown in Figure \ref{fig:r}, we perform mental rotation to decide if the tilted R has the correct handedness, not to recognize that it is an R. 
% % \begin{figure}[h]
% % 	\centering
% % 	\includegraphics[width=0.5\linewidth]{gfx/R.pdf}
% % 	\caption{Mental rotation.}
% % 	\label{fig:r}
% % \end{figure}
% We need to do mental rotation to decide handedness because it is difficult to compute the handedness of a coordinate transformation by looking at the individual elements of the matrix (the handedness is the sign of the determinant of the matrix relating the object to the viewer). To avoid this computation, we rotate to upright preserving handedness, then we look to see which way it faces when it is in its familiar orientation. If we had individual neurons that represented a whole pose, we would not have a problem with identifying handedness, because these neurons would have a handedness. This is an indication that these relations are not modelled by a single neuron but they are instead spread over many numbers, a whole bunch of active neurons called \emph{capsule}, just like in computer graphics.

These theories lead to the idea that our brain does a sort of \textit{inverse computer graphics transformation} to perceive shapes~\citep{hinton-dr}. Computer vision is inverse computer graphics, so the higher levels of a vision system should look like the representations used in graphics. Graphics programs use \textit{hierarchical models} in which viewpoint invariant matrices model spatial structure. These matrices represent the transformation from a \textit{coordinate frame} (or coordinate system) embedded in the whole to a coordinate frame embedded in each part. Hierarchical modeling creates complex real-world objects by combining simple primitive shapes into more complex aggregate objects. Adopting this representation, it is easy to compute the relationship between a part and the retina from the relationship between a whole and the retina. Let's see a more formal definition.

\begin{figure}
	\centering
	\includegraphics[width=0.55\linewidth]{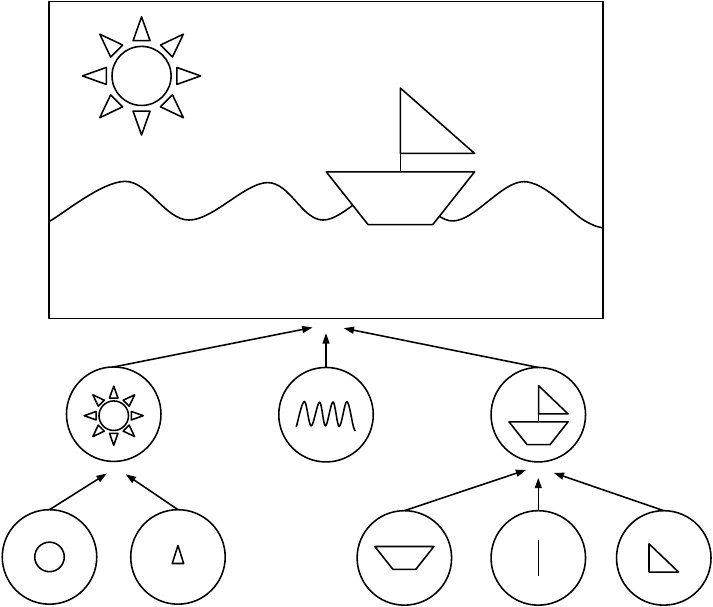}
	\caption{Example of a scene graph.}
	\label{fig:scenegraph}
\end{figure}

We refer to the \textit{position} and \textit{orientation} of an object as its \textit{pose}. Every time we describe the pose of an object, we must specify its relation to some coordinate frame. The pose can be described for example through a rotation and translation transformation, which brings the object from a reference pose to the observed pose.
\textit{Affine transformations} are geometric transformations that preserve collinearity; namely, they map points to points, lines to lines and planes to planes, but not the origin of the space. An affine transformation $f$ acting on a vector $\vx$ can be represented as the composition of a \textit{linear transformation} $\mL$ (a mapping that preserves the operation of addition and scalar multiplication) and a \textit{translation} by a vector $\vb$:
\begin{equation}
    f(\vx) = \mL\vx + \vb.
\end{equation}
Examples of linear transformations are reflections, scalings, rotations and shearing. Because matrix multiplication is associative, we can represent the composition of a sequence of linear transformations more efficiently with a single matrix. Therefore, we can encapsulate and store complex transforms in a compact and convenient form.
However, since translation is not a linear transformation, we need a different representation.
We can represent affine transformations in a fully matrix-multiplication form using \textit{homogeneous coordinates} to determine a frame (4D vector $[p_{x}, p_{y}, p_{z}, p_{0}]^{\top}$). With this notation, any point $P$ can be written uniquely as $P = \alpha_{x}p_{x} + \alpha_{y}p_{y} + \alpha_{z}p_{z} + p_{0}$. The standard convetion sets $p_{0}$ to $1$.
Then any sequence of such operations can be multiplied into a single $4x4$ matrix, allowing simple and efficient processing.
The general form of a transformation matrix is defined as:
\begin{equation}\label{eq:transmatrix}
\mT = 
 \begin{bmatrix}
  n_{x} & o_{x} & a_{x} & v_{x} \\
  n_{y} & o_{y} & a_{y} & v_{y} \\
  n_{z} & o_{z} & a_{z} & v_{z} \\
  0 & 0 & 0 & 1 
 \end{bmatrix},
\end{equation}
where the upper $3x3$ matrix defines the linear transformation and the last column the translation. Note that the columns $n$, $o$ and $a$ of the transform matrix specify the coordinates of the $x$, $y$ and $z$ axes of the new coordinate frame relative to the old frame; namely, they are unit vectors. The third column is the origin of the new coordinate frame concerning the previous frame.
For example, in computer graphics terms, we say that $\mT$ converts points from the \textit{object frame} to the \textit{camera frame}. The inverse of $\mT$ takes us from the \textit{camera frame} to the \textit{object frame}. These transformations play an important role in hierarchical modeling in computer vision. As mentioned before, complex objects are made up of simpler objects, and simpler objects are made up of geometric primitives that can be drawn directly. We can think of these objects as if each has its coordinate system. The components of a complex scene (for example, an image) can be represented using a tree-like structure called \textit{scene graph} (see Figure \ref{fig:scenegraph}). 
Each node can contain any number of children and represents either an object or a part of an object. The root represents the entire scene or the world. Each connection represents one occurrence of the object in its parent object, and it can be associated with a transformation. 

Therefore, our vision system use rectangular coordinate frames embedded in objects and it has the notion of a scene graph that specifies how each node is related to the viewer, namely an \textit{hierarchy of parts} exists. 
CNNs do not use coordinate frames. This is why they do not explain why we assign intrinsic coordinate frames to objects and why they have many effects on our perception, namely using a different frame totally changes the percept.

\section{Invariance and Equivariance}\label{sec:equivariance}
\begin{figure}[h]
	\centering
	\includegraphics[width=0.99\linewidth]{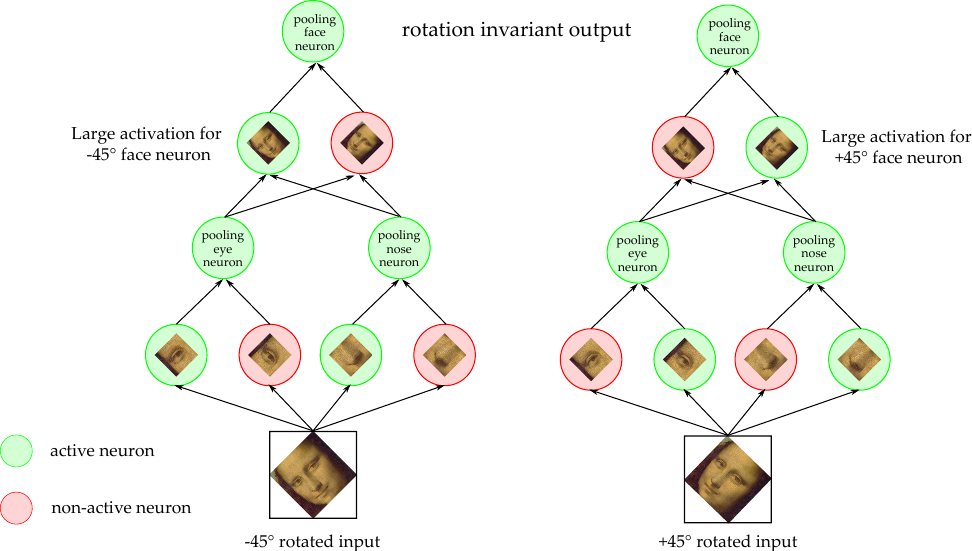}
	\caption[CNNs can learn transformation invariances pooling over different feature detectors.]
	{
Max-pooling over spatial regions produces invariance to translation but not to other transformations, such as rotations. If we pool over the outputs of separately parametrized convolutions, the features can learn which transformations to become invariant. Here we show how a CNN can learn rotation invariances thanks to pooling layers applied to many feature detectors.}
	\label{fig:cnn_invariance}
\end{figure}

CNNs try to make the neural activities \emph{invariant} to small transformations thanks to pooling layers (see Figure~\ref{fig:cnn_invariance}). This is motivated by the fact that in certain tasks, such as image classification, the final label needs to be viewpoint invariant.
As shown in Figure \ref{fig:cnn_invariance}, CNNs use multiple neurons or feature detectors to detect different variants of the features, namely CNNs realize the face matches a variant that is rotated 45 degrees. 

However, when we look at a face, we do not merely recognize it, but we also know its position, orientation, and scale precisely. With the pooling operation, we loose such information, and it is impossible to compute precise spatial relationships. In other words, we would like a model that detects that the face is rotated 45 degrees thanks to corresponding changes in its activations. This way of learning representations enables the development of deep learning models that require fewer trainable parameters and less training data, also avoiding exhaustive data augmentation with all possible transformations and viewpoints. 

\begin{figure}[h]
	\centering
	\includegraphics[width=0.8\linewidth]{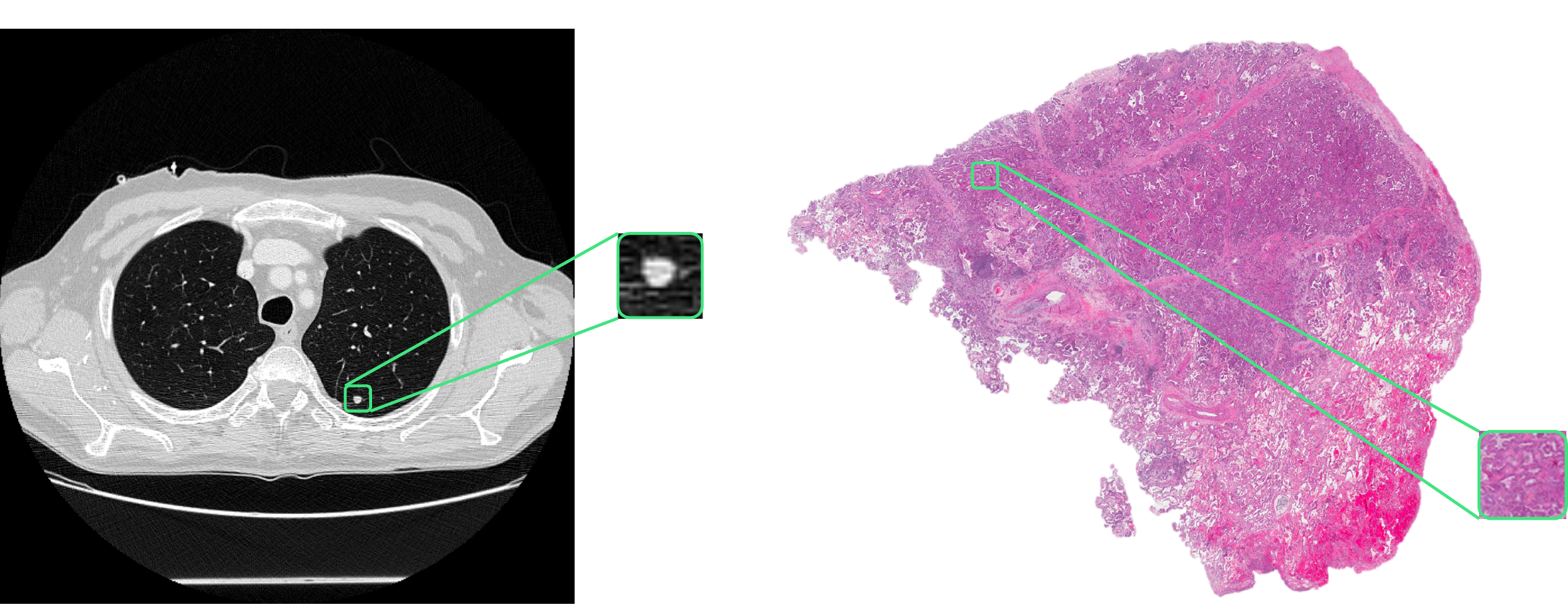}
	\caption{Lung nodules in CT scans (left) and histopathology slices (right) have translation, rotation, reflection, and scaling symmetries.}
	\label{fig:medicaldata_symmetry}
\end{figure}

\begin{figure}[h]
	\centering
	\includegraphics[width=0.7\linewidth]{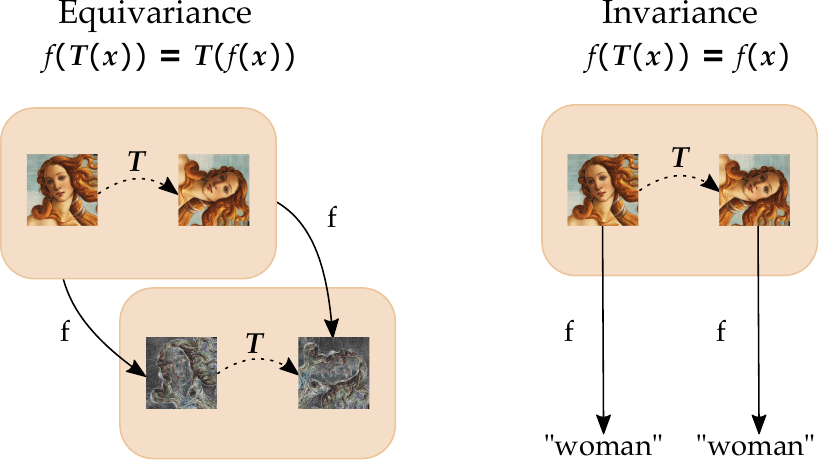}
	\caption{Example of invariance and equivariance when $\mT$ model a +90° rotation transformation and $f$ is the Sobel filter.}
	\label{fig:invariance_vs_equivariance}
\end{figure}

\citeauthor{hinton-trans-autoencoders} \citep{hinton-trans-autoencoders} claim that it is better to aim for \emph{equivariant} neural activities: we want representations where changes in viewpoint lead to corresponding (smoothly) changes in neural activities.
There are some symmetries in our data, for example, in medical images (Figure~\ref{fig:medicaldata_symmetry}), and building a model that retains the input symmetries in its outputs can be helpful. 
Therefore, generating equivariant embeddings can help improve the accuracy, robustness, and generalization of the neural network's predictions. This is effective for a broader range of applications, including supervised classification~\citep{pmlr-v48-cohenc16, winkels2018d} and segmentation~\citep{veeling2018, han2020} tasks, and self-supervised learning pre-training procedures~\citep{dangovski2022equivariant,xie2022equivariant, garrido2023selfsupervised}, where sensitivity to transformations is a desirable property.
Furthermore, cognitive scientists discovered that in the inferior temporal cortex, the same neurons are selective for edges in different orientations, scales, positions, colors, textures, or even for more complex shapes~\citep{schwartz, kandel2012principles}.
More formally, we say that a function $f$ is
\begin{itemize}
	\item \emph{invariant} to a transformation $\mT$ if $f(\mT(\vx)) = f(\vx)$. If the input changes, the output does not change.
	\item \emph{equivariant} to a transformation $\mT$ if $f(\mT(\vx)) = \mT(f(\vx))$. If the input changes, the output changes in a coordinated way. 
\end{itemize}
Invariance is a special case of equivariance when $\mT$ is the identity function. Figure~\ref{fig:invariance_vs_equivariance} shows how rotation invariance and equivariance work. CNNs are not inherently designed to recognize the same objects at different rotations without relying on data augmentations and many feature detectors.

\citeauthor{hinton-dr} \citep{hinton-dr} distinguishes between two types of equivariance.
\begin{enumerate}
	\item{\spacedlowsmallcaps{Place-coded:}} A discrete change in a property of a visual entity leads to a
discrete change in which neurons are used for encoding that visual entity.
	\item{\spacedlowsmallcaps{Rate-coded:}} A real-valued change in a property of a visual entity leads to a
real-valued change in the output of some of the neurons used for coding that visual entity, but there is no change in which neurons are used.
\end{enumerate}

\begin{figure}[h]
	\centering
	\includegraphics[width=0.7\linewidth]{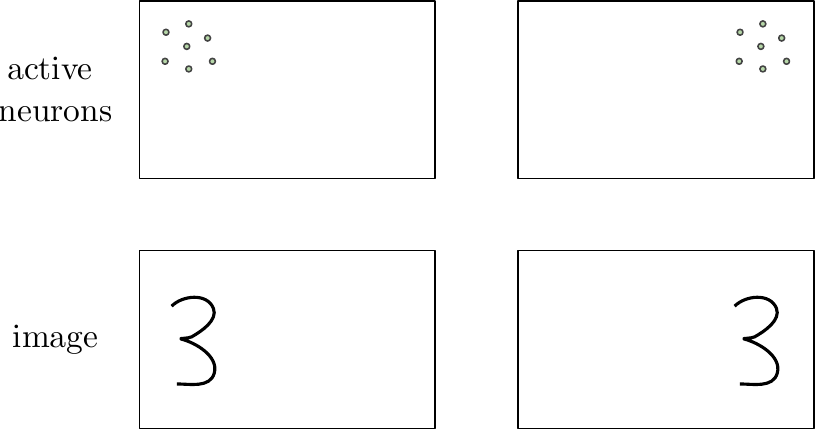}
	\caption{Example of place-coded equivariance when translating an image in a CNN.}
	\label{fig:convequivariance}
\end{figure}

In the case of convolution, the particular form of parameter sharing causes the
layer to be naturally place-coded equivariant only to translation \citep{Goodfellow:2016}. 
(see Figure \ref{fig:convequivariance}). 
% \begin{figure}
% 	\centering
% 	\includegraphics[width=0.8\linewidth]{Parts/01_LearningWithCapsNets/gfx/place_rate_equivariance.pdf}
% 	\caption{Each feature vector is a group of neurons. The activation of a vector is its magnitude, the pose is its orientation. Local information is place-coded for lower features (different neurons encode the transformed visual entities) and rate-coded for higher ones (the same neurons encode the transformed visual entities but with different activations). \riccardo{da cassare, too much} }
% 	\label{fig:place_rate_equivariance}
% \end{figure}
A way to achieve rate-coded equivariance is to use a group of neurons that do quite a lot of internal computation (using non-linear \emph{recognition units}) and encapsulate the results of this computation into a low dimensional output~\citep{hinton-trans-autoencoders}.
At low levels, there are small domains, so if a part moves to a very different position, it will be represented by a different group of neurons (place-coded equivariance). At higher levels, we have bigger domains so if a part only moves a small distance, it will be encoded by the same group, but its activations will change (rate-coded equivariance).
Therefore, we would like that in deep learning models, like in the perceptual system, viewpoint-invariant knowledge is embedded in the weights, not in the
neural activities, which, in contrast, encode the equivariance property. 
% An example of a model with low-level place-coded neurons and high-level rate-coded neurons is depicted in Figure~\ref{fig:place_rate_equivariance}.

\section{Linear Manifold}\label{sec:linmanifold}
If humans are presented with an image from a viewpoint different from the usual one, we do not have difficulty recognizing the object. Therefore, unlike CNNs, we do not have to see the same object in a thousand different poses to learn to recognize it. A model is needed that learns the relationships between the parts so that it is viewpoint invariant.
CNNs with pooling are not built on the \textit{manifold hypothesis}. A \textit{manifold} is a set a of points associated with a neighborhood around each point and from any given point it locally appears to be an Euclidean space \citep{Goodfellow:2016}. According to the manifold hypothesis, valid high-dimensional input data (such as images) tend to lie near low-dimensional manifolds. The concept of neighborhood surrounding each point implies the existence of linear transformations that can be applied to move on the manifold from one point to a neighboring one. For example, in image space we can move on the manifold changing the lights or rotating the objects.

CNNs do not take advantage of this linear natural manifold that handles the largest source of variance in images. In order to achieve high accuracy in shape recognition, they learn different models for different viewpoints and this requires a lot of training data, for example employing data augmentation methods.

\begin{figure}[h]
	\centering
	\includegraphics[width=0.9\linewidth]{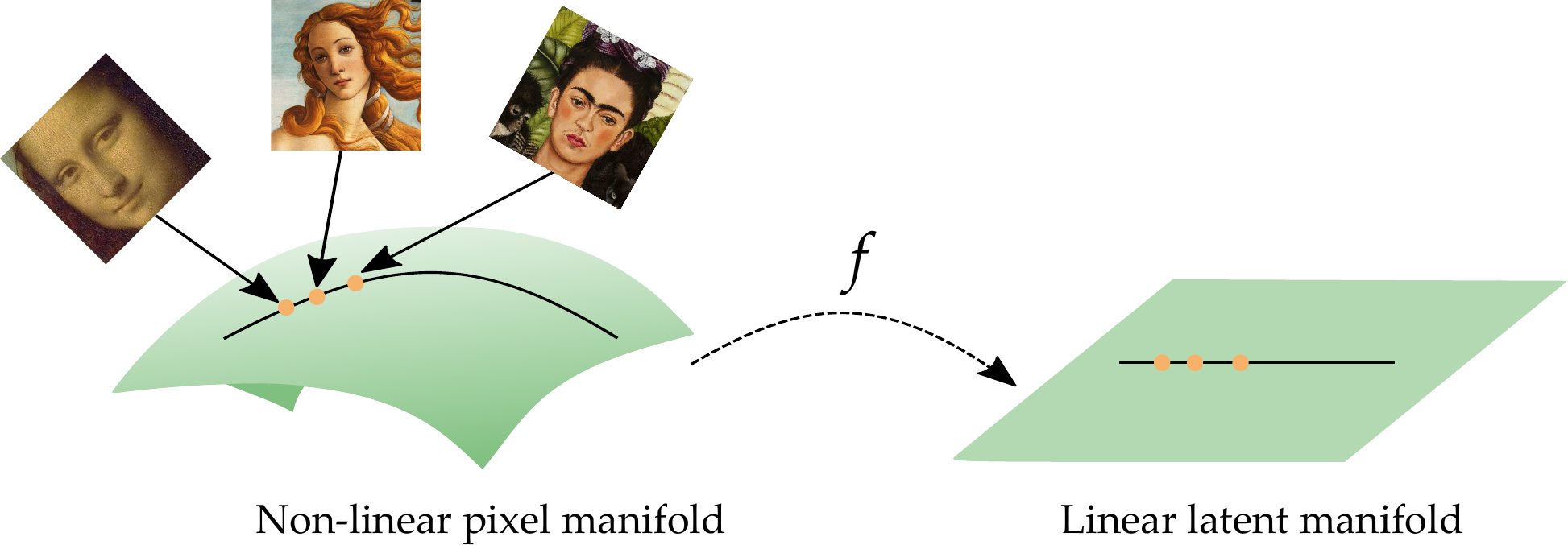}
	\caption{Locally near high-dimensional images lies in a non-linear manifold. If we map their parts into low-dimensional embeddings in a linear manifold, we can exploit linear transformations as in computer graphics to generalize on novel viewpoints.}
	\label{fig:manifold}
\end{figure}

\begin{figure}[h]
	\centering
	\includegraphics[width=1\linewidth]{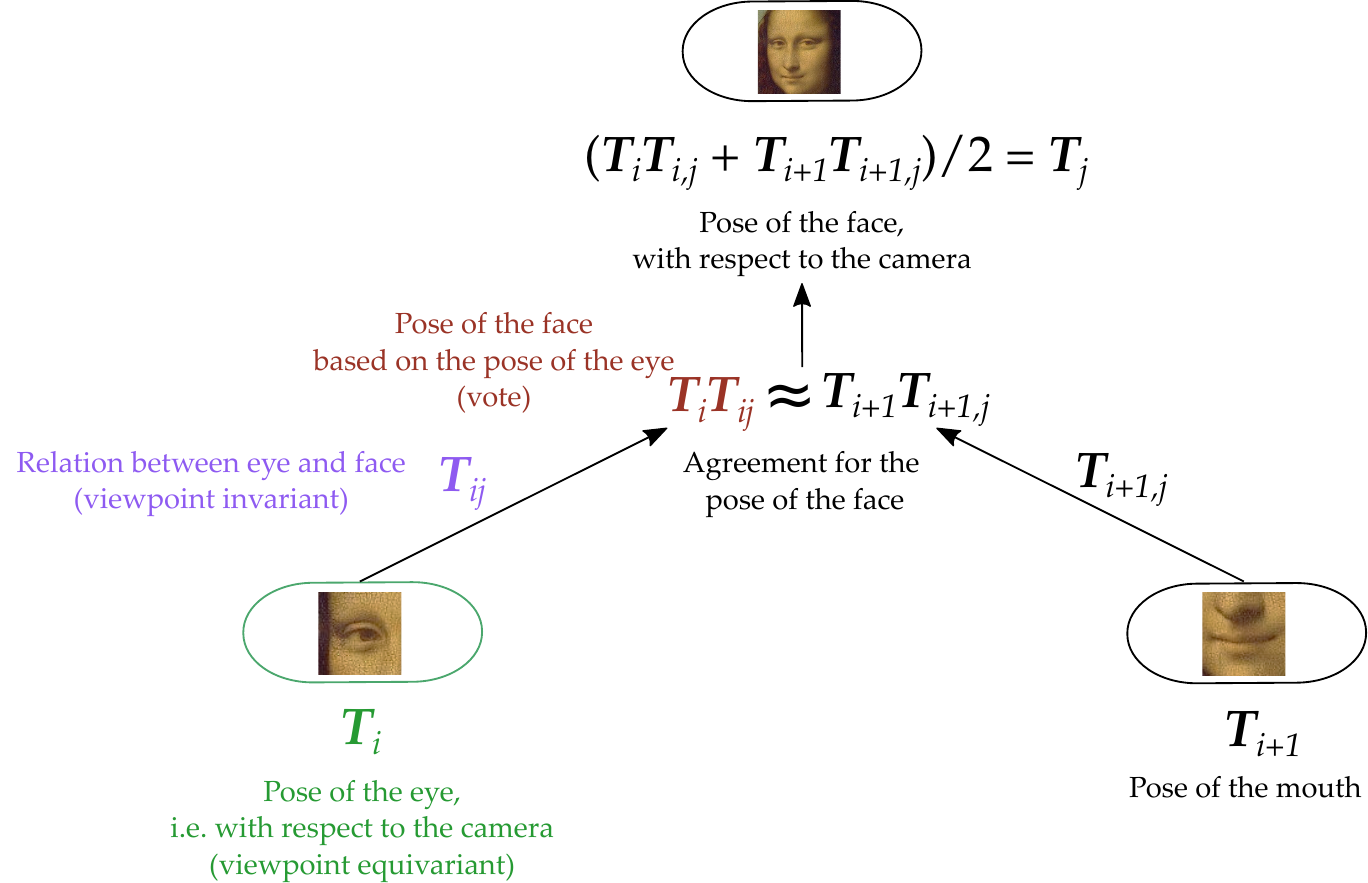}
	\caption[Coincidence filtering in a two layers hierarchy of parts.]
	{Coincidence filtering in a two layers hierarchy of parts. Example of how to compute the pose of a face (i.e., with respect to the camera) from the poses of an eye and a mouth. There is an agreement between the predictions of the poses of the eye and the mouth for the pose of the face, so a face is detected.}
	\label{fig:coincidence_filtering}
\end{figure}

We know that the manifold of images of the same rigid shape is highly-non linear in the space of pixel intensities. As shown in Figure~\ref{fig:manifold} we want to transform pixel intensities into a space in which the manifold is globally linear, just like the graphical representation that uses explicit pose coordinates. Therefore, if we can get pose coordinates from raw pixels, we can do massive extrapolation of new poses from this linear manifold by applying linear transformations. This lead to better generalization over novel viewpoints, training our network on fewer input data.
Generalization is achieved using the (linear) invariant transformation matrices $\mT$ described in Section \ref{sec:perception}. It does not matter how much the pose changes, it is the same matrix of weights that takes the pose of the part and gives the pose of the whole, or vice versa. Therefore, with inverse graphics, we mean that from raw pixels we obtain the poses of the parts (encoded in a bunch of neurons), we multiply these poses by $\mT$ to get the pose of the whole. But can a deep learning model detect objects exploiting this linear manifold?
In Figure~\ref{fig:coincidence_filtering} we can see how a coincidence filtering mechanism using the linear manifold can detect familiar objects in a hierarchy of parts. Each part makes a \textit{vote} for the object: how the pose of the object will be based on the pose of that part. Then, an object can be detected by looking for agreement between these votes for its pose matrix, computed for example with the mean of these predictions. Finding these agreements is not easy when there are many parts and objects. We need a more complex operation than pooling over the lower activations.
\section{Routing}\label{sec:routing}
As mentioned in Section~\ref{sec:perception}, given an input image, we would like to carve out a scene graph. Therefore, parts of objects in one layer need to choose which object in the layer above they belong to. This process of part-object relationship extraction is called routing. It is based on the coincidence filtering mechanism when there are many parts and objects.

\begin{figure}
	\centering
	\includegraphics[width=0.9\linewidth]{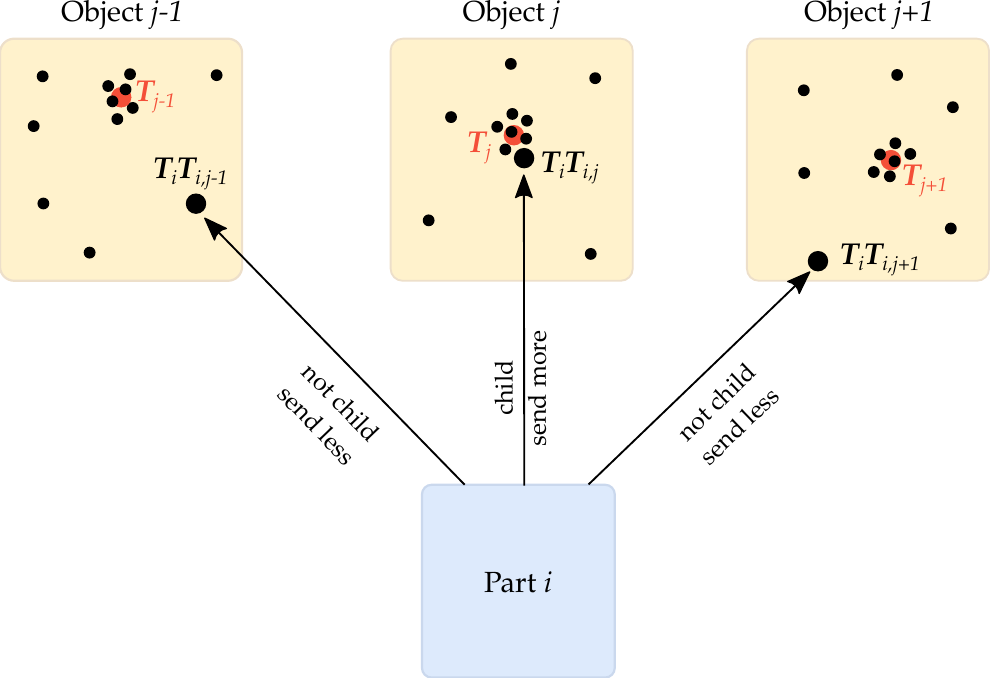}
	\caption[Routing naturally resembles clustering.]
	{Routing naturally resembles clustering. Each dot is a lower-level part prediction for a given higher-layer object. We want to send more feedback to parts that agree more, namely those in the same cluster. The final pose of a high-level object will be the centroid of this cluster.}
	\label{fig:routing}
\end{figure}

CNNs do a basic primitive routing through max-pooling: the pooling unit selects the most active neuron only based on its magnitude. For complicated shapes, replicating the knowledge across all locations with a small stride is not a good solution.  A good routing principle is to send the information about a specific part to the object in the layer above that can account more for it. Routing, as shown in Figure~\ref{fig:routing}, naturally resembles clustering: a higher-layer parent object $j$ (cluster) comprises many lower-layer child parts predictions. 
A higher-level object asks for more feedback from lower-level parts that vote for its cluster and less feedback from lower-level parts that vote for its outliers. 
Finding the best routing is equivalent to parsing the image to obtain a parse tree where each node has only one parent~\citep{hinton-dr}. The total weight on the bottom-up inputs supplied by one lower-level part is 1 to satisfy this ''single parent'' constraint. The parse tree representations are similar to scene graphs in computer graphics. 
So how can we detect objects without pooling? 
We can think of this process as an assembly mechanism that takes as input lower-level parts and outputs higher-level objects.
This process will be deepened in Chapter~\ref{ch:capsnetsbackground}.

% \riccardo{spostare tutta sta parte in what are capsules}
% First, each part cast a vote, a prediction, for the pose of more complex objects based on its own pose. If multiple parts cast very similar votes for the same object, this object is very likely to be present in the input.
% Then, based on these agreements, we increase the connections between the correct part-object relationship and decrease the connections of the incorrect ones to get the final weighted object pose. The corresponding neurons become active.

%*****************************************
%*****************************************
%*****************************************
%*****************************************
%*****************************************

\cleardoublepage
\chapter{Background on Capsule Networks}\label{ch:capsnetsbackground}

\section{Notation}
This Section reports the notation used in the following Chapters for capsule networks.
Superscript ($\xi$) will denote the $\xi^{th}$ input example while superscript $[l]$ will denote the $l^{th}$ capsule layer.

\begin{itemize}
    \item $\Xi$ input dataset
    \item $\vx^{(\xi)} \in \Xi$ the $\xi^{th}$ input example represented as a feature vector
    \item $Y$ label dataset
    \item $y^{(\xi)} \in Y$ the corret label for the $\xi^{th}$ example
    \item $\hat{y}^{(\xi)} \in Y$ the predicted label for the $\xi^{th}$ example
    \item $\Gamma$ label set
    \item $\gamma \in \Gamma$ generic label
    \item $\mT \in \mathbb{R}^{4 \times 4}$ generic affine transformation
    \item $\mT_{i,j}$ generic affine transformation between part $i$ and object $j$
    \item $L$ number of capsule layers in the network
    \item $\Omega^{[l]}$ set of capsules in layer $l \in \{1,2,...,L\}$
    \item $|\Omega^{[1]}|$ number of primary capsules
    \item $|\Omega^{[L]}|$ number of class capsules
    \item[] When $L=2$, it is possible to denote $|\Omega^{[1]}|=I$ and $|\Omega^{[L]}|=J$
    \item $a \in \mathbb{R}$ generic activation of a capsule
    \item $\vu \in \mathbb{R}^{D}$ generic $D$-dimensional feature vector of a capsule
    \item $a^{[l]}$ generic activation of a capsule in layer $l$
    \item $\vu^{[l]} \in \mathbb{R}^{D^{[l]}}$ generic $D^{[l]}$-dimensional feature vector of a capsule in layer $l$
    \item $b^{[l]}_{i,j}$ log prior probability that capsule $i$ in layer $l$ should be coupled to capsule $j$ in layer $l+1$
    \item $\mU^{[l]}$ matrix containing all feature vectors of all capsules in layer $l$
    \item $\mC^{[l]}$ matrix containing coupling coefficients between capsules in layer $l$ and capsules in layer $l+1$
    \item $c^{[l]}_{i,j}$ coupling coefficient between capsule $i$ in layer $l$ and capsule $j$ in layer $l+1$
    \item $c^{[l](\xi)}_{i,j}$ coupling coefficient between capsule $i$ in layer $l$ and capsule $j$ in layer $l+1$ for the $\xi^{th}$ example
    \item $\tilde{c}^{[l]}_{i,j}$ quantized coupling coefficient between capsule $i$ in layer $l$ and capsule $j$ in layer $l+1$
    \item $\mW^{[l]}_{i,j} \in \mathbb{R}^{D^{[l]} \times D^{[l+1]}}$ trainable weight matrix between a capsule $i$ in layer $l$ and a capsule $j$ in layer $l+1$
    \item $\hat{\mU}^{[l+1]} \in\mathbb{R}^{|\Omega^{[l]}| \times |\Omega^{[l+1]}| \times D^{[l+1]}}$ contains all the votes from capsules in layer $l$ for capsules in layer $l+1$
    \item $\hat{\vu}^{[l+1]}_{i,j} \in \mathbb{R}^{D^{[l+1]}}$ vote of a capsule $i$ in layer $l$ for a capsule $j$ in layer $l+1$
    \item $r^{[l]}$ number of routing iterations between capsule layers $l$ and $l+1$
\end{itemize}

When there are only two capsule layers, we can omit the $l$ notation since $i$ is used for lower-level capsules and $j$ for higher-level capsules. In this scenario, we also use the index $j$ to denote the $j$-th class label. For abuse of notation, we often suppress the index $\xi$.

\section{What are capsules}\label{sec:intro:whatarecaps}
In Chapter \ref{ch:limitconvnets} we discussed some of the limitations of CNNs, emphasizing which are the main characteristics an artificial neural system should hold to perceive shapes as humans, using less data and parameters. We also introduced the notion of parts/children and objects/parents when parsing an image and the relationships between them. How can we encode this idea directly into the architecture of an ANN? In standard neural networks, there is no explicit notion of an entity. 

In 1990, a first attempt to embed structural relations into a general model of object recognition inspired by inverse computer graphics was pursued by~\citeauthor{hintonTRAFFIC}~\cite{hintonTRAFFIC}. The concept of a capsule was introduced later in 2011 \citep{hinton-trans-autoencoders}, where a capsule is defined as a group of neurons within a layer.

Capsules are neurobiologically inspired by the structure of \emph{cortical minicolumns} discovered by Mountcastle  in 1957, a group of neurons in the cortex of the brain that encode similar features. Cells in 50 $\mu m$ minicolumn all have the same receptive field; adjacent minicolumns may have different fields. Cortical minicolumns contains from $80$ to $100$ neurons and they are organized in \textit{macrocolumns}. There are nearly $100$ minicolumns in a macrocolumn.
\begin{figure}[h]
	\centering
	\includegraphics[width=0.8\linewidth]{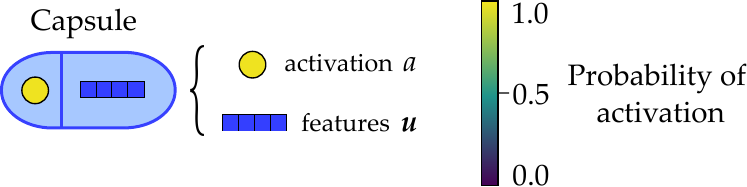}
	\caption[Capsule components.]
	{Capsule components..}
	\label{fig:capsule}
\end{figure}

Parts and objects are indeed visual entities. Each capsule learns to recognize an implicitly defined visual entity over a limited domain of viewing conditions and deformations \citep{hinton-trans-autoencoders}. The neural network will learn what the entities are and how they interact with each other. 
A Capsule Network (CapsNet) consists of several layers $l$ of capsules, where $l \in \{1, 2,...,L\}$. The set of capsules in layer $l$ is denoted
as $\Omega^{[l]}$. We will refer to parts as lower-layer capsules $i \in \Omega^{[l]}$ and to objects as higher-layer capsules $j \in \Omega^{[l+1]}$. For classification tasks, the number $|\Omega^{[L]}|$ of capsules in the last layer is equal to $|\Gamma|$ output labels. 

A capsule in layer $l$ is a set of neurons $\{a^{[l]}, \vu^{[l]}\}$ where 
% the presence and the \textit{instatiation parameters} of a multi-dimensional entity of the type that the capsule detects.
% A capsule consists of two association parameters:
\begin{enumerate}
	\item $a^{[l]} \in\mathbb{R} $ is the activation of the capsule. It is the probability that an object of that type is present;
	\item $\vu^{[l]} \in\mathbb{R}^{D^{[l]}}$ is the feature vector of the capsule. It encodes the instantiation properties of a multi-dimensional entity of the type that the capsule detects. These properties can include its pose (position, size, orientation), deformation, velocity, albedo, hue, texture, etc.
\end{enumerate}
The matrix $\mU^l \in\mathbb{R}^{|\Omega^{[l]}| \times D^{[l]}}$ contains all $D^{[l]}$-dimensional capsule vectors in layer $l$, the vector $\va^{[l]}$ holds the corresponding activations.

\begin{figure}[h]
	\centering
	\includegraphics[width=1\linewidth]{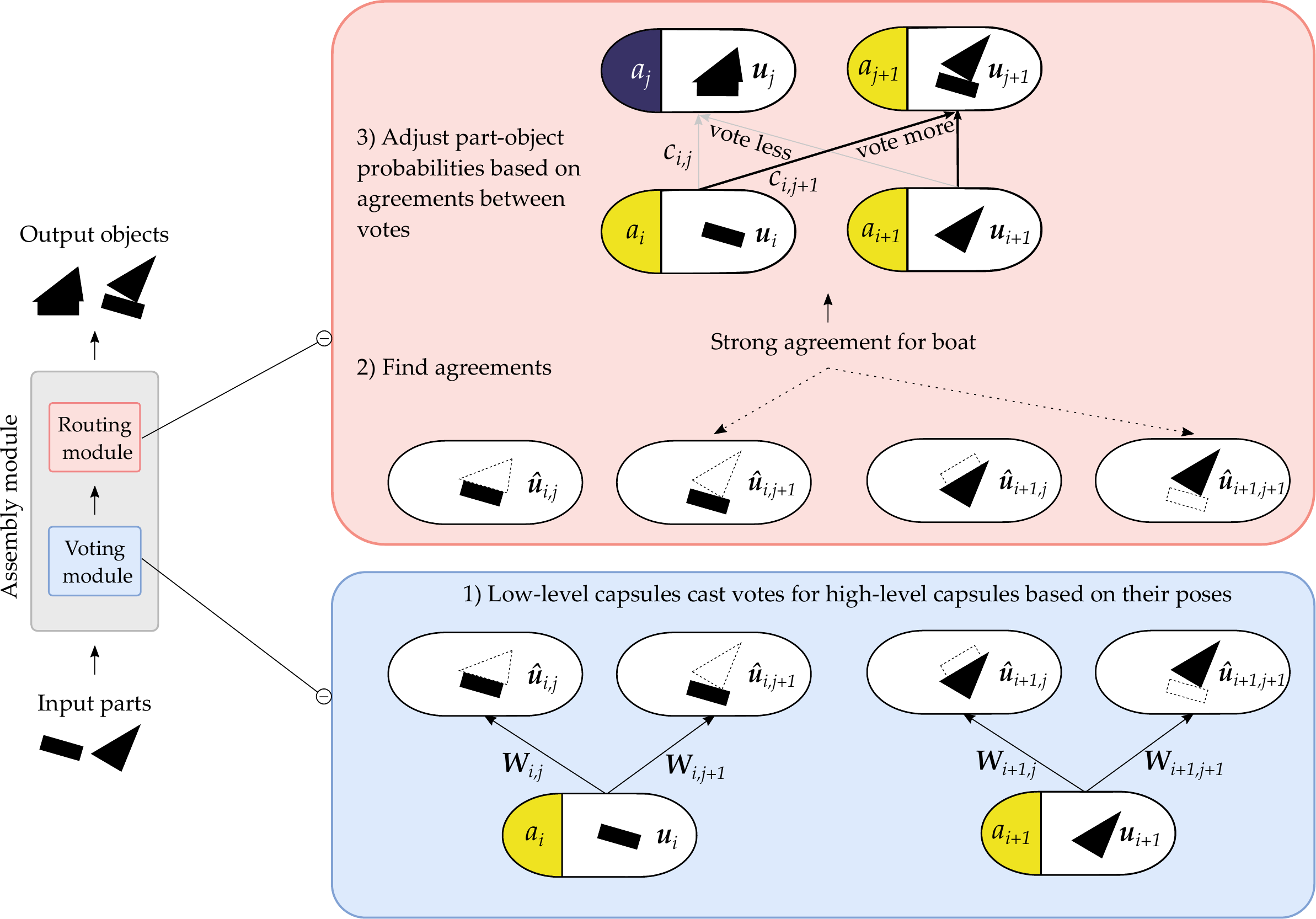}
	\caption[Assembly module.]
	{The assembly module comprises two sub-modules: 1) the votes module, which computes the object-capsules predictions from the input part-capsules, and 2) the routing module, which adjusts the part-object connections based on the agreements between these votes to compute the output capsules.}
	\label{fig:assembly}
\end{figure}

As mentioned in Section~\ref{sec:routing}, assembling parts to build objects can be divided into two phases: a voting procedure and a routing procedure. Similarly, CapsNets use this assemble module to compute higher-layer capsules $j$ from lower-layer capsules $i$ based on coincidence filtering. Figure~\ref{fig:assembly} gives an overview of this module. Most capsule layers in the literature share the same voting module and they differ mainly on how the agreements are found in the routing module.

High-dimensional coincidences do not happen by chance: a higher-order capsule receives multi-dimensional predictions from capsules in the layer below on how its pose should be and it looks for a tight cluster of predictions. If it finds such cluster, namely the predictions are in strong agreement, then it outputs:
\begin{enumerate}
	\item A high probability $a^{[l+1]}_j$ that an entity of its type exists in its domain. We say that the low-level capsules of the cluster are in the right spatial relationship to activate the high-level capsule $j$.
	\item The center of gravity of the cluster, which represents the instantiation parameters $\vu^{[l+1]}_j \in\mathbb{R}^{D^{[l+1]}}$ of the high-level capsule $j$. 
\end{enumerate}
The prediction $\hat{\vu}^{[l+1]}_{i,j}$, or vote, of a capsule $i$ in a layer $l$ for a capsule $j$ in layer $l+1$ is produced by multiplying its features $\vu^{[l]}_{i}$ by a weight matrix $\mW^{[l]}_{i,j} \in \mathbb{R}^{D^{[l]} \times D^{[l+1]}}$:

\begin{equation}\label{eq:vote}
	\hat{\vu}^{[l+1]}_{i,j} = \mW^{[l]}_{i,j}\vu^{[l]}_{i}.
\end{equation}

The matrix $\hat{\mU}^{[l+1]} \in\mathbb{R}^{|\Omega^{[l]}| \times |\Omega^{[l+1]}| \times D^{[l+1]}}$ contains all the votes of two consecutive capsule layers.
In fully-connected or dense capsule layers there is a different weight matrix for each possible pair ($i$, $j$). 
We learn them through backpropagation and we hope that they will be the invariant viewpoint transformation matrices $\mT_{i,j}$ seen in computer graphics. As the viewpoint changes, the poses of the parts and the whole will change in a coordinated way so that any agreement between votes from different parts will persist \citep{hinton-dr}. As we mentioned in Section \ref{sec:equivariance}, a capsule works properly when its activation is locally invariant, and the instantiation parameters are equivariant.
The part-whole relations are modeled by the
coupling coefficients $c^{[l]}_{i,j} \in\mathbb{R}$, which are dependent on the input and not learned during backpropagation.
After the voting procedure, we need to route the votes to higher-layers to output object-capsules $\vu^{[l+1]}_{j}$ and adjust their connection $c^{[l]}_{i,j}$ based on the agreements.
A ``routing softmax'' determines the coupling coefficients (single-parent constraint), whose initial logits $b^{[l]}_{i,j}$ are the log prior probabilities that capsule $i$ should be coupled to capsule $j$
\begin{align}\label{eq:softmax}
		c^{[l]}_{i,j} = \sigma(\vb^{[l]}_{i,:})_j = \frac{e^{b^{[l]}_{i,j}}}{\sum_{k=1}^{|\Omega^{[l+1]}|}e^{b^{[l]}_{i,k}}}, &&
		\sum_{j=1}^{|\Omega^{[l+1]}|} c^{[l]}_{i,j} = 1.
\end{align}
During the routing procedure, each $c^{[l]}_{i,j}$ is refined by measuring the agreement between the output $\vu^{[l]}_{j}$ of a capsule $j$ and the prediction $\hat{\vu}^{[l+1]}_{i,j}$ 
Therefore, routing finds clusters of agreements between the predictions to solve the problem of assigning parts to wholes \citep{hinton-dr}. 
Routing can be viewed as a parallel \emph{attention} mechanism that allows each capsule at one level to attend to some active capsules at the level below and to ignore others \citep{hinton-dr}. It is an attention mechanism in the opposite direction as the competition is between the higher-level capsules that a lower-level capsule might send its vote to.

\begin{figure}
	\centering
	\includegraphics[width=1\linewidth]{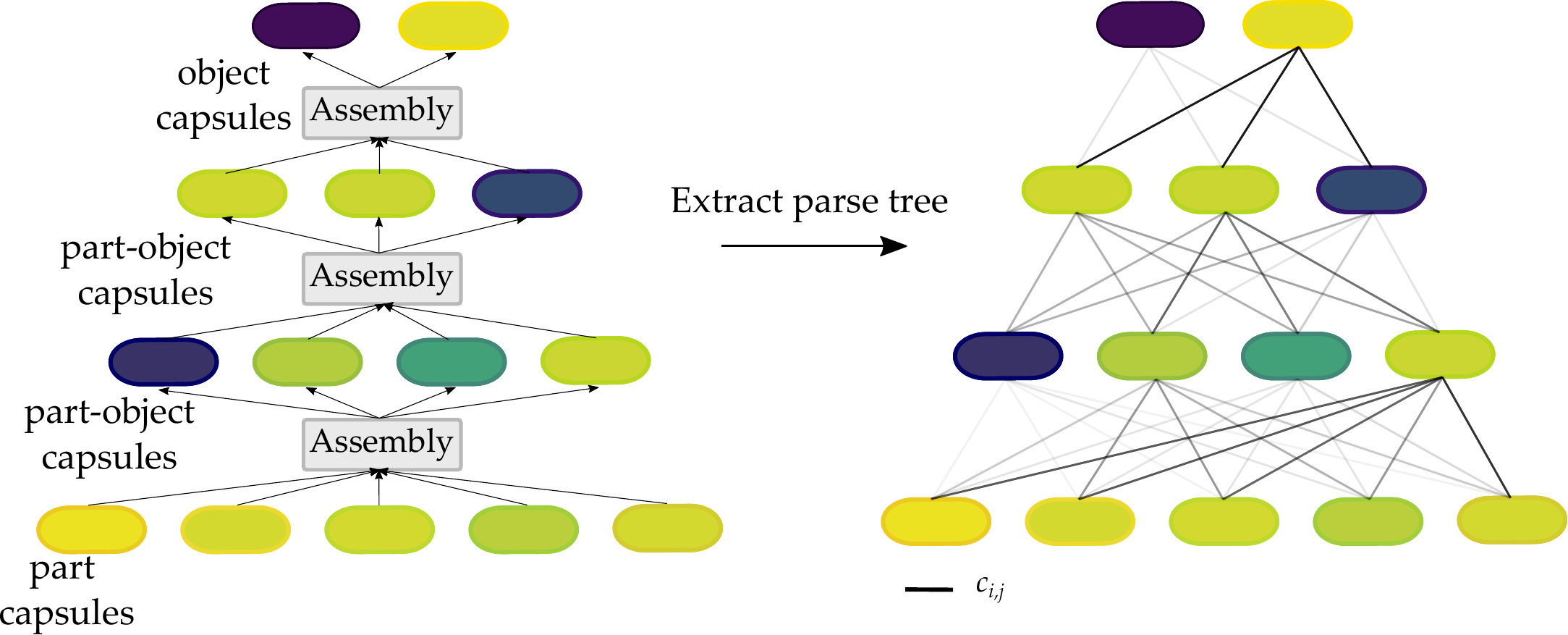}
	\caption[Extraction of the parse tree for an input image.]
	{Stacking multiple assembly modules on top of each other create a deeper hierarchy of objects. For a given input image, we can extract a parse tree where each node is a part or object capsule, and each connection is the probability of the part-whole (or child-parent) relation.}
	\label{fig:parsetree}
\end{figure}

At the end of the routing process, we extract a parse tree of the image using the coupling coefficients. Since we are dealing with deep learning models, the parse tree representations approximate the scene graphs in computer graphics. Figure~\ref{fig:parsetree} shows an example of a parse tree for a given input image. 

Capsules can be implemented in many different ways. In the next sections we will explore different implementations of capsules and routing.

\section{Capsule Networks Fundamentals}\label{sec:dr-capsnets}
This section gives an overview of a general architecture of a CapsNet. Then, we review the first routing algorithm proposed by~\citeauthor{hinton-dr}~\citep{hinton-dr} and other routing implementations.

\section{General Architecture}\label{sec:general_architecture}
\begin{figure}[h]
\centering
\includegraphics[width=1\textwidth]{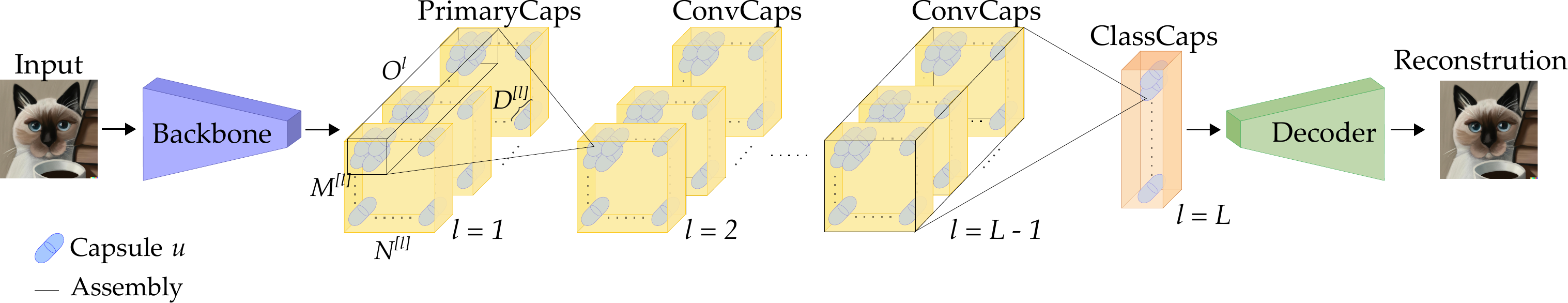}			
\caption{An example of a general CapsNet architecture composed by a convolutional backbone, a PrimaryCaps layer (which is a convolutional layer with squash activation and then reshaping), several ConvCaps layers (where a capsule in layer $l$ is computed using only a subset of capsules in layer $l-1$), and a ClassCaps (or FcCaps, where each capsule in layer $l$ is computed using all the capsules in layer $l-1$) layer. A decoder is used to reconstruct the input image.\label{fig:capsnet_general_architecture}}
\end{figure}

A CapsNet is composed of an encoder and a decoder part. The encoder consists of several capsule layers $l$ stacked on top of a convolutional backbone. There are three groups of capsule layers: primary capsules (PrimaryCaps, the first one, built upon convolutional layers), convolutional capsules (ConvCaps, each capsule has its own receptive field), and fully-connected capsules (FcCaps, each capsule is connected with all the capsules in the previous layer). Typically, a CapsNet comprises at least two capsule layers, PrimaryCaps and FcCaps (also called ClassCaps when it is the last layer, with one output capsule for each object class).
The decoder network is used to reconstruct the input image from the capsule of the target class (during training) or with highest activation (during inference) and it is used as a regularization that helps the network to encode better the objects instantiation features. However, many CapsNets implementations rely only on the encoder part since the decoder improves performances mostly on simple datasets. 
Figure~\ref{fig:capsnet_general_architecture} shows an example of a general CapsNet architecture. We can see that primary and convolutional capsules are organized in three dimensions and located in a $M^{[l]} \times N^{[l]} \times O^{[l]}$ grid structure, where $O^{[l]}$ is the number of capsule types. Therefore, in this scenario, $|\Omega^{[l]}|=M^{[l]} \times N^{[l]} \times O^{[l]}$.
As shown in Table~\ref{tab:caps_params}, the dimensions of the weight matrix $\mW^{l}$ between capsules in layer $l$ and capsules in layer $l+1$ can vary based on the groups of these two layer.

\begin{landscape}
\begin{table}[]
\begin{tabular}{llll}
\hline
\textbf{Caps layer $l$}    & \textbf{Caps layer $l+1$}       & \textbf{Weight sharing} & \textbf{$\mW^{[l]}$ dimensions} \\ \hline
{PrimaryCaps+Flatten, ConvCaps+Flatten, FcCaps} & FcCaps                     & False                   & $|\Omega^{[l]}| \times |\Omega^{[l+1]}| \times D^{[l]} \times D^{[l+1]}$                     \\
{PrimaryCaps+Flatten, ConvCaps+Flatten}, FcCaps & FcCaps                     & True                    & $|\Omega^{[l+1]}| \times D^{[l]} \times D^{[l+1]}$                     \\
{PrimaryCaps, ConvCaps} & FcCaps                     & False                   & $M^{[l]} \times N^{[l]} \times O^{[l]} \times |\Omega^{[l+1]}| \times D^{[l]} \times D^{[l+1]}$                     \\
{PrimaryCaps,  ConvCaps} & FcCaps                     & True                    & $O^{[l]} \times |\Omega^{[l+1]}| \times D^{[l]} \times D^{[l+1]}$                     \\
{PrimaryCaps, ConvCaps} & ConvCaps                     & False                   & $K^{[l]} \times K^{[l]} \times O^{[l]} \times O^{[l+1]} \times D^{[l]} \times D^{[l+1]}$                     \\
{PrimaryCaps,  ConvCaps} & ConvCaps                    & True                    & $K^{[l]} \times K^{[l]} \times O^{[l+1]} \times D^{[l]} \times D^{[l+1]}$                     \\ \hline
\end{tabular}
\caption{The dimensions of the weight matrix in two consecutive capsule layers in different configurations, where $K^{[l]}$ denotes the kernel size in convolutional capsule layers. Here, we assume that $|\Omega^{[l]}| \neq |\Omega^{[l+1]}|$.}
\label{tab:caps_params}
\end{table}
\end{landscape}

\subsection{Dynamic Routing}
In 2017~\citeauthor{hinton-dr}~\citep{hinton-dr}, Hinton presented the first CapsNet architecture along with the first description of the routing algorithm based on dynamic agreements (DR-CapsNet) using the scalar product. Due to its relevance for our work and for sake of simplicity, we will go into details only for this routing algorithm.

A capsule is implemented as an activity vector $\vu$ whose overall length  $\lVert \vu \rVert_2$ represents the probability of existence of the entity and its orientation represents its instantiation parameters. 

% Fig.~\ref{fig:dr-capsnet} shows the DR-CapsNet architecture proposed in~\cite{hinton-dr}, consisting of an encoder and a decoder part.
% The encoder part consists of one convolutional layer and two capsule layers, the decoder consists of three fully-connected layers.
% \begin{figure}[h]
% 	\centering
% 	\includegraphics[width=0.9\textwidth,]{Parts/01_LearningWithCapsNets/gfx/dr-capsnet.pdf}
% 	\caption[CapsNet architecture.]
% 	{CapsNet architecture described in \cite{hinton-dr}. There are one convolution layer (Conv1) and two capsule layers (PrimaryCaps and DigitCaps). The dynamic routing algorithm controls the information flow between capsule layers.}
% 	\label{fig:dr-capsnet}
% \end{figure}

% The first layer (Conv1) is a convolutional layer that converts pixel intensities to the \textit{activities} that are given in input to the first capsule layer.

% The  PrimaryCaps layer is the first capsule layer ($l = 1$). It is implemented as a reshaped convolutional layer with a $k_1 \times k_1$ filters and $T \times D_1$ channels where $T$ is the number of \textit{primary capsules} types. Note that each primary capsule has a spatial position $(m,n,o)$ in the grid. Overall, there are $|\Omega^{1}| = T \times M \times N$ primary capsules. The output layer (\textit{ClassCaps}) comprises of $|\Omega^{2}| = J$ $D_2$-dimensional \textit{class capsules}, one for each output class.\\
The information flow between two capsules is governed by the iterative routing algorithm described in \cite{hinton-dr} and shown in Alg.~\ref{alg:dynamicrouting}. 
\begin{algorithm}[ht]
\caption{Dynamic routing algorithm}\label{alg:dynamicrouting}
\begin{algorithmic}[1]
\Procedure{Dynamic-Routing($\hat{\mU}^{[l+1]}$, $r^l$)}{}
\State for each capsule $i \in |\Omega^{1}|$ and capsule $j  \in |\Omega^{1+1}|$: $b^{[l]}_{i,j} \gets 0$
\For {$r^l$ iterations}
	\State for each capsule $i$: $\vc^{[l]}_{i,:} \gets \sigma(\vb^{[l]}_{i,:})$ 
	\State for each capsule $j$: $\vs^{[l+1]}_{j} \gets \sum\limits_{i} c^{[l]}_{i,j} \hat{\vu}^{[l+1]}_{i,j}$
	\State for each capsule $j$: $\vu^{[l+1]}_{j} \gets g(\vs^{[l+1]}_{j})$
\State for each capsule $i$ and capsule $j$: $b^{[l]}_{i,j} \gets b^{[l]}_{i,j} + \vu^{[l]}_{j} \cdot \hat{\vu}^{[l+1]}_{i,j}$ \label{lst:line:agreement}
\EndFor
\State \Return $\mU^l_{j}$
\EndProcedure
\end{algorithmic}
\end{algorithm}
This procedure is used to dynamically compute the poses $\vu^{[l+1]}_j$ of capsules $j$ in layer $l+1$ given the vote matrix $\hat{\mU}^{[l+1]}$, which holds the predictions $\hat{\vu}^{[l+1]}_{i,j}$, computed using Equation~\ref{eq:vote}.
At the beginning of the routing algorithm (line 2), each logit $b^{[l]}_{i,j}$ is initialized to zero. 
The core of the routing algorithm is depicted in lines 3-8. At every iteration, the routing softmax (line 4, Equation~\ref{eq:softmax}) is applied to the logits $\vb^{[l]}_{i,:}$ to obtain the corresponding coupling coefficient $c^{[l]}_{i,j}$.
Then, the total unnormalized input $\vs^{[l+1]}_{j}$ of capsule $j$ is computed as the weighted average of the input predictions (line 5) as
\begin{equation}
    \vs^{[l+1]}_{j} \gets \sum\limits_{i} c^{[l]}_{i,j} \hat{\vu}^{[l+1]}_{i,j}.
\end{equation}\label{eq:sj}
Each vote $\hat{\vu}^{[l+1]}_{i,j}$ is weighted by the corresponding coupling coefficient $c^{[l]}_{i,j}$. The vector $\vu^{[l+1]}_{j}$ is the normalized \textit{squashed} $\vs^{[l+1]}_{j}$ (line 6), where the squashing function $g$ map the magnitude of a vector into the range [0, 1) 
 \begin{equation}\label{eq:squashing}
	\vu = g(\vs) = \frac{\|\vs\|^{2}}{1 + \|\vs\|^{2}} \frac{\vs}{\|\vs\|}.
\end{equation}
Then, each $b^{[l]}_{i,j}$ is refined by measuring the agreement between the output $\vu^{[l]}_j$ of a capsule $j$ and the prediction $\hat{\vu}^{[l+1]}_{i,j}$ (line 7). Therefore, if there is a strong agreement, the corresponding link strength $b^{[l]}_{i,j}$ between capsules $i$ and $j$ is increased, decreased otherwise. The agreement is simply the scalar product $\vu^{[l]}_{j} \cdot \hat{\vu}^{[l+1]}_{i,j}$.
Finally, after $r^{[l]}$ iterations of lines 4-7, the routing algorithm output the final pose $\vu^{[l]}_j$ for each digit capsule. 

The activity vector of the correct digit capsule is used to reconstruct the input image. During testing we take the capsule with the highest magnitude. This capsule is fed into the decoder which consists of three traditional fully-connected layers. The decoder encourage the digit capsules to encode the instantiation parameters \citep{hinton-dr}.

Note that this iterative process is executed both on training and testing time only between two consecutive capsule layers.

In order to train the network, two losses are used: the \textit{margin loss} and the \textit{reconstruction loss}.

The margin loss is based on the idea that the class capsule for a label $\gamma$ should have high magnitude if and only if that the object with label $\gamma$ is present in the input image. We have a separate margin loss $\mathcal{M}_{\gamma}$ for each label $\gamma$:

\begin{equation}
	\mathcal{M}_{\gamma} = T_{\gamma} \textrm{max}(0, m^{+} - \|\vu^{[l]}_{\gamma}\|_2)^2 + \lambda(1 - T_{\gamma})\textrm{max}(0, \|\vu^{[l]}_{\gamma}\|_2 - m^{-})^2
\end{equation}

where  $T_{\gamma} = 1$ if and only if the object of label $y$ is present. The $\lambda$ term prevents initial learning from shrinking the lengths of the capsules of all the classes.

The reconstruction loss $\mathcal{R}$ is the sum of squared differences between the reconstructed image and the original one. It is used a regularization term during training.

The total margin loss is defined as:
\begin{equation}
	\mathcal{L}_{\mathcal{M}} = \sum\limits_{\gamma} \mathcal{M}_{\gamma} + \beta \mathcal{R}
\end{equation}

DR-CapsNets achieved higher classification performance than the convolutional baseline on MNIST~\citep{mnist}, Fahion-MNIST~\citep{fashion-mnist}, CIFAR-10~\citep{CIFAR10} and SVHN~\citep{SVHN} datasets. They also showed more robustness to affine transformations (affNIST) and novel viewpoints (smallNORB). Capsules are also very good for dealing with segmentation highly overlapping digits (MultiMNIST~\citep{hinton-dr}).
\section{Drawbacks}
Capsule networks have some limitations.
\begin{enumerate}
	\item Capsules make a very strong representational assumption \citep{hinton-dr}: at each location in the image, there is at most one instance of the type of entity a capsule represents. As a matter of fact, the only thing that binds the instantiation units together is the fact that they are the pose outputs of the same capsule. Therefore, it is not possible for a capsule to represent more than one instance of its visual entity at the same time \citep{hinton-dr}. This is a serious weakness because it will
always be possible to confound the system by putting two instances of the same
visual entity with slightly different poses in the same region. This phenomenon is known as \textit{crowding} and it occurs also in human vision.
	\item Capsules, like other generative models, account for everything in the image so they have poorer performances when the backgrounds are too much varied, like in CIFAR10, and the size of the network is not big enough to model the clutter.
	\item Training capsule networks is computationally expensive because of the large transformation matrices and because of the routing algorithm which can not benefit from the parallel computation of modern GPUs, increasing both training and testing time respect to a standard CNN. This is one of the reason why they have not been trained yet on datasets with high resolution such as ImageNet~\cite{deng2009imagenet}.
	\item A capsule network has many hyperparameters like the number of types of capsules, their dimensions in every layer, the shape of the kernels etc. This could make fine-tuning in capsule network more difficult than in standard CNNs.
\end{enumerate} 

\section{Capsule Networks Follow-Ups}

\paragraph{Capsule Networks Architectures} 
Capsule networks were first introduced by ~\citet{hinton-dr}, and since then, much work has been done to improve the routing stage and build deeper models. Regarding the routing algorithm, \citet{hinton-em} replaces the dynamic routing with Expectation-Maximization, adopting matrix capsules instead of vector capsules. \citet{Wang2018AnOV} model the routing strategy as an optimization problem. \citet{nn-encaps} use master and aide branches to reduce the complexity of the routing process. \citet{peer2019gammacapsules} use inverse distances instead of the dot product to compute the agreements between capsules to increase their transparency and robustness against adversarial attacks. 
\citet{self-routing} incorporates a \textit{self-routing} method such that capsule models no longer require agreements.  \citet{desousa} replace the routing algorithm with a variational inference of part-object connections in a probabilistic capsule network, leading to a significant speedup without sacrificing performance. 
\citet{ribeiro2020capsule} propose a new routing algorithm derived from Variational Bayes for fitting a mixture of transforming Gaussians.
\citet{subspacecaps} model entities through a group of capsule subspaces without any form of routing.
Since the CapsNet model introduced by \citet{hinton-dr} is a shallow network, several works attempted to build deep CapsNets. 
 \citet{deepcaps} propose a deep capsule network architecture that uses a novel 3D convolution-based dynamic routing algorithm to improve the performance of CapsNets for more complex image datasets. \citet{res-caps} introduce residual connections to train deeper capsule networks.
The above-mentioned CapsNets architectures rely on the fact that capsules model specific location-dependent features of objects in the image. However, \citep{hinton-stacked-caps} claims that it is better to model capsules as \textit{set} of parts or objects. This assumption ensures that if the parts can be detected equivariantly, then the inferences for the objects will also be
equivariant \cite{generative-caps-model}. Therefore, \citet{hinton-stacked-caps} propose an unsupervised stacked capsule autoencoder (SCAE) consisting of two stages. First, it segments an image into constituent parts, predicting the presence and poses of part templates. Then, it tries to organize discovered parts and their poses into a smaller set of objects, which are then used to reconstruct part poses. Similarly, \citet{generative-caps-model} consider a scene or an image composed of a set of objects in different poses, which are instations of affine-transformed fixed templates. They show that the parts detected in the first stage by SCAE are not equivariant to rotation. Therefore, they propose a generative model and derive a variational algorithm for inferring the transformation of each model object in a scene and the assignments of observed parts to the objects. The model is interpretable and admits prior knowledge about an object. Furthermore, their formulation is also composable in that models for individual objects can be learned separately and then combined together at inference time. Alternatevely to capsules, other object-centric representation learning approaches using neural networks exist in the literature, for example, slot-based models~\cite{locatello}. Slots are ``universal'' capsules that can bind to any object in the input; they provide a sort of working memory with a fixed capacity which can be used to access independent object representations simultaneously.

Recently, \citeauthor{glom}~\cite{glom} described how to represent part-whole hierarchies in neural networks, which attempts to merge the advantages of capsules, contrastive learning, distillation and transformers into a single theoretical system (known as GLOM). GLOM aims at mimicking how humans parse visual scenes. Still, these approaches are not tested on more complicated datasets such as ImageNet, which limits their use in downstream tasks, and consequently, capsules have not yet gained widespread adoption in the literature.  

\paragraph{Sparse Capsule Networks} 
A naive solution to reduce uncertainty within the routing algorithm is to run more iterations. As shown by~\citet{paik} and \citet{gu}, the routing algorithms tend to overly polarize the link strengths, namely a simple route in which each input capsule sends its output to only one capsule and all other routes are suppressed. On the one hand, this behavior is desirable because the routing algorithm computes binary decisions to either connect or disconnect objects and parts. On the other hand, running many iterations is computationally expensive, and it is only helpful in the case of networks with few parameters, as we will see in Chapter~\ref{ch:icann_routing_annealing}. \citet{sparse-caps} trained CapsNets in an unsupervised setting, showing that the routing algorithm no longer discriminates among capsules: the coupling coefficients collapse to the same value. Therefore, they sparsify latent capsule layers activities by masking output capsules according to a custom ranking function. 
\citet{hinton-stacked-caps} impose sparsity and entropy constraints into capsules, but they do not employ an iterative routing mechanism. \citet{ladder-caps} introduced a structured pruning layer called ladder capsule layers, which removes irrelevant capsules, namely capsules with low activities. \citet{3D-Caps} solve the task of 3D object classification on point clouds with pruned capsule networks. They aimed to compress robust capsule models to deploy them on resource-constrained devices.

\section{Comparison with vision transformers}\label{sec:capsnetvstrans}
In this section, we compare Vision Transformers (ViTs)~\cite{dosovitskiy2021an} and CapsNets architectures, focusing on their similarities and differences.
\begin{figure}[h]
\centering
\includegraphics[width=1\textwidth]{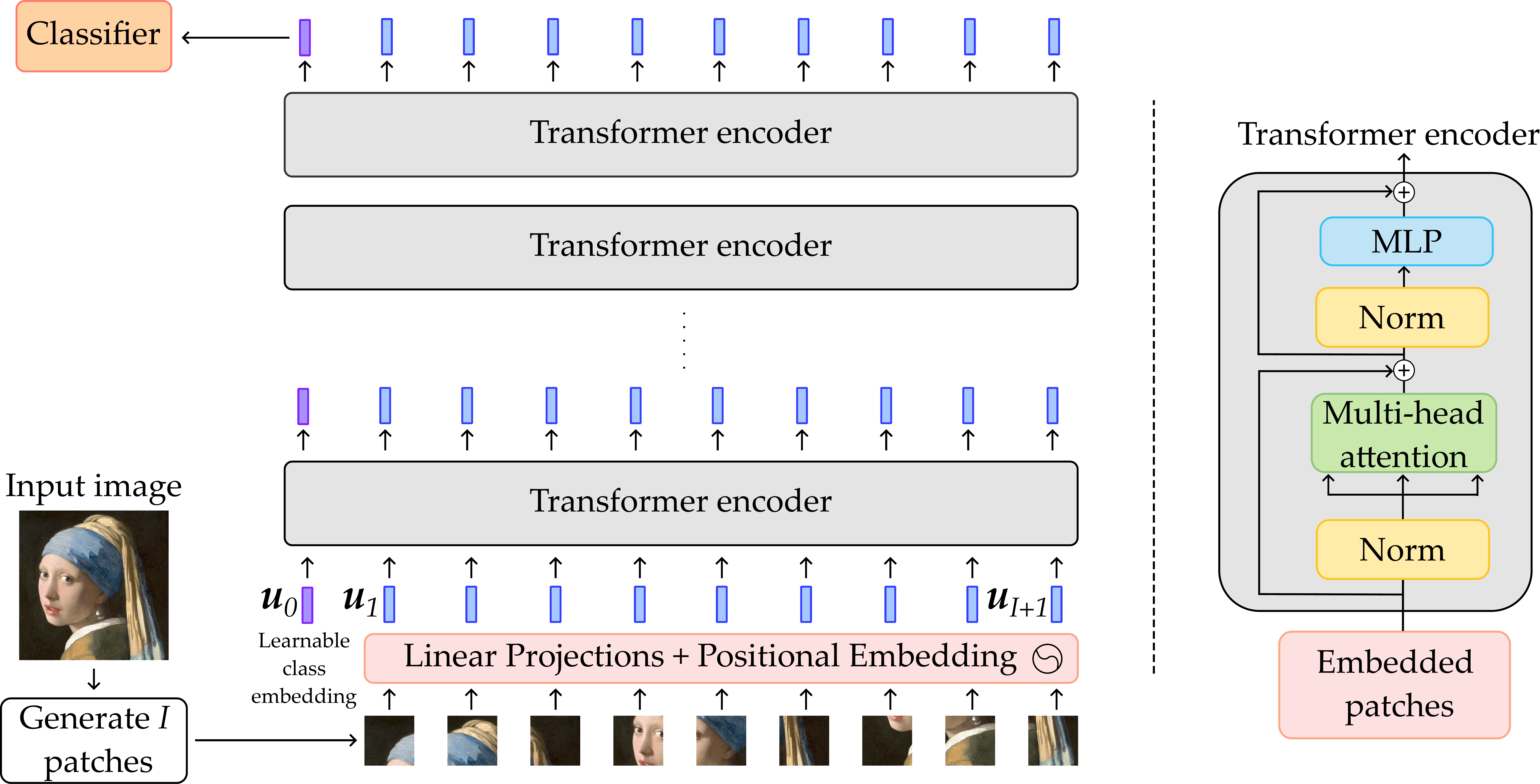}	
\caption{Overview of the ViT architecture illustrated in \cite{dosovitskiy2021an}. Fixed-size patches are extracted from an input image; then, they are linearly embedded with the addition of position embeddings to form tokens, which are then fed to a standard transformer encoder. Multiple blocks of transformer encoder can be stacked together for deeper models.
An extra learnable classification token is added to the input sequence. The embedding vector of this token in the last layer can be seen as a compact representation of the input image. Therefore, it is fed to a multilayer perceptron network (MLP) to perform classification. An alternative is to have a global average pool over all the output embedding vectors.\label{fig:vit_general_architecture}}
\end{figure}

\begin{figure}[h]
\centering
\includegraphics[width=0.8\textwidth]{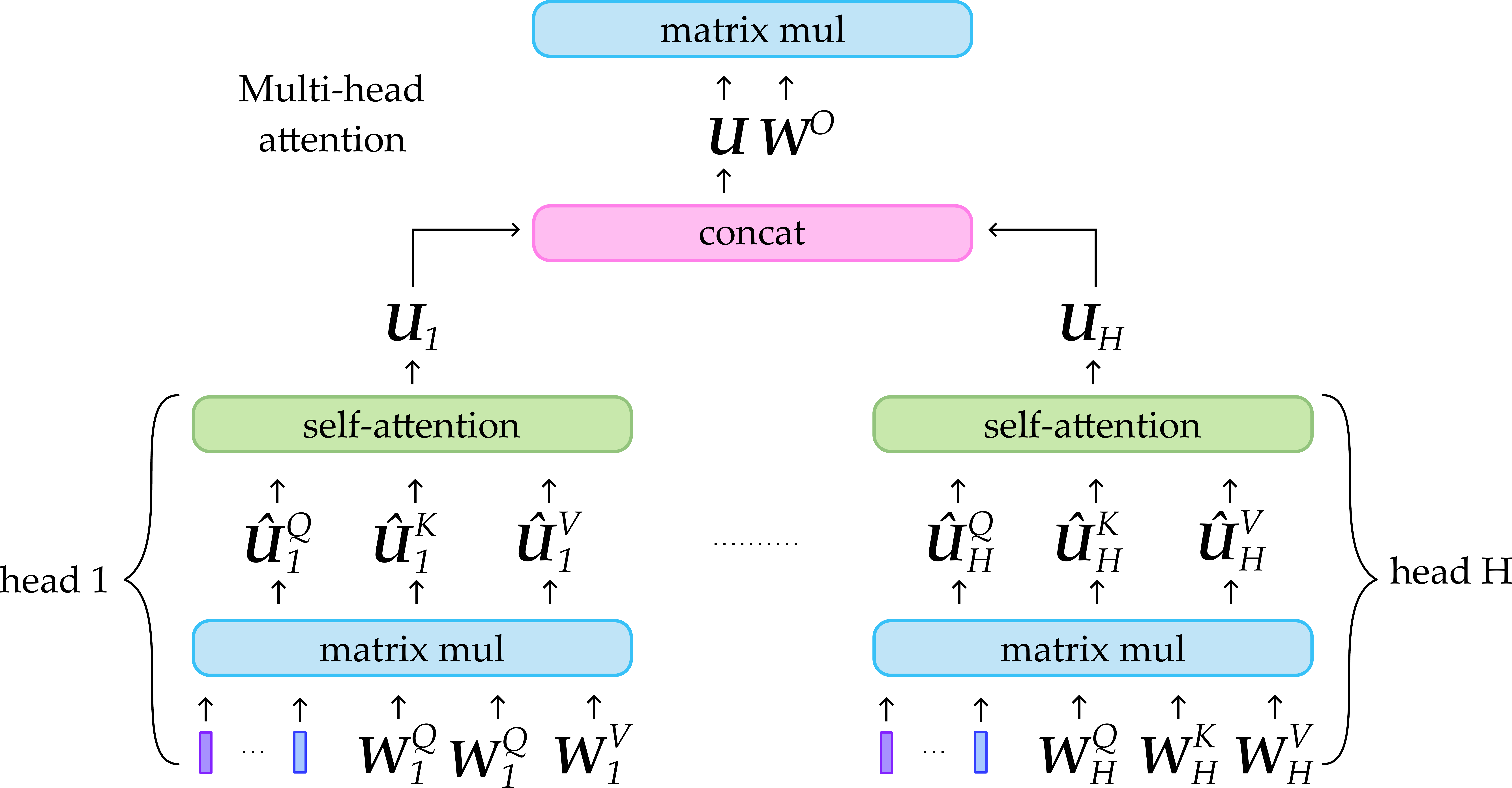}			
\caption{Multi-head attention is a module that runs MLPs and self-attentions several times in parallel. The independent attentions computed by different heads are then concatenated and linearly transformed into the expected dimension. Intuitively, multiple attention heads allow for attending to parts of the input differently.\label{fig:vit_multihead}}
\end{figure}

Even though capsule networks have been around for a while, they have not received as much attention as transformers. These networks, introduced around the same time as CapsNets, have become quite popular and are widely used in various applications, including large language models and computer vision. ViTs, which are transformers designed for computer vision, generally consist of only the encoder part of a standard transformer. An overview of ViT architecture is depicted in Figure~\ref{fig:vit_general_architecture}, while Figur~\ref{fig:vit_multihead} depicts the multi-head attention layer that extends the idea of single-head attention by performing self-attention computations multiple times in parallel.

In this Section, we use the same notation we adopted when describing capsule network components. For example, we denote with $\vu^{[l]} \in \mathbb{R}^{D^{[l]}}$ the $D$-dimensional embedding vectors in layer $l$, as we did with a capsule vector. Since we consider only two consecutive layers, let $\mU^{[l]} = (\vu^{[l]}_{1}, ..., \vu^{[l]}_{I})$ denote a matrix of input embedding vectors in layer $l$, and $\mU^{[l+1]} = (\vu^{[l+1]}_{1}, ..., \vu^{[l+1]}_{J})$ denote a matrix of output embedding vectors in layer $l+1$ (or capsules), with $\vu^{[l]}_{i} = (u^{[l]}_1, ..., u^{[l]}_{D^{[l]}})$ and $\vu^{[l+1]}_{j} = (u^{[l+1]}_1, ..., u^{[l+1]}_{D^{[l+1]}})$. We can omit the $l$ notation since $i \in \{1, 2, ..., I\}$ is used for inputs of the single-head attention and $j \in \{1, 2, ..., J\}$ for the outputs. We also use $D = D^{[l]}$ and $P = D^{[l+1]}$.

In scaled dot product self-attention~\cite{dosovitskiy2021an} and shown in Figure~\ref{fig:vit_multihead}, $I$ input embedding vectors $\mU_i \in \mathbb{R}^{I \times D}$ are transformed into query, key and value matrices for each head $h \in \{1, 2, ..., H\}$ as 
\begin{align}\label{eq:querykeyvalue}
		\hat{\mU}^{Q}_{h} = \mU_i\mW^Q_h, && \hat{\mU}^{K}_h = \mU_i\mW^K_h, && \hat{\mU}^{V}_h = \mU_i\mW^V_h,
\end{align}
where $\mW^Q_h$, $\mW^K_h$ and $\mW^V_h$ are the trainable $D \times P$ transformation matrices. Therefore, $\hat{\mU}^{Q}_h$, $\hat{\mU}^{K}_h$ and $\hat{\mU}^{V}_h$ are  $I \times P$ dimensional. Here, we assume that the new query, key, and value vectors have the same dimension as the embeddings (usually, they are smaller). Let be $\mW^Q$, $\mW^K$ and $\mW^V$ the weight matrices considering all attention heads, these are $H \times D \times P$ dimensional, where usually $H < J$. Note that typically, between two consecutive fully-connected capsule layers (so capsules in these layers are flattened and not organized in a grid, without weight sharing), as also shown in Table~\ref{tab:caps_params}, the transformation weight matrix $\mW^Q$ is $I \times J \times D \times P$ dimensional, therefore the vote matrix $\hat{\mU}$ is $I \times J \times P$ dimensional. The self-attention mechanism computes the attention weights $c_{i,j,h}$ (or coupling coefficients in capsules) between each pair $(i,j)$ of embedding vectors in head $h$ as
\begin{align}\label{eq:vit_softmax}
		c_{i,j, h} = \sigma\left(\frac{{\hat{\mU}^{Q}_h}{{}\hat{\mU}_{h}^{K}}^{\!\!\top}}{\sqrt{D}}\right)_{ij}, && 
		\sum_{i=1}^{I} c_{i,j,h} = 1,
\end{align}
where $\sigma$ is the softmax function. Then, a single output embedding (or capsule) $\vu_{j, h}$ in a head $h$ for token $j$ is computed as a weighted average of the input token vectors
\begin{equation}\label{eq:outputemb}
    \vu_{j,h} = \sum\limits_{k=1}^{I} c_{j,k,h} {\hat{\mU}^{V}_{h,k}}
\end{equation}
Note that in ViTs, the number of embedding vectors for the input tokens is the same in each layer, namely $I=J$. Therefore we can rewrite Equation~\ref{eq:outputemb} as 
\begin{equation}\label{eq:outputemb_original}
    \vu_{i,h} = \sum\limits_{j=1}^{I} c_{i,j,h} {\hat{\mU}^{V}_{h,j}}
\end{equation}
Therefore, a significant distinction between capsules and transformer embeddings lies in how they utilize context. In the case of capsules, to calculate the vector of an output capsule (and attention weights), all other output capsules are required as context (remember that in the vote matrix, there are $I \times J$ predictions of the input capsule vectors for the output capsule vectors). In contrast, in ViTs, the vector of an output embedding is computed by considering all the other input embeddings as context (there are three ``vote'' matrices, query, key and value, so the total number of votes is $I \times 3$). Note that, in the language of transformers, the output capsules at iteration $r-1$ act as the query matrix at iteration $r$, while capsule votes act as the key and value matrices. See Figure 23 in \cite{ribeiro2022learning} for comparing capsule routing and self-attention in Pytorch code.

So far, we have seen that CapsNets and ViTs have a common ground regarding learned representations. For example, both methodologies organize neuron activities into vectors to represent concepts, and they compute the agreement between these vectors as a filtering mechanism to refine concepts based on global context~\cite{ribeiro2020capsule}. Moreover, both models can be stacked on top of a CNN backbone to build hybrid architectures. Note that typically, a ResNet model is preferred since it has only one max pooling layer, so that the spatial relationships are not completely lost when going deeper in the network. In the case of a ViT, as an alternative to raw image patches, the input sequence can be formed from feature maps of a CNN, and then patches are extracted from a CNN feature map\cite{dosovitskiy2021an}. This hybrid model inherits, for example, the ability to learn mutual regions at different scales in a CNN, thanks to multiple convolutional layers with different kernel sizes. As we will see in Chapter~\ref{ch:icip_effcapsnets}, in the case of a CapsNet, the CNN backbone (without pooling layers) can help in extracting effective low-level part descriptions (primary capsules) since CNNs are very good at learning low-level features, such as edges, corners and basic local shape information. 

However, CapsNets and ViTs have different \textit{inductive biases}. Recall that inductive biases in deep learning architectures refer to the implicit assumptions or constraints that are built into the design of a neural network model. These biases guide the learning process by favoring certain types of solutions or representations over others.

\paragraph{Locality}
Primary and convolutional capsule layers are organized in a 3-dimensional grid, and the sliding window approach (as in CNNs) assumes that important features are present in small, local regions of an image and that these features can be detected irrespective of their position within the image. On the contrary, ViTs do not have a built-in notion of spatial locality and layers are not organized in a grid structure. Images are tokenized into smaller patches, which is a different inductive bias than the sliding window approach. They use positional embeddings to encode the spatial information of the input image, and these positional embeddings provide information about where each token (representing a patch of the image) is located in the image. 

\begin{figure}[h]
\centering
\includegraphics[width=0.8\textwidth]{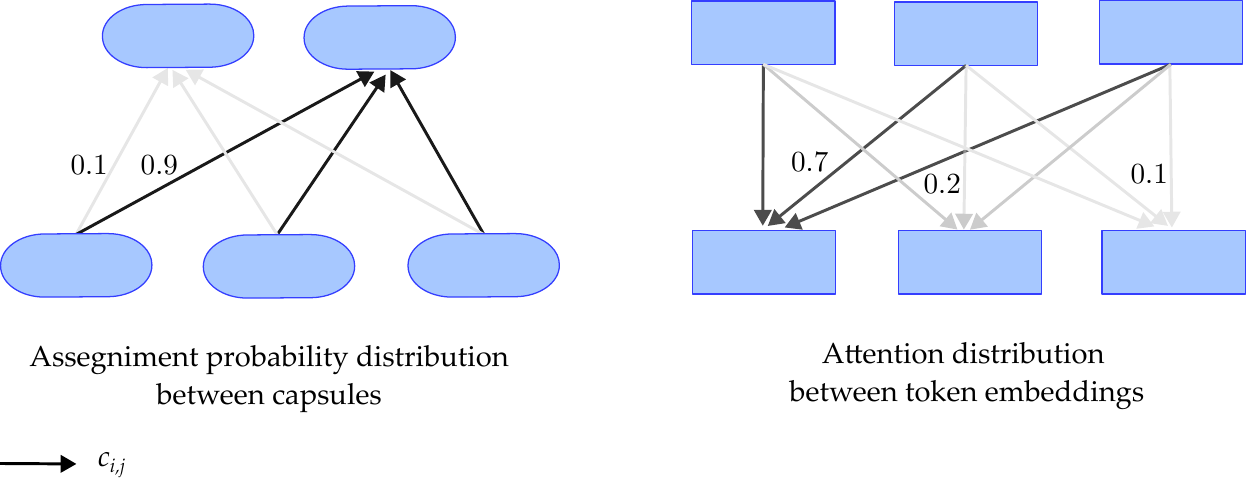}
\caption{In CapsNets, the competition is between capsules in the higher layer (bottom-up attention), while in ViTs, the competition is between embeddings in the lower one (top-down attention).\label{fig:routing_vs_attention}}
\end{figure}

\paragraph{Single-parent assumption}
Both routing and self-attention compute attention weights between capsules or embeddings. However the semantics of this attention weights is different. In fact, in CapsNets, since routing is a clustering mechanism, $c_{ij}$ is an assignment probability, namely to what extent object $i$ is part of object $j$. This is also known as the single-parent assumption (see the softmax in Equation~\ref{eq:softmax}), since the goal of a CapsNet is to build part-whole hierarchies.  While in ViT, $c_{ij}$ models how to attend to different parts of the input and how information from different parts contributes to the updates of representations. We can see in Figure~\ref{fig:routing_vs_attention} the opposite directions of the connections in capsules (bottom-up) and token embeddings (top-down). 
% One trick we can use in capsules is to employ a self-attention layer between capsules in layer $l$ before the voting stage, using a $D \times P$ dimensional matrix. In this case, in the voting procedure, we do not employ raw input capsules but contextualized versions since each input capsule is transformed according to all the other input capsules.
\paragraph{Number of objects}
Figure~\ref{fig:routing_vs_attention} also shows that, when assembling capsules, we often have a smaller number of output capsules than input capsules ($J < I$) because we assume that, in an input image, the number of objects is lower than the number of parts, since an object is composed of many parts. This also means that we do not have any prior assumption on the representations of the higher layer capsules, what they are supposed to represent is not known in advance. In contrast, in ViTs, the number of embeddings in each layer is constant ($J=I$), corresponding to the number of patches. This is because an embedding can be considered a contextualized representation of the same patch in each layer.
% This is equivalent to the assumption that the higher layer is first initialized with the representations from the lower layer, and then it is updated based on the attention probabilities computed by comparing these initial representations with the representations from the lower layer\footnote{\url{https://samiraabnar.github.io/articles/2019-03/capsule}}

\paragraph{Capsule types and attention heads}
Different capsule types in CapsNets and attention heads in ViTs are akin to the diverse kernels in CNNs.
They aim to process lower-layer representations from multiple viewpoints when computing higher-layer representations. In CapsNets, different transformation matrices exist between pairs of capsule types from adjacent layers. This way, each higher-layer capsule sees the entity in the lower-layer capsule from a different point of view. Similarly, in ViTs, multiple attention heads employ distinct transformation matrices to compute key, value, and query projections, allowing each head to focus on a different perspective of lower-layer representations. In real implementations\footnote{\url{https://pytorch.org/docs/stable/generated/torch.nn.MultiheadAttention.html}}, each head focuses on a different portion of the input vector, namely $D$ will be split across $H$ heads (i.e. each head will have dimension $D // H$). However, the key difference between the two architectures is that, in the end, in CapsNets, assignment probabilities for a lower-layer capsule are normalized over all higher-layer capsules, irrespective of their types, resulting in one assignment distribution per lower-layer capsule (as seen in Figure~\ref{fig:routing_vs_attention}). In contrast, in ViTs, each attention head independently processes its input, yielding separate attention distributions for each position in the higher layer. The attention head outputs are combined in the last step by concatenating and linearly transforming them to compute the final output of the multi-headed attention block.

\paragraph{Invariance and Equivariance}
As explained in Section~\ref{sec:equivariance} and Section~\ref{sec:intro:whatarecaps}, CapsNets take advantage of the linear manifold that handles the source of variance in images thanks to the transformation matrices $\mW$s used to compute the votes $\hat{\vu}$s. Therefore, we say that the vote transformation matrix $\mW$ is \textit{biased} towards encoding invariance to viewpoint transformations, and capsule vectors are \textit{biased} towards capturing equivariance of neural activities~\citet{ribeiro2022learning}. In ViTs, only MLP layers are local and translationally equivariant~\cite{dosovitskiy2021an}.

So, we have seen that ViTs inductive biases are more relaxed compared to CapsNets, or even CNNs. However, inductive biases are essential because they help neural networks generalize from limited training data and make predictions in situations they haven't seen before. Thus, ViTs do not generalize well when trained on insufficient amounts of data~\cite{dosovitskiy2021an} and must learn all spatial relations between the patches from scratch. However, when trained on large amounts of data, large-scale training trumps inductive bias~\cite{dosovitskiy2021an}. Recently, \citet{glom}\citep{glom} proposed a single theoretical framework (called GLOM), which attempts to merge the advantages of capsules and transformers to parse visual scenes. However, research is still in the early stages, and as of now, a high-performing GLOM network pretrained on ImageNet has not surfaced yet.
\cleardoublepage
\ctparttext{
This part delves into the intriguing aspects of capsule networks, focusing on three key aspects that aim to unlock their full potential:

\begin{itemize}
    \item \textbf{Effectiveness of the routing algorithm}: We introduce a novel technique that anneals the number of routing iterations during training, striking better performance, especially for architectures with fewer parameters.
    \item \textbf{Extraction of low-level part descriptors}: We propose a method to extract better first-layer capsules promoting sparse topologies. We improve their computational efficiency by reducing the number of capsules.
    \item \textbf{Cardinality of part-whole relationships}: We show that capsules with low entropy hold the potential to extract more succinct and more discriminative part-relationships than traditional capsule networks. This can be achieved without accounting for everything in the input image, even with reasonable network sizes.
\end{itemize}
}
\part{Demystifying Capsule Networks Components}\label{part:capsnet-components}
%*****************************************
\chapter{Capsule networks with routing annealing}\label{ch:icann_routing_annealing}
%*****************************************
The research findings and insights presented in this chapter were originally published in the following work: \fullcite{renzulli-ra}.
\section{Introduction}
\label{sec:intro}
Recently, the contribution of the routing algorithm to DR-CapsNets generalization ability and robustness to affine transformations has been questioned~\cite{gu,paik}. Typically, the number of routing iterations $r$ is fixed once and for all during training and inference. Paik et al.~\cite{paik} highlight that running just one iteration of the routing algorithm (assigning the connection strengths uniformly or randomly) leads to better results. This is explained as more iterations of the routing algorithms do not change the classification result but polarize the link strengths \cite{paik}. Gu et al.~\cite{gu} mitigate this problem with a simple but effective solution in which the transformation matrices are shared between all capsule types. However, these works do not change the number of iterations during the training process nor the number of capsule types and their dimensions, which, as we will see, they do have a strong impact on the number of iterations of the routing algorithm that achieves better generalization.
Therefore, it is unclear whether the routing algorithm improves the performance of DR-CapsNets and the optimal number of iterations.

This chapter provides new evidence on the benefits of routing proposing
\textit{Routing Annealing (RA)}, a novel technique where the number of routing iterations is iteratively found at training time.
With RA, the number of iterations of the routing algorithms increases whenever the network performance reaches a loss plateau.
We observed that, for the same number of routing iterations, a gradual ramp thereof allows to reach better minima of the loss function.
Our experiments over multiple datasets show better performance when using RA, especially when the number of capsules in the network is limited, i.e. where DR-CapsNets performance is weaker.
We also found that the number of routing iterations depends on the number of capsules, their dimensions and on the dataset itself.
% The rest of this chapter is organized as follows. 
% In Section~\ref{sec:ra_meth} we present \textit{Routing Annealing} (RA), our proposed training procedure for DR-CapsNets. Then, Section~\ref{sec:ra_architecture} describe the architecture used to test RA.
% Finally, in Sec. \ref{secra_:experiments} we experiment with RA over multiple datasets, highlighting the benefits of the former.
% Section~\ref{sec:ra_conclusion} drawn the conclusions and discusses further developments of this work.

\section{Methodology}
\label{sec:ra_meth}

This section first describes the standard methodology training algorithm and \textit{Routing Annealing} (RA), the routing training technique we propose in this work, and then discusses its relation with the simulated annealing. For simplicity of notation, we assume there are only two capsule layers (primary capsules and class capsules), and we suppress the index $l$ for capsule vectors $\vu$ and routing iterations $r$. However, RA can be applied to any multilayered CapsNet whose connections are governed by any iterative routing algorithm and where the number of routing iterations can vary for different layers.

\subsection{Training with Fixed Routing}\label{subsec:fixed-routing}

As a reference, Alg.~\ref{alg:fixed-routing} shows the standard strategy for training a DR-CapsNet.
The network parameters are optimized with standard backpropagation of the error gradients for a number of epochs until some stop criterion is met.
For each epoch, the forward pass (line 5-11) is computed, followed by error gradients backpropagation and parameter update (line 12).
The training procedure ends when the loss computed over a validation set does not decrease for $p$ epochs in a row ($p$ is usually termed as \textit{patience}).
The algorithm returns the network (i.e., the learned parameters  set) that yields the lowest loss on the validation set.  
In this procedure, as can be noted from line~\ref{lst:line:fixed-routing} which refers to Alg. \ref{alg:dynamicrouting}, the number of routing iterations $r$ is fixed once for all (usually, $r$=3), so we refer to this technique as \textit{Fixed Routing} (FR).
Notice that when the trained network is deployed for inference, the routing algorithm is executed for $r$ iterations, as well.
A standard procedure towards optimising the iterations number would be to optimize $r$ with a grid-search strategy: one runs as many simulations as $r$ values to test, during which $r$ is kept constant.
However, we experimentally show that this approach leads to sub-optimal performance, which motivates the design of our routing technique.

\begin{algorithm}[ht]
\caption{Training with \textit{Fixed Routing}: learns the network parameters for a fixed number of iterations $r$.}
\label{alg:fixed-routing}
\begin{algorithmic}[1]
\Procedure{Fixed-Routing($r$, $p$)}{}
\State initialize DR-CapsNet
\State $e \gets 0$;~~ $\mathcal{L}^\star \gets 0$; $e^\star \gets 0$ 
\While {$e - e^\star < p$}
	\State compute all primary capsules poses $\vu_i$ and votes $\hat{\vu}_{i,j}$
	\State compute all class capsules poses: $\vu_{j} \gets$ \Call{Routing($\hat{\vu}_{i,j}$, $r$)}{} \label{lst:line:fixed-routing}
	\State evaluate current loss $\mathcal{L}$ on the validation set
    \If{$\mathcal{L} < \mathcal{L}^\star$}
        \State $\mathcal{L}^\star \gets \mathcal{L}$; $e^\star \gets e$
    \EndIf
	\State $e \gets e+1$
	\State backpropagate error gradients and update parameters
\EndWhile
\State \Return DR-CapsNet of epoch $e^\star$ with the best loss value
\EndProcedure
\end{algorithmic}
\end{algorithm}

\subsection{Training with Routing Annealing}\label{subsec:annealing}
In this section we propose \textit{Routing Annealing} (RA), an iterative method to jointly optimize the number of routing iterations $r^\star$ and the network parameters.
In a nutshell, RA finds $r^\star$ adaptively during training for a given capsule architecture over a given dataset and is described in pseudo-code as Alg.~\ref{alg:routingannealing}.
The algorithm takes as input: $r_0$, the initial value of $r$; $r_T$, the maximum value for $r$; $s$, the schedule used to increase $r$; the patience $p$, in number of epochs.
Let us denote as $r_{k}$ the value of $r$ at step $k$: we say that every time $r$ increases, an \emph{annealing step} is performed.
We denote as $\mathcal{L}_{k}^\star$ and  $e_{k}^\star$ the lowest losses achieved so far and the corresponding epochs for each $r_{k}$.
The main difference between Alg.~\ref{alg:fixed-routing} and \ref{alg:routingannealing} lies in line~\ref{lst:line:annealing-while} where we loop over the possible values of $r$ instead over the number of epochs and in line~\ref{lst:line:annealing-step} where the number of routing iterations is increased. In line 5, Alg.~\ref{alg:dynamicrouting} is used as core routing algorithm.
At step $k$, we increase $r$ by $s$ if the validation loss $\mathcal{L}_{k}^\star$ does not decrease for $p$ epochs (lines 7-8).
Every time $r$ is increased, the training does not start from scratch again.
Instead it is resumed with the network weights with the best loss achieved with the previous value of $r$, namely the network at epoch $e_{k-1}^\star$ (line 9).
Here we assume that we save the network weights for each epoch. When $r$ reaches the maximum allowed $r_T$, the training procedure ends and best network obtained during training along with the corresponding number of routing iterations is returned (lines 16-17).
\begin{algorithm}[ht]
\caption{Training with \textit{Routing Annealing}: learns the number of iterations $r^\star$ jointly with the network parameters.}
\label{alg:routingannealing}
\begin{algorithmic}[1]
\Procedure{Routing-Annealing($r_{0}$, $r_{T}$, $s$, $p$)}{}
\State $\vr$; $r \gets r_{0}$; $\mathcal{L}^\star_0 \gets +\infty$; $e^\star_{0} \gets 0$ , $e \gets 0$; $k \gets 0$ 
\While {$r \leq r_{T}$} \label{lst:line:annealing-while}
	\State compute all primary capsules poses $\vu_i$ and votes $\hat{\vu}_{i,j}$
	\State compute all class capsules poses: $\vu_{j} \gets$ \Call{Routing($\hat{\vu}_{i,j}$, $r$)}{}
	\State evaluate loss $\mathcal{L}$ on the validation set
	\If{$(\mathcal{L} \geq \mathcal{L}^\star_k)  ~~ $\textbf{and}$ ~~ (e - e^\star_k \geq p)$} \label{lst:line:annealing-if}
	    \State $k \gets k + 1$; $r \gets r + s$; $r_k \gets r$; $\mathcal{L}^\star_k \gets +\infty$; $e^\star_{k} \gets 0$ \label{lst:line:annealing-step}
        \State load DR-CapsNet network of epoch $e^\star_{k-1}$ \label{lst:line:annealing-restart}
    \ElsIf{$\mathcal{L} < \mathcal{L}^\star_k$} \label{lst:line:annealing-isbest}
        \State $\mathcal{L}^\star_k \gets \mathcal{L}$; $e^\star_k \gets e$ \label{lst:line:annealing-updatebest}
    \EndIf
	\State $e \gets e+1$
	\State backpropagate error gradients and update parameters
\EndWhile
\State $k^\star \gets \argmin\limits_{k} \mathcal{L}^\star$; $r^\star \gets r_{k^\star}$ 
\State \Return DR-CapsNet network of epoch $e^\star_{k^\star}$ and $r^\star$
\EndProcedure
\end{algorithmic}
\end{algorithm}
\noindent
To summarize, RA increases the value of $r$ 
when the validation loss does not decrease for $p$ epochs in a row. As an upper bound for the number of routing iterations, we stop the training when $r$ reaches its maximum value $r_{T}$. When $r$ increases, the training restarts with the weights of the network with the best validation loss obtained with its previous value.
By comparison, using the standard training procedure mentioned in Sec.~\ref{subsec:fixed-routing}, the weights need to be reinitialize for every simulation with a a different value for $r$.
\subsection{Rationale}
RA takes inspiration from the \textit{simulated annealing} (SA) algorithm, a probabilistic technique used in combinatorial-optimization problems to minimize a cost function.
\begin{figure}[h]
    \centering
    \includegraphics[width=0.9\linewidth]{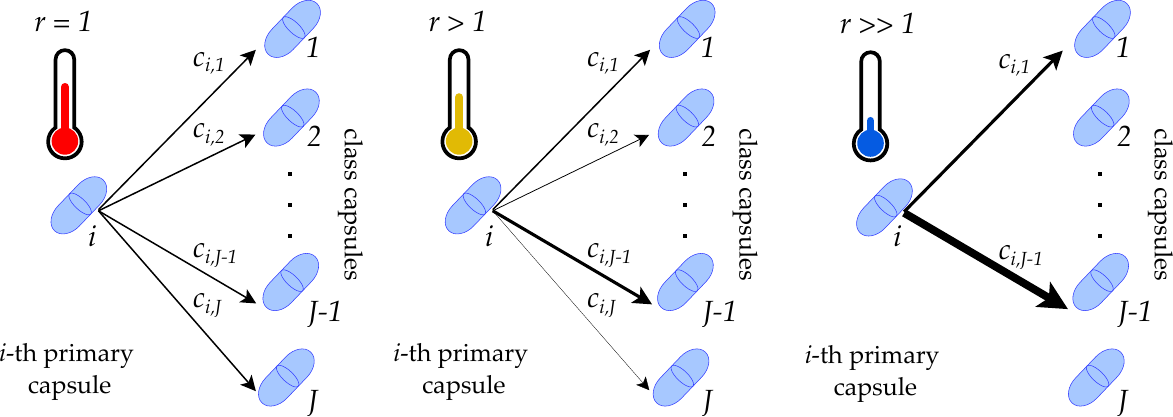}
    \caption{Routing coupling coefficients between the $i$-th primary capsule and $J$ class capsules. The higher the line weight, the higher the corresponding coupling coefficient. When $r=1$ (left), the coupling coefficients have all the same value, while increasing $r$ (center), some of the coupling parameters survive, while the others are almost zero. When $r$ is high (right), $J^\star=2$ and $\mathcal{J}_i = \{1, J-1\}$.}
    \label{fig:annealing}
\end{figure}

In our approach, we relate the temperature of our system being inversely proportional to the number of routing iterations $r$: the highest $r$, the highest the agreement between the capsules and the lowest the noise.\\
The number of routing iterations relates to the distribution of the coupling coefficients $c_{i,j}$. According to Alg.~\ref{alg:dynamicrouting}, when $r$ is low, the agreement is low as well. When $r=1$, all the coupling coefficients will have the same value, $\frac{1}{J}$. Increasing the routing iterations, a certain number of coupling coefficients becomes dominant over others, since Alg.~\ref{alg:dynamicrouting} looks for capsule's agreement. Considering that $c_{i,j}$ are normalized values, we can say that, for the $i$-th capsule $\sum_{j \in \mathcal{J}_i}c_{i,j} \rightarrow 1\nonumber$ and $\sum_{j \in \overline{\mathcal{J}_i}}c_{i,j} \rightarrow 0$, where $\mathcal{J}_i$ is a subset of the $J$ coupling coefficients for the $i$-th primary capsule and $\overline{\mathcal{J}_i}$ is its complementary set. When $r = 1$, the cardinality of $\mathcal{J}_i$ is exactly $J$, but increasing $r$, its cardinality drops to some optimal value $J_i^\star$: this means that the $i$-th primary capsule will be coupled to $J_i^\star$ digit capsules only, avoiding noisy coupling to the others (which are $J-J_i^\star$). A visual representation of this effect is displayed in Figure~\ref{fig:annealing}. As  $r$ increases, many coupling coefficients drop to zero, while others converge to higher coupling values. In this way, the routing algorithm learns how to build relationships between primary and digit caps, discarding noisy information, which helps in improving the generalization of the model. In the next section we are going to test on-the-field our RA strategy, observing in particular the generalization capability of the RA models compared to the other state-of-the-art approaches.

\section{Experiments}
\label{sec:ra_experiments}

In this section we compare our proposed Routing Annealing (RA) method in Alg. \ref{alg:routingannealing} against the reference method in Alg. \ref{alg:fixed-routing}.
First we show that, with RA, DR-CapsNet performs better as the number of routing iterations $r$ improves, whereas this is not the case with the reference algorithm.
Then, we further validate RA on multiple datasets and settings showing that it delivers best gains especially where the number of parameters the network can afford is low, i.e. where DR-CapsNets performance is weaker.

\section{Architecture}

\begin{figure}[h]
	\centering
	\includegraphics[width=0.7\textwidth,]{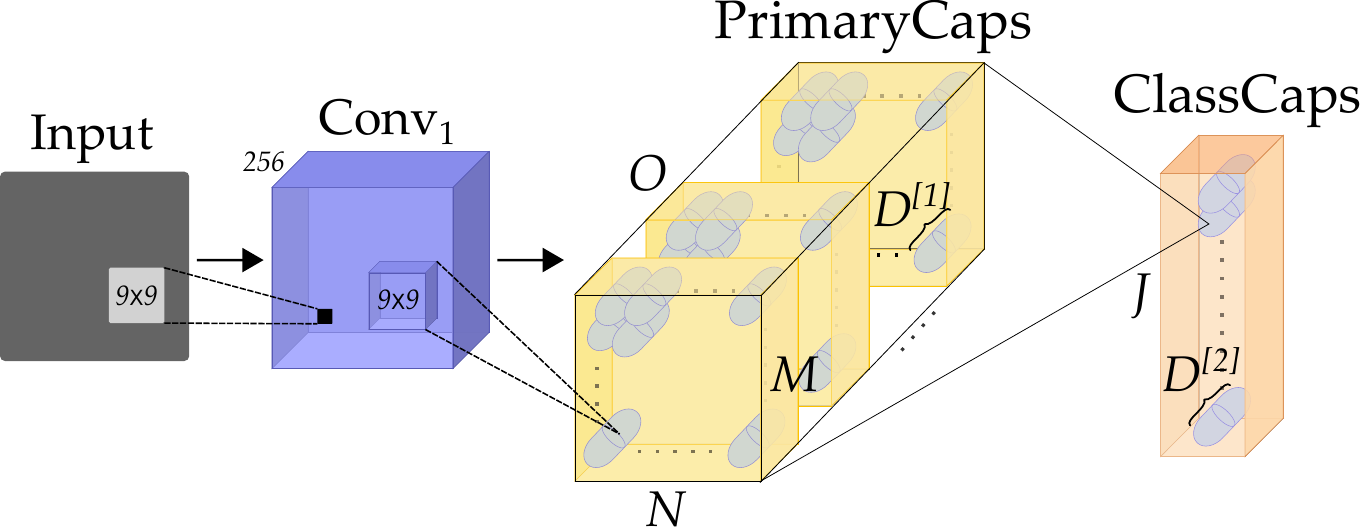}
	\caption[DR-CapsNet.]
	{DR-CapsNet architecture. There are one convolution layer (Conv1) and two capsule layers (PrimaryCaps and ClassCaps). The dynamic routing algorithm controls the information flow between capsule layers.}
	\label{fig:vectorcapsnet}
\end{figure}

Figure~\ref{fig:vectorcapsnet} shows the DR-CapsNet architecture used in the experiments in this chapter, consisting of one convolutional layer and two capsule layers. 

The first layer (Conv\textsubscript{1}) is a $9 \times 9$ convolutional layer that converts pixel intensities to the activities that are given in input to the first capsule layer. The PrimaryCaps layer is implemented as a convolutional layer with $9 \times 9$ filters and $O \times D^{[1]}$ channels where $O$ is the number of primary capsules types and $D^{[1]}$ is the dimension of a capsule vector. Overall, there are $O \times M \times N$ primary capsules. Capsule vectors are normalized with the squashing function defined in Equation~\ref{eq:squashing}. The output layer (ClassCaps) comprises $J$ $D^{[2]}$-dimensional class capsules $\vu_j$, one for each output class. 
The information flow between primary and class capsules is governed by the dynamic routing described in Algorithm~\ref{alg:dynamicrouting}.

\subsection{Experimental Setup}

We experiment with the DR-CapsNet in Figure~\ref{fig:vectorcapsnet} at classifying natural images in a fully supervised scenario.
We consider the MNIST, Fashion-MNIST and CIFAR-10 datasets.
For all datasets, 5\% of the training set samples are reserved for validation.
MNIST and Fashion-MNIST are composed of 28x28 images; concerning CIFAR-10, we randomly crop the original 32x32 images into 24x24 patches for training whereas crops from the image center are used for testing as done in \cite{hinton-dr}.
Our experiments consider several flavors of the architecture in Figure~\ref{fig:vectorcapsnet} with different types $O \in \{1, 2, 4, 8, 16, 32\}$ and dimensions  $(D^{[1]}, D^{[2]}) \in \{(2,4), (4,8), (8,16)\}$ of capsules.
We train the network minimizing a \textit{margin loss} \cite{hinton-dr} with the Adam optimizer \cite{adam} with a constant learning  rate equal to 0.001 and a batch size of 128.
No  weight decay, dropout or other regularization techniques were used.

Concerning the proposed RA method,  we train the network with the procedure in Alg. \ref{alg:routingannealing}, with the following configuration: $r_{0}=1$, $r_{T}=50$, $s=1$ and $p=10$.
As we discussed in Sec. \ref{subsec:annealing}, RA can be applied to any iterative routing algorithm but this work use as base routing algorithm the one described in Alg. \ref{alg:dynamicrouting}.

About the reference method, we use FR which employs the procedure in Alg. \ref{alg:fixed-routing},  i.e. where the number of routing iterations $r$  is fixed (common values in literature are $r=1$ \cite{gu} or $r=3$ \cite{hinton-dr}).

\subsection{Preliminary analysis on MNIST}

Preliminary, we assess the effect of the number of routing iterations $r$ on MNIST for a minimal capsule network where the PrimaryCaps layer has only $O=1$ capsule types and vectors have dimension $D^{[1]}=2$ while the ClassCaps layer vectors have $D^{[2]}=4$ elements.
This network has only 65k parameters, which  helps  isolating  the effect of $r$, whereas the architecture  in \cite{hinton-dr} has 6M parameters (8.2M with the decoder).

\begin{figure}[h]
	\centering
	\includegraphics[width=1\textwidth,]{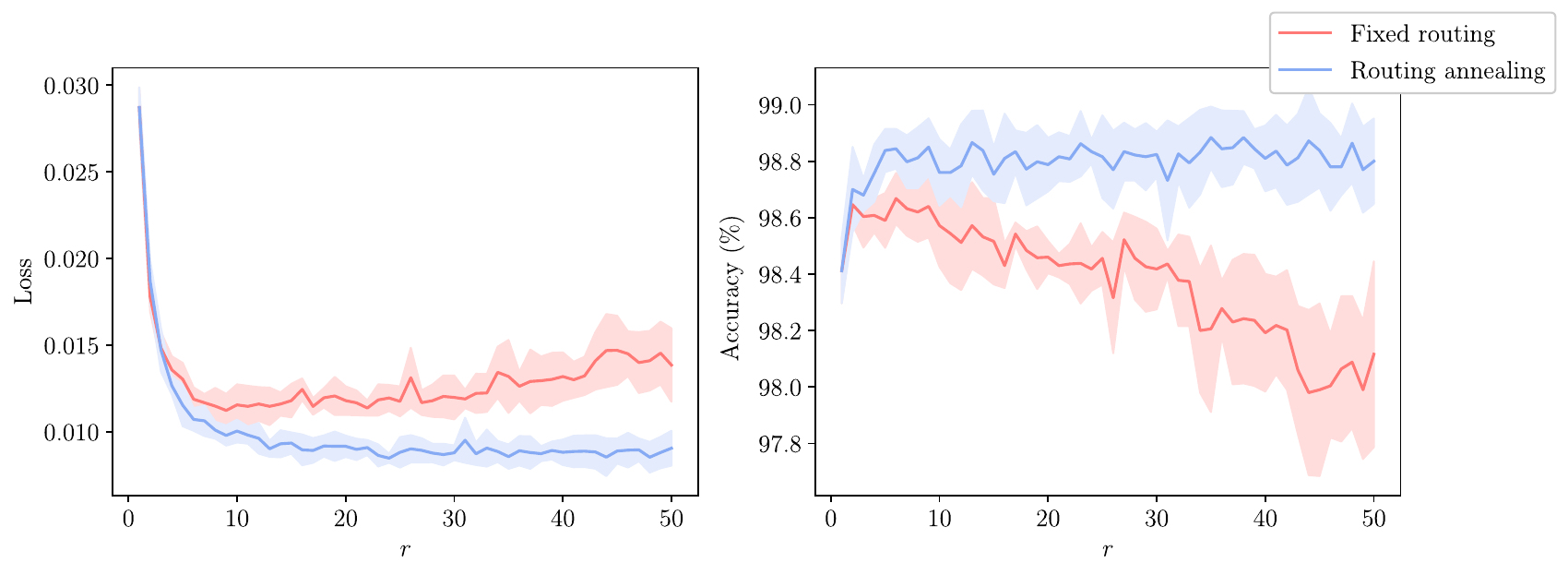}
	\caption[Fixed vs annealing.]
	{Loss function (left) and classification accuracy (right) on MNIST test set for a DR-CapsNet with $O=1$, $D^{[1]}=2$, $D^{[2]}=4$ (means and stds of 5 seeds) as a function of the number of routing iterations $r$.}
	\label{fig:fix_vs_ann}
\end{figure}

In Figure~\ref{fig:fix_vs_ann} we report the learning curves for FR and RA. 
For FR, we train a new DR-CapsNet from scratch for each and every value of $r$. In the case of RA, instead, we train one model only, where we gradually increase the number of routing iterations (when the network loss reaches a plateau). 
We plot the best loss and accuracy values for every $r$. 
Figure~\ref{fig:fix_vs_ann} shows that
as $r$ increases, the  proposed RA enables decreasing loss that reflects into higher classification accuracy.
Conversely, with a fixed routing strategy, the loss function diverges as $r$ increases. 
We explain the gap between the two loss curves with the following hypothesis. Each iteration of the routing algorithm strengthens or weakens the connections between a capsule of the primary layer and all the capsules of the digit layer. Therefore, imposing high $r$ for all the training epochs leads the DR-CapsNet to be overconfident on its predictions on the link strengths, preventing the network form learning the correct connections between the capsules.

\subsection{Results}
\label{sec:results}
Next, we experiment with the more complex datasets Fashion-MNIST and CIFAR-10.
Figure~\ref{fig:fix_vs_ann} showed that RA performs better than the fixed routing reference for large $r$ values.
\begin{figure}
	\centering
	\begin{subfigure}{\textwidth}
	    \centering
	    \includegraphics[width=\textwidth]{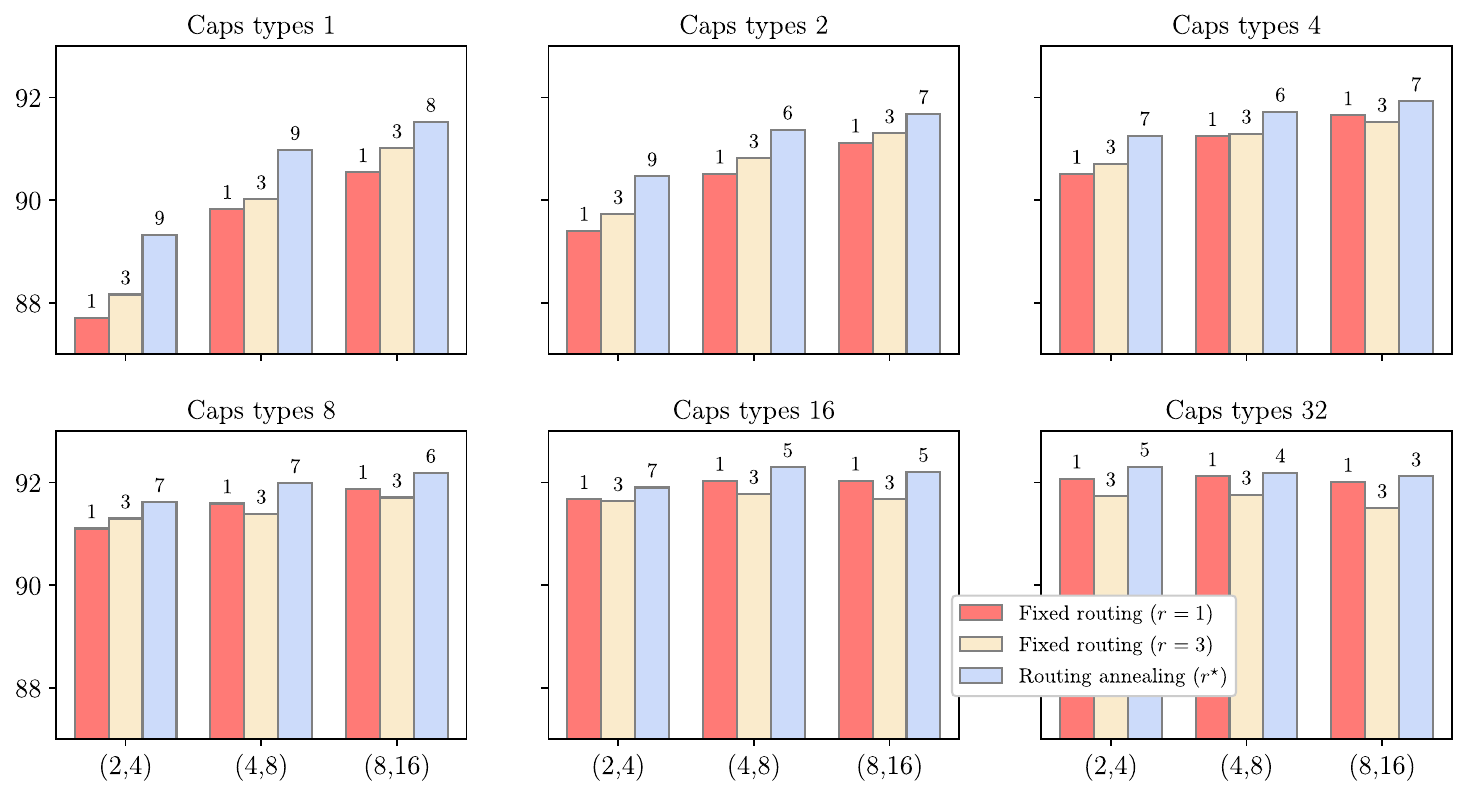}
	    \caption{Fashion-MNIST}
	    \label{fig:fashion}
	\end{subfigure}
	\begin{subfigure}{\textwidth}
	    \centering
	    \includegraphics[width=\textwidth]{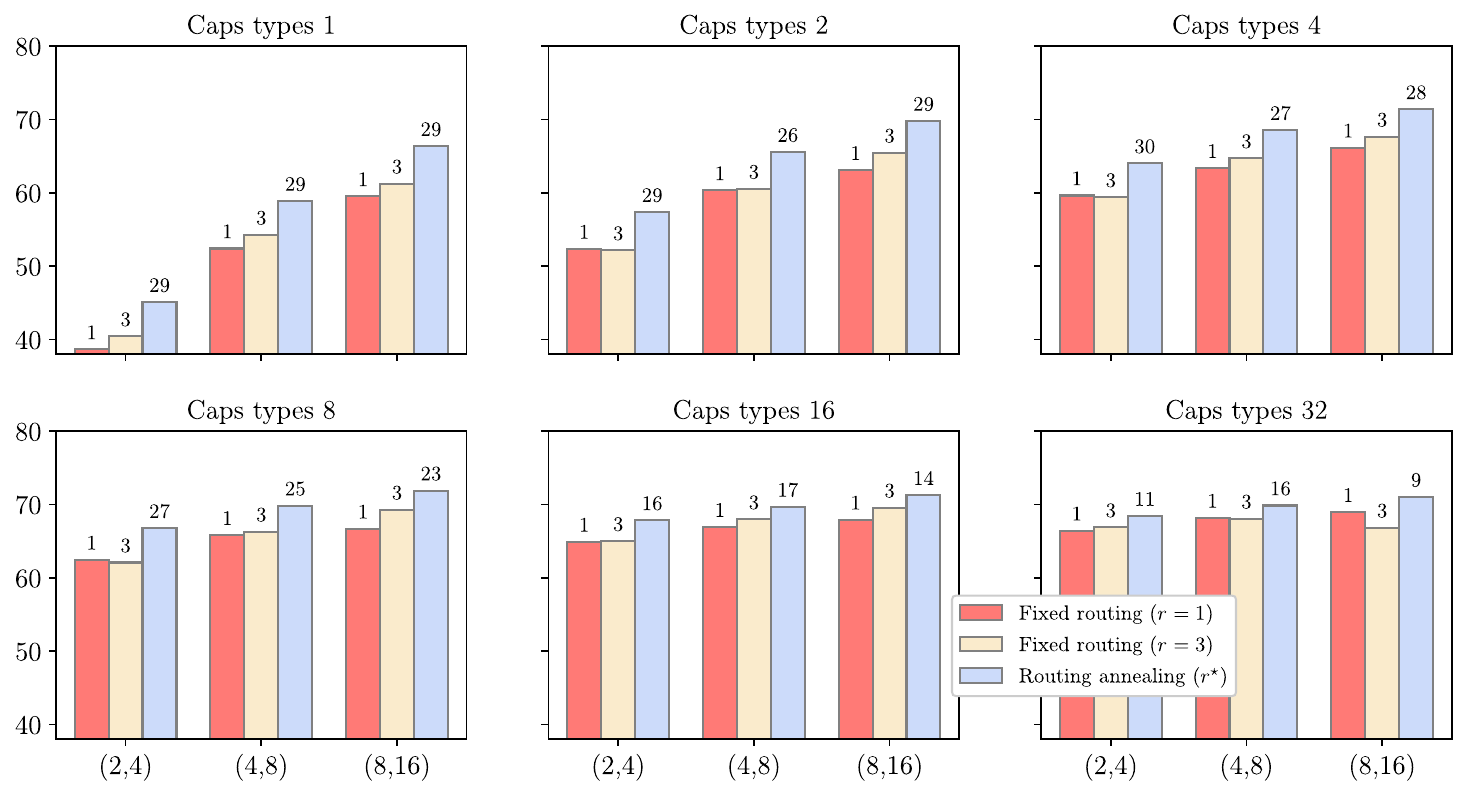}
	    \caption{CIFAR1-0}
	    \label{fig:cifar10}
	\end{subfigure}
	\caption{Classification accuracy (\%) on Fashion-MNIST (a) and CIFAR-10 (b) test set for different capsule types $T$ and dimensions $(D^{[1]}, D^{[2]})$. On top of each bar it is shown the number of iterations $r$ used during training/inference, for RA it is shown the median value of $r^\star$.}
\end{figure}
\noindent 
For fixed routing experiments we only consider $r=1$ and $r=3$, as done in much of the recent literature.
Figure \ref{fig:fashion} and \ref{fig:cifar10} show that RA performs better than fixed routing (both $r=1$ and $r=3$) in all the settings.
Such experiments brought us to the following observations. First, coherently with our previous findings on MNIST, RA delivers the most appreciable gains when the network can afford only few learnable parameters.
We recall here that for each capsule we have a matrix of weights $\mW_{ij}$ and these matrices have shapes $D^{[1]} \times D^{[2]}$, namely the dimensions of the capsule vectors. This means that the number of capsules types $T$ and their dimensions, along with the convolutional layers, drive the number of parameters of the network.
This behaviour can be explained observing that finding agreements between many high-dimensional capsule is not trivial.
Running more iterations of the routing algorithm tends to polarize the coupling coefficients, namely the link strengths between capsules, such that it results in a simple route where each primary capsule sends its output to only one digit capsule.
Therefore, when there are a lot of capsules, introducing some level of uncertainty with a low value of $r$ helps the network to not be overconfident on its predictions and to not overfit on the training data.
As a matter of fact, Figure~\ref{fig:fashion} shows that with 32 capsule types of 8-dimensional primary capsules and 16-dimensional digit capsules, our proposed method RA finds $r^\star=3$, namely the value used in the original formulation of CapsNets in \cite{hinton-dr}.
Second, in high-dimensional settings the same conclusions about routing as in \cite{gu} and in \cite{paik} hold for the fixed routing procedure, which achieves higher accuracy with $r=1$ than $r=3$.
Nevertheless, RA always achieves better performance in all cases, sometime even with fewer routing iterations. Third, Figure~\ref{fig:cifar10} shows that $r^\star$ for CIFAR-10 is not the same as for Fashion-MNIST in Figure~\ref{fig:fashion} for identical network conditions.
This means that despite $r^\star$ differs from dataset to dataset, nevertheless our method can find it. 

\section{Summary}
\label{sec:ra_summary}
In this chapter we presented a novel training technique for iterative routing-based CapsNets where the number of iterations is iteratively found at training time rather than being fixed.
We also showed experiments on DR-CapsNets in settings with a different number of capsule types and their dimensions, namely the network capacity in terms of trainable parameters, and on several datasets.
We showed that this value depends heavily on the size of the network and the dataset used. Typically, the smaller the network, the higher the number of iterations it requires to improve its generalization capability.
Given the potentiality of our technique, in future works, we plan to apply RA on more complex and sophisticated routing algorithms such as EM routing \cite{hinton-em}.
\cleardoublepage
%*****************************************
\chapter{Towards Efficient Capsule Networks}\label{ch:icip_effcapsnets}
%*****************************************
The research findings and insights presented in this chapter were originally published in the following work: \fullcite{renzulli-effcaps}.
\section{Introduction}
From the moment NNs dominated the scene for image processing, the computational complexity needed to solve the targeted tasks skyrocketed: against such an unsustainable trend, many strategies have been developed, ambitiously targeting performance's preservation. Promoting sparse topologies, for example, allows the deployment of deep neural networks models on embedded, resource-constrained devices. 
\begin{figure}[h]
\centering
\includegraphics[width=0.75\columnwidth]{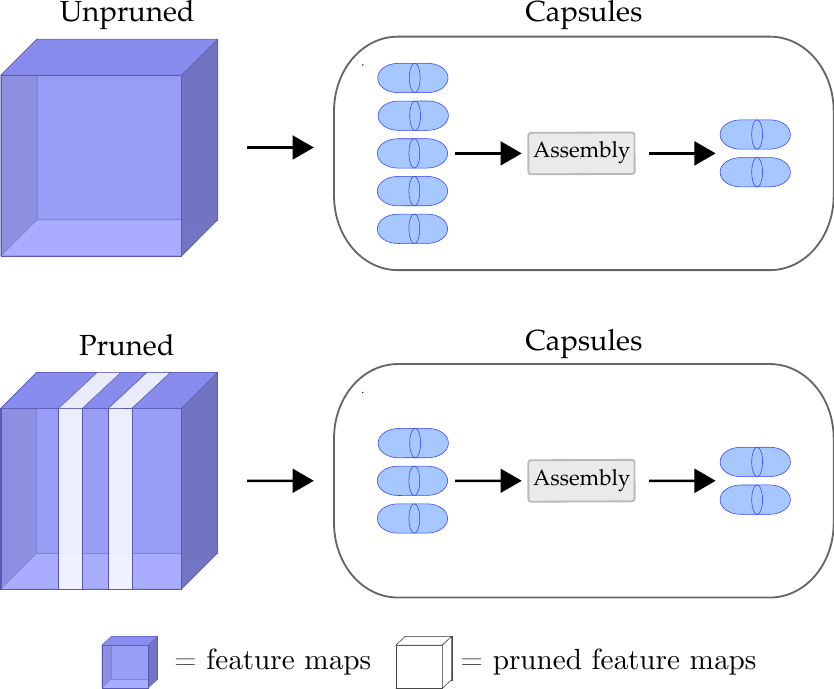}
%\def\svgwidth{480pt}
%\input{figures/pipeline.eps_tex}
%\normalsize			
\caption{A standard CapsNet with many capsules (top). We exploit structured pruned features (bottom) to reduce the number of capsules.\label{fig:teaser}}
\end{figure}
As mentioned in previous chapters, CapsNets were proposed to overcome the shortcomings of CNNs. Despite CapsNets being successfully applied to various tasks~\cite{zhao20193d,afshar2018brain}, the research is still in its early stages and CapsNets are far from replacing CNNs as state-of-the-art neural architectures due to their low scalability. Their applicability is limited because stacking multiple capsule layers increases the computational effort~\cite{light-weight-caps, Q-CapsNets}. Applying CapsNets on high-resolution data with complex backgrounds is also critical: stacking many capsule layers or adding more capsules can lead to instability during training because of the delicate routing algorithm~\cite{deepcaps}. 
However, the problem of over-parametrization and the difficulty of deploying networks in scenarios with limited memory and resources is not peculiar only to CapsNets, but also to CNNs. To this end, pruning methods (which reduce the size and complexity of a neural network model) have gained more and more attention lately~\cite{tartaglione2021serene}. Learning sparse topologies can effectively reduce the network footprint and speed up execution with negligible performance loss~\cite{bragagnolo2021simplify}. So, why not exploit these techniques also for CapsNets? Reducing their complexity, especially in early layers, will enable their deployment on complex datasets.
In this chapter, we exploit structured pruning approaches to reduce the complexity of CapsNets without performance loss and test them on data with non-straightforward background such as CIFAR-10 and high dimensional data such as Tiny ImageNet. Figure~\ref{fig:teaser} provides an overall idea of the functioning of our approach.
 We show how employing structurally sparse backbones, extracting features in early layers for CapsNets, affects the number of capsules in the overall network. Our experiments show that reducing the complexity of the backbone is an effective way to achieve high generalization with less memory requirements, energy consumption, training and inference time.

\section{Methodology}
In this section, we show first how to extract primary capsules with a backbone model, then we propose a method to reduce the complexity of a capsule network with pruned backbones.
\label{sec:method}
\subsection{Primary capsules extraction}\label{sec:primary_caps_extraction}
Following the work of~\cite{self-routing, paik}, we extract $I$ primary capsule activity vectors by means of a backbone network. With this approach, we can build deep capsule networks in an efficient way since the input space of the capsule network is significantly reduced. 
\begin{figure}[h]
\centering
\includegraphics[width=1\columnwidth]{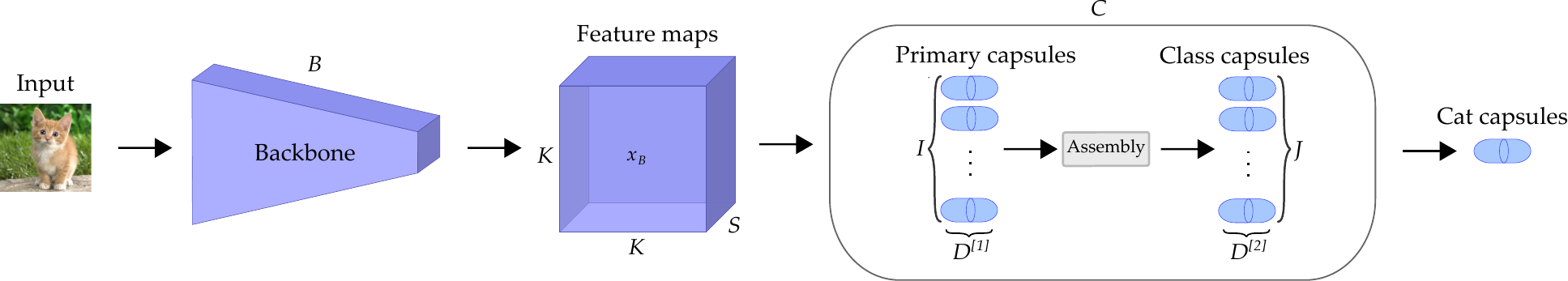}
%\def\svgwidth{480pt}
%\input{figures/pipeline.eps_tex}
%\normalsize			
\caption{The architecture used in this chapter is composed by a backbone part and a capsule part. Primary capsules space has 8 real dimensions  while class capsules are 16-dimensional vectors ($D^{[1]} = 8$ and $D^{[2]} = 16$).\label{fig:capsnet}}
\end{figure}
Figure~\ref{fig:capsnet} shows the architecture used in this chapter. We can build a CapsNet on top of the backbone by replacing the last two layers by a primary capsule (PrimaryCaps) and fully-connected capsule (ClassCaps) layers, respectively. Note that PrimaryCaps is a convolutional capsule layer with $I$ channels of convolutional
$D^{[1]}$ capsules, namely each capsule contains $D^{[1]}$ convolutional units with a $K \times K \times S$ kernel. 
The kernel size used is wide as the spatial dimensions of the output of the backbone network, which means the number of capsules is dynamically reduced.
In our experiments we use backbones both as fixed feature extractor, namely we freeze their weights except for the capsule layer, and as trainable feature extractor, where we finetune all layers.
\subsection{Effect of pruned backbones to capsule layers}
%Since the aim of our work is to...\enzo{parte mia da aggiungere domattinba}
In order to reduce the computational complexity of CapsNets as in Figure~\ref{fig:capsnet}, one possible strategy is to deploy pruned backbones. Let us define the backbone network $\mathcal{B}(\boldsymbol{x}_\mathcal{M}, \boldsymbol{w}_\mathcal{B})$ where $\boldsymbol{w}_\mathcal{B}$ are the backbone parameters and $\boldsymbol{x}_\mathcal{M}$ is the input; it produces as output the tensor $\boldsymbol{x}_\mathcal{B} \in \mathbb{R}^{K \times K \times S}$. Then, tensor $\boldsymbol{x}_\mathcal{B}$ is fed into the CapsNet $\mathcal{C}(\boldsymbol{x}_\mathcal{B}, \boldsymbol{w}_\mathcal{C})$, where $\boldsymbol{w}_\mathcal{C}$ are the capsule parameters. 
% \begin{equation}
%     |\Omega_{\mathcal{I}}| = \frac{\|\boldsymbol{x}_\mathcal{I}\|_0}{\|\boldsymbol{u}_i\|_0}
% \end{equation}
We can concatenate the two parts expressing the end-to-end model $\mathcal{N}$ as 
\begin{equation}
 \mathcal{N}(\boldsymbol{x}_\mathcal{M}, \boldsymbol{w}_\mathcal{B}, \boldsymbol{w}_\mathcal{C}) = \mathcal{C}\left[\mathcal{B}(\boldsymbol{x}_\mathcal{M}, \boldsymbol{w}_\mathcal{B}), \boldsymbol{w}_\mathcal{C})\right].   
\end{equation}
As we saw in Sec.~\ref{sec:primary_caps_extraction}, $D^{[1]}$-dimensional primary capsules poses $\boldsymbol{u}_i$ are built upon convolutional layers.
We define the number of primary capsules as 
\begin{equation}\label{eq:num_primary}
    I = \left\lfloor\frac{S}{D^{[1]}}\right\rfloor.
\end{equation}
From this, we observe that reducing the complexity of the backbone would result in the overall reduction of the complexity for the entire model $\mathcal{N}$, and towards this end pruning has already proved to be an effective approach~\cite{tartaglione2021serene, bragagnolo2021simplify}.\\
Pruning approaches can be divided into two groups. \textit{Unstructured} pruning methods aim at minimizing the cardinality $\|\boldsymbol{w}\|_0$ of the parameters in the model, regardless the output topology~\cite{han2015learning, tartaglione2018sensitivity, lobster}. 
%In fact, the ratio of parameters pruned with these method is usually high, but the output tensor is randomly sparse so it does not necessarily translate into smaller memory footprints. 
On the other hand, \textit{structured} approaches drop groups of weight connections entirely, such as entire channels or filters, imposing a regular pruned topology~\cite{tartaglione2021serene, li2020eagleeye}. As an effect, they minimize the cardinality of some $i$-th intermediate layer's output $\|\boldsymbol{x}_i\|_0$. Bragagnolo~\emph{et~al.}~\cite{bragaICIP} showed that 
structured sparsity, despite removing significantly less parameters from the model, yields lower model's memory footprint and inference time. When pruning a network in a structured way, a simplification step which practically reduces the rank of the matrices is possible; on the other side, encoding unstructured sparse matrices lead to representation overheads~\cite{bragagnolo2021simplify}.\\
Considering the structure of the primary capsules layer, we are interested in the reduction of the cardinality for the input of the primary capsules $\boldsymbol{x}_\mathcal{B}$. Indeed, as we can see in \eqref{eq:num_primary}, the minimization of $\|\boldsymbol{x}_\mathcal{B}\|_0$ results in a proportional reduction of $S$, which in turn reduces $I$, resulting in a reduction of the votes and the complexity of the routing algorithm.

\section{Experiments}\label{sec:experiments}
We performed several experiments with different backbones on CIFAR-10 and Tiny ImageNet. We choose ResNets and MobileNets models since they are used as common baselines in traditional computer vision tasks. In fact, they offer a good compromise between both efficiency and performance. As CapsNet architecture, we employed DR-CapsNet introduced in \cite{hinton-dr}. Furthermore, they are mainly composed of convolutional layers so it is easy to replace the first convolutional layer in a DR-CapsNet with these models.

\subsection{Experimental Setup}
We resized CIFAR-10 images to $64 \times 64$ resolution so as to add a deeper backbone network as in \cite{deepcaps}. We used the 5\% of the training set as validation set for hyper-parameters tuning. We also run experiments on Tiny Imagenet ($224 \times 224$ images, 200 classes). Here we used 10\% of the training set as validation set and the original validation set as test set.
Our experiments consider several flavors of the architecture in Figure~\ref{fig:capsnet} with different number of primary capsules $I$, backbones (ResNet-50 e MobileNetV1) and training methods (whether the backbone is finetuned or not). 
%As regards pruning backbones, we use pretrained models pruned with EagleEye \cite{li2020eagleeye}, one of the state-of-the-art structured pruning strategies.
We train all the networks minimizing a \textit{margin loss}~\cite{hinton-dr} with the Adam optimizer and a batch size of 128.  The experiments were run on a Nvidia Ampere A40 equipped with 48GB RAM, and the code uses PyTorch~1.10.
We also used PyNVML library to compute the GPU memory
consumption.
As pretrained pruned backbones, we have used both ResNet-50 and MobileNetV1 pruned with EagleEye~\cite{li2020eagleeye}. EagleEye is a state-of-the-art, open-source, structured pruning strategy which identifies relevant features at the level of intermediate outputs $\boldsymbol{x}_i$.

\subsection{Results and discussion}
\begin{figure}
\centering
\includegraphics[width=0.95\columnwidth]{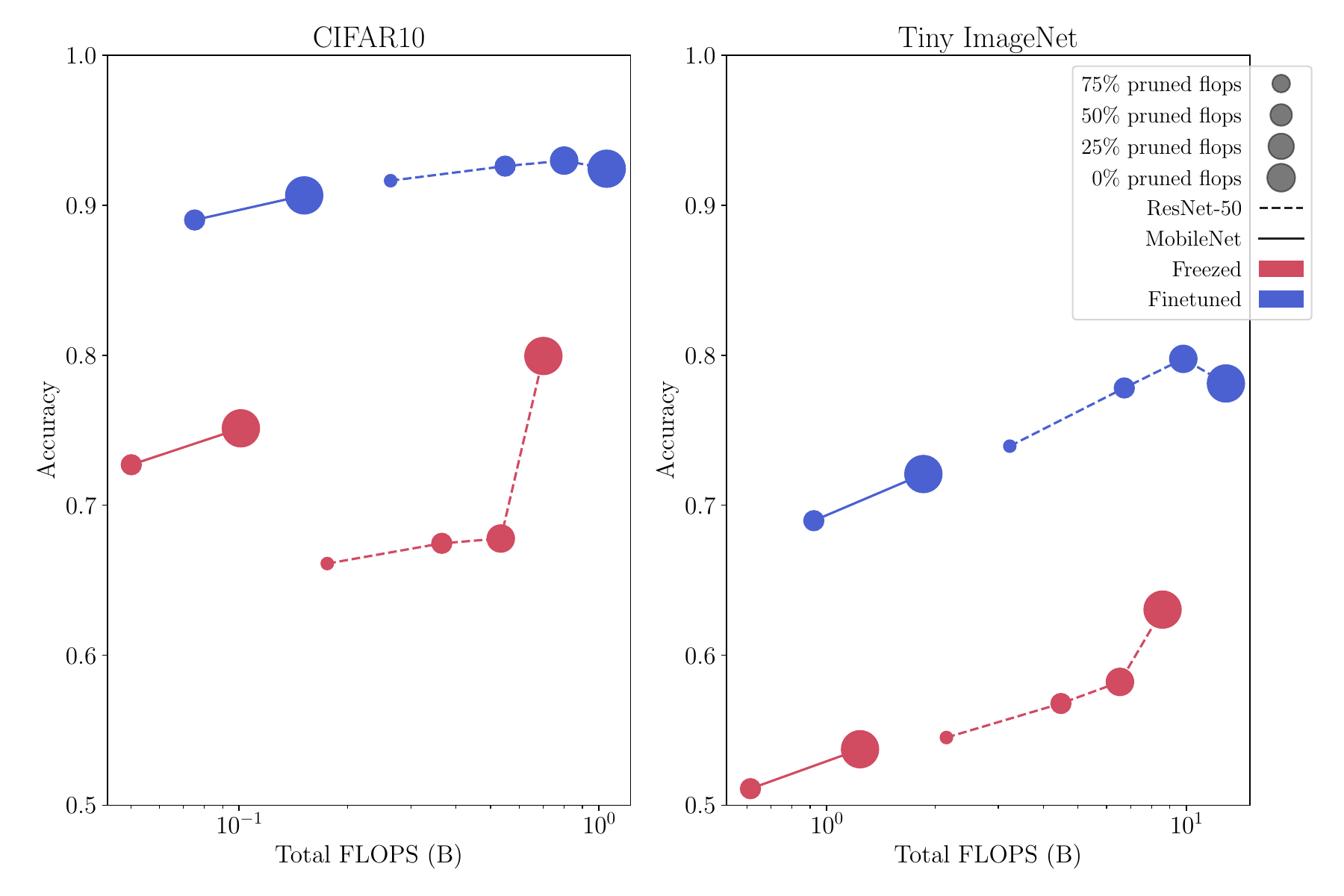}
%\def\svgwidth{480pt}
%\input{figures/pipeline.eps_tex}
%\normalsize			
\caption{Accuracies using ResNet-50 and MobileNetV1 as backbones for CIFAR-10 (left) and Tiny ImageNet (right) with several pruned FLOPS configurations.\label{fig:flops_vs_acc}}
\end{figure}
The main results are shown in Figure~\ref{fig:flops_vs_acc}. Each configuration differs from the dataset used, percentage of pruned parameters in the backbone network (which affects the number of primary capsules), and training method (backbone finetuned or not). In Figure~\ref{fig:flops_vs_acc} we report the accuracy as a function of the total number of FLOPS (of both backbone and capsule part) for training the network.
%  Note that the configuration with 75\% of pruned parameters and where the backbone is used as fixed feature extractor is the one with the lowest number of FLOPS. As opposite, the one with 0\% of pruned parameters and where the network is finetuned entirely is the one with the highest number of FLOPS. 
 We can see that when the backbone is freezed (in red) the networks achieve poor performances, but as the number of pruned backbone FLOPS decreases, namely the number of primary capsules is higher, the accuracy increases. The poor performance can be explained by the fact that the output features $\boldsymbol{x}_\mathcal{B}$ of the backbone are not directly optimized to represent objects poses. Therefore, they can not be used directly as input to capsule layers. To overcome such a drawback, we can either add more primary capsules or train the whole network. In fact, when the backbone is finetuned (in blue) the performance is higher since the backbone is able to map more suitable features to the capsules space. We can see in this case that adding more capsules does not always lead to an improvement in the classification accuracy. In Tab.~\ref{tab:performance-dynamiccaps} the performance for Tiny~ImageNet with a pretrained ResNet-50 backbone is reported. Employing a backbone with 50\% of pruned FLOPS yields similar accuracy as with a full backbone, with less GPU memory consumption and training time. With 25\% of pruned FLOPS we can even improve the accuracy compared to the unpruned one. The partial improving in the performance in a low pruning regime, despite being at a first glance surprising, is not new to the literature and it is twofold. First, Rajasegaran~\emph{et~al.}~\cite{deepcaps} shows that building too many capsules inhibits learning because the coupling coefficients are too small, preventing the gradient flow. %and inhibiting learning.
Furthermore, Han~\emph{et~al.} observed such a phenomenon for unstructured pruning~\cite{han2015learning}. Recently, it has been showed that the average entropy in the bottleneck layer for pruned backbones (in our case, $\boldsymbol{x}_\mathcal{B}$) is higher than in non-pruned ones~\cite{bragagnolo2020icann}: this results in the propagation of less specific and more general information, which prevents features overfit, on top of which capsules layers can extract much more accurate information.
\begin{table}[h]
\centering
\resizebox{\textwidth}{!}{
\begin{tabular}{ccccccc}
\hline
\textbf{\makecell{Backbone\\pruned FLOPS (\%)}} &
\textbf{\makecell{Bottleneck\\ size}} &
\textbf{\makecell{Primary\\ caps}} &
\textbf{\makecell{Total\\ FLOPS (B)}} &
\textbf{\makecell{GPU memory \\consumption (GB)}} & \textbf{\makecell{Training time\\ (epoch, s)}} & \multicolumn{1}{c}{\textbf{Accuracy}} \\ \hline
0 & 2048 & 256 & 12.9 & 47.02          & 462          & 0.78          \\
\textbf{25} & 1970 & 246 &  \textbf{9.8} & \textbf{43.50} & \textbf{450} & \textbf{0.80} \\
\textbf{50} & 1947 & 243 & \textbf{6.7} & \textbf{39.42} & \textbf{410} & \textbf{0.78} \\
75 & 944 & 118 & 3.2 & 20.11          & 214          & 0.74          \\ \hline
\end{tabular}}
\caption{\label{tab:performance-dynamiccaps}Performances using ResNet-50 as backbone (finetuned, TinyImageNet).}
\end{table}

\section{Summary}
\label{sec:majhead}
In this chapter, we presented a method to improve the scalability and reduce the computational effort of CapsNets on complex datasets deploying backbones with structured sparsity. 
CapsNets with many capsules are difficult to train: in such a scenario, the routing algorithm, a key mechanism for CapsNets, struggles to find the necessary agreements between capsules. Therefore, employing pruned backbones (so fewer capsules) for high-resolution datasets leads to competitive results with no performance loss.
% Interestingly, we have observed that with a backbone with half of the FLOPS, we can achieve competitive results with no performance loss. 
% We showed that in low pruning regimes, the performances can be even improved.
We also showed how extracting features in early layers with sparse networks improves the efficiency, memory consumption, inference, and training time of the overall CapsNet. Therefore, this chapter opens the way to apply CapsNets on resource-constrained devices.
Future works involve specific methods concerning pruning capsule layers in a structured way.
\cleardoublepage
%*****************************************
\chapter{Capsule Networks Do Not Need to Model Everything}\label{ch:rem}
%*****************************************
The research findings and insights presented in this chapter are currently under review: \fullcite{renzulli-rem}.
\section{Introduction}\label{sec:rem-introduction}

As mentioned in the previous chapters, CapsNets can learn hierarchical representations from the input images thanks to a routing algorithm, which carves out a parse tree composed of part-object relationships. They are also more robust to viewpoint changes and affine transformations than traditional CNNs. 
\begin{figure}[h]
\centering
\includegraphics[width=0.8\columnwidth]{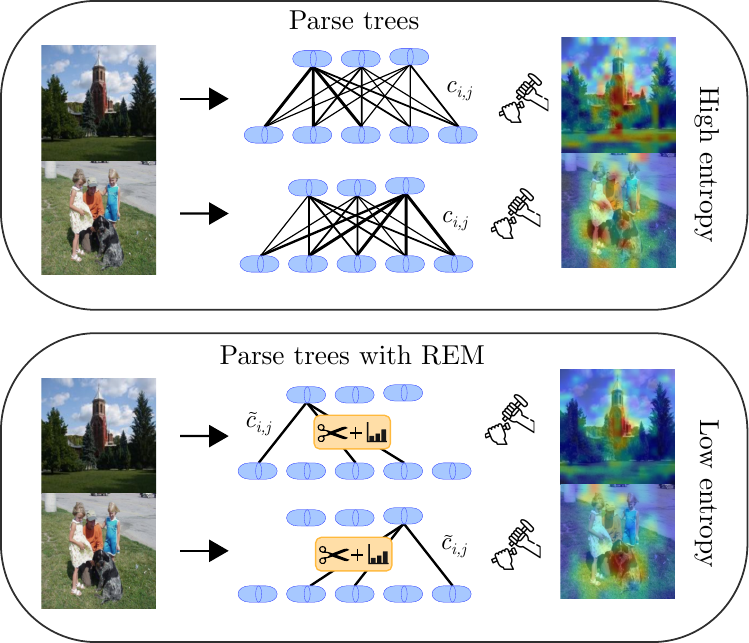}	
\caption{
A hierarchy of parts (parse tree) is \textit{carved out} of a CapsNet, like a sculpture is carved from a rock. 
% Capsule activations and coupling coefficients can be exploited to output saliency maps. 
REM enriches a CapsNet with pruning and quantization. REM, with low entropy configurations, reduces modeling noisy backgrounds and extracts more discriminative features.
\label{fig:rem-teaser}}
\end{figure}

This chapter explores a drawback of CapsNets identified by \citet{hinton-dr}: these networks perform better when they can model all elements present in an image; therefore, larger network sizes are required to handle image clutter, such as intricate backgrounds or complex objects. Therefore, the entropy of the connections of the parse trees is high. Recently, many pruning methods were applied to CNNs to reduce the complexity of the networks, enforcing sparse topologies~\citep{tartaglione2018sensitivity, sparsevd, louizos2017learning}. These considerations lead us to the following question: \textit{is it possible to tailor pruning and quantization approaches to not only reduce the size of CapsNets but aid the extraction of more discriminative features (namely fewer parse trees)?}

Here we introduce REM (Routing Entropy Minimization), which moves some steps toward extracting more succinct relationships from CapsNets with negligible performance loss and fewer parameters. To encourage this, as shown in Figure~\ref{fig:rem-teaser}, we impose sparsity and entropy constraints. In low pruning regimes, noisy couplings cause the entropy to increase considerably. By contrast, in high sparsity regimes, pruning can effectively reduce the overall entropy of the connections of the parse tree-like structure encoded in a CapsNet. 
%We used LOBSTER \ref{lobster} as the main pruning technique since it can prune many parameters without performance loss and it can be easily coded into the training process.
We collect the coupling coefficients studying their frequency and cardinality, observing lower intra-class conditional entropy: the pruned version \emph{adds a missing explicit prior} in the routing mechanism, grounding the coupling of the unused capsules and disallowing fluctuations under the same baseline performance on the validation/test set. This implies that the parse trees are significantly less, hence more stable for the pruned models, focusing on more relevant features (per class) selected from the routing mechanism. We also introduce two visualization methods of the parse trees based on the type of last capsule layer.

The rest of the chapter is organized as follows: in Section \ref{sec:rem-method} we describe our technique REM, in Section \ref{sec:rem-results} we show the effectiveness of our approach on many datasets and finally we discuss the summary of the proposed approach.

\section{Methodology}
\label{sec:rem-method}
The coupling coefficients computed by the routing mechanism model are the part-whole relationships between capsules of two consecutive capsule layers. %Figure~\ref{fig:capsnet_vs_remcaps}. 
Assigning parts to objects (namely, learning how each object is composed) is challenging. 
One of the main goals of the routing algorithm is to extract a parse tree of these relationships. %By definition, a parse tree is a tree where each node can have multiple children but only one parent
%Using the CapsNets notations, 
For example, given the $\xi$-th input of class $j$, an ideal parse tree for a capsule $i$ detecting one of the discriminative parts of the entity in the input $\xi$ would ideally lead to
\begin{equation}
    \boldsymbol{c}_{i,:}^{[L-1](\xi)} = \mathbb{1}_{\hat{y}^{(\xi)}},
\end{equation}
where $\mathbb{1}_{\hat{y}^{(\xi)}}$ is the one-hot encoding for the target class $y^{(\xi)}$ of the $\xi$-th sample. This means that the routing process can carve a parse tree out of the CapsNet which explains perfectly the relationships between parts and wholes.
One of the problems of this routing procedure is that there is no constraint on how many parse trees there should be. %As a consequence, a primary capsule can be connected with high probabilities to more than one output capsule. Furthermore, CapsNets have a high number of trainable parameters and there might be too many active primary capsules, so in this setting finding the agreements between capsules can be problematic \cite{renzulli}. 

\begin{figure}
    \centering
    \includegraphics[width=1\columnwidth]{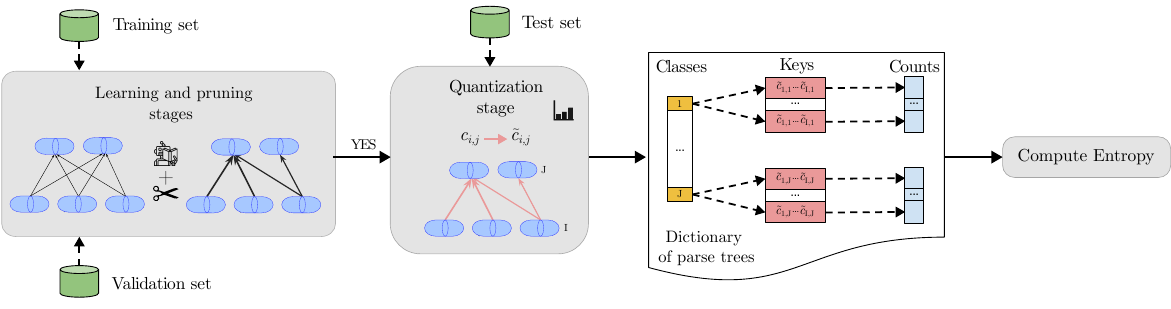}
    %\def\svgwidth{480pt}
    %\input{Parts/02_DemystifyngCapsNetsComponents/gfx/rem/pipeline.eps_tex}
    %\normalsize			
    \caption{Pipeline\label{fig:pipeline} of REM. After the learning stage, the coupling coefficients of the CapsNet are quantized, and the obtained parse trees are collected in a dictionary.}
\end{figure}

In this section, we present our technique REM, first showing how to extract a parse tree and then how to extract fewer parse trees.
The pipeline of our method is depicted in Figure~\ref{fig:pipeline}.

\subsection{Parse Trees Extraction} \label{sec:rem-parse-tree-extraction}
%\footnote{also known as dictionaries or hash tables in traditional computer programming} for each object class. Given a sequence of these coupling coefficients (that we term dictionary)
Once we have a trained CapsNets model, to analyze the routing mechanism, we extract all the possible routing coupling coefficients and build a parse tree. Towards this end, we want to define a metric that helps us decide if the relationships captured by the routing algorithm resemble a parse tree or not. Therefore, we organize the coupling coefficients into \textit{associative arrays} so that we can compute the number of occurrences of each coupling sequence to measure the entropy of the whole dictionary. We refer to this entropy as the \textit{simplicity} of the parse tree. In other words, we refer to the number of keys in the dictionary as the number of unique parse trees that can be carved out from the input dataset. 
In the following paragraphs, we explain how to generate these sequences by discretizing the coupling coefficients and how to create the dictionary.

\textbf{Quantization.} During the quantization stage, we first compute the \textit{continuous} coupling coefficients $c_{i,j}^{[l](\xi)}$ for each $\xi$-th input example. It should be noted that these are the coupling coefficients obtained after the forward pass of the last routing iteration. Then, we quantize them into $K$ discrete levels through the uniform quantizer $q_K(\cdot)$, obtaining
\begin{equation}
    \tilde{c}_{i,j}^{[l](\xi)} = q_K({c}_{i,j}^{[l](\xi)}).
\end{equation}
%We sample our discrete levels from a uniform distribution $X \sim \mathcal{U}(0,1)$. 
We choose the lowest $K$ such that the accuracy does not deteriorate. Here, we will refer to CapsNet+Q as trained CapsNet where the coupling coefficients are quantized.

\textbf{Entropy.} Given the quantized coupling coefficients of a CapsNet+Q, we can extract the parse tree (and create a dictionary of parse trees) for each class $j$, where each entry is a string composed of the quantization indices of the coupling coefficients $\boldsymbol{\tilde{c}}_{:,j}^{[L-1](\xi)}$. 
%\marco{Se scriviamo input $x_n$ forse si capisce meglio... ma allora teniamo la notazione con la dipendenza da n}
%If $\xi_{j}(n)$ is already present in the dictionary, we update the corresponding count value for that string.\enzo{banale}
Given a dictionary for the coupling coefficients of a CapsNet+Q, we can compute the entropy for each class as 
% \begin{equation}
%     \mathbb{P}(\xi \mid y = j) = \frac{\sum\limits_{n=1}^{N_j} [\xi_{\hat{y}_n}(n) = \xi]} {\sum\limits_{j=1}^{N} \mathbb{1}_{\hat{y}^n, \hat{y}^j}}
% \end{equation}
% \enzo{dubbi sul numeratore...} \marco{$N_j$ non definito ... anche se si capisce meglio essere pedanti} where $\mathbb{1}$ is the indicator function. 
\begin{equation}
    \label{eq::Hj}
        \mathbb{H}_j = - \sum\limits_{\xi}\big\{ \mathbb{P}\big(\boldsymbol{\tilde{c}}_{:,j}^{[L-1](\xi)} \mid y^{(\xi)}=j\big) \cdot 
        \log_2 \big[\mathbb{P}(\boldsymbol{\tilde{c}}_{:,j}^{[L-1](\xi)} \mid y^{(\xi)} = j)\big]\big\},
% \begin{split}
% a &= \left( \frac12 + \frac13 + \frac14 \right. \\
%   &\quad \left. {}+ a + b + c \vphantom{\frac12}\right)
% \end{split}
\end{equation}
where $\mathbb{P}(\boldsymbol{\tilde{c}}_{:,j}^{[L-1](\xi)} \mid y^{(\xi)} = j)$ is the frequency of occurrences of a generic string $\xi$ for each \emph{predicted} class $y^{(\xi)}$.
When stacking multiple capsule layers, we apply the quantization stage to each layer and compute the entropy values on the last layer $L$.
Finally, the entropy of a dictionary for a CapsNet+Q on a given dataset is the average of the entropies $\mathbb{H}_j$ of each class
\begin{equation}
    \label{eq::H}
        \mathbb{H} = \frac{1}{J}\sum\limits_{j} \mathbb{H}_j.
\end{equation}
Intuitively, the lower \Eqref{eq::H}, the fewer the number of parse trees carved out from the routing algorithm. 

\subsection{Parse Trees Visualization}\label{sec:rem-extract_salmap}
As mentioned in Section~\ref{sec:general_architecture}, there are three types of capsule layers. Primary capsules and convolutional capsules have a spatial connotation, while fully-connected capsules do not. We describe here two different visualization methods of the extracted parse trees based on the type of the capsules in layer $L-1$. In both scenarios, we follow only the connections starting from the capsule representing the predicted object label, we refer to this method as ``backtracking''.

\textbf{Primary and convolutional capsules.}
\begin{figure}
\centering
\includegraphics[width=1\columnwidth]{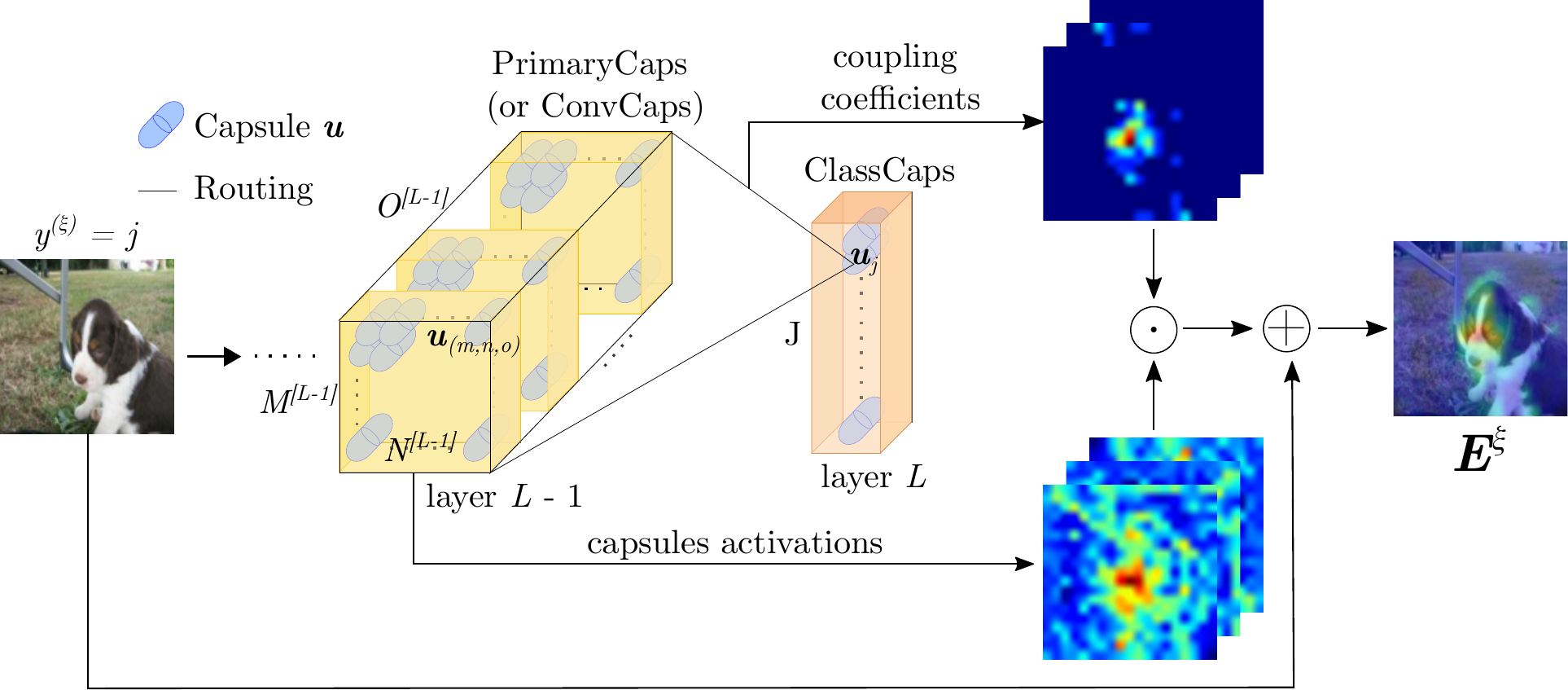}
%\def\svgwidth{480pt}
%\input{Parts/02_DemystifyngCapsNetsComponents/gfx/rem/pipeline.eps_tex}
%\normalsize			
\caption{\textcolor{black}{Extraction of a saliency map.}\label{fig:rem_heatmap_extraction}}
\end{figure}
In Figure~\ref{fig:capsnet_general_architecture}, we see that primary and convolutional capsules are organized in three dimensions and located in a $M \times N \times O$ grid structure. Therefore, in this scenario, we exploit the coupling coefficients $c_{i,j}^{[L-1]}$ for the predicted class $j$ as visual attention built-in explanation to carve out the part structure discovered by a capsule model. We say built-in explanation because we rely only on the forward pass of the network. While Grad-CAM~\cite{gradcam} weighs neurons activations by gradients computed in the backward pass, we weigh each capsule activation by the corresponding coupling coefficient. We refer to $\vu_{(m,n,o)}^{(\xi)}$ to indicate the pose of the capsule in position $(m,n,o)$ for a given input $\xi$. With $\tilde{c}_{(m,n,o),j}^{[L-1](\xi)}$, we refer to the quantized coupling coefficient between a capsule in position $(m,n,o)$ of layer $L-1$ and the predicted capsule-class $j$ of layer $L$  for a given input $\xi$. We denote with $\mE^{(\xi)}$ the saliency map for a given input $\xi$.
We follow \citep{gracapsnets}, where the coupling coefficients of the predicted class $j$ of a trained model for a given input are used as an attention matrix. Unlike \citep{gracapsnets}, we also weight each coupling coefficient by the activity of capsule $i$. 
Therefore, each element of the saliency map is computed as 
\begin{equation}
    E_{(m,n)}^{(\xi)} = \frac{1}{O}\sum\limits_{o}(\|\vu_{(m,n,o)}^{[L-1](\xi)}\| \cdot \tilde{c}_{(m,n,o),j}^{[L-1](\xi)}).
\end{equation}
Then we upsampled the saliency map to the input size with the bilinear method.
Figure~\ref{fig:rem_heatmap_extraction} depicts our method's visualization of extracting a saliency map from an input image given a CapsNet model.

\textbf{Fully-connected capsules.} CapsNets can also be organized in fully connected capsule layers. The saliency map method can not be employed in this scenario, as capsules in layer $L-1$ do not have a spatial connotation. Therefore, similarly to~\citet{mitterreiter2023capsule}, we propose a different visualization of the parse tree. Each capsule is a node where the color is its activation probability, and the strength of the connection is depicted with fuzzy edges of different strokes. We also denote with $deg^-(\vu_j)$ the indegree of a capsule $j$ in layer $l$, which is the number of coupling coefficients starting from capsules $i$ in layer $l-1$ that are above some threshold. We only show the connections from capsule $j$ to capsules $k$ in layer $l+1$ if $deg^-(\vu_j)$ is at least 1.
\subsection{Unconstrained Routing Entropy}
\label{sec:rem-ure}
In this subsection, we are going to more-formally analyze the distribution of the coupling coefficients
\begin{equation}
    \label{eq::standardcij}
    c_{i,j} = \frac{e^{b_{i,j} + \sum_{r=1}^{r_T} \boldsymbol{u}_j^{(r)} \boldsymbol{u}_{i} \boldsymbol{W}_{i,j}}}{\sum_k e^{b_{i,k} + \sum_{r=1}^{r_T} \boldsymbol{u}_k^{(r)} \boldsymbol{u}_i \boldsymbol{W}_{i,k}}},
\end{equation}
where $r_T$ indicates the target routing iterations. We suppress the indices $\xi$ and $l$ for abuse of notation. Let us evaluate the $c_{i,j}$ over a non-yet trained model: as we also saw in Section \ref{sec:rem-parse-tree-extraction}, we have
\begin{equation}
    \label{eq::uniformcij}
    c_{i,j} \approx \frac{1}{J}\ \forall i,j.
\end{equation}
When updating the parameters, following~\cite{gu}, we have 
%\begin{equation}
%    \frac{\partial L}{\partial \boldsymbol{W}_{ij}} = 
%    \left\{
%    \begin{array}{ll}
%        \frac{\partial L}{\partial \boldsymbol{v}_j} \frac{\partial \boldsymbol{v}_j}{\partial \boldsymbol{s}_j}\cdot \boldsymbol{u}_i \cdot c_{ij}& c_{ij} \approx 1\\
%        C \cdot \frac{\partial L}{\partial \boldsymbol{v}_j} \frac{\partial \boldsymbol{v}_j}{\partial \boldsymbol{s}_j}\cdot \boldsymbol{u}_i \cdot c_{ij}& c_{ij} \approx 0\\
%    \end{array}
%    \right .
%\end{equation}
%where $C$ is a positive, non-null constant. Computing the gradient through back-propagation, we observe that
\begin{equation}
    \frac{\partial \Ls}{\partial \boldsymbol{W}_{i,j}} = \Bigg[\frac{\partial \Ls}{\partial \boldsymbol{u}_j} \frac{\partial \boldsymbol{u}_j}{\partial \boldsymbol{s}_j} \cdot c_{i,j} + 
    \sum_{m=1}^M \Bigg( \frac{\partial \Ls}{\partial \boldsymbol{u}_m} \frac{\partial \boldsymbol{u}_m}{\partial \boldsymbol{s}_m} \cdot \boldsymbol{\hat{u}}_{i,m}\frac{\partial c_{i,m}}{\boldsymbol{\hat{u}}_{i,m}}\Bigg) \Bigg]\cdot \boldsymbol{u}_i,
\end{equation}
where we can have the gradient for $\boldsymbol{W}_{i,j}\approx 0$ in a potentially-high number of scenarios, despite $c_{i,j}\neq\{0, 1\}$. Let us analyze the simple case in which we have perfect outputs, matching the ground truth. Hence we are close to a local (or potentially the global) minimum of the loss function:
\begin{equation}
    \left\|\frac{\partial \Ls}{\partial \boldsymbol{u}_m} \right\|_2 \approx 0\ \forall m.
\end{equation}
Looking at \Eqref{eq:softmax}, we see that the right class is chosen, but given the squashing function, we have as an explicit constraint that, given the $j$-th class as the target one, we require 
\begin{equation}
    \label{eq::constraint}
    \left\|\boldsymbol{u}_j\right\|_2 \gg \left\|\boldsymbol{u}_m\right\|_2\ \forall m\neq j
\end{equation}    
on the $\boldsymbol{W}_{i,j}$, which can be accomplished in many ways, including:
\begin{itemize}
    \item having sparse activation for the primary capsules $\boldsymbol{u}_i$: in this case, we have constant $\boldsymbol{W}_{i,j}$ (typically associated with no-routing based approaches); however, we need heavier deep neural networks as they have to force sparse signals already at the output of the primary capsules. In this case, the coupling coefficients $c_{i,j}$ are also constant by definition;
    \item having sparse votes $\boldsymbol{\hat{u}}_{i,j}$: this is a combination of having both primary capsules and weights $\boldsymbol{W}_{i,j}$ enforcing sparsity in the votes and the typical scenario with many routing iterations.
\end{itemize}
Having sparse votes, however, does not necessarily result in having sparse coupling coefficients. According to \Eqref{eq:sj}, the coupling coefficients are multiplied with the votes, obtaining the output capsules. The distribution of the coupling coefficients requires \Eqref{eq::constraint} to be satisfied only: if $\boldsymbol{W}_{i,j}$ is not sparsely distributed, we can still have sparse votes. However, this is the main reason we observe high entropy in the coupling coefficient distributions: %(Figure~\ref{fig:cij_nopruning}) 
as the votes, $\boldsymbol{\hat{u}}_{i,j}$ are implicitly sparse (yet also disordered, as we are not explicitly imposing any structure in the coupling coefficients distribution), the model is still able to learn, but it finds a typical solution where $c_{i,j}$ are not sparse. However, we would like sparsely distributed, recurrent couplings to the same $j$-th output caps $\boldsymbol{c}_{:,j}$, establishing stable relationships between the features extracted at the primary capsule layer.

Minimizing explicitly the entropy term in \Eqref{eq::Hj} is an intractable problem due to the non-differentiability of the entropy term and the quantization step (in our considered setup) and due to the huge computational complexity to be introduced at training time. Hence, we can try to implicitly enforce routing entropy minimization by forcing a sparse and organized structure in the coupling coefficients. Towards this end, one efficient solution is to enforce sparsity in the $\boldsymbol{W}_{i,j}$ representation by compelling a vote between the $i$-th primary capsule and the $j$-th output caps to be exactly zero for any input, according to \Eqref{eq::standardcij}
\begin{equation}
        c_{i,j} = \frac{1}{\sum_k e^{b_{i,k} + \sum_{r=1}^t \boldsymbol{u}_k^{(r)} \boldsymbol{u}_i \boldsymbol{W}_{i,k}}}.
\end{equation}
In this way, having a lower variability in the $c_{i,j}$ values (and hence building more stable relationships between primary and output capsules), straightforwardly, we are also explicitly minimizing the entropy of the quantized representations for the coupling coefficients. In the following subsection, we will tailor a sparsity technique to accomplish such a goal. 

\subsection{Enforcing REM with Pruning}
CapsNets are trained via standard back-propagation learning, minimizing some loss functions like margin loss. Our ultimate goal is to assess to what extent a variation of the value of some parameter $\theta$ would affect the error on the network output. In particular, the parameters not affecting the network output can be softly pushed to zero, meaning we can apply an $\normltwo$ penalty term. Several approaches have been proposed, especially in recent years~\citep{louizos2017learning,molchanov2019importance,lee2018snip}. One recent state-of-the-art approach, LOBSTER~\citep{lobster} proposes to penalize the parameters by their gradient-weighted $\normltwo$ norm, leading to the update rule
\begin{align}
    \theta^{t+1} = &\theta^t - \eta G\left[\frac{\partial \Ls}{\partial \theta^t}\right] - \lambda \theta^t \textrm{ReLU}\left[1- \left | \frac{\partial \Ls}{\partial \theta^t}\right| \right],
    \label{eq:oleupdateRule}
\end{align}
where $G\left[\frac{\partial \Ls}{\partial \theta^t}\right]$ is any gradient-based optimization update (for SGD it is the plain gradient, but other optimization strategies like Adam can be plugged) and $\eta, \lambda$ are two positive hyper-parameters.

Such a strategy is particularly effective on standard convolutional neural networks and easy to plug into any back-propagation-based learning system. Furthermore, LOBSTER is a regularization strategy that can be plugged at any learning stage, as it self-tunes the penalty introduced according to the learning phase. This non-intrusiveness in the complex and delicate routing mechanism for CapsNets resulted in a suitable choice to enforce REM.

\section{Experiments and Results}
\label{sec:rem-results}
This section reports the experiments and results we performed to test REM. We first show the effects on the MNIST dataset with dynamic routing CapsNets  \mbox{(DR-CapsNets)}, reporting how the entropy and the accuracy values change during training. Then, we test REM on more complex datasets such as Fashion-MNIST, CIFAR-10, SVHN, Dogs vs. Cats~\citep{dogs-vs-cats} and Imagenette~\citep{imagewang}. We used the same architectures configurations and augmentations described in \citet{hinton-dr}. We also conducted experiments applying our technique to $\gamma$-CapsNets~\citep{peer2019gammacapsules},
DeepCaps~\citep{deepcaps}, Eff-CapsNets~\citep{mazzia2021efficient} to test the efficacy of REM to some other variants of capsule models, including different architectures, routing algorithms, and the number of trainable parameters. We used a custom architecture called Eff-ConvCapsNet for higher-resolution images such as Dogs vs. Cats and Imagenette. This model has three capsule layers (primary, convolutional and fully-connected) and shares a backbone network similar to Eff-CapsNets.
We trained all models with five random seeds. We report the classification accuracy (\%) and entropy (averages and standard deviations), the \textit{sparsity} (percentage of pruned parameters, median), and the number of keys in the dictionary (median).
% \footnote{ The code will be released upon acceptance of the paper.}

\subsection{Ablation Study}
\label{sec:rem-abl}
To assess our REM technique, we analyze in-depth the benefits of pruning towards REM on the MNIST dataset. Despite its outdatedness, MNIST remains an omnipresent benchmark for CapsNets~\citep{hinton-dr, gu, sparse-caps, hinton-stacked-caps, VAEs_capsules}.

\begin{figure}
\centering
\includegraphics[width=1\columnwidth]{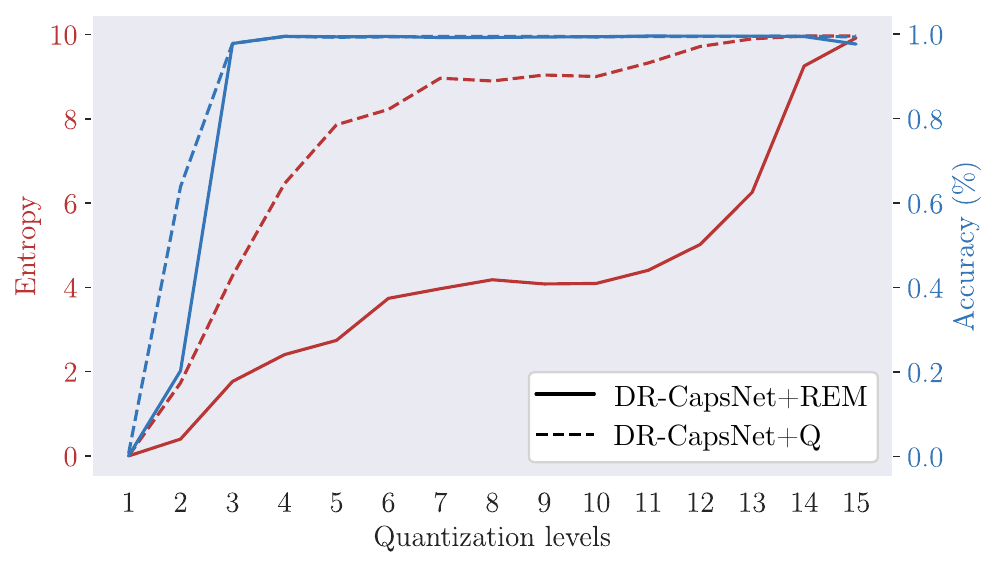}
%\def\svgwidth{480pt}
%\input{figures/pipeline.eps_tex}
%\normalsize			
\caption{\textcolor{black}{Entropy and accuracy for DR-CapsNet+Q and DR-CapsNet+Q with different quantization levels on MNIST (test set).}\label{fig:quantization levels}}
\end{figure}

\textbf{Choice of quantization levels.}\label{sec:choice_quant_levels}
The routing algorithms used in the models employed in this paper are performed between two consecutive capsule layers. As we can see in Figure~\ref{fig:quantization levels}, the choice of the number of quantization levels $K$ for the coupling coefficients computed by a routing algorithm of a DR-CapsNet affects the performance of the network. We select the value for $K$ that achieves the best accuracy value with relatively low entropy. In this case, when $K$=11, DR-CapsNet+Q achieves $99.47\%$ accuracy and $9.32$ entropy, while DR-CapsNet+REM achieves $99.57\%$ accuracy and $4.40$ entropy.

\textbf{Entropy at different epochs}. 
\begin{figure}
\centering
\includegraphics[width=0.8\columnwidth]{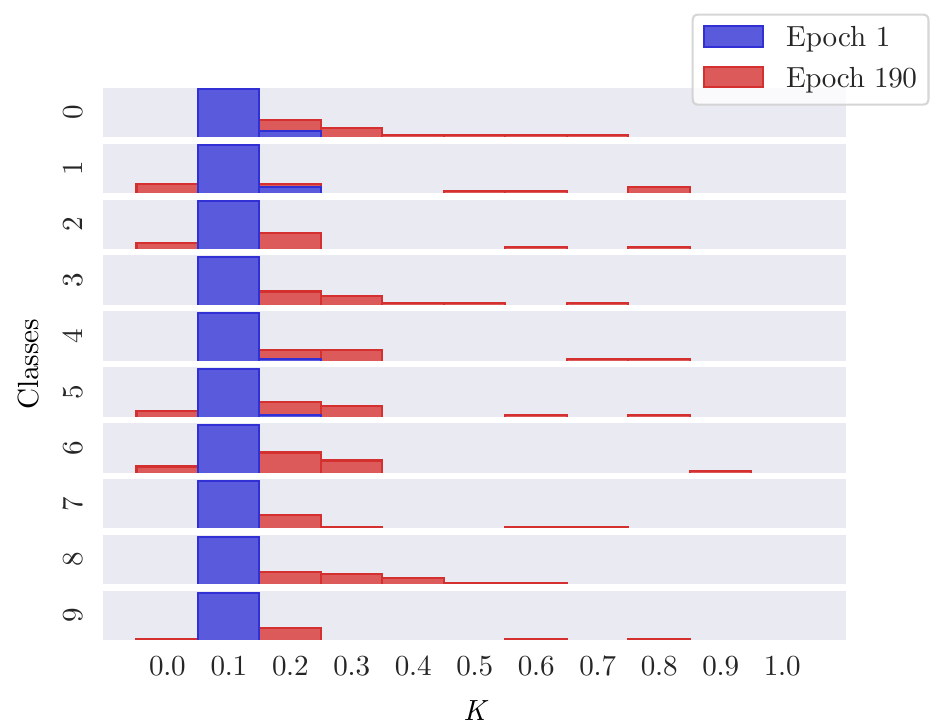}		
\caption{Coupling coefficients distributions for each class of two DR-CapsNets+Q at epochs 1 and 190 on MNIST (test set).\label{fig:cij_nopruning_all}}
\end{figure}
\begin{table}
\footnotesize
\centering
\begin{tabular}{ccc}
\hline
\textbf{Model}         & \textbf{Accuracy} & \textbf{Entropy} \\ \hline
$\text{DR-CapsNet}_{1}\text{+Q}$   & $97.53$      & $1.83$  \\
$\text{DR-CapsNet}_{190}\text{+Q}$ & $99.55$      & $9.41$  \\ \hline
\end{tabular}
\caption{\centering\label{tab:acc_nopruning_epochs}Classification results (\%) and entropy of two DR-CapsNets+Q at epochs 1 and 190 on MNIST (test set).}
\end{table}
Let us now generate the parse trees
on the CapsNet model trained on the MNIST dataset, investigating their evolution during training. To this end, we compute the distribution of the coupling coefficients on the test set for two models: the model trained after the first epoch and the one that achieved the lowest loss value on the validation set.
Figure~\ref{fig:cij_nopruning_all} shows the distributions for the two models and Table~\ref{tab:acc_nopruning_epochs} reports the corresponding accuracy and entropy values. We can observe that after the first epoch, DR-CapsNet is far from optimality, both in terms of performance (accuracy of 97.4\%) and parse tree discriminability: indeed, all coupling coefficients are almost equal to the value selected for initialization, i.e., $1/J$, where $J$ is the number of output capsules. 
On a given dataset, we target a model with high generalization but low entropy, namely a low number of extracted parse trees.
\begin{figure}
  \centering
  \includegraphics[width=1\columnwidth]{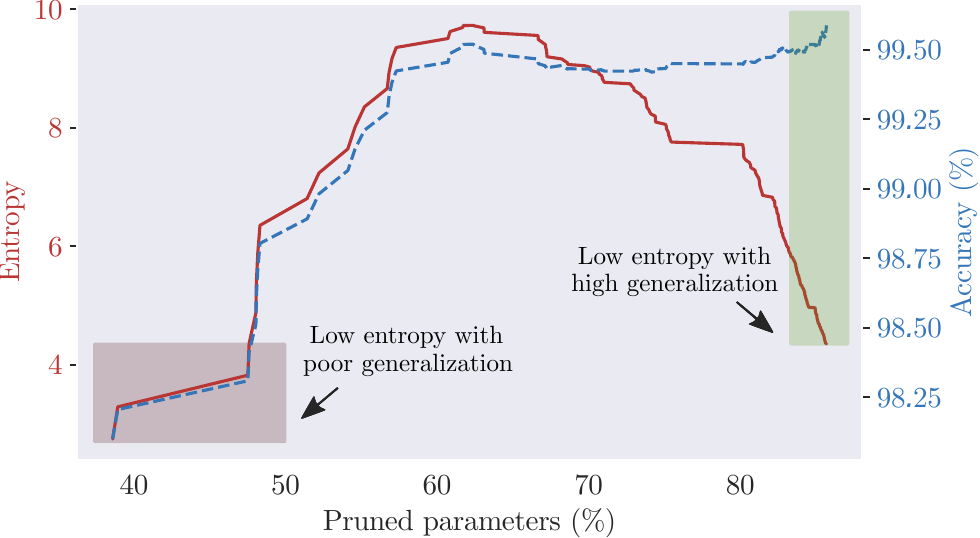}
  \caption{Accuracy and entropy vs. pruned parameters on MNIST (test set).}
  \label{fig:acc_vs_pp}
\end{figure}
Figure~\ref{fig:acc_vs_pp} shows how the entropy (red line) and classification accuracy (blue dotted line) change as the sparsity increases during training. At the beginning of the training stage, we can see that the entropy is low (1.83) because the routing algorithm has not yet learned to correctly discriminate the relationships between the capsules (97.53\% accuracy). This effect is almost the same when we train a DR-CapsNet with $t=1$ as \citet{gu}, where its entropy is zero, but capsules are uniformly coupled. However, at the end of the training process, we can get a model trained with REM with higher performances (99.60\% accuracy) and still low entropy (4.31). 
Figure~\ref{fig:capsnetq_vscapsnetrem} illustrates the distributions of the coupling coefficients for a DR-CapsNet+Q and a DR-CapsNet+REM following the method described in Section~\ref{sec:rem-parse-tree-extraction}. We can see that the distributions of the DR-CapsNet+REM model are sparser than those for the DR-CapsNet+Q model. 
\begin{figure}[h]
      \centering
      \includegraphics[width=0.8\columnwidth]{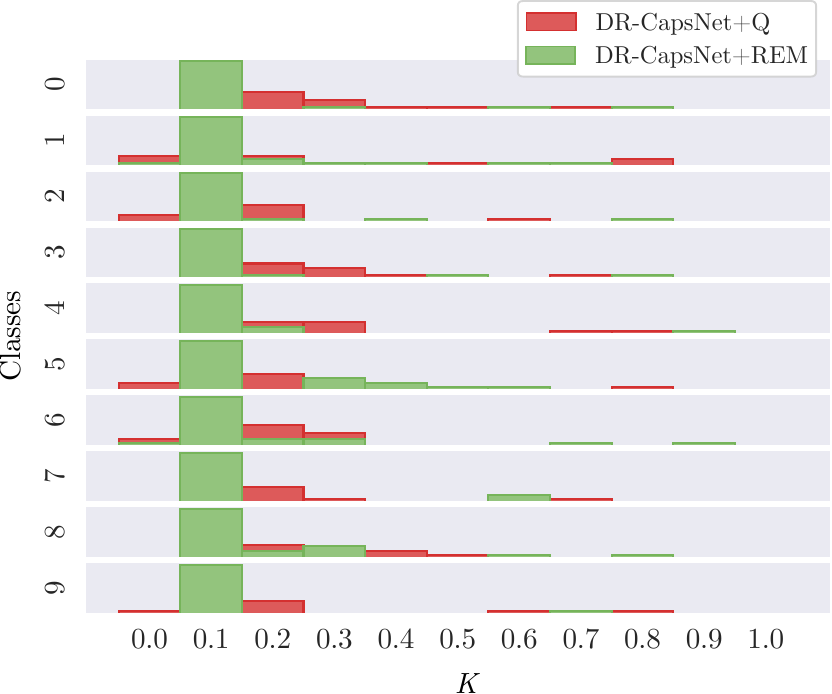}    \caption{\centering\label{fig:capsnetq_vscapsnetrem}Coupling coefficients distributions on MNIST (test set).}
\end{figure}
% \textbf{Strength of parse trees}. In Figure~\ref{fig:capsnetq_vscapsnetrem}, we plot the distributions of the coupling coefficients for a CapsNet+Q and a CapsNet+REM following the method described in Section~\ref{sec:rem-parse-tree-extraction}. We can see that the distributions of the CapsNet+REM model are sparser than those for the CapsNet+Q model. Namely, we can carve out parse trees with stronger part-whole relationships, achieving high generalization.
% \begin{figure}[h]
%       \centering
%       \includegraphics[width=1\columnwidth]{Parts/02_DemystifyngCapsNetsComponents/gfx/rem/cij_distributions_prunedvsunpruned_all.pdf}    \caption{\centering\label{fig:capsnetq_vscapsnetrem}Coupling coefficients distributions on MNIST (test set).}
% \end{figure}

\begin{figure}[h]
 \centering
 \includegraphics[width=0.8\columnwidth, trim={0 0 0 0},clip] {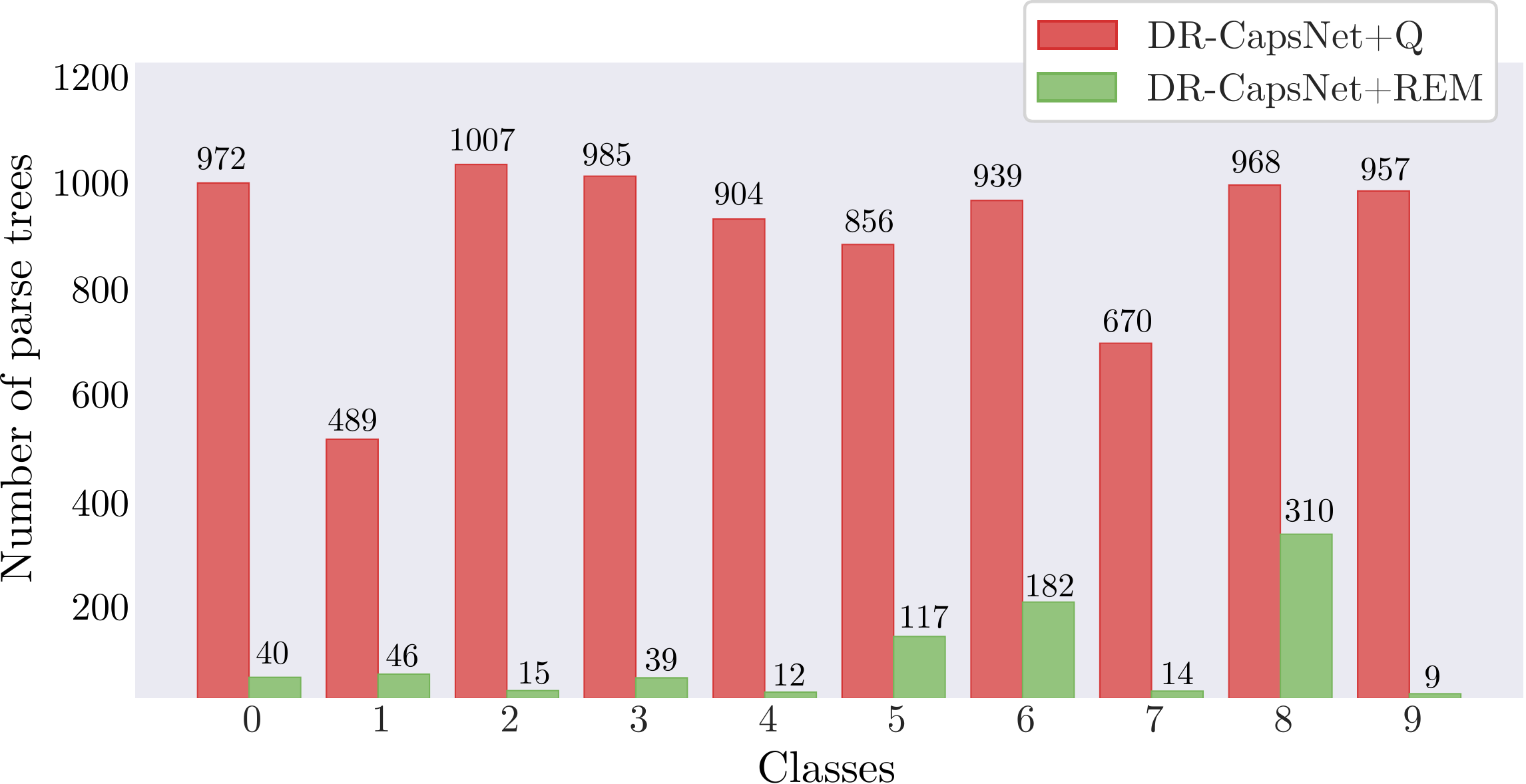}
% %\def\svgwidth{480pt}
  %\input{Parts/02_DemystifyngCapsNetsComponents/gfx/rem/pipeline.eps_tex}
% %\normalsize			
 \caption{\centering\label{fig:mnist-dictionary-test}Number of keys for each class on MNIST (test set).}
 \end{figure}

\textbf{Number of parse trees.} Figure~\ref{fig:mnist-dictionary-test} shows the number of intra-class parse trees (collected in a dictionary) for DR-CapsNets+REM and a DR-CapsNets+Q, namely a DR-CapsNet where the quantization is applied without pruning the network during training. We can see that the number of keys in the dictionary for DR-CapsNets+REM is lower than the one for DR-CapsNets+Q for each class. Also, the entropy measure for DR-CapsNets+REM is lower than DR-CapsNets+Q; namely, REM has successfully extracted fewer parse trees on the MNIST test set.

\begin{figure}[h]
		\centering
		\includegraphics[width=0.7\columnwidth]{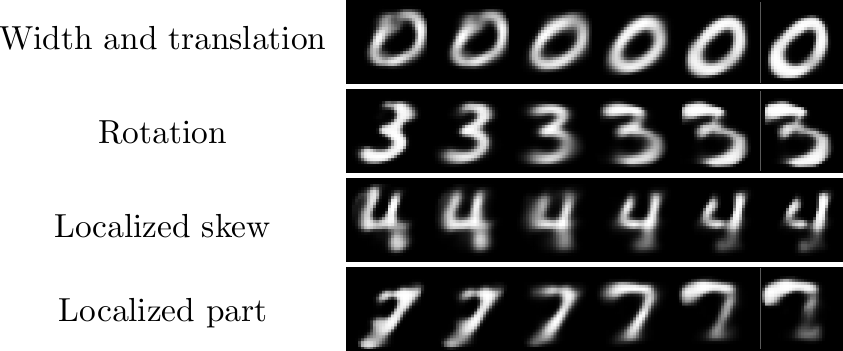}
        \caption{MNIST perturbation reconstructions of a frozen DR-CapsNet+REM.}% with a decoder stacked on top of it.}
        \label{fig:reconstruction}
	\end{figure}

\textbf{Decoder.}\label{subsec:decoder}
A CapsNet is typically composed of an encoder and a decoder part, where the latter is a reconstruction network with three fully connected layers \cite{hinton-dr}. In the previously-discussed experiments, we have removed the decoder. One limitation of our work arises when computing the entropy of CapsNets trained with the decoder. 
Tables ~\ref{tab:decoder-acc-supp} and ~\ref{tab:decoder-ent-supp} report the classification results and entropies values when we trained the encoder and the decoder part together.
We observed that the entropy of a DR-CapsNets+REM is almost the same as that of a DR-CapsNet+Q. Indeed, when the decoder is used, the activity vector of an output capsule encodes richer representations of the input. ~\citet{hinton-dr} introduced the decoder to boost the routing performance on MNIST by enforcing the pose encoding a capsule. They also show that such perturbation affects the reconstruction when a perturbed activity vector is fed to the decoder. \textcolor{black}{So capsule representations are \emph{approximately equivariant}, meaning that even if they do not come with guaranteed equivariances, transformations applied to the input can still be described by continuous changes in the output vector.} To verify if output capsules of a trained DR-CapsNet+REM without the decoder (so with low entropy) are still approximately equivariant, we stacked on top of it the reconstruction network, without training the encoder. 
The MNIST dataset's decoder comprises three fully connected layers of 512, 1024, and 784 neurons, respectively, with two RELU and a final sigmoid activation function. This network is trained to minimize the Euclidean distance between the image and the output of the sigmoid layer.
We can see in Figure~\ref{fig:reconstruction} that DR-CapsNets+REM with low entropy are still approximately equivariant to many transformations. 
\begin{table}[h]
\footnotesize
\centering
\begin{tabular}{cccc}
\midrule
\textbf{Model}           & \textbf{MNIST}               & \textbf{F-MNIST}            & \textbf{CIFAR-10}            \\ \midrule
DR-CapsNet+Q      & $99.58\pm_{0.03}$ & $92.57\pm_{0.39}$ & $72.40\pm_{0.54}$ \\ 
DR-CapsNet+REM    & $99.63\pm_{0.02}$ & $92.76\pm_{0.38}$ & $76.00\pm_{0.63}$ \\ \midrule
\end{tabular}
\caption{\centering\label{tab:decoder-acc-supp}Classification results with the decoder on MNIST, Fashion-MNIST, CIFAR-10 (test set).}
\end{table}

\begin{table}[h]
\footnotesize
\centering
\begin{tabular}{cccc}
\midrule
\textbf{Model}        & \textbf{MNIST}             & \textbf{F-MNIST}           & \textbf{CIFAR-10}          \\ \midrule
DR-CapsNet+Q    & $9.88\pm_{0.06}$ & $8.49\pm_{1.50}$ & $4.55\pm_{1.13}$ \\
DR-CapsNets+REM & $9.40\pm_{0.55}$ & $6.15\pm_{2.32}$  & $3.85\pm_{0.54}$  \\ \midrule
\end{tabular}
\caption{\centering\label{tab:decoder-ent-supp}Entropies for models trained with the decoder on Fashion-MNIST and CIFAR-10 (test set).}
\end{table}

\textbf{Other networks}
Table~\ref{tab:mnist-pruned-params} shows the performances on MNIST of DR-CapsNets, $\gamma$-CapsNets, DeepCaps and Eff-CapsNets. We notice that $\gamma$-CapsNet and $\gamma$-CapsNet+REM have the lowest entropy values since $\gamma$-CapsNets employ a scaled-distance-agreement routing algorithm that enforces the single parent constraint. With our technique REM we can successfully lower the entropy even more.
\begin{table}[h]
\footnotesize
\centering
\begin{tabular}{*{5}{c}}
\toprule
\textbf{Model}  &\textbf{Parameters}   & \textbf{Accuracy (\%)} & \textbf{Sparsity (\%)} & \textbf{Entropy}\\ \midrule
DR-CapsNet+Q & 6.8M & $99.56\pm_{0.03}$   & $-$     & $9.53\pm_{0.54}$                \\
DR-CapsNet+REM & 0.9M & $99.56\pm_{0.02}$      & $85.53$    & $4.16\pm_{1.59}$         \\
$\gamma$-CapsNet+Q & 7.7M & $99.50\pm_{0.07}$   & $-$     & $1.87\pm_{1.38}$                \\
$\gamma$-CapsNet+REM & 0.8M& $99.50\pm_{0.05}$      & $89.71$    & $1.34\pm_{1.09}$         \\
DeepCaps+Q & 8.4M & $99.51\pm_{0.24}$   & $-$     & $5.26\pm_{2.00}$                \\
DeepCaps+REM & 2.4M & $99.61\pm_{0.23}$      & $71.73$    & $3.10\pm_{1.07}$         \\
Eff-CapsNets+Q & 161k & $99.55\pm_{0.31}$   & $-$     & $4.38\pm_{1.59}$                \\
Eff-CapsNets+REM & 43k & $99.58\pm_{0.64}$      & $73.15$    & $2.60\pm_{1.72}$         \\\bottomrule
\end{tabular}
\caption{\centering\label{tab:mnist-pruned-params}Results for DR-CapsNets, $\gamma$-CapsNets, DeepCaps and Eff-CapsNets on MNIST (test set).}
\end{table}
\subsection{Experiments}
\label{sec:rem-exp}
 Considering the broad heterogeneity of proposed architectures and the adaptability of REM to other architectures and datasets, we have chosen to test DR-CapsNets $\gamma$-CapsNets, DeepCaps, Eff-CapsNets and Eff-ConvCapsNets to other datasets other than MNIST.

\textbf{Datasets.} We trained and tested CapsNets on: i) Fashion-MNIST, 28$\times$28 grayscale images (10 classes); (ii) SVHN, 32$\times$32 RGB images (10 classes); iii) CIFAR-10, 32$\times$32 RGB images (10 classes); (iv) Dogs vs. Cats (images resized to 128$\times$128); (v) Imagenette, a subset of 10 classes from the Imagenet dataset resized to 128$\times$128.
For MNIST, Fashion-MNIST, and CIFAR-10, we used 5\% of the training set as a validation set. For Dogs vs. Cats, Imagenette and Tiny Imagenet, we used 10\% of the training set as validation set. To test the robustness of novel azimuths on smallNORB, we train all models on 1/3 of the training data with azimuths of 0, 20, 40, 300, 320, and 340 degrees and test them on 2/3 of test data with remaining azimuths never seen during training. To test the robustness of our technique on novel elevations, we trained models on 1/3 of the training data with elevations of 30, 35, and 40 degrees from the horizontal. We tested on 2/3 of the test data with the remaining elevations. 
Finally, to test the robustness to affine transformations, we used expanded MNIST training and validation sets (40×40 padded and translated MNIST images) and the affNIST test set, in which each example is an MNIST digit with a random small affine transformation.

\textbf{Model architectures.}\label{sec:model_architectures}
All models employed in this work were tested using the same architectures (number of layers, capsule dimensions, number of routing iterations, etc.) presented in the original papers. Therefore, for DR-CapsNets, $\gamma$-CapsNets, DeepCaps and Eff-CapsNets we used the same architectures configurations as in \citet{hinton-dr, peer2019gammacapsules, deepcaps, mazzia2021efficient} respectively.
Eff-ConvCapsNet consists of three capsule layers stacked on top of a backbone network similar to Eff-CapsNets. We use the notation Conv(\textit{output-channels}, \textit{kernel}, \textit{stride}) for standard convolutional layers. For convolutional capsule layers, the notation is ConvCaps(\textit{output-capsule-dimension}, \textit{output-capsule-types}, \textit{kernel}, \textit{stride}). For fully-connected capsule layers, we use FcCaps(\textit{output-capsule-dimension}, \textit{output-capsule-types}). The backbone network utilizes four standard convolutional layers Conv(32,5,2), Conv(64,3,1), Conv(64,3,1), and Conv(256,3,2), followed by a fifth depthwise convolutional layer Conv(256,3,1) to extract 16 primary capsule types of dimension 16. Then, a convolutional capsule layer ConvCaps(16,16,3,2) is stacked on top of primary capsules. Finally, a fully-connected capsule layer FcCaps(16, 10) (FcCaps(16, 2) for Dogs vs. Cats) with shared transformation matrices is stacked to output the class capsules. We used the dynamic routing method introduced by ~\citet{hinton-dr}.

To visualize the parse trees in multilayer DR-CapsNets, we stacked five capsule layers, 16 capsules for the primary capsule layer, 16 capsules for the three hidden layers and ten capsules for the last layer. Each capsule is composed by 8 neurons.

\textbf{Training.}
For models without REM, we take the checkpoint that achieved the lowest loss on the validation set, while for models with REM we take the model on the last epoch. We checked the loss on the validation set and used an early-stop of $200$ epochs. The models were trained on batches of size 128 using Adam optimizer with its PyTorch~1.12 default parameters, including an exponentially decaying learning rate factor of $0.99$.

\textbf{Quantization levels}
We used $\gamma$-CapsNets with 3 capsule layers as in~\citet{peer2019gammacapsules}. For $\gamma$-CapsNets+Q and $\gamma$-CapsNets+REM we found $K=11$ for the first two capsule layers and $K=6$ for the last two. For DeepCaps+Q and DeepCaps+REM we used $K=11$ for all the capsule layers where the number of routing iterations is greater than one. For Eff-CapsNets+Q and Eff-CapsNets+REM we also used $K=11$. For Eff-ConvCapsNets+Q and Eff-ConvCapsNets+REM we used $K=21$ on Dogs vs. Cats and Imagenette.

\begin{table*}
\footnotesize
\centering
\resizebox{\textwidth}{!}{
\begin{tabular}{c c c cc ccc}
\toprule

% CapsNet+Q  & Tiny ImageNet & $58.50\pm_{0.25}$ & $0$ & $5.18\pm_{0.67}$\\
% CapsNet+REM  & Tiny ImageNet & $54.02\pm_{0.14}$ & $44.27$ & $3.15\pm_{0.81}$ \\ \midrule

\multirow{2}{*}{\textbf{Dataset}} & \multirow{2}{*}{\textbf{Architecture}} &  \multicolumn{1}{c}{\bf Vanilla} &\multicolumn{2}{c}{\bf Quantized (Q)} & \multicolumn{3}{c}{\bf REM}\\

&& accuracy & entropy & accuracy & entropy & accuracy & sparsity\\
\midrule

\multirow{4}{*}{Fashion-MNIST} & 
DR-CapsNet              & $92.76\pm_{0.2}$  & $8.64\pm_{1.2}$ & $92.46\pm_{0.2}$  & $4.80\pm_{1.7}$ & $92.62\pm_{0.1}$ & $80.71$\\
& $\gamma$-CapsNet   & $92.59\pm_{1.1}$  & $3.98\pm_{0.8}$ &$92.43\pm_{1.1}$  & $1.45\pm_{0.7}$  & $93.01\pm_{1.1}$ & $87.07$ \\
& DeepCaps           & $92.36\pm_{0.2}$  & $7.15\pm_{1.3}$ & $92.33\pm_{0.2}$  & $6.08\pm_{1.3}$ & $94.61\pm_{0.1}$ & $83.29$\\
& Eff-CapsNet & $93.31\pm_{0.2}$  & $3.88\pm_{1.1}$ & $93.22\pm_{0.2}$  & $1.10\pm_{0.5}$ & $92.98\pm_{0.4}$  & $63.29$ \\ \midrule

\multirow{4}{*}{SVHN} & 
DR-CapsNet              & $93.30\pm_{0.2}$  & $7.13\pm_{1.2}$ & $92.20\pm_{0.2}$  & $5.23\pm_{0.7}$ & $91.71\pm_{0.5}$  & $74.40$\\
& $\gamma$-CapsNet   & $89.02\pm_{0.1}$  & $7.15\pm_{0.9}$ &$87.42\pm_{1.2}$  & $5.65\pm_{1.2}$ & $88.36\pm_{0.2}$  & $73.89$ \\
& DeepCaps           & $93.32\pm_{0.3}$  & $11.06\pm_{0.6}$ & $93.20\pm_{0.4}$  & $3.97\pm_{1.5}$ & $93.06\pm_{0.2}$ & $80.50$\\
& Eff-CapsNet & $93.64\pm_{0.1}$  & $0.53\pm_{0.6}$ & $93.62\pm_{0.1}$  & $0.24\pm_{0.4}$ & $93.12\pm_{0.1}$  & $47.80$ \\ \midrule

\multirow{4}{*}{CIFAR-10} & 
DR-CapsNet              & $79.93\pm_{0.2}$  & $6.26\pm_{0.6}$ & $78.42\pm_{2.1}$  & $4.15\pm_{0.6}$ & $79.25\pm_{0.6}$ & 81.17\\
& $\gamma$-CapsNet  & $74.02\pm_{0.3}$  & $3.67\pm_{0.70}$  &$73.08\pm_{0.48}$  & $3.22\pm_{0.7}$ & $74.89\pm_{0.2}$ & 74.89\\
& DeepCaps           & $90.80\pm_{0.1}$  & $8.99\pm_{0.5}$ & $90.47\pm_{0.1}$  & $7.07\pm_{1.1}$ & $90.35\pm_{0.1}$ & 46.83 \\
& Eff-CapsNet  & $81.53\pm_{0.6}$  & $0.25\pm_{0.3}$ & $81.51\pm_{0.5}$  & $0.005\pm_{0.1}$ & $81.49\pm_{0.5}$ & 53.79 \\
\midrule

\multirow{1}{*}{Dogs vs. Cats} & 
Eff-ConvCapsNet             & $97.74\pm_{0.3}$  & $4.05\pm_{0.3}$ & $99.2\pm_{0.3}$ & $2.03\pm_{0.7}$ & $99.1\pm_{0.6}$ & 71.92\\
\midrule

\multirow{1}{*}{Imagenette} & 
Eff-ConvCapsNet              & $83.11\pm_{0.1}$  & $7.63\pm_{0.5}$ & $82.98\pm_{0.6}$ & $5.85\pm_{0.3}$ & $83.02\pm_{0.8}$ & 63.52\\
\bottomrule
\end{tabular}}
\caption{\centering\label{tab:results-otherdatasets}Accuracy (\%), entropy and sparsity on Fashion-MNIST, SVHN, CIFAR-10, Dogs vs. Cats and Imagenette (test set).}
\end{table*}

\textbf{Generalization ability.} As shown in Tables \ref{tab:results-otherdatasets}, a CapsNet+REM has a high percentage of pruned parameters with a minimal performance loss. So this confirms our hypothesis that CapsNets are over-parametrized.
We also report the entropy of the dictionary of the last routing layer for the quantized models. The entropy is successfully lower for all datasets when REM is applied to all architectures, even with fewer parameters than CapsNets, such as Eff-CapsNets. 
Compared to other models, Eff-CapsNets have lower entropy since, in the original implementation, the kernel size for the PrimaryCaps has the same dimension as the backbone output, namely PrimaryCaps, in this case, has no spatial resolution. Furthermore, we noticed that apart from simple datasets such as MNIST or Fashion-MNIST, the coupling coefficients distributions collapse to similar values, namely the non-iterative self-attention mechanism does not build a proper parse tree.

\textbf{Robustness to affine transformations.}\label{sec:affine_transformations} 
To test the robustness to affine transformations of DR-CapsNets+REM, we used expanded MNIST: a dataset composed of padded and translated MNIST, in which each example is an MNIST digit placed randomly on a black background of 40$\times$40 pixels. We used the affNIST dataset as a test set, in which each example is an MNIST digit with a random small affine transformation. We tested an under-trained DR-CapsNet with early stopping which achieved 99.22\% accuracy on the expanded MNIST test set as in ~\citet{hinton-dr, gu}. We also trained these models until convergence. We can see in Table~\ref{tab:results-affnist} that the under-trained network entropies are high. Instead, a well-trained DR-CapsNet+REM can be robust to affine transformations and have low entropy.
\begin{table}[h]
\footnotesize
\centering
    \begin{tabular}{ccccc}
        \toprule
                    & \textbf{expanded MNIST} & \textbf{affNIST} & \textbf{affNIST} & \textbf{affNIST}  \\ 
        \textbf{Model} & \textbf{Accuracy (\%)} & \textbf{Accuracy (\%)} & \textbf{\begin{tabular}[c]{@{}c@{}}Sparsity (\%)\end{tabular}} & \textbf{Entropy} \\ \midrule
        DR-CapsNet+Q   & $99.22$                       & $77.93\pm_{0.55}$           & $0$          & $8.64\pm_{1.15}$         \\
        DR-CapsNet+REM & $99.22$                       & $81.81\pm_{0.81}$           & $71.26$          & $8.45\pm_{1.10}$          \\ \midrule
        DR-CapsNet+Q   & $99.36\pm_{0.05}$                       & $83.14\pm_{0.24}$           & $0$          & $8.45\pm_{0.99}$          \\
        DR-CapsNet+REM & $99.48\pm_{0.02}$                       & $85.23\pm_{0.11}$           & $87.32$          & $5.93\pm_{1.39}$          \\ \midrule
    \end{tabular}
        \caption{\centering\label{tab:results-affnist}{Results on affNIST test set for under-trained and well-trained models.}}
% 	\begin{minipage}{0.35\linewidth}
% 		\centering
% 		\includegraphics[width=\textwidth]{figures/mnist_reconstructions.pdf}
%         \captionof{figure}{MNIST perturbation reconstructions of a freezed CapsNet+REM.}% with a decoder stacked on top of it.}
%         \label{fig:reconstruction}
% 	\end{minipage}
\end{table}

\textbf{Robustness to novel viewpoints.}\label{sec:novel_viewpoints}
CapsNets are well known for their generalization ability to novel viewpoints~\citep{hinton-dr, hinton-em}.
We conducted further experiments on the smallNORB dataset to test the robustness of novel viewpoints of our technique.
We employed Eff-CapsNets, as they are the state-of-the-art models on this dataset with few trainable parameters. 
% Therefore, instead of stacking capsule layers on top of a single convolutional layer as in ~\cite{hinton-dr}, we extract primary capsules using 7 convolutional layers with no shortcut connections. This network has fewer parameters (1.3M) compared to the one with a single convolutional layer (6.6M) since it reduces the spatial size of the input tensor of the primary capsules.
%, namely we train all models on 1/3 of training data with azimuths of 0, 20, 40, 300, 320, and 340 degrees and test them on 2/3 of test data with remaining azimuths never seen during training. To test the robustness of our technique on novel elevations, we trained models on 1/3 of the training data with elevations of 30, 35, and 40 degrees from the horizontal, and tested on 2/3 of the test data with the remaining elevations. 
We used $K=11$ quantization levels for Eff-CapsNets+Q and Eff-CapsNets+REM.
In Table~\ref{tab:novelviewpoints}, we can see that Eff-CapsNets+REM are indeed robust to novel viewpoints with low entropy. Notice that even if our work does not target state-of-the-art generalization, with REM, we manage to achieve a maximum value of accuracy of 86.99\% with only 27k trainable parameters on novel elevations (98.43\% for familiar elevations) and 90.37\% with only 37k trainable parameters on novel azimuths (97.53\% for familiar azimuths). 
\begin{table}[H]
\footnotesize
\centering
\begin{tabular}{lccccc}
\toprule
\multicolumn{1}{c}{\multirow{2}{*}{\textbf{Model}}} & \multicolumn{2}{c}{\textbf{Familiar}}                               & \multicolumn{2}{c}{\textbf{Novel}}                                  & \multicolumn{1}{l}{\multirow{2}{*}{\textbf{Sparsity}}} \\ %\cline{2-5}
                        & \textbf{Accuracy} & \multicolumn{1}{l}{\textbf{Entropy}} & \multicolumn{1}{l}{\textbf{Accuracy}} & \multicolumn{1}{l}{\textbf{Entropy}} & \multicolumn{1}{l}{}                          \\ \midrule
% \textcolor{black}{Eff-CapsNet} & $90.16\pm_{0.07}$ & - & $80.34\pm_{0.010}$ & - & $0$ \\ \hline
%\textcolor{black}{Eff-CapsNet ($\phi$)} & $90.55\pm_{0.05}$ & - & $80.68\pm_{0.009}$ & - & $0$ \\ 
%\textcolor{black}{Eff-CapsNet+Q ($\phi$)} & $89.79\pm_{0.08}$ & $2.03\pm_{0.28}$ & $78.25\pm_{0.011}$ & $2.38\pm_{0.17}$ & $0$ \\ 
%\textcolor{black}{Eff-CapsNet+REM ($\phi$)} & $90.20\pm_{0.11}$ & $1.11\pm_{0.27}$ & $78.18\pm_{0.013}$ & $1.16\pm_{0.06}$ & $55.34$ \\ \midrule
%\textcolor{black}{Eff-CapsNet ($\psi$)} & $90.16\pm_{0.07}$ & - & $80.34\pm_{0.010}$ & - & $0$ \\ 
%\textcolor{black}{Eff-CapsNet+Q ($\psi$)} & $89.22\pm_{0.09}$ & $1.94\pm_{0.93}$ & $79.52\pm_{0.010}$ & $1.88\pm_{1.03}$ & $0$ \\ 
%\textcolor{black}{Eff-CapsNet+REM ($\psi$)} & $88.85\pm_{0.06}$ & $1.09\pm_{0.94}$ & $78.69\pm_{0.015}$ & $1.05\pm_{0.94}$ & $47.91$\\ \midrule
\textcolor{black}{DR-CapsNet+Q ($\phi$)} & $90.51\pm_{0.29}$ & $6.25\pm_{1.15}$ & $77.40\pm_{0.64}$ & $5.01\pm_{1.45}$ & $0$ \\ 
\textcolor{black}{DR-CapsNet+REM ($\phi$)} & $90.38\pm_{0.53}$ & $3.35\pm_{1.18}$ & $76.98\pm_{0.45}$ & $2.47\pm_{0.96}$ & $50.13$ \\ \midrule
\textcolor{black}{DR-CapsNet+Q ($\psi$)} & $87.44\pm_{0.51}$ & $5.42\pm_{1.34}$ & $72.29\pm_{0.57}$ & $5.02\pm_{1.07}$ & $0$ \\
\textcolor{black}{DR-CapsNet+REM ($\psi$)} & $87.35\pm_{0.58}$ & $3.38\pm_{1.35}$ & $71.89\pm_{0.65}$ & $2.75\pm_{1.44}$ & $58.61$ \\\midrule
\textcolor{black}{$\gamma$-CapsNet+Q ($\phi$)} & $89.62\pm_{0.51}$ & $1.78\pm_{0.82}$ & $75.54\pm_{0.52}$ & $2.72\pm_{1.08}$ & $0$ \\
\textcolor{black}{$\gamma$-CapsNet+REM ($\phi$)} & $89.34\pm_{1.53}$ & $1.46\pm_{0.75}$ & $74.40\pm_{0.62}$ & $2.05\pm_{0.80}$ & $47.50$ \\\midrule
\textcolor{black}{$\gamma$-CapsNet+Q ($\psi$)} & $85.98\pm_{0.43}$ & $1.87\pm_{0.58}$ & $71.33\pm_{1.22}$ & $2.55\pm_{0.89}$ & $0$ \\ 
\textcolor{black}{$\gamma$-CapsNet+REM ($\psi$)} & $85.26\pm_{1.24}$ & $1.52\pm_{0.42}$ & $71.12\pm_{0.025}$ & $2.17\pm_{1.61}$ & $49.61$ \\ \midrule

\textcolor{black}{DeepCaps+Q ($\phi$)} & $95.01\pm_{0.58}$ & $7.32\pm_{1.18}$ & $83.18\pm_{1.61}$ & $7.28\pm_{1.66}$ & $0$ \\ 
\textcolor{black}{DeepCaps+REM ($\phi$)} & $94.62\pm_{0.52}$ & $6.75\pm_{1.41}$ & $82.49\pm_{1.53}$ & $6.12\pm_{1.91}$ & $34.45$ \\ \midrule
\textcolor{black}{DeepCaps+Q ($\psi$)} & $90.16\pm_{0.27}$ & $6.53\pm_{1.77}$ & $79.36\pm_{0.76}$ & $5.14\pm_{1.45}$  & $0$ \\
\textcolor{black}{DeepCaps+REM ($\psi$)} & $90.13\pm_{0.13}$ & $5.55\pm_{1.49}$ & $78.66\pm_{1.21}$ & $3.92\pm_{1.36}$ & $36.06$ \\ \midrule
\textcolor{black}{Eff-CapsNet+Q ($\phi$)} & $97.81\pm_{0.45}$ & $1.10\pm_{0.42}$ & $85.03\pm_{1.01}$ & $1.33\pm_{0.53}$ & $0$ \\ 
\textcolor{black}{Eff-CapsNet+REM ($\phi$)} & $98.08\pm_{0.25}$ & $0.26\pm_{0.14}$ & $86.50\pm_{0.31}$ & $0.25\pm_{0.11}$ & $75.09$ \\ \midrule
\textcolor{black}{Eff-CapsNet+Q ($\psi$)} & $97.07\pm_{0.52}$ & $1.69\pm_{0.28}$ & $87.95\pm_{1.67}$ & $1.71\pm_{0.34}$ & $0$ \\ 
\textcolor{black}{Eff-CapsNet+REM ($\psi$)} & $97.31\pm_{0.31}$ & $0.38\pm_{0.15}$ & $88.48\pm_{1.38}$ & $0.33\pm_{0.12}$ & $68.85$\\ 
\bottomrule                      
\end{tabular}
\caption{\centering\label{tab:novelviewpoints}Accuracy (\%) and entropy values on the smallNORB test set on familiar and novel viewpoints (elevations $\phi$ and azimuths $\psi$) seen and unseen during training respectively.}
\end{table}
We also show in Table~~\ref{tab:novelviewpoints_full_original} the performances of these networks without quantization. All the models are trained with our implementations when the source code is unavailable.
\begin{table}[H]
\footnotesize
\centering
\begin{tabular}{lccc}
\toprule
\multicolumn{1}{c}{\textbf{Model}} & \textbf{Familiar} & \textbf{Novel} & \textbf{Sparsity} \\ \midrule
\textcolor{black}{DR-CapsNet ($\phi$)} & $90.62\pm_{0.21}$ & $77.51\pm_{0.43}$ & $0$ \\ 
\textcolor{black}{DR-CapsNet ($\phi$)} & $90.51\pm_{0.45}$ & $77.03\pm_{0.38}$ & $50.13$ \\ \midrule
\textcolor{black}{DR-CapsNet ($\psi$)} & $87.90\pm_{0.50}$ & $72.37\pm_{0.43}$ & $0$ \\
\textcolor{black}{DR-CapsNet ($\psi$)} & $86.81\pm_{0.61}$ &  $71.99\pm_{0.66}$ & $58.61$ \\\midrule
\textcolor{black}{$\gamma$-CapsNet ($\phi$)} & $90.15\pm_{0.39}$ & $75.89\pm_{0.53}$ & $0$ \\
\textcolor{black}{$\gamma$-CapsNet ($\phi$)} & $89.92\pm_{0.91}$ & $74.96\pm_{0.71}$  & $47.50$ \\\midrule
\textcolor{black}{$\gamma$-CapsNet ($\psi$)} & $86.11\pm_{0.63}$ & $72.55\pm_{0.83}$  & $0$ \\ 
\textcolor{black}{$\gamma$-CapsNet ($\psi$)} & $85.35\pm_{0.61}$ &  $71.35\pm_{1.31}$  & $49.61$ \\ \midrule

\textcolor{black}{DeepCaps ($\phi$)} & $95.32\pm_{0.48}$ &  $83.13\pm_{0.91}$ & $0$ \\ 
\textcolor{black}{DeepCaps ($\phi$)} & $94.48\pm_{0.31}$ &  $82.42\pm_{1.53}$ & $34.45$ \\ \midrule
\textcolor{black}{DeepCaps ($\psi$)} & $91.11\pm_{0.27}$ &  $79.53\pm_{0.78}$  & $0$ \\
\textcolor{black}{DeepCaps ($\psi$)} & $90.15\pm_{0.91}$ & $78.83\pm_{1.12}$ & $36.06$ \\ \midrule
\textcolor{black}{Eff-CapsNet ($\phi$)} & $97.83\pm_{0.41}$  & $85.04\pm_{1.06}$ & $0$ \\ 
\textcolor{black}{Eff-CapsNet ($\phi$)} & $98.10\pm_{0.33}$ &  $86.50\pm_{0.35}$ & $75.09$ \\ \midrule
\textcolor{black}{Eff-CapsNet ($\psi$)} & $97.07\pm_{0.52}$  & $87.98\pm_{1.64}$ & $0$ \\ 
\textcolor{black}{Eff-CapsNet ($\psi$)} & $97.29\pm_{0.34}$ &  $88.54\pm_{1.35}$ & $68.85$ \\ 
\bottomrule                      
\end{tabular}
\caption{\centering\label{tab:novelviewpoints_full_original}Accuracy (\%) on the smallNORB test set on familiar and novel viewpoints (elevations $\phi$ and azimuths $\psi$) for DR-CapsNets, $\gamma$-CapsNets, DeepCaps and Eff-CapsNets without quantization.}
\end{table}

\begin{figure}
\centering
\begin{subfigure}{0.8\columnwidth}
  \centering
  \includegraphics[width=1\textwidth]{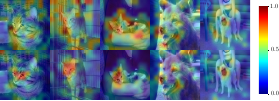}
  \caption{\centering{Dogs vs. Cats.}}
  \label{fig:attention_matr_dogscats}
\end{subfigure}\\
\begin{subfigure}{0.8\columnwidth}
  \centering
  \includegraphics[width=1\textwidth]{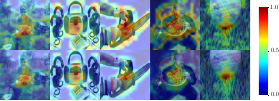}
  \caption{\centering{Imagenette.}}
  \label{fig:attention_matr_imagenette}
\end{subfigure}%
\caption{Saliency maps for Eff-ConvCapsNets without and with REM, first and second rows, respectively.}
\label{fig:saliency_maps}
\end{figure}

\textbf{Improved visualizations with REM.}
% \enzo{inserire qui le visualizzazioni, e potenzialmente i vari parse trees e discussione su essi, facendo vedere che in generale le capsule si concentrano sull'oggetto, e non solo- se possibile, far vedere ogni primary capsule su quale paret dell'oggetto si concentra!}
Figures~\ref{fig:attention_matr_dogscats} and  \ref{fig:attention_matr_imagenette} show the saliency maps overlayed on Dogs vs. Cats and Imagenette images. We chose these datasets for the saliency maps visualization method since the images come with higher resolution and more variation than the other datasets.
% \begin{figure}
% \centering
%          \includegraphics[width=1\columnwidth]{Parts/02_DemystifyngCapsNetsComponents/gfx/rem/CIFAR-10_heatmaps_10.pdf}
% \caption{\textcolor{black}{Saliency maps for CIFAR-10 for CapsNet+Q (above) and CapsNet+REM (below).}}
% \label{fig:attention_matr}
% \end{figure}
When REM is applied (second rows), we can see that the network predominantly concentrates on the object of interest, dismissing extraneous background noise or irrelevant objects. This characteristic of REM leads to the extraction of more concise part-whole hierarchies. As illustrated in Figure~\ref{fig:attention_matr_dogscats}, only the discriminative features specific to a cat or a dog are prominently highlighted in red. In contrast, features corresponding to people or cages play a diminished role in the classification process. Similarly, in Figure~\ref{fig:attention_matr_imagenette}, the model focuses mostly on the tench, the cassette player, the chain saw, the French horn, and the golf ball, disregarding trees, headphones, hands, the man, and grass.

\begin{figure}
\centering
         \includegraphics[width=1\columnwidth]{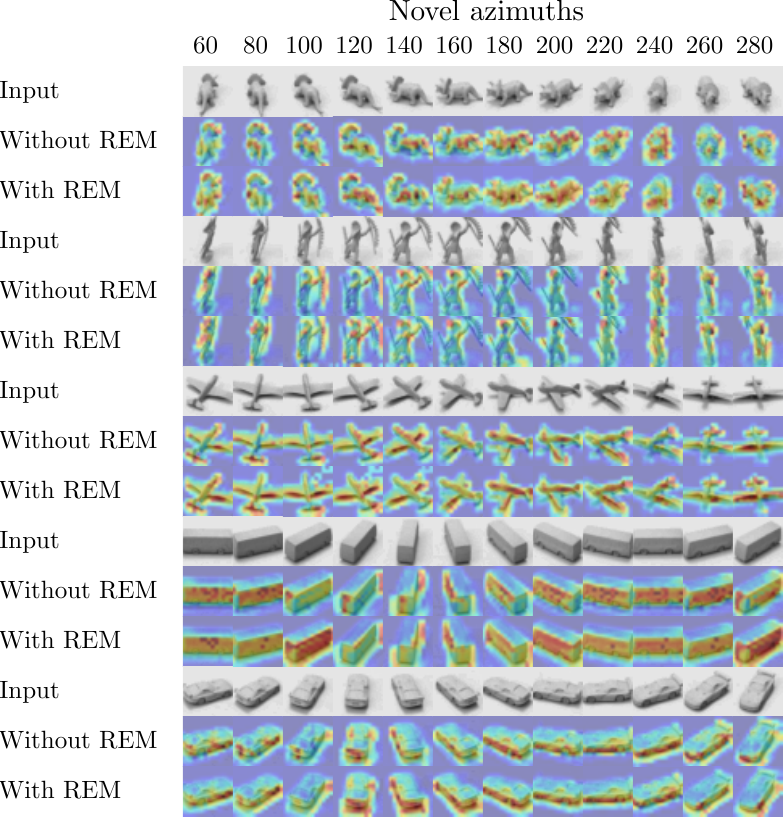}
\caption{\textcolor{black}{Saliency maps for smallNORB for DR-CapsNet+Q and DR-CapsNet+REM (novel azimuths).}}
\label{fig:saliency_maps_sn_az}
\end{figure}

\begin{figure}
\centering
         \includegraphics[width=0.7\columnwidth]{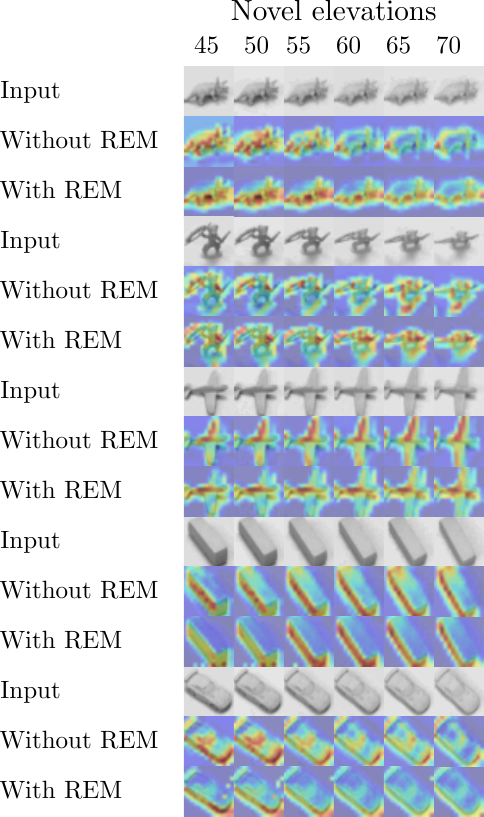}
\caption{\textcolor{black}{Saliency maps for smallNORB for DR-CapsNet+Q and DR-CapsNet+REM (novel elevations).}}
\label{fig:saliency_maps_sn_el}
\end{figure}

In Figures~\ref{fig:saliency_maps_sn_az} and \ref{fig:saliency_maps_sn_el} we show the saliency maps for smallNORB when tested on novel azimuths and elevations, respectively. We used the method described in Appendix~\ref{sec:rem-extract_salmap}. We show one object instance for each category. We can see from these visualizations on smallNORB that the differences between saliency maps of REM and without REM are less prominent. This is because smallNORB is less complex than Fashion-MNIST or CIFAR-10. However, we can still notice that some of the saliency maps extracted with REM are more prominent and consistent between each category.

\begin{figure}
\centering
         \includegraphics[width=0.8\columnwidth]{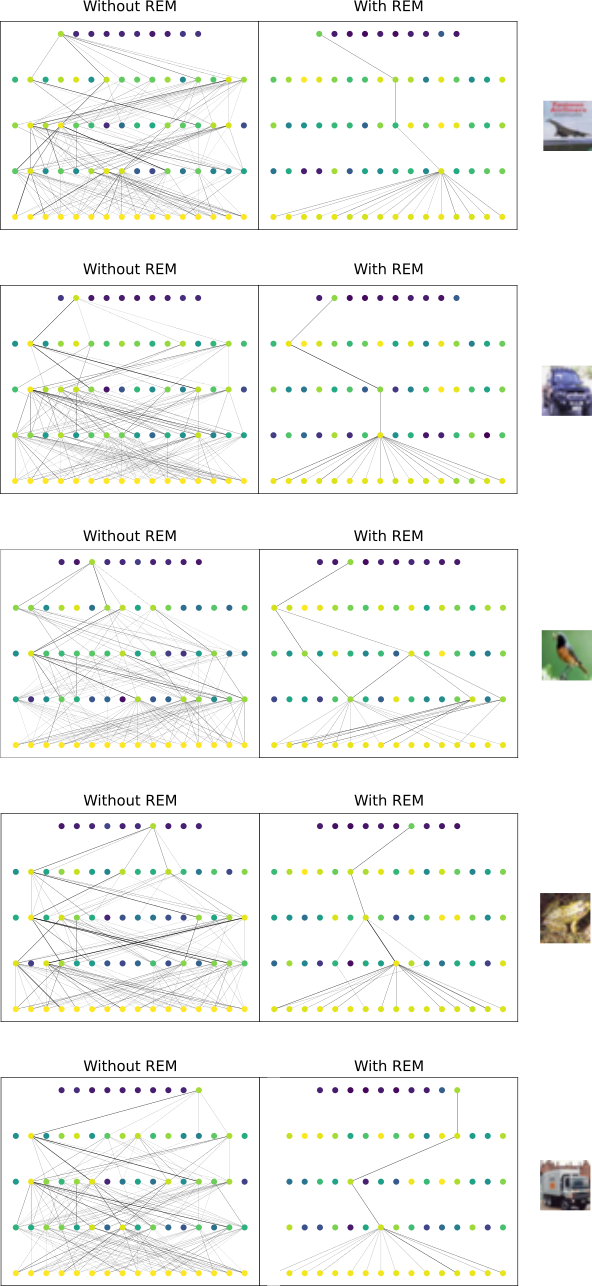}
\caption{Parse trees for some CIFAR-10 test images for multilayered DR-CapsNet and DR-CapsNet+REM with backtracking.}
\label{fig:parse_trees_cifar10_samples}
\end{figure}

Figure~\ref{fig:parse_trees_cifar10_samples} shows the parse trees extracted from a fully-connected DR-CapsNet for some CIFAR-10 images of the test set. 
 We can see that fewer capsules are part of the parse tree extracted with REM: the network relies on fewer object parts to detect the objects.

\section{Summary}
\label{sec:rem-conlcusion}
 This paper introduced REM, a technique that drives the model parameters distribution toward low entropy configurations. With REM, capsule networks do not need to model irrelevant objects or noise in the images to achieve high generalization. Therefore, even a reasonable sized capsule network can now learn more discriminative features. A novel visualization method for capsule networks is employed to show which features play a major role in the classification process. Extensive experiments on different datasets and architectures confirm the effectiveness of the proposed approach.
%  We first showed how to extract the parse tree of a capsule network by discretizing its connections and then collecting the possible parse trees in associative arrays.
% Capsule networks show high entropy in the parse tree structures, as an explicit prior on the distribution of coupling coefficients is missing. Indeed, the number of intra-class generated parse trees is relatively high. We showed how pruning in low pruning regimes naturally reduces such entropy and the cardinality over the possible parse trees, testing such a phenomenon on several datasets. We also showed that REM could carve out parse trees with stronger part-whole relationships, achieving high generalization. Furthermore, we empirically observe that with REM, models remain robust to affine transformations and novel viewpoints. 
Recent works show that capsule networks do not scale~\citep{mitterreiter2023capsule} and suffer vanishing activations when adding many capsule layers~\citep{everett2023vanishing}, leading to poor performance. Therefore, carving out an ideal parse tree from complex images with perfectly interpretable capsules is not possible with capsule networks, yet. However, REM opens research pathways toward the distillation of parse trees and model interpretability, including the design of a pruning technique specifically designed for capsules. %Future works include the study of the distilled parse tree structures and scaling up the CapsNets+REM to more complex tasks in computer vision, beyond image classification.  \riccardo{frase conclusiva lavori futuri?...}

\cleardoublepage
\ctparttext{
In the previous parts, we presented the fundamentals of capsule networks along with their extensions, primarily focusing on image classification. However, it's important to note that the utility of capsule networks extends well beyond image classification. Indeed, their versatile architectures can be adapted to tackle various tasks, such as image similarity matching and segmentation. This part delves into the applications of capsule networks beyond the realm of image recognition. Specifically, we will delve into the following intriguing applications of capsule networks:
\begin{enumerate}
    \item \textbf{Localization of unmanned aerial vehicles without GPS reliance}: By harnessing satellite and drone images, capsule networks can play a pivotal role in extracting feature vectors that facilitate the verification of whether a pair of satellite and drone images share the same geographical region. This innovation offers a novel unmanned aerial vehicle localization approach that doesn't rely on traditional GPS systems.

    \item \textbf{Encoding rotations of synthetic datasets with quaternions}: Many applications such as robot grasping need to know the exact position and orientation of an object. However, real-world datasets do not always include these informations. We generate Blender images of various meshes displayed from different angles, labeled with object classes and rotations represented using quaternions, which offer advantages over rotation matrices. We explore the idea of using capsule networks to encode quaternions, building on previous work that combined quaternions and capsules to capture object poses effectively. The ultimate goal is to leverage synthetic datasets annotated with quaternion labels to generalize to real-world datasets. 

    \item \textbf{Lung nodules segmentation}: Capitalizing on the inherent strengths of capsule architectures, it becomes feasible to expand them into both encoder and decoder components. This extension proves invaluable in the task of lung nodule segmentation. The architecture can classify each pixel within an image, differentiating between regions belonging to a nodule and those not. This advancement holds substantial promise for enhancing medical imaging diagnostics and analysis.
\end{enumerate}
Incorporating capsule networks into these diverse applications underscores their adaptability and potential to revolutionize various fields beyond the confines of image classification.}
\part{Capsule Networks Applications}\label{part:capsnets-applications}
%*****************************************
\chapter{LSVL: Large-scale season-invariant visual localization for UAVs
}\label{ch:lsvl}
%*****************************************
The research findings and insights presented in this chapter are the results of a visiting period at Aalto University (Helsinki, Finland), supervised by Prof. Ville Kyrki. The collaboration led to the following publication: \fullcite{kinnari2022lsvl}.
\section{Introduction}
The ability of a Unmanned Aerial Vehicle (UAV) to robustly estimate its position is one of the basic requirements of autonomous flight. In missions in which the UAV needs to collaborate with other agents operating in the same environment, information of position in a shared, global frame of reference is needed.

There are several possible ways to estimate position. A localization solution can rely on infrastructure built for this purpose. In outdoor UAV operations, by far the most common is to rely on Global Navigation Satellite Systems (GNSS). In ideal conditions, GNSS receivers provide measurements of position. However, GNSS, as well as other radio beacon systems, are susceptible to spoofing and jamming attacks \citep{7445815, RR-2970-DHS}, which may be used for denying operation of UAVs in an area.

A sensor system commonly carried by UAVs is the combination of a camera and an  Inertial Measurement Unit IMU. Several works have focused on using this sensor combination for resolving the position of the UAV by tracking the difference to a known starting position by methods of VIO \citep{Scaramuzza2020}. As VIO implementations integrate noisy signals, they suffer from drift over long flights \citep{8460664}. Simultaneous Localization and Mapping (SLAM) methods \citep{7747236} can help in partially compensating for this drift by detecting loop closures, but only if the mission contains reentry to a previously visited area. In addition, visual-inertial odometry VIO and SLAM approaches are not robust to random failures in position tracking; temporary failures may lead to loss of position information in the global frame with only a small chance of recovery.

If a map is available, matching camera images acquired by a UAV to the map allows compensating for drift induced by odometry methods and provides an estimate of position with respect to the map. This approach is called visual localization \citep{10.1145/3281548.3281556, COUTURIER2021103666}. Using a georeferenced map allows, in principle, finding the position of a UAV with respect to the map, even in the case of no prior information on position at start of mission or after random failures in positioning. 

There are, however, multiple challenges in visual localization. Appearance difference between the UAV image and the map may be significant due to changes in season --- an image acquired by a UAV in winter looks very different from a satellite image acquired in summer. Moreover, the environment in which the UAV is operating may be naturally ambiguous, \eg{} when flying over vast areas of forests. In order for the localization solution to support recovery from random localization failures, the size of the map must correspond to the operating area of the UAV. This means that the localization solution must be computationally efficient enough to run on an onboard computer even on large maps, and it must tolerate the natural ambiguity.

\begin{figure}
     \centering
     \includegraphics[width=\linewidth]{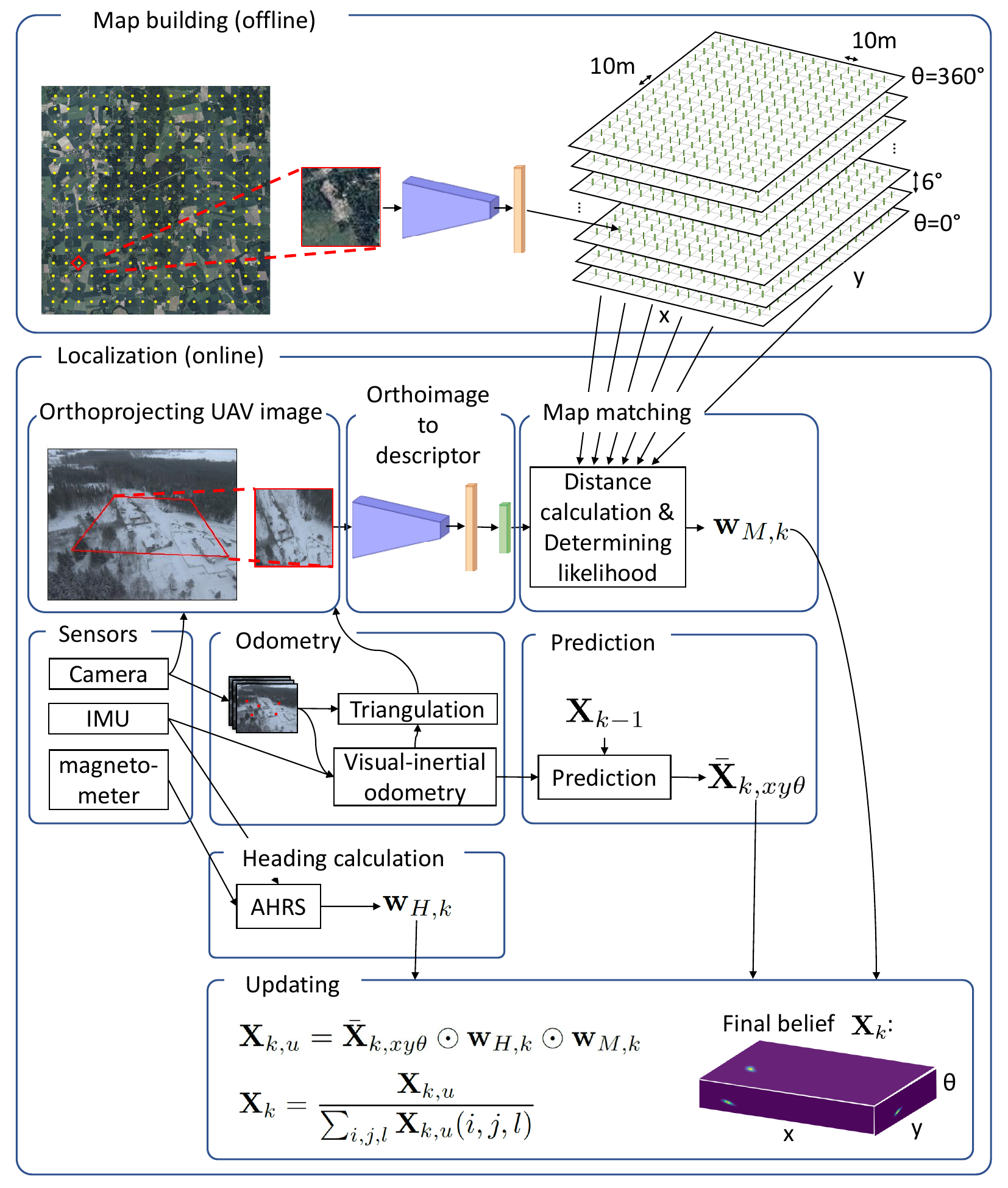}
     \caption{Block diagram of proposed localization solution. We train a neural network that computes embedding vectors. We precompute a map of embeddings offline for a grid of position and orientation hypotheses that cover the full considered state space. During flight, we run a localization algorithm that tracks UAV egomotion by VIO and computes an embedding of an orthoprojected section of the UAV image. We predict belief over time steps in accordance with VIO, and update belief based on likelihood computed from distance in embedding space between image observation and map hypotheses. We optionally include heading measurements.}
     \label{fig:lsl-first_page_image}
\end{figure}

In this work we propose LSVL, which addresses all of the above mentioned challenges of robust visual localization at large scale. The main contributions of this work are:
\begin{enumerate}
    \item We propose an approach for localizing a UAV which is able to handle natural ambiguities and determine the location of the UAV even when no prior information is available about its starting pose over very large areas. Our extensive testing demonstrates that, starting from an uncertainty corresponding to \maparea{} area, our approach can converge to an average translation error of \translerrorafterconvergence{} after no more than \numberofstepstoconvergence{} updates using UAV camera observations, under significant seasonal appearance difference between UAV observation and map, when flying over terrains containing natural ambiguity. Importantly, our approach does not rely on a digital elevation model and can operate effectively using only 2D maps.
    \item We propose a UAV image to map matching solution based on compact descriptors learned in a manner that provides invariance to seasonal appearance difference. We train the descriptor networks using only satellite images, without requiring any labeled data such as semantic classification of the terrain. The use of compact descriptors enables fast computation of a large number of hypotheses.
    \item We present the problem of UAV visual localization as state estimation utilizing a point mass filter, and integrate it to our map matching approach. This combination allows running our algorithm in real time, with constant time and memory consumption onboard a UAV at sufficiently large scale for UAV missions.
    \item We explore different architecture choices for image description, vector dimensionality, and likelihood computation, and evaluate their impact on probability to convergence, time to convergence and positioning error after convergence. We compare our solution to two \sota{} approaches for UAV localization, and demonstrate the operation of our localization solution in real time onboard a commercial UAV.
    \item We present a simple method allowing online assessment of integrity of pose estimate, enabling self-diagnostics of the localization solution, which is a key component in the robust global localization problem.
\end{enumerate}
We believe this is the first work in the visual localization area that is able to find true position of the UAV, starting from a scale of uncertainty of \maparea{}, over ambiguous terrains, under significant seasonal appearance change. A block diagram of our algorithm is shown in Figure~\ref{fig:lsl-first_page_image}.

The paper is structured as follows. Section~\ref{sec:lsl-related_work} goes over related work. Section~\ref{sec:lsl-preliminaries} defines preliminaries of the localization problem. We introduce our method in Section~\ref{sec:lsl-method} and detail a vital part of the solution, map matching, in Section~\ref{sec:lsl-map_matching}. Section~\ref{sec:lsl-experiments} describes our localization experiments and we conclude the paper with discussion and conclusions in Section~\ref{sec:lsl-discussion} and Section~\ref{sec:lsl-lsl-conclusions}, respectively.

\section{Related work}
\label{sec:lsl-related_work}
Visual localization of UAVs is a topic which has attracted interest especially over past few years. The \sota{} is covered by relatively recent surveys \citep{10.1145/3281548.3281556, COUTURIER2021103666}. Within the area of visual localization, most works are limited by one or more of the following assumptions: accurate initialization is required \citep{6942633,9196606,9357892,8793558}, the size of the operating area is limited \citep{7418753,9636705}, the movements of the UAV are constrained to specific paths \citep{9357892}, the operation takes place in conditions that are very close to map in terms of appearance \citep{6942633,7418753}, the map resolution and detail requirement is significant,  such as requiring a topography map \citep{9636705}, or a very high flying altitude is required for successful georeferencing \citep{9170807}. The following overview of related work focuses on the works that---similarly to LSVL---do not require of knowledge of initial pose and that use an easily obtainable planar 2D map for localization.

A common choice in UAV localization is to detect semantic features such as roads and intersections \citep{Dumble2015AirborneVN,volkova2018more,9184263,9170807} or buildings \citep{9341682}. Choi \etal \citep{9341682} proposed a method where the UAV image is semantically segmented to find buildings in the camera view. Based on detected buildings, a rotation invariant descriptor, building ratio map (BRM), is computed from the proportion of building pixels visible in camera view, and a precomputed map with similar descriptors is used for localization. The authors demonstrated convergence of position estimate on a 6.17 $km^2$ map after 27 updates with 25 meters of translation between each update, with 12.01 m root-mean-square (RMS) error after convergence. The demonstration flight takes place over a residential area. The main drawback of this approach is the requirement that features of a specific semantic class (\ie{} buildings) are required for successful localization, which may be unavailable when flying over natural environments.

\citet{MANTELLI2019304} demonstrate localizing a UAV on a map of size 1.34 \km2{} using a particle filter, where particle likelihood is determined based on a handcrafted descriptor called abBRIEF. The authors initialize the particle filter with 50 000 particles and show robustness of their localization solution in comparison to BRIEF \citep{calonder2010brief} descriptors on trajectory lengths up to 2.4 km, showing convergence in less than 50 m to an average translation error of 17.78 m. The descriptor is developed such that it is tolerant to illumination changes and allows fast computation of particle likelihood over a large number of pose hypotheses.

We evaluate against BRM and abBRIEF and show superior performance.

We build on the idea of compressing UAV camera observation into a compact embedding space, to allow fast testing of pose hypotheses on a large map. This has been proposed recently by \eg{} Bianchi \etal{} \citep{9357892} who use a bottleneck autoencoder approach to compress visual observations to descriptor vectors of dimension 1000. Also Samano \etal{} \citep{9562005} and Couturier \etal{} \citep{10.1117/12.2585986} train Resnet models \citep{7780459} to project map tiles and UAV images to a low-dimensional (16D) embedding space. Only Samano \etal \citep{9562005} allow movements of the UAV outside precomputed paths by precomputing a grid of embedding vectors, on which hypothesis testing is performed by interpolating a vector from this grid by each pose hypothesis and measuring distance of observed image descriptor to interpolated descriptor vector in embedding space. Authors of \citep{9562005} demonstrate with simulated flight experiment the convergence of pose estimate to less than 95 m translation error in 78.2\% of simulated flights by 200 updates. We take a similar approach but forgo interpolation in embedding space, to avoid unjustified implicit assumption of smoothness of embedding space, and explore other learned descriptor architectures. Compared to \citep{9562005}, we show significantly faster convergence, tolerance to seasonal variation and we demonstrate performance with real experiments.

The topic of localization over large areas under visual appearance change occurs also under the topics of  visual place recognition (VPR) \citep{7339473} and SLAM (\eg{} \citet{7353721}). In VPR, the problem of localization is commonly solved as an image retrieval task (\eg{} \citet{Hausler2021PatchNetVLADMF}) or as refinement with respect to a preconstructed Structure-
from-Motion (SfM) map of an area \citep{Sarlin_2019_CVPR}; in SLAM, the sensor system of the robot is used for building the map. In contrast, we propose a localization solution that does not require having previously visited or acquired imagery in the area of operation. Instead, we focus on the problem of localization with respect to an orthophoto map. In this way, using only low-cost orthoimage maps, the size of the region of operations is not constrained by cost of data acquisition for map building.

Some variations due to weather can be removed by denoising \citep{jimaging7030052, Tang2023}. Our work does not target removal of effects of adverse weather on the imaging process, but methods of denoising could be combined as a preprocessing step to our work.

\section{Preliminaries}
\label{sec:lsl-preliminaries}

We want to resolve the pose of an UAV using measurements available at the UAV during flight, with no dependency on localization or communication infrastructures. Formally, we define the pose of the UAV pose in a common, global reference frame as the state
\begin{equation}
\continuousX_k = \begin{bmatrix}x_k & y_k & \theta_k\end{bmatrix}^T
\end{equation}
where $x$, $y$, and $\theta$ represent longitude, latitude, and heading, respectively, in a Cartesian coordinate system and $k$ is index for time. We consider localization of the UAV in a limited region in longitude and latitude: we assume $x_k \in [x_{min}, x_{max}]$, $y_k \in [y_{min}, y_{max}]$, $\theta_k \in [0, 2\pi)$. The localization problem is to compute the marginal posterior distribution $\cond{\continuousX_k}{\bigY{1:k},\mathcal{M}}$ of state $\continuousX_k$, given history of measurements $\bigY{1:k}$ and map $\mathcal{M}$.

In this work, we concentrate on the \emph{wake-up robot problem}, \ie{} at the start of a flight, we assume an uniform prior distribution $p(\continuousX_0)$ over all values of $x_k$ and $y_k$, across all values of heading. This represents a situation where the only initial information about the UAV position is that it is located within the area of a map defined over a rectangular area in latitude and longitude. Our formulation allows inclusion of more informed initialization.

We take the typical approach of considering localization on preacquired map as a Bayesian filtering problem (see \eg{} \citep{thrun2002probabilistic}). This amounts to maintaining a representation of belief of current state and updating that representation when new measurements are available. At each time step $k$, we obtain three types of measurements from the onboard sensors of the UAV: an odometry measurement $\mathbf{u}_k$, an heading measurement $v_k$, and an image from the UAV camera $\mathcal{I}_k$ to be used for map matching. Given the measurment $\bigY{k} = \begin{Bmatrix}
\mathbf{u}_k, v_k, \mathcal{I}_k
\end{Bmatrix}$, at each sampling time $k$, we first perform a prediction step based on the odometry measurement:
\begin{equation}
\label{eq:prediction}
    \cond{\continuousX_k}{\mathbf{u}_k,\bigY{1:k-1}} =
    \int \cond{\continuousX_k}{\mathbf{u}_k,\continuousX_{k-1}} \cond{\continuousX_{k-1}}{\bigY{1:k-1}} d\continuousX_{k-1}
\end{equation}
We then use the heading and map matching measurements as our observation model:
\begin{equation}
\label{eq:state_estimate_continuous}
\cond{\continuousX_k}{\bigY{1:k},\mathcal{M}} = \frac{1}{\eta_k} \cond{\continuousX_k}{\mathbf{u}_k,\bigY{1:k-1}} \cond{v_k, \mathcal{I}_k}{\continuousX_k,\mathcal{M}}
\end{equation}
where $\eta_k$ is a normalizing constant. The heading and map matching measurements are considered independent, and likelihood of heading measurement is not conditional on map:
\begin{equation}
\cond{v_k, \mathcal{I}_k}{\continuousX_k,\mathcal{M}} = \cond{v_k}{\continuousX_k}\cond{\mathcal{I}_k}{\continuousX_k,\mathcal{M}}
\end{equation}

\section{Method}
\label{sec:lsl-method}

In order to find the pose on a large map in presence of matching ambiguities presented in the introduction, a method is needed for utilizing a sequence of as many UAV images as needed in order to converge to a single, correct pose estimate. We propose a recursive \emph{localization method} consisting of the following components, illustrated in Figure~\ref{fig:lsl-first_page_image}. We use an \emph{odometry measurement} for predicting belief of state $\bar{\boldx}_{k,xy\theta}$ at time $k$ based on state $\boldx_{k-1}$ at previous time instant $k-1$. We use a \emph{map matching measurement} for computing likelihood of pose hypotheses $\weightmatrix_{M,k}$ based on a single UAV image and, optionally, a \emph{heading measurement} for computing the likelihood of pose hypotheses $\weightmatrix_{H,k}$ based on a compass heading measurement. The localization method updates belief of state using all the measurements, providing $\boldx_{k}$, belief of state at time $k$.

In this section we describe each of these components in detail, putting particular attention in describing how the proposed solutions target the complexity that arises from both dealing with a large map and with considerable appearance change, central limitations of current methods that we address in this work. Section~\ref{sec:lsl-measurements} lists all measurements we use, and Section~\ref{sec:lsl-localization_method} specifies our localization method.

\subsection{Measurements}
\label{sec:lsl-measurements}
\subsubsection{Odometry measurement}
\label{sec:lsl-odometry_measurement}

We assume that the UAV is running a VIO algorithm which provides a measurement of incremental motion in six degrees of freedom. We marginalize over roll and pitch, which are directly observable in the global frame \citep{5959226}, and altitude, which we abstract with the map matching approach presented in Section~\ref{sec:lsl-map_matching}. The odometry algorithm provides, at time $k$, the measurement $\mathbf{u}_k = \begin{bmatrix}u_{k,x} & u_{k,y} & u_{k,\theta} & u_{k,o}\end{bmatrix}^T$, a measurement of translation, rotation and distance traveled since time $k-1$. $u_{k,x}$ and $u_{k,y}$ are translation since time instant $k-1$ in $x$ and $y$ coordinates, respectively, with respect to pose at time $k-1$. Similarly, $u_{k,\theta}$ is rotation around vertical axis. $u_{k,o}$ is the integral of distance traveled since instant $k-1$ according to odometry. The measurements are visualized in Figure~\ref{fig:lsl-odometry_measurements}.

We approximate the odometry pose uncertainty with a multivariate normal distribution. We further assume the covariance of pose is isotropic in $x$, $y$ directions and that rotation noise is independent from translation noise. More formally, we approximate the posterior density with a multivariate normal distribution with covariance $\mathbf{\Sigma}_u$. This gives the motion model

\begin{equation}\label{eq:2d_odometry_model}
\begin{bmatrix}
    x_{k+1} \\
    y_{k+1} \\
    {\theta}_{k+1}
\end{bmatrix}
=
\begin{bmatrix}
    x_{k} \\
    y_{k} \\
    {\theta}_{k}
\end{bmatrix}
+
\begin{bmatrix}
    cos {\theta}_k & -sin {\theta}_k & 0 \\
    sin {\theta}_k & cos {\theta}_k & 0 \\
    0 & 0 & 1
\end{bmatrix}
\bar{\mathbf{u}}_k
+
\mathcal{N}(\mathbf{0},\mathbf{\Sigma}_{u}(u_{k,o}))
\end{equation}
where the mean of the measurement
$\bar{\mathbf{u}}_k = \begin{bmatrix}
    \bar{u}_{x} &
    \bar{u}_{y} &
    \bar{u}_{{\theta}}   
\end{bmatrix}^T$
is specified in the odometry frame at time $k$ and $\mathbf{\Sigma}_{u}(u_{k,o})$ specifies the covariance of the measurement, and $\mathbf{\Sigma}_u(t)$ is a diagonal matrix:
\begin{equation}
    \mathbf{\Sigma}_u(t) = diag(\sigma_{u,xy}(t)^2, \sigma_{u,xy}(t)^2, \sigma_{u,\theta}(t)^2).   
\end{equation}
While the isotropic noise for translation and independent noise in heading are simplified approximations of the true distribution \citep{long2013banana}, we consider this a sufficient upper bound approximation to the odometry noise. This approximation enables a computationally fast method of prediction as elaborated in Section~\ref{sec:lsl-using_odometry_for_prediction}.

In addition, we assume the movements of the UAV produce sufficient IMU excitation to make scale observable or that the UAV is equipped with additional sensors with which scale can be resolved.

\subsubsection{Map matching measurement}

The purpose of map matching is to provide a means for verifying or disputing pose hypotheses, given camera image $\mathcal{I}_k$ and map $\mathcal{M}$. A method for computing the likelihood $\cond{\mathcal{I}_k}{\continuousX_k,\mathcal{M}}$ is thus needed. Our work focuses especially on finding means for computing the likelihood that work well over ambiguous terrains under significant seasonal appearance difference, in a computationally efficient way. Since the method used for map matching and likelihood computation based on a UAV image is a considerable part of our contribution, a separate section (Section~\ref{sec:lsl-map_matching}) has been given to the detailed description of this component.

\subsubsection{Heading measurement}
\label{sec:lsl-heading_measurement}

We assume the UAV is equipped with an heading ref-
erence system (AHRS), relying on fusion of a compass and an IMU, which provides a measurement of heading, $\theta$, with respect to map East, corrupted by Gaussian noise $n_\theta \sim \mathcal{N}(0,\sigma_{v}^2)$:
\begin{equation}
\label{eq:heading_measurement}
    v_k = v(t_k) = \theta(t_k) + n_{\theta}
\end{equation}

\subsection{Localization method}
\label{sec:lsl-localization_method}
We have to consider how to formulate the pose estimation problem in a computationally feasible way. In choosing the estimation approach, we need to consider characteristics of the problem. Specifically, as we start from a very uninformed state and expect natural ambiguities in the environment we are operating in, it is expected that we need to be able to track a large number of multiple hypotheses before converging to the correct pose, and before convergence, the state is expected to not conform to a parametric probability distribution model. This leads us to consider nonparametric Bayesian filtering approaches, and in the localization context, two commonly used solutions are particle filtering and point mass filtering \citep{thrun2002probabilistic}.

\subsubsection{Choosing representation for state}

The use of a particle filter in UAV localization is a common choice \citep{9562005, 7418753, MANTELLI2019304}. However, in cases of very uninformed initialization, a risk exists that a particle is initially not placed in vicinity of the true pose, leaving the probability of converging on the correct pose over time to chance. Furthermore, during a flight, it is possible that the close proximity of true state is left without sufficient particle density in cases such as when flying for long periods of time over ambiguous areas, or in case of large but local map inconsistencies (e.g. forest clearcutting having taken place between map acquisition and flight), leading to divergence in a poorly predictable way. Since the particle filter is stochastic, different instantiations of the filter using the same data may provide different results and, depending on selected resampling scheme, computation time may vary.

Instead of representing belief through the use of particles, we choose to use a point mass filter and compute the belief on a discrete grid. This selection ensures coverage of full state space throughout the mission, at the resolution specified by our choice of grid, and offers deterministic performance and constant computational time and memory requirement. Early adaptations of point mass filtering-based approaches on 2D robot localization include the work by \citet{10.5555/1864519.1864520} and in terrain navigation of flying platforms by \citet{bergman1999recursive}.

\subsubsection{Definition of state using point mass filter}
We approximate belief $p(\continuousX)$ on the continuous state space of $\continuousX$ by decomposing it into a grid of regions of equal size with resolution $r_x = r_y = r_{xy}$ in translation and $r_{\theta}$ in heading. Each region is a voxel $s(i,j,l) = \begin{bmatrix}s_x(i),s_y(j),s_\theta(l)\end{bmatrix}$ in the state space with bounds
\begin{equation}
\begin{aligned}
    s_{xv}(i) &\leq s_x(i) < s_{xu}(i) \\
    s_{yv}(j) &\leq s_y(j) < s_{yu}(j) \\
    s_{\theta v}(l) &\leq s_\theta(l) < s_{\theta u}(l)
\end{aligned}
\end{equation}
with lower bound $s_{bv}(i) = b_{min} + i \times r_{b}$ and upper bound $s_{bu}(i) = s_{bv}(i) + r_{b}$ for each axis $b \in \{x,y,\theta\}$ and $i, j, l \in \mathbb{N}$ such that the whole state space is covered.

We approximate belief over state at time instant $k$ as a piecewise constant probability matrix $\boldx_k \approx p(\continuousX_k)$
where each element of matrix $\boldx_k[i,j,l]$ assigns a probability to each voxel $s(i,j,l)$ in state space.

\subsubsection{Using odometry measurement for prediction}
\label{sec:lsl-using_odometry_for_prediction}
\begin{figure}
\centering
     \begin{subfigure}[t]{0.8\linewidth}
         \centering
         \includegraphics[width=\linewidth]{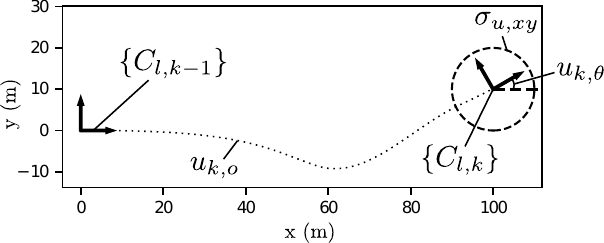}
         \caption{Odometry measurements are stated with respect to frame at previous update, \localcoordinateframe{k-1}. $u_{k,o}$ states distance traveled since previous update, according to odometry.}
         \label{fig:lsl-odometry_measurements}
    \end{subfigure}
    
    \begin{subfigure}[t]{\linewidth}
        \centering
        \includegraphics[width=0.7\linewidth]{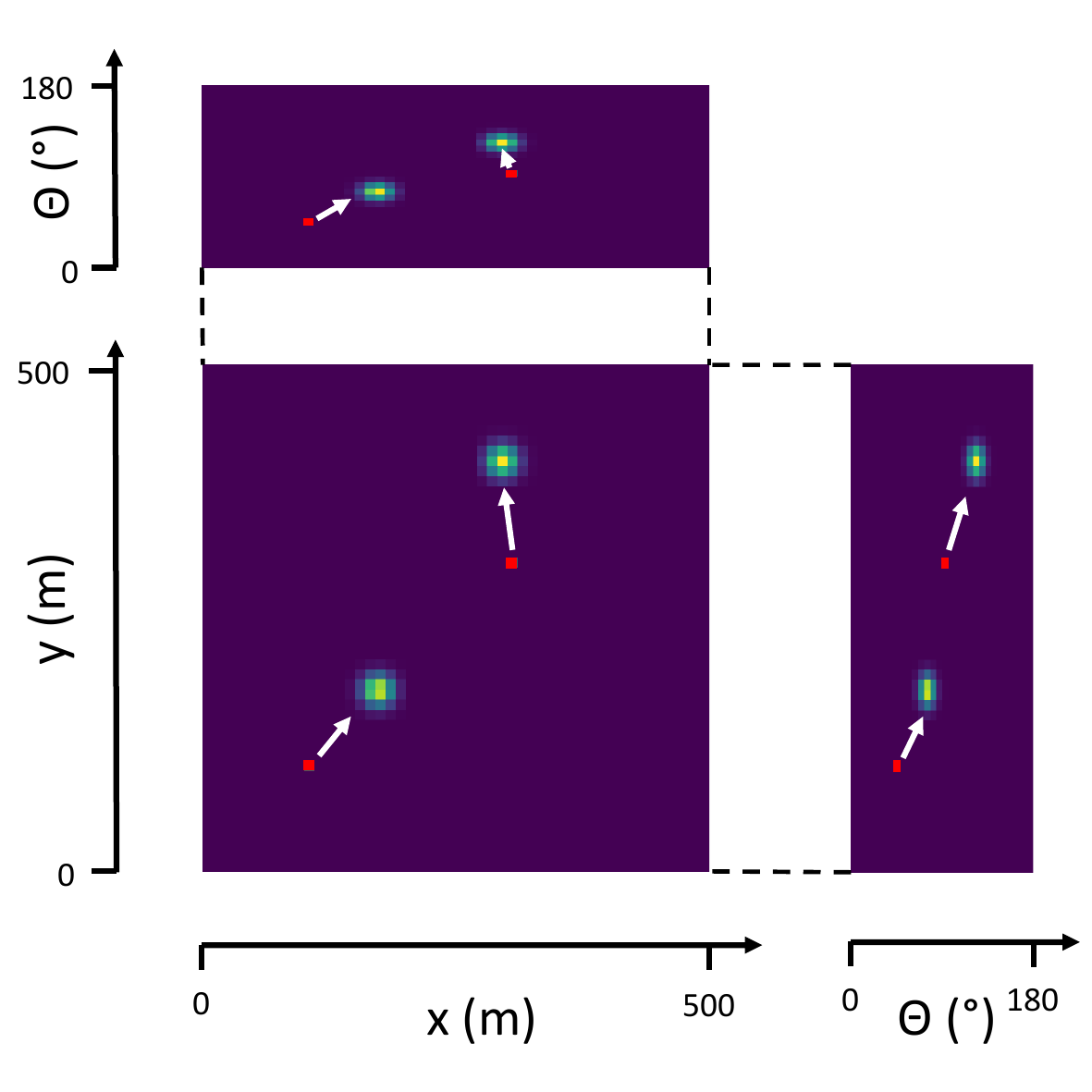}
        \caption{Part of $\boldx$ after marginalizing different axes individually. Two voxels at $(100,100,45\degree)$ and $(300,300,90\degree)$ representing belief before prediction are shown in red. Voxels highlighted in green and yellow represent belief after prediction. White arrows show how prediction using these odometry measurements with the assumed noise shifts and smooths $\boldx$ according to the measurement, and that shifts are performed in direction determined by odometry measurement and voxel's $\theta$ value.}
        \label{fig:lsl-x_marginalized_odometry}
    \end{subfigure}
    
    \caption{Example of prediction based on odometry. In this example, $u_{k,x}=100$ m, $u_{k,y}=10$ m, $u_{k,\theta}=30\degree$, $\sigma_{u,xy} = 10$ m, $\sigma_{u,\theta} = 5\degree$. Figure~\ref{fig:lsl-odometry_measurements} visualizes odometry measurements between times $k-1$ and $k$. Figure~\ref{fig:lsl-x_marginalized_odometry} visualizes how odometry measurements are used in prediction.}
    \label{fig:lsl-odometry_updating_example}
\end{figure}

To approximate the prediction step \eqref{eq:prediction} with our state representation, using the VIO measurement model presented in Section~\ref{sec:lsl-odometry_measurement}, we formulate a method in which the probability mass contained in a voxel in $\boldx_{k-1}$ is projected to other voxels in the belief grid at time $k$ as dictated by the odometry measurement. We repeat this operation for each voxel in the belief grid. An example visualization is shown in Figure~\ref{fig:lsl-odometry_updating_example}. Isotropic \xy{} odometry noise and independent $\theta$ odometry noise enable us to run the prediction computationally efficiently as three consecutive 1D convolutions.

We compute offsets $o_x(\alpha)$, $o_y(\alpha)$, $o_\theta$ and kernel vectors $\kernel_\theta$, $\kernel_x(\alpha)$ and $\kernel_y(\alpha)$ such that for a given initial heading $\alpha$ in the global frame, the kernels span a region of at least four standard deviations around mean. We then perform prediction by running 1D convolutions in sequence: \begin{subequations}
    \begin{equation}
        \label{eq:x_convolution}
        \bar{\boldx}_{k,x}[i,j,l] = \sum_{h=1}^{q_x} \boldx_{k-1}[i - o_x(\bar{s}_\theta(l)) - h,j,l] \kernel_x(\bar{s}_\theta(l))[h]
    \end{equation}
    \begin{equation}
        \label{eq:y_convolution}
        \bar{\boldx}_{k,xy}[i,j,l] = \sum_{h=1}^{q_y} \bar{\boldx}_{k,x}[i,j - o_y(\bar{s}_\theta(l)) - h,l] \kernel_y(\bar{s}_\theta(l))[h]
    \end{equation}
    \begin{equation}
        \label{eq:theta_convolution}
        \bar{\boldx}_{k,xy\theta}[i,j,l] = \sum_{h=1}^{q_\theta} \bar{\boldx}_{k,xy}[i,j,l - o_\theta - h] \kernel_\theta[h]
    \end{equation}
\end{subequations}
Here, $\bar{s}_\theta(l)$ returns the centerpoint of the angle corresponding with index $l$ in $\boldx$. Note that the offsets and convolution kernel vectors in $x$ and $y$ direction depend on the value of $\alpha$, \ie{} probability mass is shifted in the direction defined by the heading value, using the nominal value of that cell. At the edges of state space, \eqref{eq:x_convolution} and \eqref{eq:y_convolution} are filled with zero for values outside state space and \eqref{eq:theta_convolution} is wrapped around to the opposite edge.

The end result of this step is $\bar{\boldx}_{k,xy\theta}$, which states how $\boldx_{k-1}$ shifted and spread according to odometry measurement and odometry noise from time $k-1$ to $k$. A clarifying visualization can be found in Figure~\ref{fig:lsl-odometry_updating_example}.

\subsubsection{Weighing belief with heading measurement}
\label{ssec:lsl-heading_measurement}
We approximate the circular Gaussian presented in Section~\ref{sec:lsl-heading_measurement} by von Mises distribution and we compute a weight matrix $\weightmatrix_{H,k}$ for all grid indices:
\begin{equation}
    \weightmatrix_{H,k}(i,j,l) =
    \Phi_{\mathcal{V}}(s_{\theta u}(l), v_k,1/\sigma_{v}^2) -
    \Phi_{\mathcal{V}}(s_{\theta v}(l), v_k,1/\sigma_{v}^2)
\end{equation}
where $\Phi_{\mathcal{V}}(t,\theta,1/\sigma^2)$ is the cumulative density function of von Mises distribution with parameters $\theta$, $1/\sigma^2$ evaluated at $t$.

\subsubsection{Weighing belief with map matching measurement}
For all grid indices, we compute a weight matrix $\weightmatrix_{M,k}$, from likelihood of the observation $\mathcal{I}_k$ representing the voxel $s_x(i), s_y(j), s_\theta(l)$, given map $\mathcal{M}$:
\begin{equation}
    \weightmatrix_{M,k}(i,j,l) = \cond{\mathcal{I}_k}{s_x(i), s_y(j), s_\theta(l),\mathcal{M}}
\end{equation}
The likelihood computation methods are described in detail in Section~\ref{sec:lsl-determining_likelihood}.

\subsubsection{Updating with all measurements}
Our updated state estimate is computed as
\begin{equation}
    \boldx_{k,u} = \bar{\boldx}_{k,xy\theta} \odot \weightmatrix_{H,k} \odot \weightmatrix_{M,k},
\end{equation}
where $\odot$ is elementwise multiplication, and finally normalized:
\begin{equation}
    \boldx_{k} = \frac{\boldx_{k,u}}{\sum_{i,j,l}\boldx_{k,u}(i,j,l)}
\end{equation}
The end result is a recursive discrete approximation of equation \eqref{eq:state_estimate_continuous}.

\subsubsection{Interval for running algorithm}
Our localization algorithm is run at fixed intervals of travel, when the UAV has traveled more than a specified distance $u_{l}$ since the latest update. The amount of travel since latest update is approximated by odometry. The value for $u_{l}$ can be chosen to balance computational load, ensure enough movement with respect to to selected grid size, and have independence between UAV images used in map matching. In general, smaller $u_{l}$ is preferred. The lower limit for $u_{l}$ comes from the requirement of independence of consecutive observations in the Bayesian filter and thus $u_{l}$ should be approximately equal to the dimension of the image used in map matching, which in our experiments is 100 m and 40 m for convergence experiments and real-time experiments, respectively. Increasing $u_{l}$ leads to less updates per distance traveled and therefore slower convergence, but allows allocating more time for running the localization algorithm. Between updates, a location estimate in navigation frame is tracked using odometry.

\section{Map matching}
\label{sec:lsl-map_matching}

Our formulation this far has considered how to fuse odometry, heading and map matching measurements using a point mass filter. We have still to define how to compute the likelihood $\cond{\mathcal{I}_k}{\continuousX_k,\mathcal{M}}$, \ie{} assess how well the observed UAV image $\mathcal{I}_k$ supports each possible value of state $\continuousX_k$, when using map $\mathcal{M}$.

Instead of detecting image features and using them as landmarks as done in \eg{} \citep{9311612}, we prefer an area-based approach, \ie{} using a large part of the observed image for discriminating between plausible and incorrect poses. The main motivation for this choice is that feature-based approaches require detection of spatially local features, which may be very sparsely detectable when flying over ambiguous terrains, especially across significant seasonal change (\eg{} after accumulation of snowfall) and when image footprint on ground is small. Using a large section of UAV image allows us to assess plausibility of matching with respect to a map even if locally distinct features cannot be detected.

Besides the targeted robustness when flying in ambiguous terrains, there are a number of other desirable characteristics for a map matching method. The method should work in matching patches of ground that are of a reasonable size; we should not develop a solution which requires a very large observed ground footprint and thus a high flying altitude in order for the solution to work. As we don't want to restrict the UAV trajectories, it is desirable to have a map matching method that is tolerant to viewpoint change (especially camera pitch and roll) and works at different flight altitudes and across a range of camera intrinsics. In addition, in developing a map matching approach trained with data samples, we cannot assume that a large amount of UAV imagery containing all types of expected variability (seasonal apperance change, camera angle with respect to ground and camera intrinsics, flight altitudes) is available for this purpose. Finally, the approach should be such that it works in the targeted scale and is computationally feasible for an onboard deployment. As a synthesis of all these needs, we propose an approach that splits the problem in two.

We first perform a rudimentary orthoprojection of the observed UAV image such that a section of the image that corresponds to a patch on ground of a specific size, at specific ground sampling distance, is generated (see Figure~\ref{fig:lsl-example_uav_image_and_projection_of_uav_image}). This acts as a means for abstracting away the flight altitude, camera intrinsic parameters and camera orientation with respect to ground and renders the problem of likelihood computation into that of finding likelihood between patches of orthoimages. This is beneficial also in training deep learning-based matching methods that are robust against seasonal variance as we demonstrated in earlier work \citep{9830867}, since our method abstracts away parameters that are irrelevant for the map matching problem (\ie{} camera intrinsics and altitude differences) and, unlike UAV image datasets, satellite images containing seasonal variation are plentiful.

As a second step, we compute a compact descriptor vector from the UAV image patch. We compute likelihood of each pose hypothesis by comparing that descriptor vector to a set of descriptor vectors that have been precomputed from map $\mathcal{M}$. Likelihood estimation is done by computing distance in the embedding space spun by the descriptor vectors. The choice of operating on compact descriptor vectors instead of template matching individual pose hypotheses as \eg{} in previous work \citep{9830867} is a key enabler for fast likelihood estimation over a large map, but raises the question of how to engineer a descriptor vector computation method in a way that allows as small vector size as possible while enabling robust localization.

In this section, we first introduce the orthoprojection method, after which we describe the methods of computing descriptor vectors, precomputing a map and finally computing the matching likelihood.

\subsection{Orthoprojecting UAV image}
\label{sec:lsl-orthoprojecting}

The map matching measurement is generated based on the view from a single UAV image, which may be tilted from nadir and thus contains perspective difference with respect to top-down view. To reduce the impact of this perspective change, we orthoproject the UAV image and make the assumption that the ground beneath the UAV is planar. Our approach resembles earlier work \citep{9659333} with a few key differences and we report an overview of the full approach for completeness.

Based on direction of gravity estimated by AHRS, we first define local frame \localcoordinateframe{k} for image sampled at time $t_k$ whose origin is at the origin of the UAV camera frame, $z$ axis points in opposite direction to gravity, and the component of camera image plane horizontal axis perpendicular to $z$ is aligned with $y$ axis.

Our localization approach is based on using an orthoprojection of the camera view of the UAV. We detect image feature points using Shi-Tomasi detector~\citep{323794} and track movement of the features across a batch of ten consecutive camera image frames around keyframe sampled at $t_k$ using a pyramidal Lucas-Kanade tracker~\citep{bouguet2001pyramidal}. Besides image sampled at $t_k$, the batch consists of four images prior to the image corresponding with time $t_k$ and five after. We estimate the 3D locations of tracked feature points in frame \localcoordinateframe{k} using the linear triangulation method in \citep{HARTLEY1997146}, using relative pose transformations between frames in batch that we compute from ground truth data. Exploration of VIO frontends is beyond the focus of this work and we believe the use of noiseless relative transformations is a sufficient approximation of a generic VIO algorithm over the sequence of ten frames.

\begin{figure}
  \centering
  \begin{minipage}[t]{.45\linewidth}
    \subcaptionbox{Original UAV image $\mathcal{I}_k$}
      {\includegraphics[width=\linewidth]{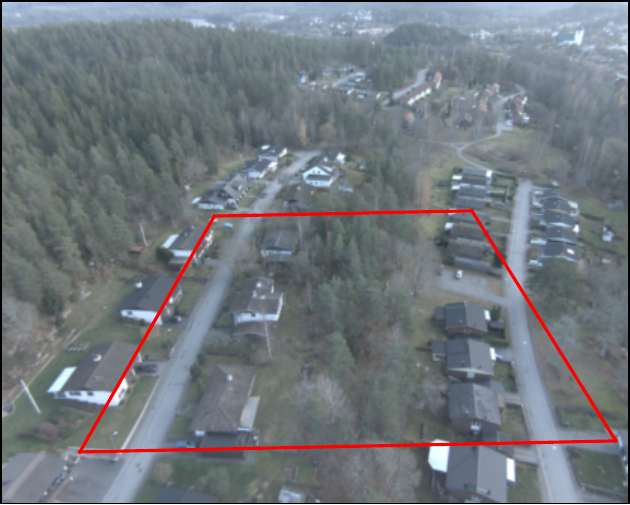}
      \label{fig:lsl-original_uav_image}
      }\\
    \subcaptionbox{Orthoprojected image $I_k$}
      {\includegraphics[width=\linewidth]{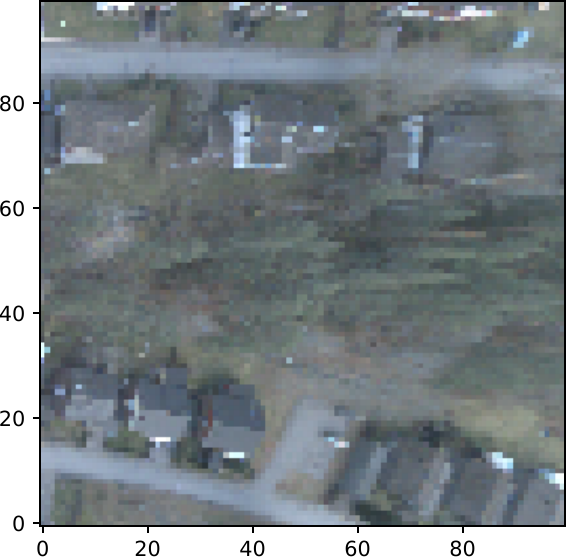}
      \label{fig:lsl-orthoprojected_uav_image}
      }%
  \end{minipage}%
  \hfill
  \begin{minipage}[b]{.50\linewidth}
    \subcaptionbox{Coordinate frame \localcoordinateframe{k}, points tracked in VIO (cyan dots), square lying on plane fit to VIO landmarks and transformation $\localcoordtosquarecoord{}$ to \squarecoordinateframe{k}}
      {\includegraphics[width=\linewidth]{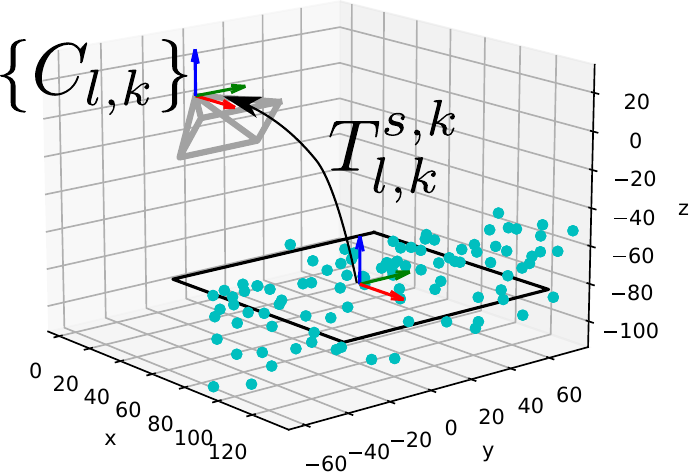}\label{fig:lsl-coordinate_frames}}%
  \end{minipage}%
  \caption
    {Example of UAV image (\squaresizem{} area used for orthoprojection highlighted in red), its orthoprojection, and a visualization of the coordinate frames.}
\label{fig:lsl-example_uav_image_and_projection_of_uav_image}
\end{figure}

We then find the best-fitting plane whose normal is aligned with $z$ axis; \ie{} we assume the ground below the UAV is planar and horizontal. We find a square with size \squaresizem{} lying on the best-fitting plane that is closest to nadir and fully visible in the camera image $\mathcal{I}_k$. We project the cornerpoints of this square into the image plane of $\mathcal{I}_k$ and, by homograpy, transform the pixels corresponding with the square in $\mathcal{I}_k$ to an \squaresizepx{} image at 1 m/px resolution which we call $I_k$. $I_k$ is the observation we use in map matching. We also compute $\localcoordtosquarecoord{}$, a transformation from frame \localcoordinateframe{k} to a frame centered in the middle of the square, which we label \squarecoordinateframe{k}. There is no rotation between \localcoordinateframe{k} and \squarecoordinateframe{k} and translation is defined such that \squarecoordinateframe{k} is at the center of the square. An example of one orthoprojected UAV image and a visualization of the coordinate frames is shown in Figure~\ref{fig:lsl-example_uav_image_and_projection_of_uav_image}.

The presented approach enables us to consider the motion of a UAV as a sequence of 2D translations and rotations while allowing extraction of observation $I_k$ at correct scale independent of fight altitude, while $\localcoordtosquarecoord{}$ contains information about position of its centerpoint relative to camera, including estimated ground plane altitude.

\subsection{From orthoimage patch to descriptor vector}
\label{sec:lsl-observation_into_embeddingspace}
Inspired by other works projecting the UAV observation into a single compact descriptor vector \citep{9357892, 9562005, 10.1117/12.2585986}, to provide a fast way to compare an observation to a large number of pose hypotheses, our map matching method is based on transforming both UAV observations and a reference map into a suitable descriptor space, and computing probability of pose hypotheses using descriptor vector values.
Based on image $I_k$, we compute a descriptor vector $\descriptorvector_k$ using function $f_i$: $\descriptorvector_k = f_i(I_k)$. This function $f_i$ is composed of a deep neural network backbone $b$ such as Resnet \citep{He_2016_CVPR} followed by a projection module $m$. Therefore $f_i(I_k)= m_i(b_i(I_k))$. The projection module is stacked on top the ResNet model where the the last average pooling and final fully connected layer specified in \citep{He_2016_CVPR} are removed. The last layer of the projection module has $D$ neurons in order to compute $D$-dimensional $l_2$-normalized descriptor vectors. The projection model thus produces vectors in unit $D$-sphere. Therefore, $f: \mathbb{R}^{h \times w \times c} \rightarrow \mathbb{R}^{D}$, where $h$, $w$ and $c$ are the height, width and channels dimensions of the input image $I_k$.
\subsubsection{Projection modules}
\label{sssec:lsl-projection_modules}

In order to extract $D$-dimensional embedding vectors, we used two different types of projection modules: one composed only of fully connected layers, $m_{FCN}$, and one composed by a Capsule Network (CapsNets) model~\citep{hinton-dr}, $m_{CAP}$. We refer to $f_{FCN}$ and $f_{CAP}$ as the models where the ResNet backbone is followed by $m_{FCN}$ and $m_{CAP}$, respectively.

\begin{figure}
     \centering
     \begin{subfigure}[b]{0.7\linewidth}
         \centering
         \includegraphics[width=\textwidth]{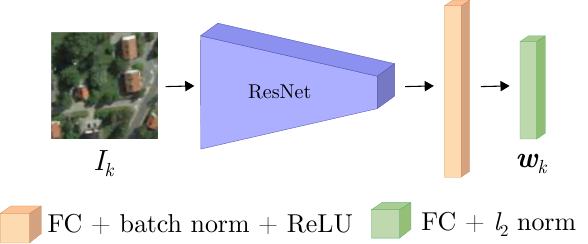}
         \caption{$f_{FCN}$}
         \label{fig:lsl-fcn_architecture}
     \end{subfigure}
     \par\bigskip
     \begin{subfigure}[b]{\linewidth}
         \centering
         \includegraphics[width=\textwidth]{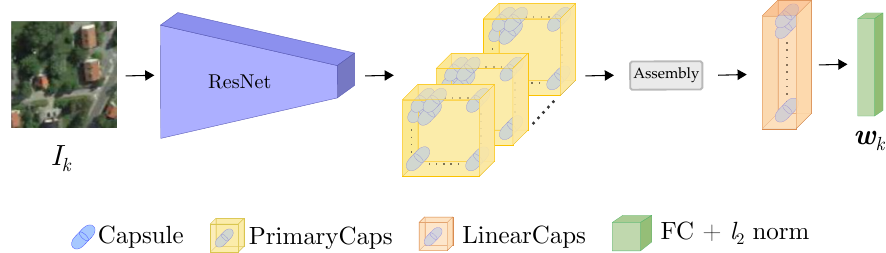}
         \caption{$f_{CAP}$}
         \label{fig:lsl-caps_architecture}
     \end{subfigure}
     \hfill     
        \caption{Architectures for fully connected and capsule networks}
        \label{fig:lsl-f_architectures}
\end{figure}

Inspired by the architecture choices proposed by \citep{9562005,10.1117/12.2585986}, the $m_{FCN}$ module is composed by two fully connected layers, one with N neurons and one with D outputs. A visualization of the network structures is shown in Figure~\ref{fig:lsl-fcn_architecture}.

CapsNets have gained great attention recently since they form more stable and robust representations with respect to input perturbations~\citep{hinton-trans-autoencoders}. Compared to convolutional neural networks (CNNs), CapsNets also achieve greater generalization with fewer trainable parameters~\citep{hinton-dr, hinton-em}. Their main innovation lies in two major distinctions from CNNs:
(i) the encoding of object poses (position, size, orientation) and visual attributes (\eg{} color, texture, deformation, hue) into groups of neurons called \textit{capsules} (ii) the routing-by-agreement mechanism, which models the connections between capsules of different layers. Namely, it models the part-whole relationships among objects without losing spatial information.

Before this work, the seasonal invariance property of CapsNets had not been tested. But since CapsNets are well known for being robust to affine transformations, novel viewpoints, and lighting conditions~\citep{hinton-trans-autoencoders}, we investigated the potentials of CapsNets to learn season invariant features. 
Furthermore, in the literature~\citep{self-routing, gracapsnets}, capsule layers have been widely stacked on top of ResNet backbones instead of traditional fully connected layers to achieve better performance. Owing to these properties, extracting descriptor vectors using $f_{CAP}$ can help to solve the UAV localization task described in this work. A visualization of $f_{CAP}$ is shown in Figure~\ref{fig:lsl-caps_architecture}.

\subsection{Precomputing descriptors for map}
\label{ssec:lsl-precomputing_map}
We precompute offline a map around the expected operating region. The map holds descriptor vector values that have been computed from a georeferenced orthophoto RGB bitmap $\mathcal{M}$. For each map coordinate $X_h = (x_h, y_h, \theta_h)$, we crop a $w$ m by $w$ m square image $I_{\mathcal{M},h}$, translated from origin by $(x_h, y_h)$ in map coordinates and rotated by $\theta_h$, from the map image $\mathcal{M}$. We then compute an embedding vector $\descriptorvector_h = f_i(I_{\mathcal{M},h})$. We compute these embedding vectors using the centerpoint coordinate of each voxel in $\boldx$. This process yields a precomputed map $\precomputedmap(i,j,l) \in \mathbb{R}^{D}$ where indices $i$, $j$, $l$ correspond with indices of grid cells in our belief representation $\boldx$. We compute a separate map for each tested network architecture. Dimensions of the image in pixels, as well as chosen resolution of images, are a design parameter that can be selected based on expected use case. In the experiments section in this work we use either $w=$100 m or $w$=40 m, as reported in each experiment, and scale each image to 1 m/px resolution.

\subsection{Determining map matching likelihood}
\label{sec:lsl-determining_likelihood}
To assess whether a UAV observation corresponds with a state hypothesis, we approximate
\begin{equation}
    \cond{\mathcal{I}_k}{s_x(i), s_y(j), s_\theta(l),\mathcal{M}} \approx \cond{\descriptorvector_k}{i,j,l,\precomputedmap}
\end{equation}
and compute $\weightmatrix_{M,k}$, which contains the weight for each state space element on the chosen voxel grid.
\begin{equation}
    \weightmatrix_{M,k}(i,j,l) = \cond{\descriptorvector_k}{i,j,l,\precomputedmap}
\end{equation}

We compare two solutions for this. The first one, labeled \emph{linear}, is similar to the choice in \citep{9562005}. It assumes that Euclidean distance between the embedding obtained from the UAV and from the map $c_k(i,j,l) = \|\descriptorvector_k-\precomputedmap(i,j,l)\|_2$ is inversely proportional to the probability of correct pose:
\begin{equation}
    \weightmatrix_{M1,k}(i,j,l) =
    \frac{2-c_k(i,j,l)}{2}
\end{equation}

% \begin{equation}
%     c_k(i,j,l) = \|\descriptorvector_k-\descriptorvector_h\|_2
% \end{equation}

The second one is the method for computing importance factor presented in earlier work \citep{9830867}, where we estimate the probability density of distances in Euclidean space for true and false matches from satellite image data and compute the probability that the observation is from "match" class, for each element in $\boldx_k$ individually. We label this weighing method \emph{bayesian} and we name weight matrix $\weightmatrix_{M2,k}$.

\section{Experiments}\label{sec:lsl-experiments}

\subsection{Overview of experiments}

We experiment the performance of our solution with respect to baseline methods on real-world datasets. We evaluate probabillity of convergence, time to convergence and localization error after convergence with flights taking place in two areas in Sweden. In both areas, we experiment with the problem of localization starting from \maparea{} uncertainty. In addition, we experiment with real-time implementation on a UAV onboard computer.

For evaluating the critical design choices in our map matching approach, we vary the projection module type (fully connected or capsule network), likelihood vector dimensionality $D$ (8, 16, 32 or 128) and likelihood conversion method (linear or bayesian), evaluate the impact of these choices on probability of convergence, time to convergence and mean localization error after convergence.

We evaluate localization performance by the criteria defined in Section~\ref{sec:lsl-evaluation_criteria_in_experiments}. The datasets we use for experimentation and model training are described in Section~\ref{ssc:datasets}. Training methods are outlined in Section~\ref{subsubsec:lsl-training_details}, baseline methods are described in Section~\ref{ssec:lsl-comparisonmethods}, followed by evaluation of localization performance in Section~\ref{ssec:lsl-evaluating_localization_performance} and learnings from real-time experiments in Section~\ref{sec:lsl-realtime_experiment}.

\subsection{Evaluation criteria in localization experiments}
\label{sec:lsl-evaluation_criteria_in_experiments}
\subsubsection{Translation error}

We compute the estimated \xycoords~ and heading using \eqref{eq:xy_estimate} and \eqref{eq:theta_estimate}, respectively.
\begin{subequations}
\begin{equation}
    \label{eq:xy_estimate}
    \widehat{X}^{xy}_{s,k} =
    \sum\limits_{i,j,l}\boldx(i,j,l)
    \begin{bmatrix}
    \bar{s}_{x}(i) & \bar{s}_{y}(j)
    \end{bmatrix}^T
\end{equation}
\begin{equation}
    \label{eq:theta_estimate}
    \widehat{X}^{\theta}_k = atan2(\sum\limits_{i,j,l}\boldx(i,j,l)
    sin(\bar{s}_{\theta}(l)),\sum\limits_{i,j,l}\boldx(i,j,l)
    cos(\bar{s}_{\theta}(l)))
\end{equation}
\end{subequations}
Here, $\bar{s}_x(i)$, $\bar{s}_y(j)$ and $\bar{s}_\theta(l)$ are the centerpoint coordinates of voxel corresponding to indices $(i, j, l)$ and $atan2$ is the 2-argument arctangent. We then transform these estimates from $\widehat{X}^{xy}_{s,k}$ to drone-centric coordinates $\widehat{X}^{xy}_{k}$.

We compute the Euclidean distance to \ground{} \xycoords{} $X^{xy}_{k,gt}$:

\begin{equation}
    \label{eq:xy_error}
    \widetilde{X}^{xy}_k = \| \widehat{X}^{xy}_k - X^{xy}_{k,gt} \|_2
\end{equation}

We also compute
$\sigma_{\widehat{xy}}$, the weighted standard deviation of Euclidean distances to $\widehat{X}^{xy}_k$ in $(x,y)$ plane, weighing with $\boldx_k$. We claim that $\sigma_{\widehat{xy}}$ is a suitable measure of convergence of the localization solution, and a large spread of uncertainty signals the need for re-initialization of the estimator.

\subsection{Datasets}\label{ssc:datasets}
\subsubsection{Satellite images}\label{subsubsec:lsl-sat_dataset}
We trained our networks on 100m by 100m samples at 1 m/pixel randomly drawn from Google Earth satellite images of size 4800 by 2987 meters collected from 9 regions, in arbitrarily selected places in southern Finland that cover urban and non-urban areas. For each region, 4 to 15 satellite images collected from the same area at different times, containing seasonal variation, are used. The Google Earth datasets are the same as in an earlier work \citep{9830867}.
\subsubsection{UAV images}\label{subsubsec:lsl-uav_dataset}
\begin{figure}
     \centering
     \begin{subfigure}[b]{0.22\linewidth}
         \centering
         \includegraphics[width=\textwidth]{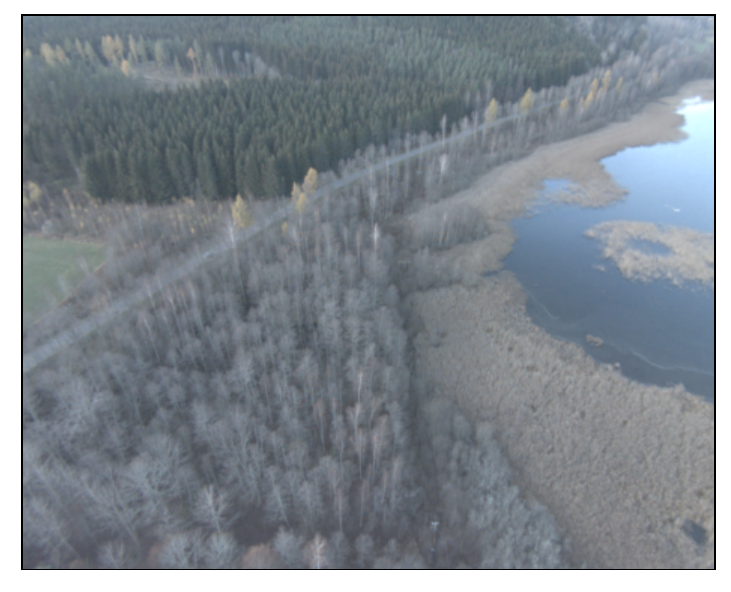} % kisa-2019-11-07-000
         \caption{Dataset 1}
         \label{fig:lsl-sampleimages_dataset1}
     \end{subfigure}
     \hfill
     \begin{subfigure}[b]{0.22\linewidth}
         \centering
         \includegraphics[width=\textwidth]{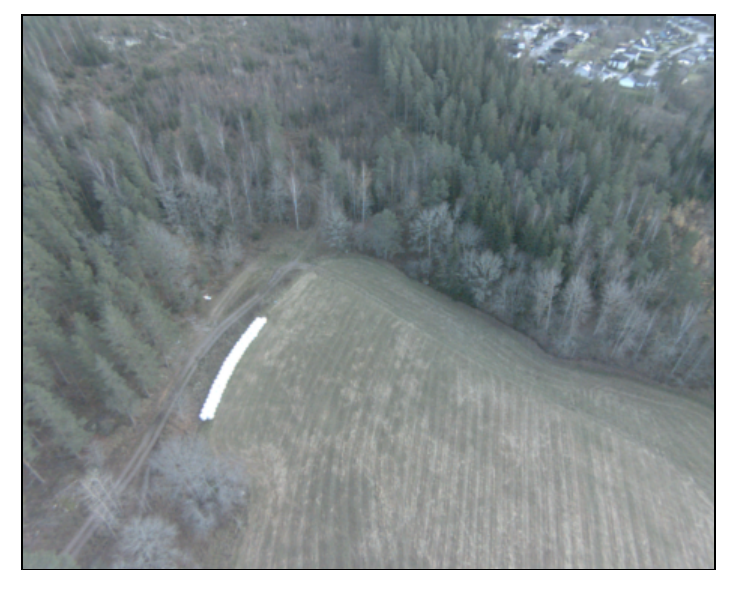} % kisa-2019-11-07-002
         \caption{Dataset 2}
         \label{fig:lsl-sampleimages_dataset2}
     \end{subfigure}
     \hfill
     \begin{subfigure}[b]{0.22\linewidth}
         \centering
         \includegraphics[width=\textwidth]{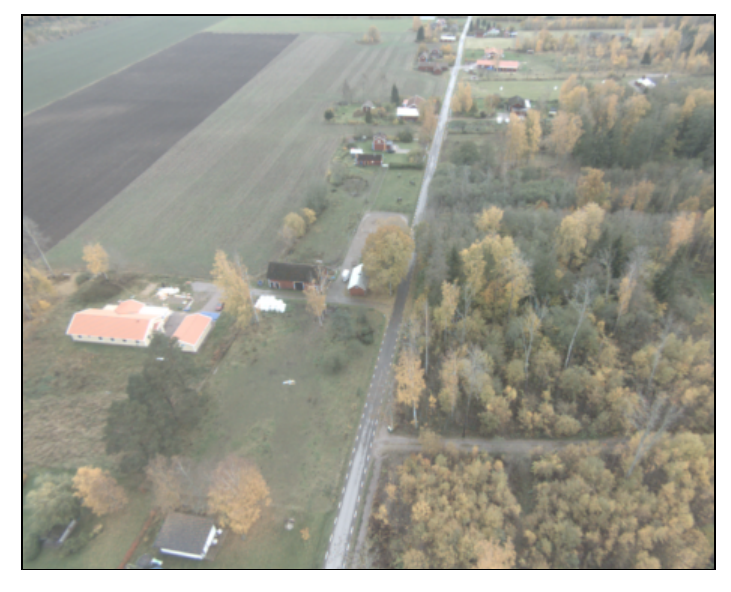} % klockrike-2019-10-18-002
         \caption{Dataset 3}
         \label{fig:lsl-sampleimages_dataset3}
     \end{subfigure}
     \hfill
     \begin{subfigure}[b]{0.22\linewidth}
         \centering
         \includegraphics[width=\textwidth]{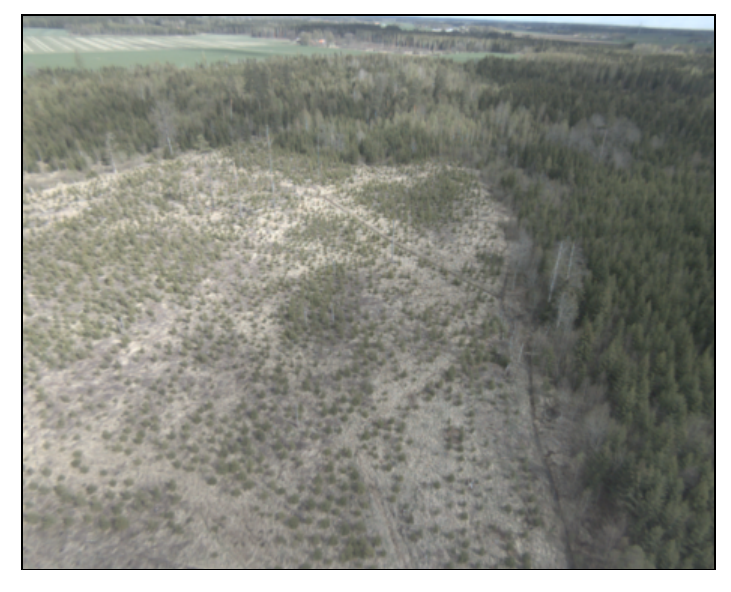} % klockrike-2020-04-29-001
         \caption{Dataset 4}
         \label{fig:lsl-sampleimages_dataset5}
     \end{subfigure}     
     \hfill
     \begin{subfigure}[b]{0.22\linewidth}
         \centering
         \includegraphics[width=\textwidth]{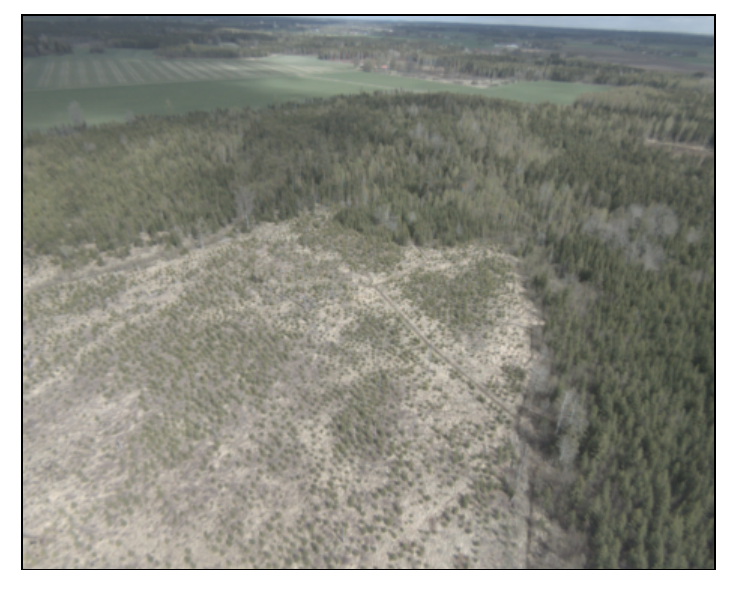} % klockrike-2020-04-29-002
         \caption{Dataset 5}
         \label{fig:lsl-sampleimages_dataset6}
     \end{subfigure}
     \hfill
     \begin{subfigure}[b]{0.22\linewidth}
         \centering
         \includegraphics[width=\textwidth]{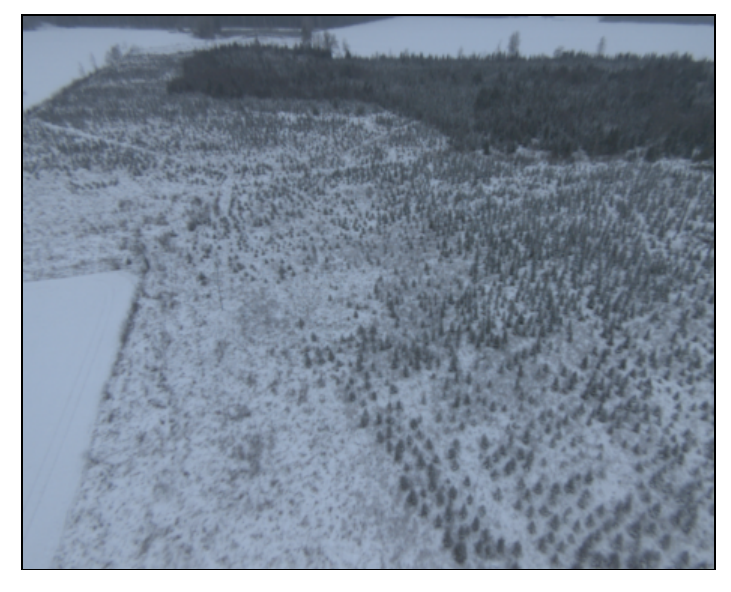} % klockrike-2021-01-19-000
         \caption{Dataset 6}
         \label{fig:lsl-sampleimages_dataset7}
     \end{subfigure}
     \hfill
          \begin{subfigure}[b]{0.22\linewidth}
         \centering
         \includegraphics[width=\textwidth]{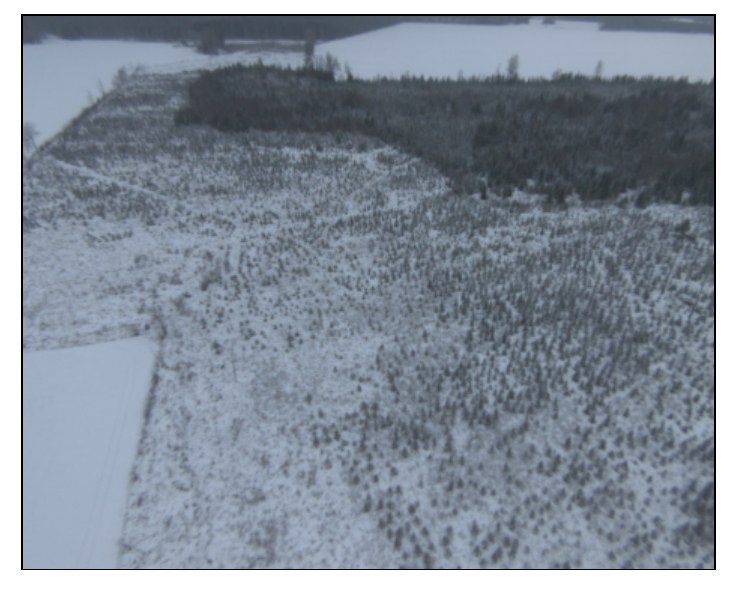} % klockrike-2021-01-19-001
         \caption{Dataset 7}
         \label{fig:lsl-sampleimages_dataset4}
     \end{subfigure}
     \hfill
     \begin{subfigure}[b]{0.22\linewidth}
         \centering
         \includegraphics[width=\textwidth]{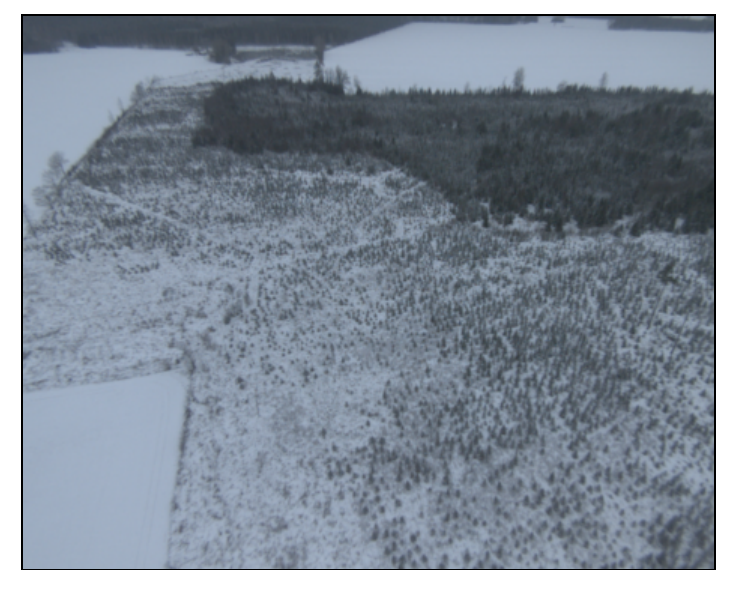} % klockrike-2021-01-19-002
         \caption{Dataset 8}
         \label{fig:lsl-sampleimages_dataset8}
     \end{subfigure}
     \hfill     
        \caption{Example images from datasets. Note difference in seasonal appearance.}
        \label{fig:lsl-example_images_from_datasets}
\end{figure}

\begin{figure}
     \centering
     \begin{subfigure}[b]{0.49\linewidth}
         \centering
         \includegraphics[width=\textwidth]{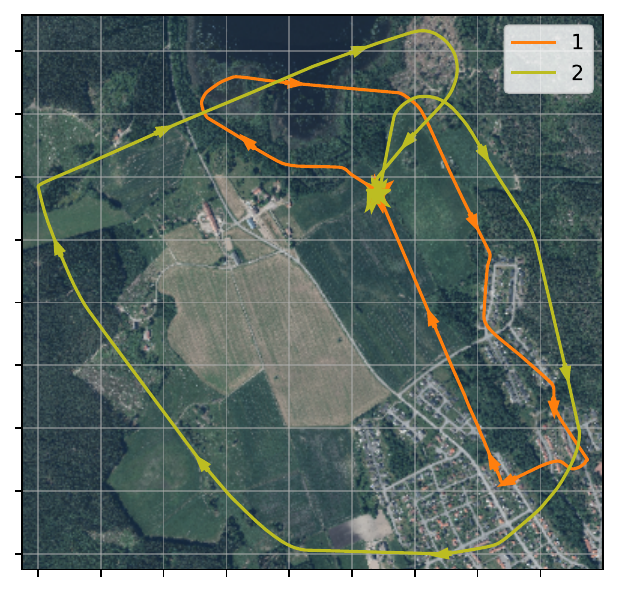}
         \caption{Kisa}
         \label{fig:lsl-area_a_trajectories}
     \end{subfigure}
     \hfill
     \begin{subfigure}[b]{0.3\linewidth}
         \centering
         \includegraphics[width=\textwidth]{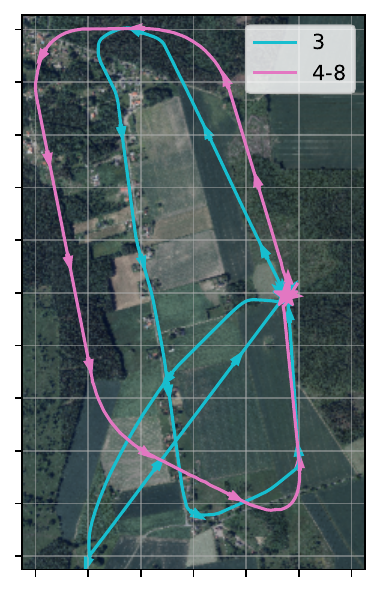}
         \caption{Klockrike}
         \label{fig:lsl-area_b_trajectories}
     \end{subfigure}
     \hfill     
        \caption{Flight trajectories of each dataset. Grid spacing 200 m. Datasets 4-8 have same trajectory with varying altitude and season.}
        \label{fig:lsl-trajectories}
\end{figure}

Our experiments are run on datasets that have been collected with a UAV in two locations in Sweden at different times\footnote{Data provided by Saab Dynamics Ab.}. The dataset contains a sequence of posed images sampled at 10 Hz. \Ground{} camera trajectory is collected with a real-time kinematic (RTK)-corrected global positioning system (GPS) in uninterfered conditions. The use of RTK-GNSS ensures position precision in the centimeter range. A listing of the flight experiments is given in Table~\ref{tab:lsl-uav_dataset_characteristics}. A representative image of each dataset is shown in Figure~\ref{fig:lsl-example_images_from_datasets} and flight trajectories over an orthophoto map are shown in Figure~\ref{fig:lsl-trajectories}. The flights take place over terrains with forest areas, a lake, agricultural fields and some residential areas. At the time of running this experiment, IMU data was not available.

\subsubsection{Maps used for localization}
Each original map $\mathcal{M}$ is an orthophoto bitmap constructed from aerial images taken over the operating area in summer 2021. We use orthophoto bitmaps with an original ground sampling distance of 0.16 m/px provided by a local map information supplier\footnote{\copyright ~Lantmäteriet, \url{https://www.lantmateriet.se/}.}.

\begin{table}
\caption{\label{tab:lsl-uav_dataset_characteristics}Characteristics of flight datasets. Trajectory lengths are computed along $(x, y)$ plane, and camera angles between nadir and camera principal axis. Altitude is with respect to starting position.}
\begin{tabular}{p{2mm} p{15mm} p{20mm} p{12mm} p{12mm} p{35mm}} 
\toprule
\# & Area & Date & Traj. length (km) & Alt. (m) & Median camera angle [range] ($^{\circ}$)\\
\midrule
%% these results generated with git hash e7f30f0cd5ecd8cf2ec52f8a778c73017a408611 on 09/13/2022, 10:14:52
1 & Kisa & 2019-11-07 & 4.1 & 91 & 59.0 [56.9 \ldots 67.7] \\ % kisa-2019-11-07-000
2 & Kisa & 2019-11-07 & 6.1 & 92 & 52.1 [49.8 \ldots 56.0] \\ % kisa-2019-11-07-002
3 & Klockrike & 2019-10-18 & 6.8 & 92 & 51.2 [49.9 \ldots 53.9] \\ % klockrike-2019-10-18-002
4 & Klockrike & 2020-04-29 & 4.8 & 53 & 59.3 [57.8 \ldots 60.5] \\ % klockrike-2020-04-29-001
5 & Klockrike & 2020-04-29 & 4.8 & 92 & 59.9 [58.4 \ldots 61.3] \\ % klockrike-2020-04-29-002
6 & Klockrike & 2021-01-19 & 4.9 & 54 & 58.9 [58.0 \ldots 60.4] \\ % klockrike-2021-01-19-000
7 & Klockrike & 2021-01-19 & 4.7 & 71 & 58.7 [57.7 \ldots 60.0] \\ % klockrike-2021-01-19-001
8 & Klockrike & 2021-01-19 & 4.9 & 84 & 58.4 [56.9 \ldots 60.9] \\ % klockrike-2021-01-19-002
\bottomrule
\end{tabular}
\figvspace{}
\end{table}
\subsection{Training descriptor generator networks}

\subsubsection{Training details}\label{subsubsec:lsl-training_details}
We trained the models $f_{FCN}$ and $f_{CAP}$ only on satellite images.\footnote{We acknowledge the computational resources provided by the Aalto Science-IT project.} At each epoch, 45k training satellite samples are generated from the 9 areas. Each sample represents a location and orientation on a map. We draw 5 images corresponding with each location and orientation. For each batch, we randomly select 10 locations. As deep metric learning for encoding the images into low dimensional vectors, we employed the triplet loss~\citep{tripletloss} with batch-all strategy and margin $0.2$, using vectors representing the same location and orientation as positive samples, and vectors representing other locations and orientations in the batch as negative samples. In addition to random selection of locations and orientations, we add a random translation offset of 0 to 35 meters with uniform distribution to each location, with the intent to add robustness for observations that do not align perfectly on the map grid. With similar motivation, we add normally distributed random rotations with standard deviation of 6 degrees to the orientation of each sample before extracting the image patch from the satellite image. In addition to these, we augment the samples with Gaussian noise, motion blur, random brightness and contrast changes, and hue and saturation changes to add tolerance for changes in imaging conditions. We used Adam optimizer~\citep{adam} with learning rate $10^{-6}$. We found empirically that low triplet loss on satellite data leads to low loss on UAVs images and better localization performance. Therefore, we exploit the loss of the network as a proxy for localization performance during training. We let the network train for 500 epochs and employed descriptors of the model that reached the lowest loss on the validation satellite images.

\subsubsection{Architecture details of neural networks}
We employed a Resnet-50 model implemented in PyTorch 1.8.0 as backbone $b$. This network is pretrained on Places365 dataset~\citep{zhou2017places}. As regards the projection module $m_{FCN}$, we used a fully connected layer of sizes 1024 and $D$, respectively. We experiment with vector length $D \in \{8, 16, 32, 128\}$. On the other hand, with regards to the module $m_{CAP}$, the PrimaryCaps layer is a capsule layer with 64 types of 16-dimensional capsules, which are obtained from a convolutional layer of 1024 $1\times1$ kernel size filters. The LinearCaps layer consists of 32 32-dimensional capsules extracted running 3 iterations of the routing algorithm. We stacked on top of the LinearCaps layer a fully connected layer of size $D$. With the capsule projection module, we explore $D \in \{8, 16, 32\}$.
For both projection modules, the $D$-dimensional output of the last layer is used to produce the $l_2$-normalized embedding vectors.

\subsection{Comparison methods}\label{ssec:lsl-comparisonmethods}

We implement two methods to work as comparison methods of map matching for our approach. To provide comparable results, we utilize the same point mass filter implementation, same odometry measurements, prediction method, heading weighing method and same measurement images $I_k$ for all methods and only replace the map matching solution with their approach.

As baseline methods for map matching, we use the solution by \citet{MANTELLI2019304}, whose formulation is scalable to large maps. To provide comparable results, we compute the descriptor vectors $\descriptorvector_k = f_i(I_{k})$ such that $\descriptorvector_k$ is the abBRIEF descriptor. In a similar manner, we precompute a grid of abBRIEF descriptors from $\mathcal{M}_b$, using equal grid spacing as with our methods, and with abBRIEF, we compute similarity as specified in \citet{MANTELLI2019304} and label this similarity computation method \emph{mantelli}.

To provide a reference to a method that operates on semantic maps and allows very compact map representation over large areas, we implement the map matching solution proposed by \citet{9341682}. We trained a U-Net~\citep{unet} network on the Massachusetts Buildings Dataset~\citep{MnihThesis} to segment buildings in the input images. We employ this network to extract the invariant feature descriptors introduced by Choi \etal{} \citep{9341682} on satellite and UAVs images: in this setting the feature vectors $\descriptorvector_k$ are built upon building ratio information.

\subsection{Localization performance}\label{ssec:lsl-evaluating_localization_performance}
\subsubsection{Experimental setting}
We perform all localization experiments on datasets that contain images collected with a UAV. In all experiments, we update our belief after at least $u_{l} = $ \translationbetweenupdates{} m of travel have occurred since the previous update, and the amount of travel after most recent update is estimated from odometry. In all experiments, we use a map grid with resolution $r_{xy}=10$ m, $r_{\theta}=6 \degree$.

\subsubsection{AHRS and VIO measurements}
The datasets used in experiments do not contain IMU or magnetometer measurements and thus we have to simulate them. We simulate odometry measurements by random sampling from distribution \eqref{eq:2d_odometry_model}. The translation standard deviation $\sigma_{u,xy}$ is approximated as 0.05 m per meter of travel and heading standard deviation $\sigma_{u,\theta}$ as $0.15 \degree$ per meter of travel, which approximately correspond with the performance of VIO algorithms reported in literature \citep{8460664}.

For heading, we sample from the distribution specified in \eqref{eq:heading_measurement}.  Manufacturers of compact commercial AHRS sensors typically report RMS error of $2\degree$ in heading \citep{xsens_mti3_datasheet, vectornav_vn_100_datasheet}. In all experiments with AHRS, we simulate the heading measurement from \ground{} orientation data and assume $\sigma_{v} = 3\degree$.

\subsubsection{Evaluating localization performance}
\begin{figure*}[t!]
    \centering
    \begin{subfigure}[t]{0.5\textwidth}
        \centering
        \includegraphics[width=\linewidth,valign=t]{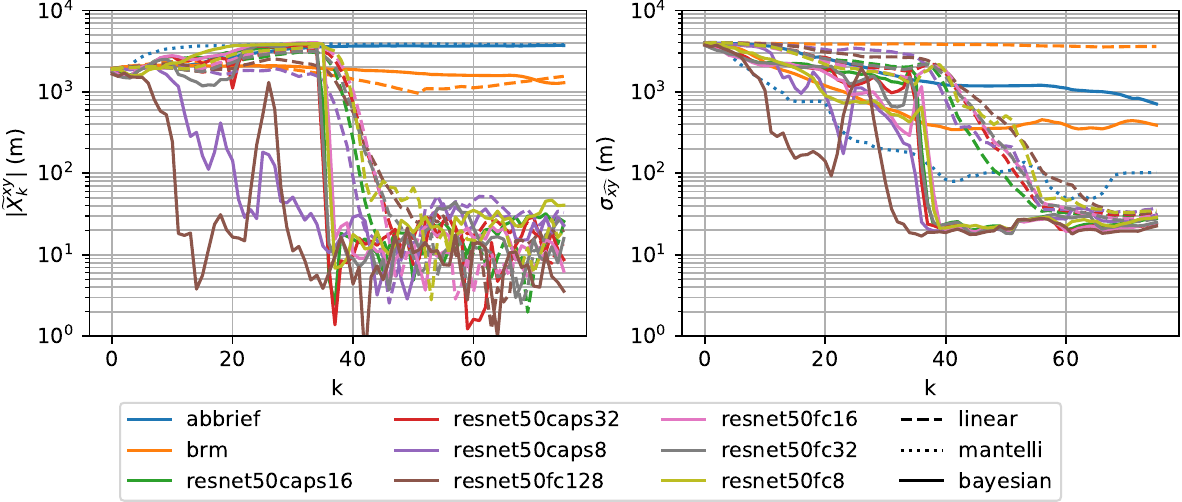} %Dataset 1 = kisa-2019-11-07-000
        \caption{Dataset 1}
        \label{fig:lsl-dataset1_graph}
    \end{subfigure}%
    ~ 
    \begin{subfigure}[t]{0.5\textwidth}
        \centering
        \includegraphics[width=\linewidth,valign=t]{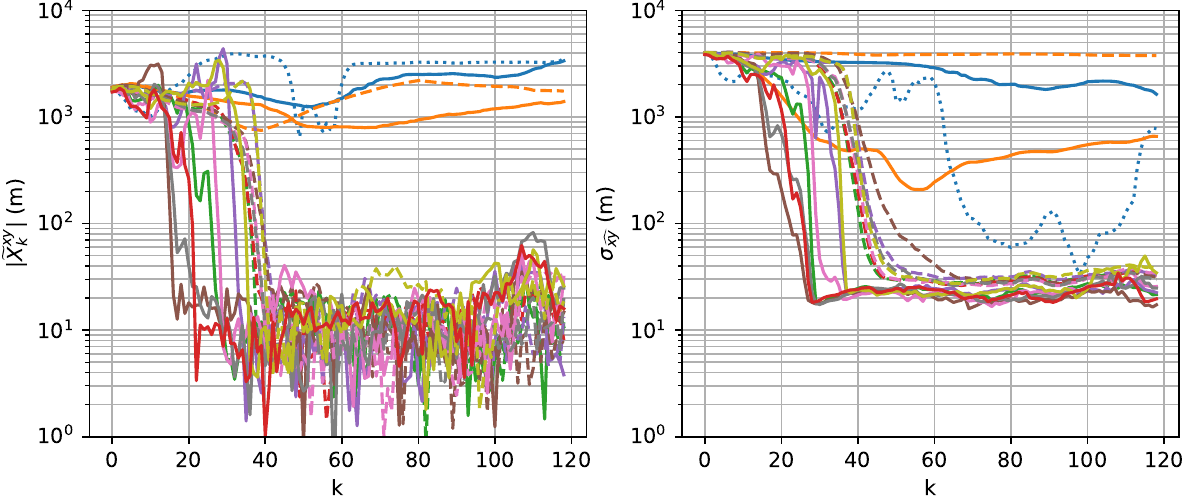} %Dataset 2 = kisa-2019-11-07-002
        \caption{Dataset 2}
        \label{fig:lsl-dataset2_graph}
    \end{subfigure}%
    \hfill
    \begin{subfigure}[t]{0.5\textwidth}
        \centering
        \includegraphics[width=\linewidth,valign=t]{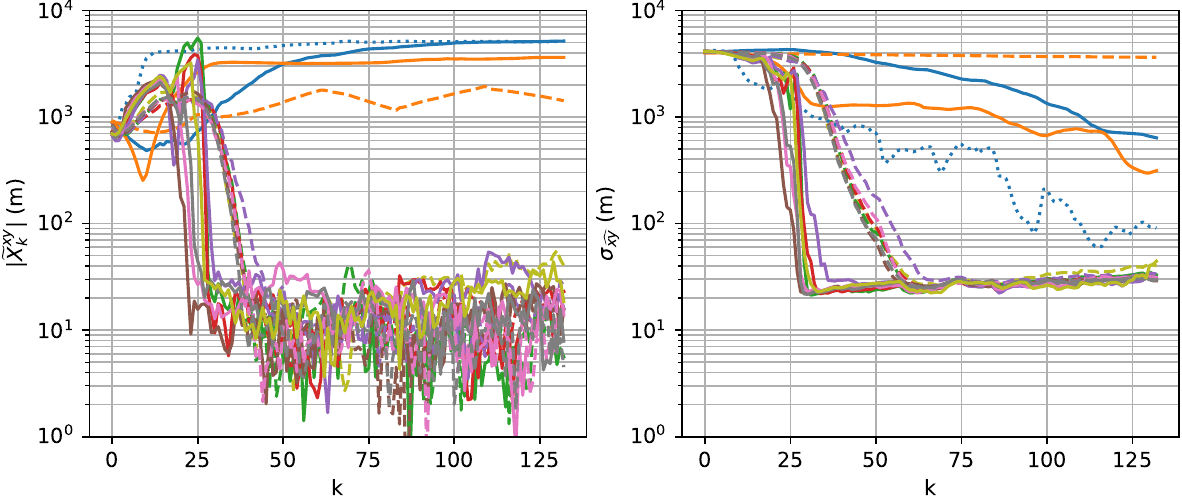} %Dataset 3 = klockrike-2019-10-18-002
        \caption{Dataset 3}
        \label{fig:lsl-dataset3_graph}
    \end{subfigure}%
    ~
    \begin{subfigure}[t]{0.5\textwidth}
        \centering
        \includegraphics[width=\linewidth,valign=t]{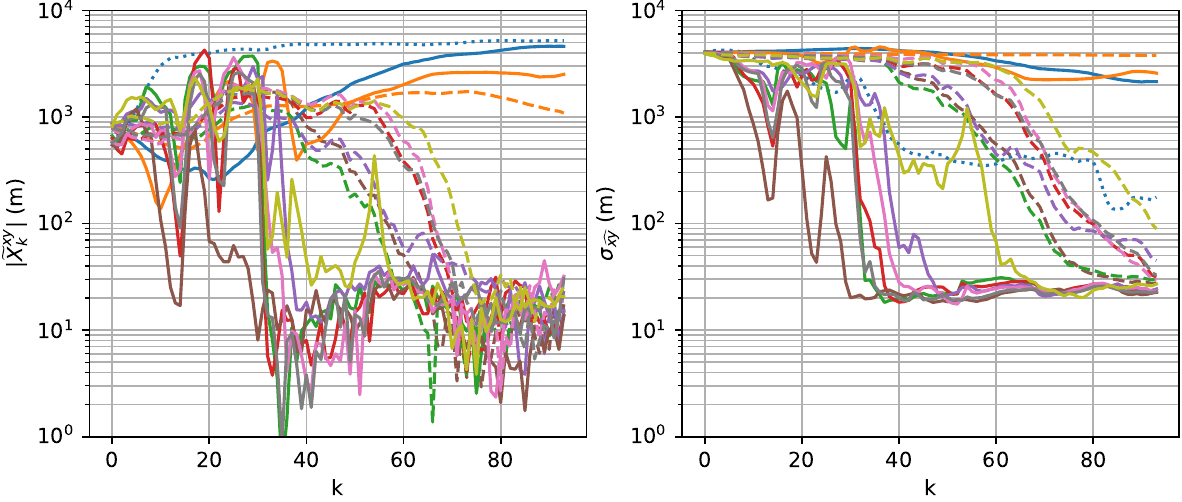} %Dataset 4 = klockrike-2020-04-29-001
        \caption{Dataset 4}
        \label{fig:lsl-dataset4_graph}
    \end{subfigure}
    \hfill
    \centering
    \begin{subfigure}[t]{0.5\textwidth}
        \centering
        \includegraphics[width=\linewidth,valign=t]{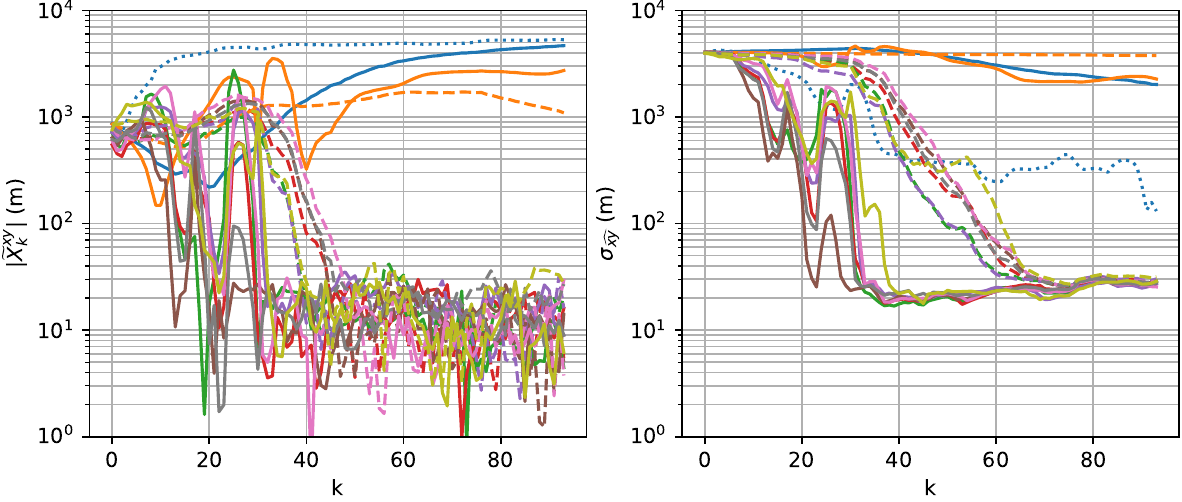} %Dataset 5 = klockrike-2020-04-29-002
        \caption{Dataset 5}
        \label{fig:lsl-dataset5_graph}
    \end{subfigure}%
    ~ 
    \begin{subfigure}[t]{0.5\textwidth}
        \centering
        \includegraphics[width=\linewidth,valign=t]{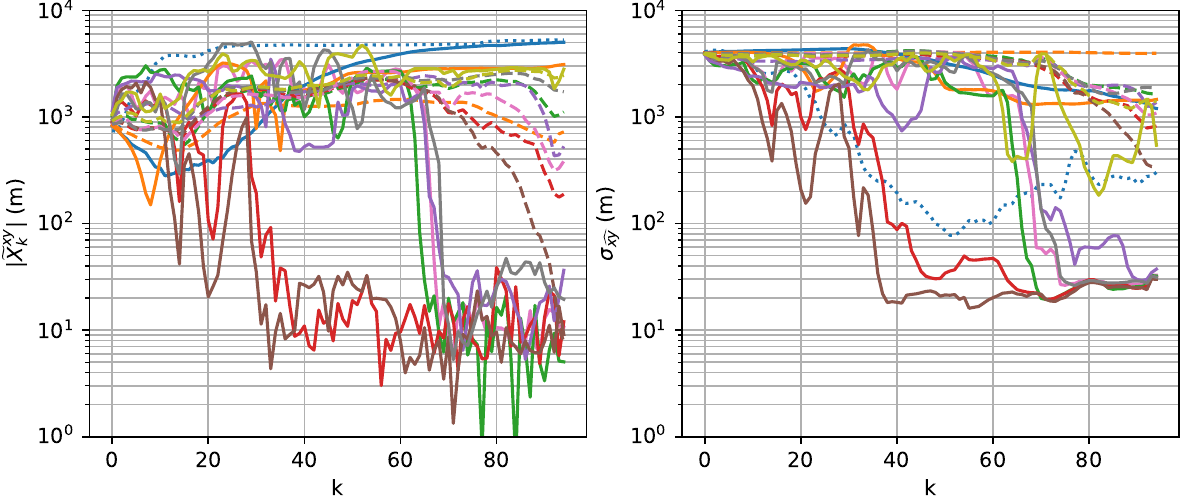} %Dataset 6 = klockrike-2021-01-19-000
        \caption{Dataset 6}
        \label{fig:lsl-dataset6_graph}
    \end{subfigure}%
    \hfill
    \begin{subfigure}[t]{0.5\textwidth}
        \centering
        \includegraphics[width=\linewidth,valign=t]{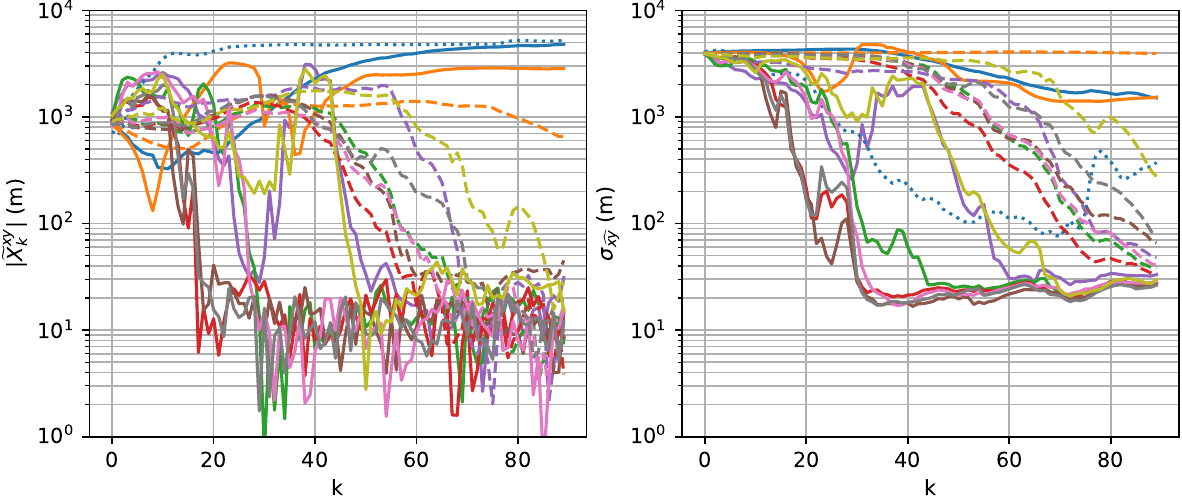} %Dataset 7 = klockrike-2021-01-19-001
        \caption{Dataset 7}
        \label{fig:lsl-dataset7_graph}
    \end{subfigure}%
    ~
    \begin{subfigure}[t]{0.5\textwidth}
        \centering
        \includegraphics[width=\linewidth,valign=t]{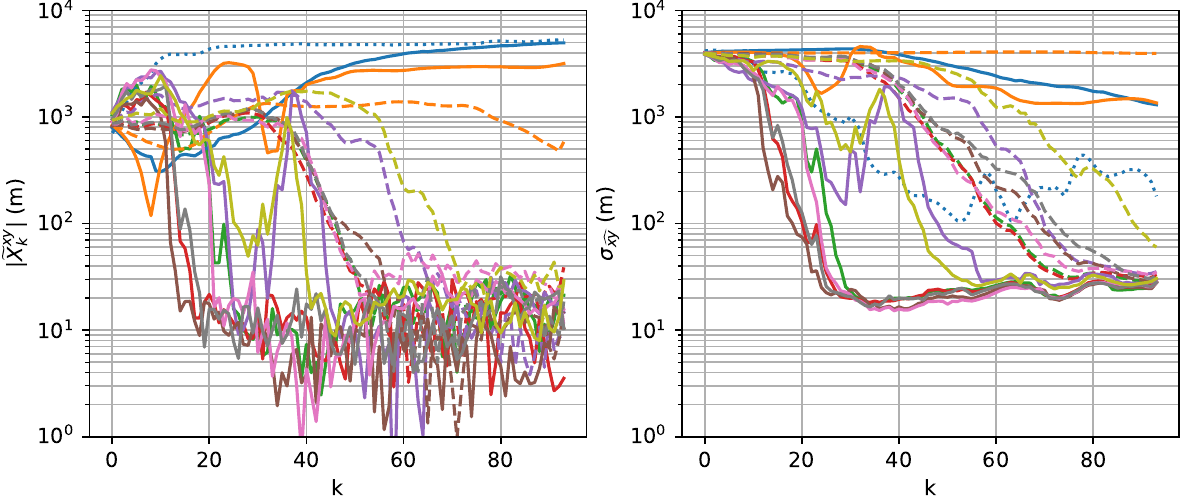} %Dataset 8 = klockrike-2021-01-19-002
        \caption{Dataset 8}
        \label{fig:lsl-dataset8_graph}
    \end{subfigure}
    \caption{Translation error $|\widetilde{X}^{xy}_k|$ and standard deviation of translation $\sigma_{\widehat{xy}}$ as function of update index $k$ in different datasets using all methods. Line color represents descriptor computation method, line style (dashed, dotted, solid) represents likelihood computation method. Logarithmic scale. Updates are made approximately every \translationbetweenupdates{} meters of travel.}
    \label{fig:lsl-errorplots}
\end{figure*}

We evaluate localization performance for each step $k$ after completing prediction, map matching and AHRS update at that step. We define that a localization solution has converged when the translation standard deviation $\sigma_{\widehat{xy}}$ is less than 100 m. We compute the mean number of updates to convergence $\bar{k}_{c}$ and mean translation error in converged state, $\bar{\widetilde{X}}^{xy}_{c}$, for each tested model and likelihood conversion method.  We compute the proportion of flights where each compared solution converges, $p_c$, and tabulate results in Table~\ref{tab:lsl-time_to_convergence_and_error}. In addition, we visualize the translation error and standard deviation of \xy{} translation in each case in Figure~\ref{fig:lsl-errorplots}.

\begin{table}
\footnotesize
\caption{Probability of convergence $p_c$, time to convergence $\bar{k}_{c}$ and mean localization error after convergence $\bar{\widetilde{X}}^{xy}_{c}$ when using our methods with various design choices and when comparing to reference method.}
    \centering
    \begin{tabular}{p{0.25\linewidth} p{0.15\linewidth} p{0.1\linewidth} p{0.1\linewidth} p{0.1\linewidth}}
\toprule
\modeltypetableheader & \likelihoodconversiontableheader & $p_{c}$ & $\bar{k}_{c}$ & $\bar{\widetilde{X}}^{xy}_{c} (m)$ \\
\midrule
%% these results generated with git hash 6d9d37d21e56816de3a1bff54acc46a3d701c609 on 11/30/2022, 08:14:40
Ours, Caps, D=8 & Linear & 0.875 & 60.9 & 15.7 \\ 
Ours, Caps, D=16 & Linear & 0.875 & 56.3 & 13.3 \\ 
Ours, Caps, D=32 & Linear & 0.875 & 58.9 & 13.9 \\ 
Ours, FCN, D=8 & Linear & 0.75 & 65.2 & 21.1 \\ 
Ours, FCN, D=16 & Linear & 0.875 & 61.1 & 13.9 \\ 
Ours, FCN, D=32 & Linear & 0.875 & 63.1 & 12.9 \\ 
Ours, FCN, D=128 & Linear & 0.875 & 62.3 & 11.2 \\ 
Ours, Caps, D=8 & Bayesian & 1.0 & 44.4 & 18.3 \\ 
Ours, Caps, D=16 & Bayesian & 1.0 & 35.2 & 12.8 \\ 
Ours, Caps, D=32 & Bayesian & 1.0 & 30.8 & 15.0 \\ 
Ours, FCN, D=8 & Bayesian & 0.875 & 43.0 & 18.7 \\ 
Ours, FCN, D=16 & Bayesian & 1.0 & 36.1 & 14.6 \\ 
Ours, FCN, D=32 & Bayesian & 1.0 & 33.2 & 15.7 \\ 
Ours, FCN, D=128 & Bayesian & 1.0 & 23.2 & 12.6 \\ 
BRM & Linear & 0.0 & N/A & N/A \\ 
BRM & Bayesian & 0.0 & N/A & N/A \\ 
abBRIEF & Mantelli & 0.625 & 63.8 & 4112.2 \\ 
abBRIEF & Bayesian & 0.0 & N/A & N/A \\ 

\bottomrule
    \end{tabular}
    \label{tab:lsl-time_to_convergence_and_error}
\end{table}

% \begin{table}
% \footnotesize
% \caption{Probability of convergence $p_c$, time to convergence $\bar{k}_{c}$ and mean localization error after convergence $\bar{\widetilde{X}}^{xy}_{c}$ when using our methods with various design choices and when comparing to reference method.}
%     \centering
%     \begin{tabular}{p{0.25\linewidth} p{0.1\linewidth} p{0.1\linewidth} p{0.1\linewidth}}
% \toprule
% \modeltypetableheader  & $p_{c}$ & $\bar{k}_{c}$ & $\bar{\widetilde{X}}^{xy}_{c} (m)$ \\
% \midrule
% %% these results generated with git hash 6d9d37d21e56816de3a1bff54acc46a3d701c609 on 11/30/2022, 08:14:40
% Ours, Caps, D=8 & 1.0 & 44.4 & 18.3 \\ 
% Ours, Caps, D=16 & 1.0 & 35.2 & 12.8 \\ 
% Ours, Caps, D=32 & 1.0 & 30.8 & 15.0 \\ 
% Ours, FCN, D=8 & 0.875 & 43.0 & 18.7 \\ 
% Ours, FCN, D=16 & 1.0 & 36.1 & 14.6 \\ 
% Ours, FCN, D=32 & 1.0 & 33.2 & 15.7 \\ 
% Ours, FCN, D=128 & 1.0 & 23.2 & 12.6 \\ 

% \bottomrule
%     \end{tabular}
%     \label{tab:lsl-time_to_convergence_and_error_cutted}
% \end{table}

\subsubsection{Statistical significance testing}
To explore what significance our results show in the various parameters we have used in the experiment configurations, we run a type II analysis of variance test on time to convergence and on localization error after convergence. We consider the embedding dimension (8, 16, 32, 128), projection module type (fully connected or capsules) and likelihood computation method (linear or bayesian) and combinations of these parameters. We use $p=0.05$ as limit for significance.

For time to convergence, we reject null hypothesis that chosen likelihood method is not significant ($p=4.2*10^{-16}$). For localization error after convergence, we reject null hypothesis that the parameter is not significant for
embedding dimension ($p=2.5*10^{-14}$),
projection module type ($p=2.1*10^{-2}$),
likelihood method ($p=3.7*10^{-4}$)
and combination of embedding dimension and projection module type ($p=5.0*10^{-5}$) and combination of embedding dimension and likelihood method ($p=1.3*10^{-2}$).

\subsection{Real-time experiment}
\label{sec:lsl-realtime_experiment}

\begin{figure}
\centering
\includegraphics[width=0.7\linewidth]{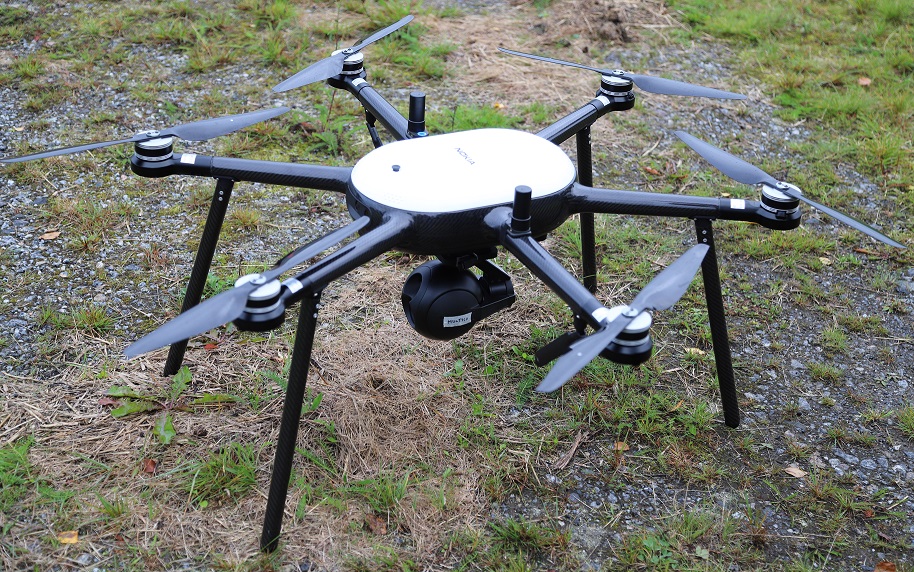}
\caption{UAV used in real time experiment}
\label{fig:lsl-image_of_nokia_drone}
\end{figure}

To understand the applicability of our solution to an embedded system, we implemented a version of our algorithm running in real time on an embedded computer on an example UAV. We use the Nokia Drone Networks drone (see Figure~\ref{fig:lsl-image_of_nokia_drone}) carrying a camera gimbal. The drone was equipped with an NVidia Jetson Nano computer on which the localization algorithm was run. The algorithm was implemented in Python and we used ROS \citep{ros} for inter-process communication. To accommodate the smaller observable ground footprint due to the narrower field of view of the drone camera at the selected flight altitude and observation parameters (50 m altitude at 45 degree camera pitch), we retrained a model with Resnet50 + FCN, D=16 architecture to work on 40m by 40m images at 1 m/px resolution. The size of the operating region and map was \kauhavamapdimensions{}. We used VINS-Mono \citep{8421746} as the VIO algorithm. We used the output of a non-GNSS-corrected AHRS algorithm implemented on the drone flight controller for heading updates.

The mean network inference time (run on CPU) was 1.02 s, mean prediction time was 0.83 s, map matching took 1.18 s and AHRS update took on average 0.12 s, and all steps took on average 3.15 s. The algorithm was configured to perform an update every 40 meters of travel, while the drone was flying at 5 m/s. There is room for speedup by proper parallelization of the algorithm. The algorithm was run fully onboard the UAV.

Some localization errors appeared over terrain patches with very uniform colouring due to scale drift and rotation estimation errors of the VIO algorithm which are not considered in our odometry noise model. In addition, the AHRS heading estimate from the flight controller appeared to contain non-Gaussian heading errors up to 15 degrees; to accommodate this, we used $\sigma_{v} = 60\degree$ in this experiment.

The memory footprint of the precomputed map in this experiment was 460.8 MB. On a \maparea{} map, the precomputed map memory requirement for D=8, D=16, D=32 and D=128 are 3.5 GB, 7.0 GB, 14.0 GB and 55.9 GB, respectively, setting further practical constraints for embedded implementation at very large scale. 

This experiment shows that the algorithm can be run in real time on an embedded computer carried by a drone in an operating region of typical size for drone operations.

\section{Discussion}
\label{sec:lsl-discussion}

To solve the wake-up robot problem at the presented scale, we derived a solution based on a recursive point mass filter instead of the more common particle filter approach. We believe this avoids issues with particle depletion, which is important in cases of significant initial uncertainty, ambiguity of terrain and potential intermittent but not random correspondence mismatches between observations and maps.

An architecture where an image observation is projected into an embedding space, with a major reduction in data dimensionality, is a key enabler for fast and memory-efficient similarity comparisons over a large number of hypotheses. Our approach for projecting data from different source domains (\ie{} UAV camera images and orthophoto maps or satellite images) into a common domain enables the use of learned descriptors. With our approach, extensive training data covering all expected variation is required in only one source domain without need for labeling of data. Our approach of using a learnable embedding appears to be efficient for localization over large areas containing natural and built environments, in scenarios where also significant seasonal appearance change occurs between flights. Other existing methods, \eg{} a handcrafted descriptor approach (abBRIEF) or learned description trained to detect pre-specified semantics (BRM), do not converge at this scale.

Bayesian likelihood conversion showed the greatest effect in time to convergence, in comparison to the more common linear approximation. The results in Table~\ref{tab:lsl-time_to_convergence_and_error} appear to hint at the possibility that increasing embedding dimensionality leads to faster convergence, but further experimentation would be required to verify statistical significance. For localization error afer convergence, embedding dimensionality was found to be one of the statistically significant parameters. Results tabulated on Table~\ref{tab:lsl-time_to_convergence_and_error} and error plots in Figure~\ref{fig:lsl-errorplots} seem to suggest that there may be a lower bound on localization error that appears independent from dimensionality and is most likely a result of other design choices and the characteristics of the operating environment. In other words, it appears that low-dimensionality descriptors are an efficient way of expressing what is important for localization and dimensionality of description is not the main hindering factor, what comes to localization error.

In order to extract low-dimensionality descriptors, we tested both traditional fully connected and capsule layers.
Our hypothesis was initially that capsule layers would improve localization performance, extracting better encodings thanks to their ability to model part-whole relationships and robustness to novel viewpoints.
Employing capsule layers lead to a slight improvement in error after convergence in comparison to fully connected layers while using less trainable parameters, as also stated in~\citep{hinton-dr}. In fact, for 16-dimensional embeddings, with fully connected layer, the network has 57M trainable parameters, while with capsule layers, it has ~42M trainable parameters. Since this improvement is only marginal, we did not conduct experiments with 128-dimensional capsule embeddings, as CapsNets are also well known to be  computationally very demanding in terms of memory consumption, training and inference times.

Experiments demonstrate that the proposed approach is suitable for real time implementation on a flying platform at a typical scale of UAV operations. Tailoring design parameters allows the implementation of the solution on a very resource-constrained platform and running it in real time together with an odometry system. Future work for reductions in computational requirements may include \eg{} North-aligning UAV observations before map matching and prediction, thereby removing the need for heading estimation from the filtering problem, which would significantly (60-fold) reduce computational and memory requirements of map matching and prediction. However, in this approach, a dependable heading estimation method would be required at all times. For reducing running time of forward pass of the neural network during flight, pruning and quantization strategies can be considered.

Our work focuses on the UAV application context. In considering applicability of the proposed approach to ground vehicular applications, it is likely that a more elaborate means for orthoprojecting the camera observations into a top-down view is required, since the simple assumption of planar ground will lead to severe slanting of objects protruding from terrain, if the camera is moving very close to ground. However, we postulate that the concept of separating the problem of geometric appearance change due to difference in viewpoint and visual appearance change due to natural changes in environment apperance, and solving those problems in sequence, enables projecting the latter of the two problems into a domain where training data for learning-based approaches are plentiful across natural apperance change. In the traditional VPR problem setting, current solutions focus on solving both of these subproblems simultaneously, inescapably leading to the costly need to collect extensive datasets that contain both these variation types. We postulate our two-step approach, given a more roubst novel view synthesis method, would be an efficient method in extreme cross-view localization cases in ground vehicular applications, and consider this a potential future research direction.

To enable error recovery, the solution provides a measure of position uncertainty by computation of standard deviation. Upon detection of prolonged high standard deviation, this enables the triggering of reinitialization of the pose estimator. The ability to detect localization failures and recover from the loss of location information on a large scale paves way for a failure-aware, failsafe UAV localization system.

\section{Summary}
\label{sec:lsl-lsl-conclusions}

We have shown that the approach utilizing our map matching method together with a point mass filter is able to resolve UAV pose even in the case of highly uninformed initialization, in conditions of significant seasonal appearance change between UAV image and map, even when flying over areas with natural ambiguity. We experiment with a map size of \maparea{} where the proposed solution converges to a localization error of \translerrorafterconvergence{} on average in \numberofstepstoconvergence{} updates, depending on the chosen architecture, while reference methods are not able to converge to the correct pose under the same circumstances. All of these contributions show that real-time localization is possible on a large scale. Going beyond the demonstrated \maparea{} will require being able to represent even larger hypothesis spaces. 
Addressing this challenge will potentially require future work in hierarchical models in order to retain the favorable characteristics of high spatial accuracy, extreme initial uncertainty, and complete coverage of the hypothesis space.

\cleardoublepage
%*****************************************
\chapter{Encoding rotation representations of synthetic datasets in quaternions-based deep learning models}\label{ch:quaternions_capsnets}
%*****************************************
This chapter represents the initial stages of our research, and it has been a collaborative effort with Alessandro Grassi, an undergraduate student. This chapter marks the beginning of our pursuit to achieve more robust and generalized results, bridging the gap between synthetic and real-world datasets.
\section{Introduction}\label{sec:quaternions_introduction}
Over recent years, although artificial intelligence research has been propelled by the continuous development of novel models and advanced algorithms, one challenge remains: data. In fact, neural networks still need a massive amount of training data to generalize to new patterns, and finding suitable datasets is not trivial, assuming they are available at all. To address this issue, we leverage computer graphics techniques to generate synthetic datasets containing images of different meshes displayed in thousands of views. Each image is labeled with the object class and rotation using quaternions. 

When representing rotations, employing quaternions offers several advantages over using rotation matrices. Firstly, unlike rotation matrices, Euler angles, and exponential maps, quaternions do not suffer from gimbal lock~\cite{grassia, pavllo:quaternet:2018}. Secondly, to properly represent a rotation, rotation matrices must be orthogonal, and quaternions must be unit vectors. However, neither of these properties can be guaranteed to be maintained after weight updates. At this juncture, maintaining orthogonality in rotation matrices becomes more challenging, while normalizing quaternions proves to be a better approach. Furthermore, previous research has demonstrated the success of quaternions in restoring spatial relationships~\cite{Matsui2004} and extracting them from images~\cite{trabelsi2018deep, gaudet2017deep}. Lastly, using quaternions reduces the number of parameters required to represent a 3D rotation from 9 to 4.

Since we are dealing with rotation representations and capsule networks can capture spatial relationships and poses of objects, one question arises: \textit{can we exploit capsule vectors to encode quaternions?} \citeauthor{quatcapsnet} \citep{quatcapsnet} incorporated the aforementioned properties of quaternions and capsules, constraining the pose representation to 3D rotations between capsules since convolutional connections are translation invariant. In the object coordinate system, the intrinsic rotations between parts and whole are constant from any viewpoint. Thus, they exploit this rotation to achieve better generalization to novel viewpoints. Furhermore, \citeauthor{quatcapsnet-pointclouds} \citep{quatcapsnet-pointclouds}  built a capsule network that disentangles geometry from pose, paving the way for more informative descriptors and a structured latent space. They architecture allows joint object classification and orientation estimation without explicit supervision of rotations. In contrast with these works, our final goal is to exploit synthetic datasets annotated with quaternion labels to generalize to real-world datasets. This chapter is the first preliminary step towards achieving this goal, where we present a method to modify network architectures to predict rotations encoded as quaternions.

This chapter compares the performances and the limitations of several neural network architectures, including multilayer perceptrons, convolutional and capsule networks, in predicting both the orientation of rendered objects and their labels. This is achieved by grouping output neurons to encode the probability of existence for each class and a valid quaternion (unit length quaternion so that it represents a valid rotation).

\section{Background on quaternions}\label{sec:quaternions_background}
Quaternions were developed in 1843 by William Rowan Hamilton and used to represent rotations and orientations. Even if Euler angles are more human-understandable, quaternions are computer-efficient and resolve several problems like Gimbal lock and ambiguity. Hamilton’s quaternions have many applications besides computer graphics. For example, they are widely used in spacecraft and computer graphics.

Quaternions use complex numbers that can be interpreted as unit-vectors. 
A quaternion $\vq$ is a 4-dimensional vector that can be written as
\begin{equation}
    \vq = xi + yj + zk + w
\end{equation}

where $x$, $y$, $z$, and $w$ are real numbers, and $i$, $j$, and $k$ are three imaginary axes. 

According to Hamilton's famous expression:
\begin{gather}
    i^2 = j^2 = k^2 = -1, \\
    ij = -ji = k, k = -kj = i, ki = -ik = j.
\end{gather}

Quaternions are also represented as an ordered pair
\begin{equation}
    \vq = [s, \vv]
\end{equation}
where $s$ is the scalar part and $\vv$ is the vector part.

\section{Methodology}\label{sec:quaternions_methodology}
\subsection{Datasets}
Each dataset was generated with Blender through a Python script that takes as input a configuration file that determines the composition of the scene. The configuration file is responsible for defining parameters such as:
\begin{itemize}
    \item \textbf{Background} — Color, transparency
    \item \textbf{Render} — Engine, CUDA, and resolutions
    \item \textbf{Camera} — Name, location, and rotation
    \item \textbf{Lights} — Energy, location, rotation, type, energy, and color
    \item \textbf{Meshes} — Start location, start rotation, and object directory
\end{itemize}

In this work, each 3D model is rotated on each axis of 20 degrees, for a total of 6859 images on each object. Each image has a resolution of $128 \times 128$ pixel with a black background, and it is paired with a file that contains the rotation of the object expressed in quaternions.

Blender's fixed scene for generating datasets includes the following details: a camera in position $(x = 4, y = 0, z = 0)$ pointing to $(x = 0, y = 0, z = 0)$; a first point light in position $(x = 6, y = 6, z = 4)$ with 1000 energy; a second light point in position $(x = 0, y = 0, z = 0)$ with 1000 energy. All renderings are made with the EEVEE rendering engine integrated into Blender to significantly increase performance at the expense of a slight loss of detail, such as the absence of advanced ray tracing techniques. 

\section{Cubes and Tetrahedrons}
To verify our method, we first built a dataset with simple shapes, such as the cube and the tetrahedron. Table~\ref{tab:quaternions_cubetetrahedron} shows some details of this dataset.

\begin{table}[h]
\footnotesize
    \centering
    \begin{tabular}{ c|c }
        \multicolumn{2}{c}{\textbf{CubeTetrahedron}} \\ [0.2ex] \hline
        Resolution & 128 $\times$ 128 $\times$ 3 \\ 
        Classes & 2 \\ 
        Objects per class & 1 \\
        Images per objects & 6859  \\
        Total images & 13718 \\
    \end{tabular}
    \caption{Details of the CubeTetrahedron dataset.}
    \label{tab:quaternions_cubetetrahedron}
\end{table}

\begin{figure}[h]
    \centering
    \includegraphics[width=0.4\columnwidth]{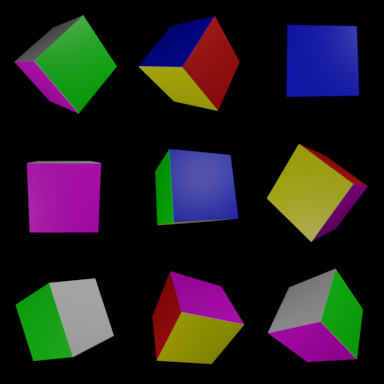}
    \includegraphics[width=0.4\columnwidth]{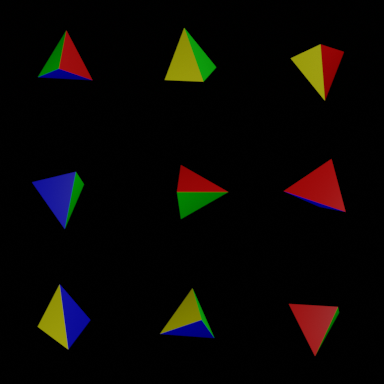}
    \caption{Some images of the CubeTetrahedron dataset without borders, on the left the cube, and on the right the tetrahedron.}
    \label{fig:quaternions_cubetetrahedronnoborders}
\end{figure}

\begin{figure}[h]
    \centering
    \includegraphics[width=0.4\columnwidth]{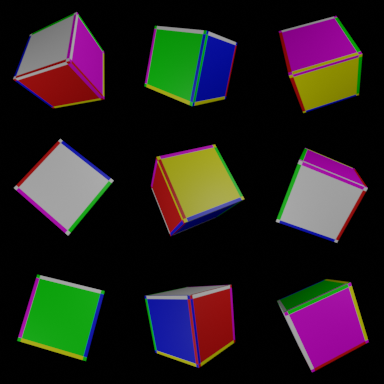}
    \includegraphics[width=0.4\columnwidth]{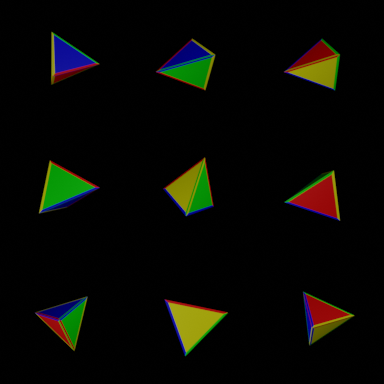}
    \caption{Some images of the CubeTetrahedron dataset with borders.}
    \label{fig:quaternions_cubetetrahedronborders}
\end{figure}

However, predicting rotations on the $x$-axis when the model shows to the camera only one face (Figure~\ref{fig:quaternions_cubetetrahedronnoborders}) is impossible because multiple solutions can exist. For this reason, we enriched the model faces with visual information about the adjacent faces, as shown in Figure~\ref{fig:quaternions_cubetetrahedronborders}.

With this solution, a neural network can predict the rotation of a single face in a deterministic way.

\section{TinyShapeNet}

\begin{figure}[h]
    \centering
    \includegraphics[width=1\columnwidth]{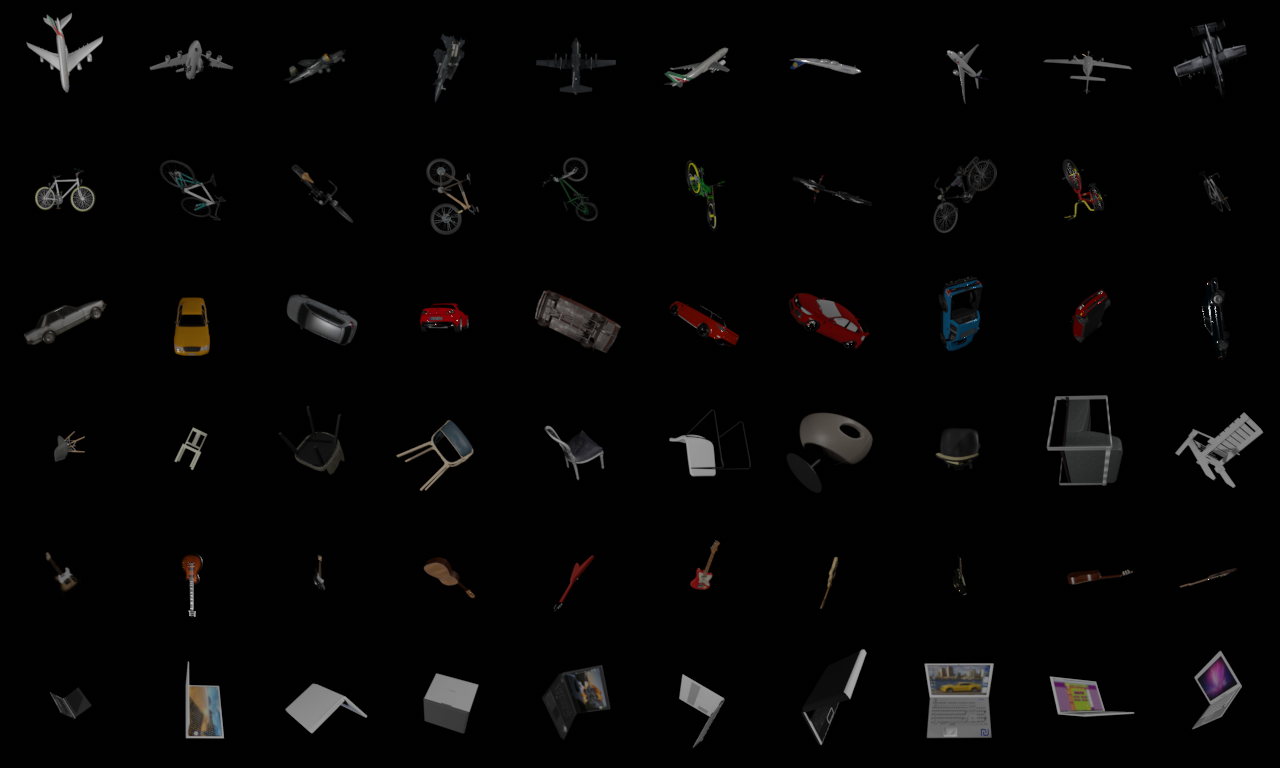}
    \caption{Some samples taken from the TinyShapeNet dataset.}
    \label{fig:quaternions_tinyshapenet}
\end{figure}

The main dataset of this chapter is partially based on the ShapeNet \cite{shapenet2015} dataset. It is a richly-annotated, large-scale dataset of 3D objects developed by researchers from Stanford University. The repository contains over 300 million models, with 220,000 classified into 3,135 classes. However, we used a small subset of ShapeNet with 60 objects divided into 6 classes: airplanes, cars, chairs, guitars, and laptops. Table~\ref{tab:quaternions_tinyshapenet} shows some details of this dataset, while Figure~\ref{fig:quaternions_tinyshapenet} shows some samples.

The models used were adjusted with Blender to achieve a more or less uniform size across all models, large enough not to lose detail and to fit the entire object into the scene. Also, each model was modified to get the same default pose across all meshes of the same class. However, no objective definition of ``default pose'' applies to every object, so a subjective interpretation was given based on the class of the objects in question.

\begin{table}[h]
\footnotesize
    \centering
    \begin{tabular}{ c|c }
        \multicolumn{2}{c}{\textbf{TinyShapeNet}} \\ [0.2ex] \hline
        Resolution & 128 x 128 x 3 \\ 
        Classes & 6 \\ 
        Objects per class & 10 \\
        Images per objects & 6859  \\
        Total images & 411540 \\
    \end{tabular}
    \caption{Details of the TinyShapeNet dataset.}
    \label{tab:quaternions_tinyshapenet}
\end{table}

For this work, the dataset was also generated at 1024 $\times$ 1024 $\times$ 4 resolution, which is a high-resolution version with an alpha channel (transparency) to take advantage of data augmentation. 
% However, it has not been used due to the fact that the networks would have taken too long to train.
\subsection{Training}
% The networks have been done using the Python programming language and PyTorch. An open-source machine learning library based on the Torch library, used for applications in Machine Learning.
Neural networks employed in this chapter are trained by means of two losses: spread loss and render loss. The spread loss is the classification loss, used to predict the object class. The render loss is used to compute the similarity between the target and predicted quaternions. In fact, each network outputs $J$ 4-dimensional vectors that represent the predicted quaternions (one for each object label $j$) along with the class probabilities.

The total spread loss $S$ is computed as the sum of the spread loss $S_j$ for the $j$-th class 
\begin{equation}
    \mathcal{S} = \sum\limits_{j} \mathcal{S}_{j} \quad\mathrm{where}\quad \mathcal{S}_j = (\text{max}(0, m-(a_t - a_j)))^2,
\end{equation}
 $a_j$ represents the activation of the capsule of class $j$ and $a_t$ the activation of the target class. If the margin between the true label and the wrong class is smaller than $m$, we penalize it by the $m-(a_t - a_j)$ square. 

The render loss is used to compare quaternions. To check if two quaternions represent the same orientation (namely, the object would be facing the same way) we define the render loss
\begin{equation}
    \mathcal{R} = 1 - |\langle \vu_j, \vq \rangle|,
\end{equation}

where $\vu_j$ is the prediction for the class capsule $j$ and $\vq$ the target quaternion.
Since $q$ and $-q$ represent the same orientation and they are unit vectors, we compute the absolute value of their product to make sure they are the same rotation.

The total loss is defined as 
\begin{equation}
	\mathcal{L} = \mathcal{S} + \lambda \mathcal{R}.
\end{equation}

\subsection{Architectures}
Here, we describe the neural network architectures used in this chapter, namely, multilayer perceptron (MLP), convolutional neural network (CNN), and capsule network with Expectation-Maximization routing~\cite{hinton-em} (EM-CapsNet).

The EM routing algorithm computes the activation and pose of the capsules in layer $l+1$, finding a cluster of similar votes of capsules in layer $l$ using the Expectation-Maximization algorithm. In our case, we apply a $l_2$ normalization to the class capsules vectors to compute the predicted quaternions. We use directly the activations computed by the EM routing as class probabilities. For a more detailed explanation of using EM routing instead of other routing algorithms, see Section~\ref{sec:quaternions_capsnet}.
Besides, for the MLP and ConvNet architectures, given the embedding of the last layer, we apply $l_2$ normalization to determine the quaternion and the squash function described in Equation~\ref{eq:squashing} to shrink the embedding so that its norm determines the class activation.

%Finally, all networks use ReLU as their activation function. 

\subsection{Multilayer perceptron}
The multilayer perceptron is the simplest network of the three used in the experiments. As shown in Table~\ref{tab:quaternions_MLP}, our implementation for the TinyShapeNet dataset involves an input of 28 $\times$ 28 $\times$ 3, a first hidden layer of 1024 neurons connected to a second layer of 512 neurons, which is in turn connected to a third layer of 256 neurons. Then, ending with 24 (4 $\times$ 6 classes) outputs. 

\begin{table}[h]
\footnotesize
    \centering
    \begin{tabular}{ c|c|c|c }
        \multicolumn{2}{c|}{\textbf{Layer}} & \textbf{Size} & \textbf{Act./Norm.} \\ [0.2ex] \hline
        Input & Image & 28$\times$28$\times$3 & - \\ 
        1 & Hidden & 1024 & ReLU \\ 
        2 & Hidden & 512 & ReLU \\ 
        3 & Hidden & 256 & ReLU \\ 
        Output & FC & 4 $\times$ 6 classes & squash \& $l_2$ \\ 
    \end{tabular}
    \caption{The multilayer perceptron architecture (TinyShapeNet) with corresponding layer type, tensor size, activation and normalization functions.}
    \label{tab:quaternions_MLP}
\end{table}

For the CubeTetrahedron dataset test, we used an MLP with a simpler structure, with 300 neurons at the first hidden layer, and 100 neurons at the second.

\subsection{Convolutional network}
As shown in Table~\ref{tab:quaternions_Conv}, the architecture of the convolutional network is composed of a convolutional layer of size 20$\times$20, with 256 channels, a 9$\times$9 filter, and a stride of 1. Then, the result of the first layer passes to a second convolutional layer of size 6$\times$6, consisting of 128 channels, with a kernel size of 9$\times$9 and a stride of 2. 
\begin{table}[h]
\footnotesize
    \centering
    \begin{tabular}{c|c|c|c|c|c|c}
    
        \multicolumn{2}{c|}{\textbf{Layer}} & \textbf{Size} & \textbf{Ch.}  & \textbf{Kernel} & \textbf{Stride} & \textbf{Act./Norm.} \\ [0.2ex] \hline
        Input & Image & 28$\times$28 $\times$ 3 & - & - & - & - \\ 
        1 & Convolution & 20$\times$20 & 256 & 9$\times$9 & 1 & ReLU \\ 
        2 & Convolution & 6$\times$6 & 128 & 9$\times$9 & 2 & ReLU \\ 
        3 & Avg. Pooling & 1$\times$1  & - & - & - & - \\ 
        Output & FC & 4$\times$6 classes & - & - & - & squash \& $l_2$ \\ 
    \end{tabular}
    \caption{The architecture of the convolutional network (TinyShapeNet) with corresponding layer type, tensor size, activation and normalization functions.}
    \label{tab:quaternions_Conv}
\end{table}

\subsection{Capsule network}\label{sec:quaternions_capsnet}
\begin{figure}[h]
    \centering
    \includegraphics[width=1\columnwidth]{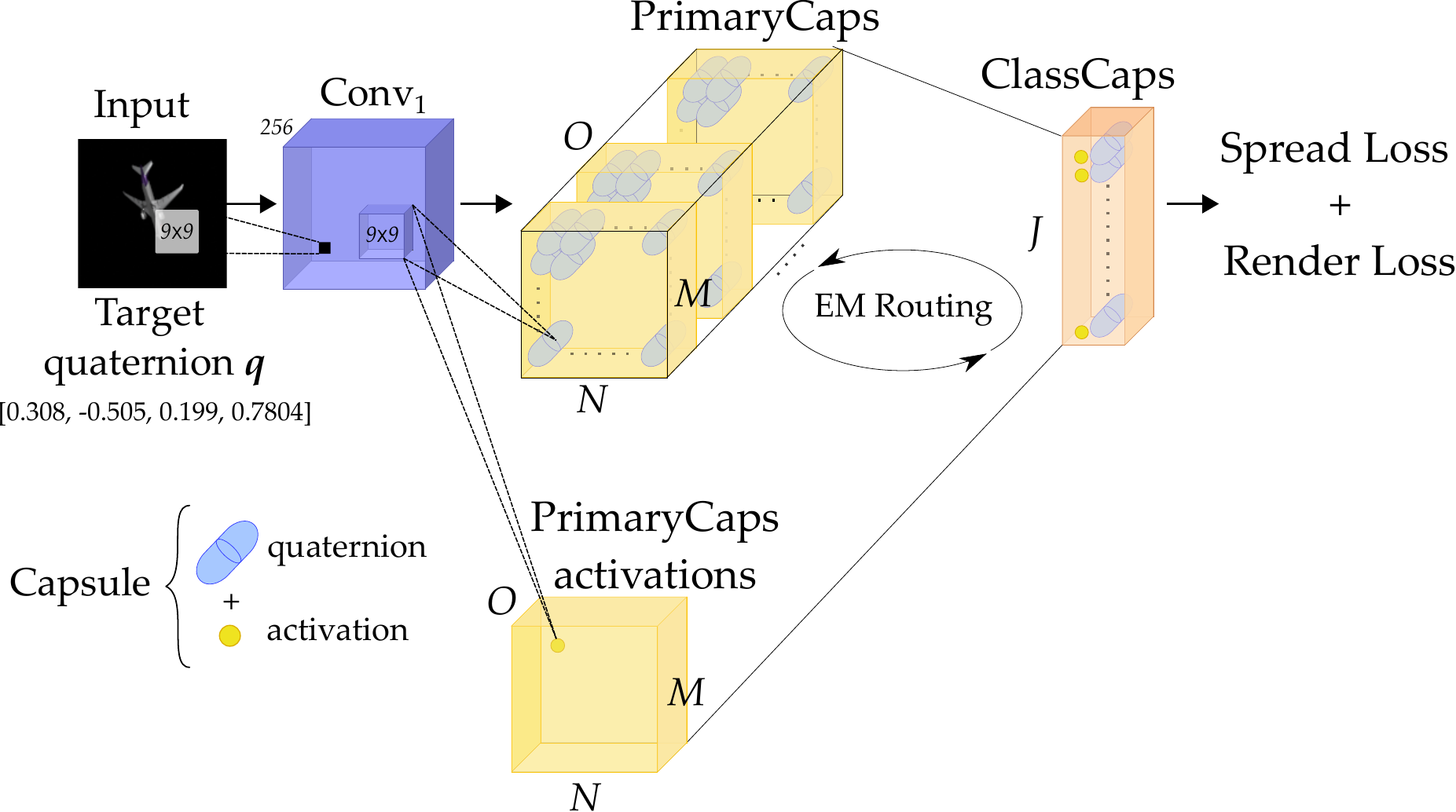}
    \caption{Visual representation of capsule network architecture, where a capsule encodes a quaternion representation. This model comprises a convolutional layer, a PrimaryCaps layer, which maps neurons to quaternions, and a ClassCaps layer, which outputs a capsule-quaternion for each class. The spread loss is computed using all the class capsule activations, while the render loss only considers the capsule with the highest activation, whose vector represents the predicted quaternion.}
    \label{fig:quaternions_capsnet}
\end{figure}

Table~\ref{tab:quaternions_capsnet} depicts the architecture used in this work for the CapsNet model. In this model, a capsule is implemented as a 4-dimensional vector to match the quaternion dimension. The first layer is a traditional ReLU convolutional layer with a size of 20$\times$20, a kernel of 9$\times$9 and stride of 1. Then, the result of the kernel is passed to the PrimaryCaps layer, which is responsible for computing both the poses and the activations of the primary capsules. This layer applies two convolutional layers (one for the poses and one for the activations) and a reshape operation to convert neurons into capsule-quaternions. Then, the primary capsules are given as input to the CapsClass layer, where, thanks to the EM routing algorithm, it outputs the quaternions predictions for each class and the corresponding probability values. We chose the EM routing instead of the dynamic routing (Algorithm~\ref{alg:dynamicrouting}) because, in the latter one, the activation of a capsule is computed using the squash function, which normalizes the vector to a length between 0 and 1. However, in our case, since we are dealing with quaternions that are unit vectors, a separation between class prediction and quaternion prediction is needed to model the additional complexity of the proposed dataset. A visual representation of the architecture is shown in Figure~\ref{fig:quaternions_capsnet}
\begin{table}[h]
\footnotesize
    \centering
    \begin{tabular}{c|c|c|c|c|c|c}
    
        \multicolumn{2}{c|}{\textbf{Layer}} & \textbf{Size} & \textbf{Ch.} & \textbf{Kernel} & \textbf{Stride} & \textbf{Act.} \\ [0.2ex] \hline
        Input & Image & 28$\times$28$\times$3 & - & - & - & - \\ 
        1 & Convolution & 20$\times$20 & 256 & 9$\times$9 & 1 & ReLU \\ 
        2 & PrimaryCaps (activations) & 32$\times$6$\times$6 & 32 & 9$\times$9 & 2 & sigmoid \\
        3 & PrimaryCaps (poses) & 4$\times$32$\times$6$\times$ 6 & 128 & 9$\times$9 & 2 & $l_2$ \\
        Output & CapsClass (activations) & 6 classes & -  & -  & - & EM \\ 
        Output & CapsClass (poses) & 4$\times$6 classes & -  & -  & - & EM \\ 
    \end{tabular}
    \caption{The architecture of the capsule network with corresponding layer type, tensor size, activation and normalization functions.}
    \label{tab:quaternions_capsnet}
\end{table}

\section{Experiments}\label{sec:quaternions_results}
This section shows the experimental results achieved on CubeTetrahedron and TinyShapeNet datasets. 
Each network uses Adam as an optimization algorithm, with a learning rate of $0.001$. 
The hyperparameter $m$ of the spread loss starts as $0.2$ and linearly increases by 0.1 after each epoch training. It stops growing after reaching the maximum of $0.9$. Starting at a lower margin helps the training to avoid too many dead capsules during the early phase \cite{hinton-em}. We weight the render loss by $\lambda = 0.1$.
We rescaled the images to a 28 $\times$ 28 $\times$ 3 resolution to dramatically decrease training times at the expense of a small loss of detail.
 
The training was executed on an Nvidia GTX 1080 GPU.   
\subsection{Cube and tetrahedron}
\begin{figure}[h]
    \centering
    \includegraphics[width=0.4\columnwidth]{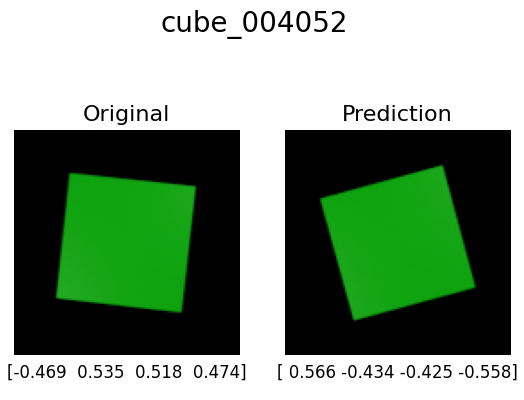}
    \includegraphics[width=0.4\columnwidth]{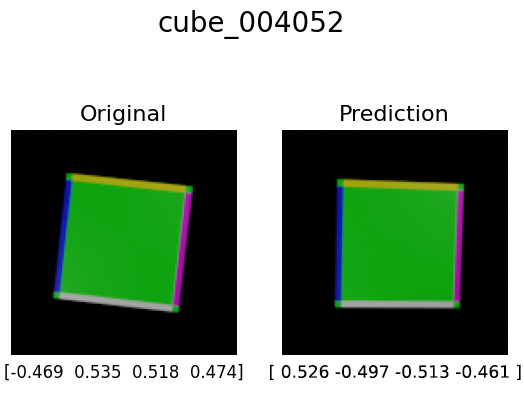}
    \caption{Visual differences in predicted quaternions between the two versions of the CubeTetrahedron dataset.}
    \label{fig:quaternions_visualdiff}
\end{figure}

We used CubeTetrahedron as a toy dataset to evaluate the feasibility of our method. We trained an MLP model on the dataset without and with faces. Each face enriches the cube or tetrahedron with visual information about the adjacent face. In Figure~\ref{fig:quaternions_visualdiff} we can see that the network, when given as input the face additional information, is able to match the ground truth quaternion better.

\subsection{TinyShapeNet}
\begin{figure}[h]
    \centering
    \includegraphics[width = 280pt]{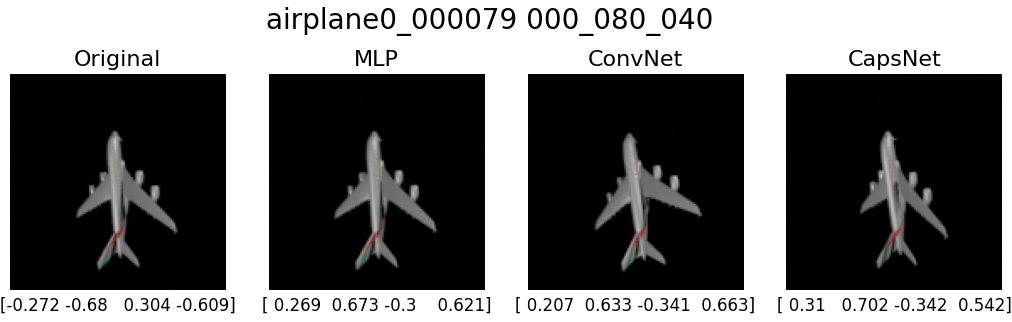} \\
    \vspace{0.5cm}
    \includegraphics[width = 280pt]{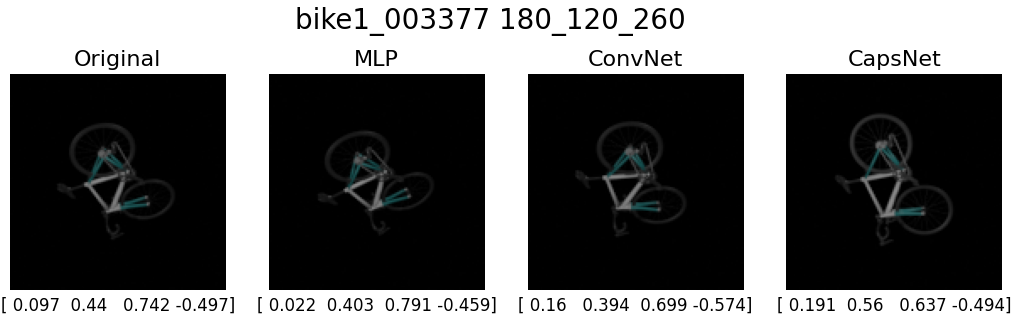} \\
    \vspace{0.5cm}
    \includegraphics[width = 280pt]{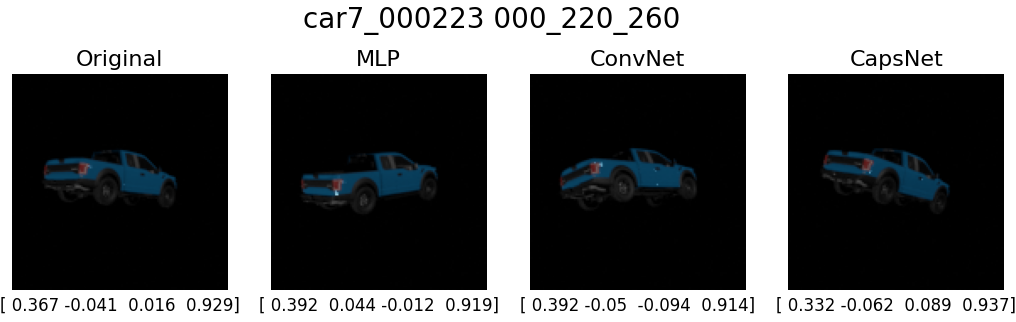} \\
    \vspace{0.5cm}
    \includegraphics[width = 280pt]{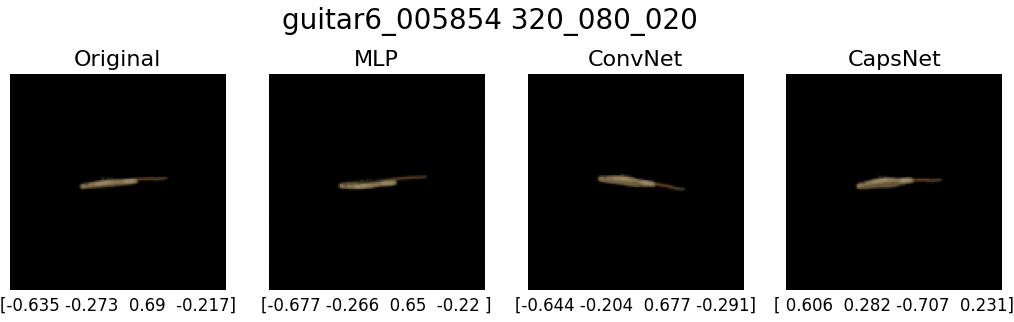}
    \caption{Visual comparisons of network predictions (MLP, ConvNet, and CapsNet) on TinyShapeNet.}
    \label{fig:quaternions_tinyshapenetresults}
\end{figure}

As mentioned above, no data augmentation was used for this work, but each image in the dataset was rescaled to decrease network training times. All models were trained for a maximum of 1000 epochs using Early Stopping with a patience of 50, monitoring the validation score. Experimental results are shown in Table~\ref{tab:quaternions_tinyshapenetresults}. We can see that the MLP network achieves the best performance in terms of test loss, while the ConvNet and CapsNet achieve the best accuracy values. Figure~\ref{fig:quaternions_tinyshapenetresults} shows some visual comparisons of the networks obtained feeding to Blender as input the predicted quaternions.

\begin{table}[h]
\footnotesize
    \centering
    \begin{tabular}{ c|c|c|c }
         & \textbf{MLP} & \textbf{ConvNet} & \textbf{CapsNet} \\ [0.2ex] \hline
        \textbf{Accuracy} & 0.99984 & 1.0 & 1.0 \\
        \textbf{Loss} & 0.00182 & 0.00384 & 0.00416 \\
        \textbf{Training time} & 4d 5h & 2d 1h & 9d 5h \\
    \end{tabular}
    \caption{Experimental results on TinyShapeNet test set for the multilayer, convolutional and capsule networks.}
    \label{tab:quaternions_tinyshapenetresults}
\end{table}

% \begin{figure}[htbp] \label{fig:loss}
%     \centering
%     \includesvg[width = 370pt, inkscapelatex=false]{Parts/03_CapsNetsApplications/gfx/quaternions/loss_train.svg}
%     \includesvg[width = 370pt, inkscapelatex=false]{Parts/03_CapsNetsApplications/gfx/quaternions/loss_valid.svg}
%     \caption{Training and validation graph of the total loss function}
% \end{figure}

% \begin{figure}[htbp] \label{fig:acc}
%     \centering
%     \includesvg[width = 370pt, inkscapelatex=false]{Parts/03_CapsNetsApplications/gfx/quaternions/accuracy_train.svg}
%     \includesvg[width = 370pt, inkscapelatex=false]{Parts/03_CapsNetsApplications/gfx/quaternions/accuracy_valid.svg}
%     \caption{Accuracy Training and validation graph}
% \end{figure} 

\section{Summary}\label{sec:quaternions_summary}
In this chapter, we showed how to generate a simple synthetic dataset with quaternions and how to modify the architecture of neural networks to handle rotation predictions.

Despite the limited complexity of the architecture used, the convolutional network still proved to be state-of-the-art regarding image recognition, both for accuracy and for very low training times.

Instead, the multilayer perceptron showed several limitations due to too high training times and a training subject to many fluctuations. Despite this, the accuracy does not differ much from the other results, and render loss is better minimized than other networks. Capsule networks appear promising but still immature architecture for object recognition. Its training was very time-consuming, especially when compared to the convolutional model. The slowness of capsule networks is mainly due to the iterative nature of the dynamic routing algorithm and the high number of hyperparameters. Further research needs to be done to optimize hyperparameters to improve render loss. However, despite all these problems, capsules provide a different and successful approach for dealing with generalization on new viewpoints compared to other networks. 

In the future, it would be interesting to continue this research by testing more complex architectures. Then, to evaluate the goodness of the networks more objectively, it is necessary to increase the dataset's size, also through data augmentation. In addition, another good option to apply is to use uniformly distributed random rotations, rather than 20-degree steps, for the generation of the object pose. Finally, splitting the data into mutually exclusive object classes for training and testing could help evaluate the architectures.

\cleardoublepage
%*****************************************
\chapter{Lung Nodules Segmentation with Capsule networks}\label{ch:lungs_nodules_capsnets}
%*****************************************
During my Ph.D. studies, I worked on the EU-funded DeepHealth project\footnote{\url{https://deephealth-project.eu/}}, which focuses on applying deep learning and high-performance computing to biomedical image processing. In this context, we released the UniToChest dataset\footnote{\url{https://zenodo.org/record/5797912}}, a collection of anonymized chest CT scan slices coupled with the proper lung nodule segmentation map. %UniToChest is described in detail in the following publication \fullcite{unitochest1}.
This chapter marks the beginning of our pursuit to improve lung nodule segmentation, requiring less data using capsule networks. It represents the initial stages of our research, and it has been a collaborative effort with Paolo Peretti, a graduate student. 
\section{Introduction}\label{lung:introduction}
In recent years, medical imaging has undergone a revolutionary transformation, primarily attributed to the remarkable progress of deep learning techniques. Deep learning has demonstrated exceptional capabilities in various domains, and its application to medical imaging has opened up new avenues for accurate and efficient diagnosis, prognosis, and treatment planning. With the ever-increasing complexity and volume of medical data, traditional image analysis methods have faced limitations in extracting meaningful insights. In contrast, deep learning models, fueled by their ability to automatically learn intricate features from data, have shown unparalleled potential in enhancing the accuracy, speed, and precision of medical image interpretation.
\begin{figure}[h]
\centering
\includegraphics[width=1\columnwidth]{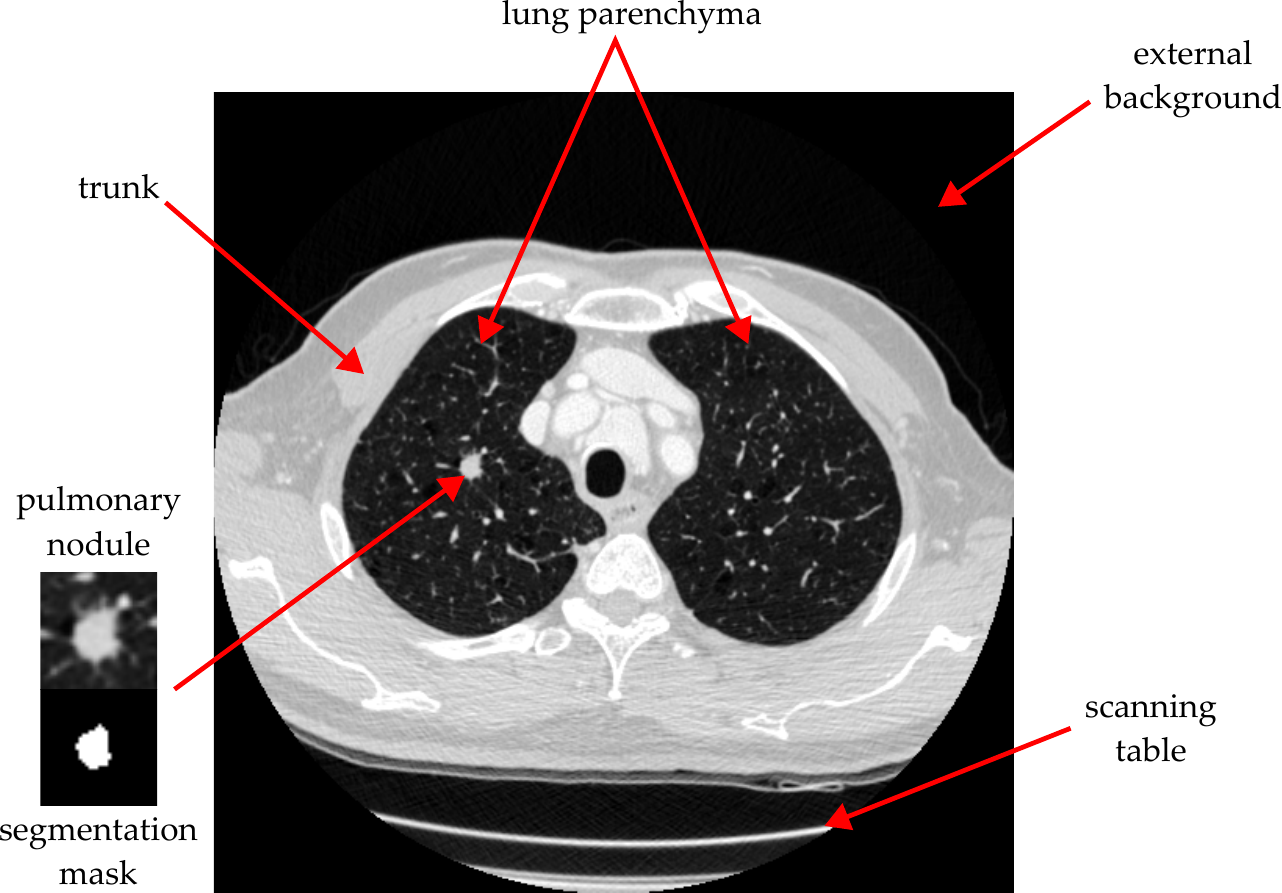}
\caption{CT image of lungs with an annotated nodule segmentation mask.}
\label{fig:lung-lung}
\end{figure}

Lung nodules are an early indicator of lung cancer, which remains one of the most prevalent and deadly forms of cancer worldwide. With such a low survival rate of 14-15\% at late stages of lung cancer, early detection and accurate diagnosis are crucial for improving patient outcomes and survival rates~\cite{blandin2017progress}. Deep learning models such as convolutional networks have shown tremendous promise in enhancing the accuracy and efficiency of lung cancer detection and diagnosis. Most common tasks include nodule detection~\cite{lung-detect-framework, lung-detect-multiscale, lung-detect-unsup}, segmentation~\cite{lung-segm-maskrcnn, lung-segm-feataware, lung-segm-adv, lung-segm-marching} and malignancy prediction~\cite{lung-malignancy-isbi, lung-malignancy-accurate, lung-malignancy-cliplung} in chest radiographs and computed tomography (CT) scans. Despite the remarkable achievements, the application of deep learning in lung cancer imaging presents certain challenges. Images and relative annotations in fact often lack in terms of quality or quantity or both, due to the cost of acquiring and annotating the images by a radiologist. Therefore, industry, academia, and health organizations work together to collect and publicly release new datasets with marked-up annotated lesions such as LIDC-IDRI~\cite{LIDC} or UniToChest~\cite{unitochest} datasets. 

Encoder-decoder models based on CNNs such as U-Net~\cite{unet} are the most commonly applied to the
medical image segmentation task, where the goal is to identify the pixels of organs or lesions.  
Though CNNs are popularly used, they have limitations due to poor robustness to affine transformations and lack of instantiation parameters (e.g., precise location information, pose, deformation, etc.).
Therefore, many capsule-based models were introduced, such as SegCaps~\cite{segcaps}, Matwo-CapsNet~\cite{Matwo-CapsNet}, 3D-UCaps~\cite{3D-UCaps} and OnlyCaps-Net~\cite{onlycapsnet}.
Capsules should be able to generalize to unseen images even when the number of images with different views used to train the networks is very low. However, to the best of our knowledge, no study has shown this property for lung nodules segmentation. To overcome this issue, this chapter aims to present a preliminary investigation on whether capsule-based methods can perform better compared to U-Net-based models in fewer data scenarios for segmenting lung nodules.

\section{Methodology}\label{sec:lung_meth}
This section describes our method to segment lung nodules, including the preprocessing stage and the architecture of CNN-based (U-Net) and capsule-based (SegCaps) networks we rely upon. 

\subsection{Dataset and preprocessing}
The Lung Image Database Consortium image collection~\cite{LIDC} (LIDC-IDRI) consists of
diagnostic and lung cancer screening thoracic CT scans with marked-up annotated lesions. This dataset contains more than 1000 cases (1018 cases).
Each subject includes many images from a clinical thoracic CT scan and an
associated XML (eXtensible Markup Language) file that records the results of the image annotation process performed by experienced radiologists, particularly by a maximum of four experienced radiologists.

Since original CT scans are in DICOM format, we preprocessed each slice, converting pixel intensity values
in Hounsfield Units (HU) and then applying a windowing operation. Since this is preliminary work, we decided not to consider all the lungs but only a portion of the image of interest. Specifically, we extract several regions of interest (ROIs) of size $128 \times 128$, including a nodule, from each scan. We applied random data rotations and horizontal flips during training.

\subsection{Architectures}
\begin{figure}
\centering
     \begin{subfigure}[t]{1\linewidth}
         \centering
         \includegraphics[width=\linewidth]{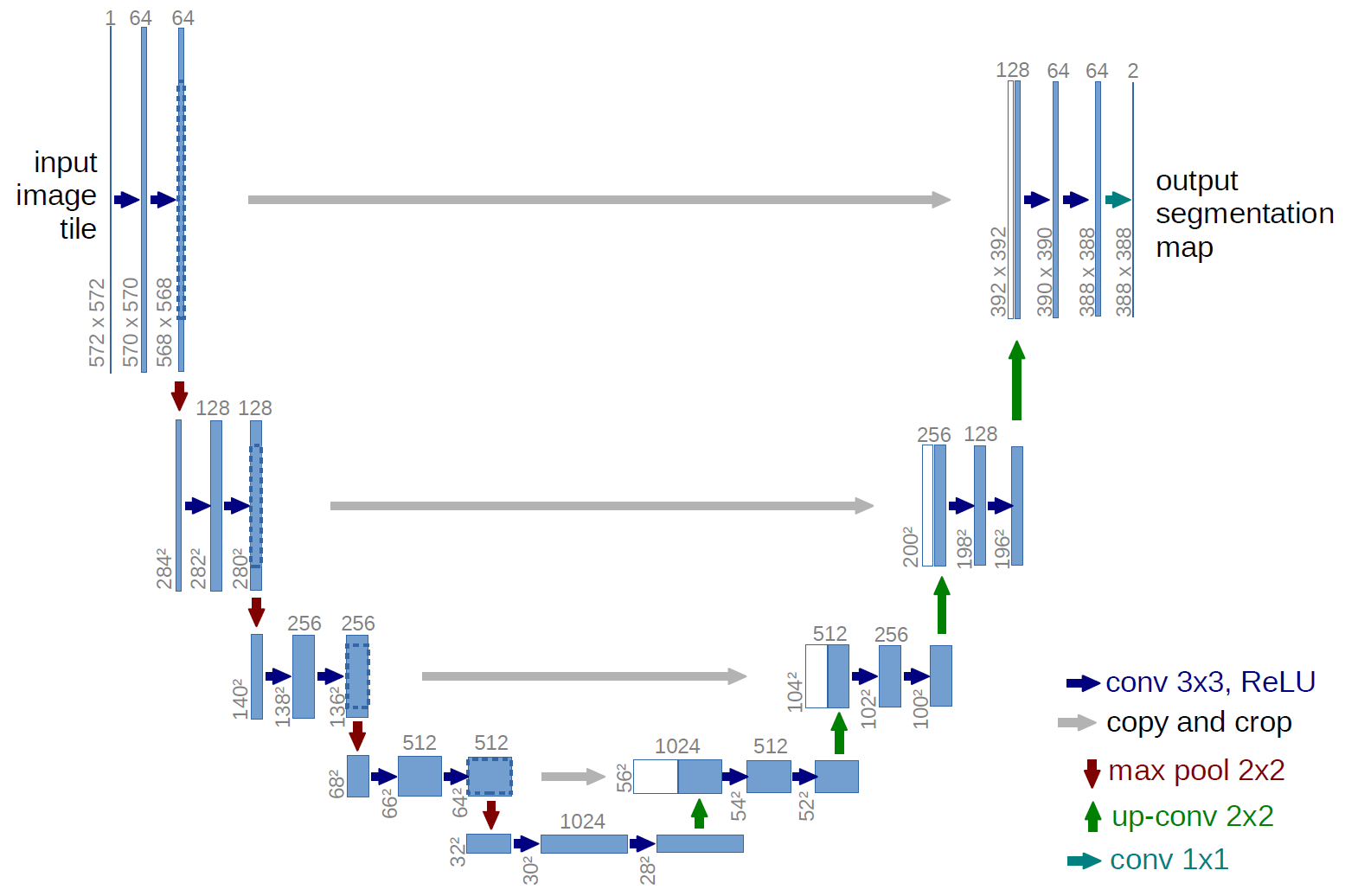}
         \caption{U-Net architecture.}
         \label{fig:lung-unet}
    \end{subfigure}
     \begin{subfigure}[t]{1\linewidth}
         \centering
         \includegraphics[width=\linewidth]{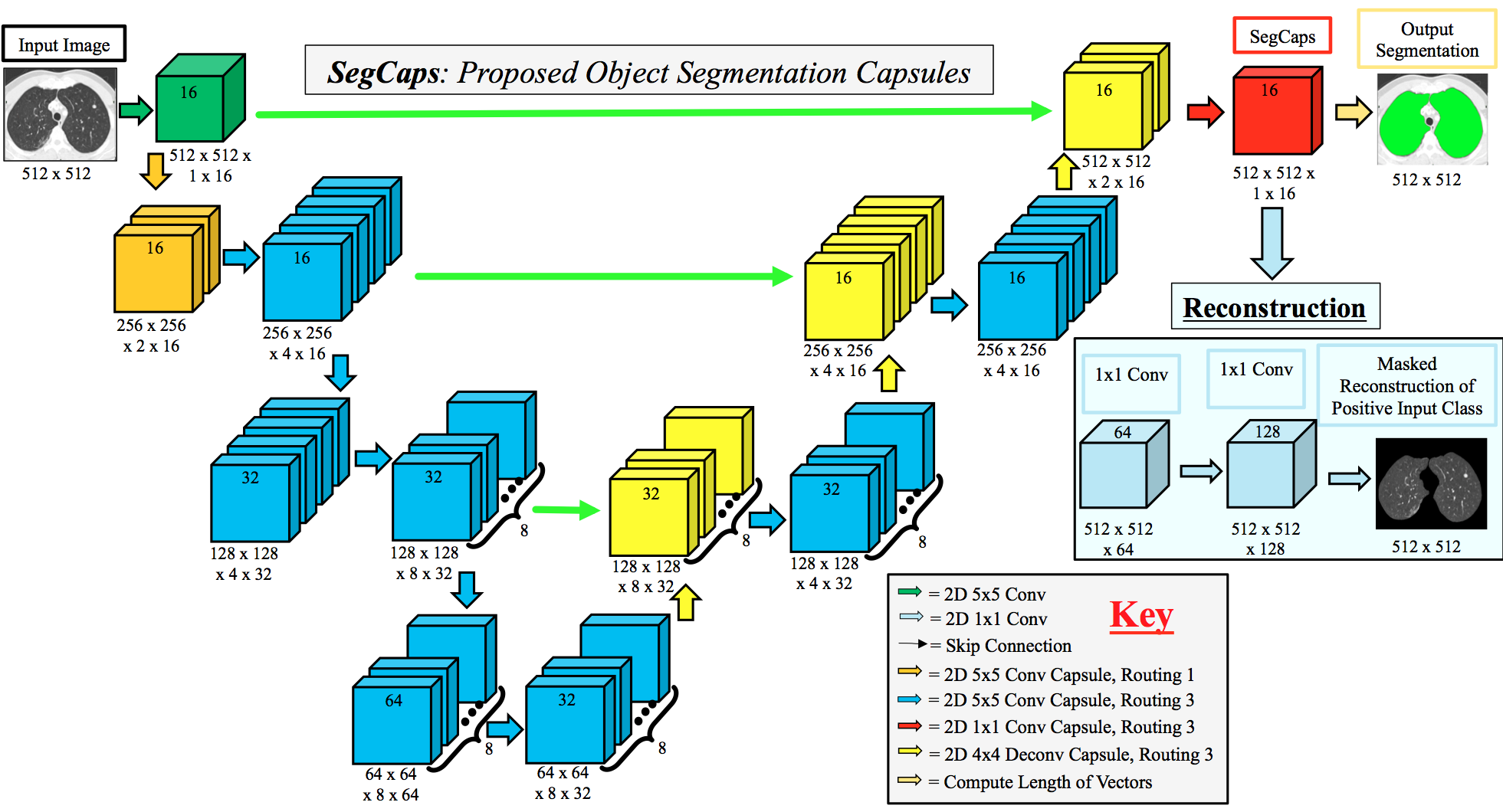}
         \caption{SegCaps architecture.}
         \label{fig:lung-segcaps}
    \end{subfigure}
    \caption{U-Net and SegCaps are the architectures used to segment lung nodules.}
    \label{fig:lung-architectures}
\end{figure}

In our experiments, we tested U-Net and SegCaps architectures, shown in Figure~\ref{fig:lung-architectures}.

U-Net consists of a contracting path (downsampling) to capture context and a symmetric expansive path (upsampling) to generate accurate segmentation maps. U-Net is known for its "U" shape and skip connections, aiding in preserving spatial information and enabling precise object boundary localization. We chose this network because it is still widely used in biomedical image analysis and has been influential in various computer vision applications. However, the number of trainable parameters of U-Net is very high (31M). Therefore, \citep{segcaps} introduced SegCaps, the first neural network architecture with fewer parameters (1.4M) that employs capsule networks for image segmentation. Capsules capture spatial hierarchies and object poses, enabling a better understanding of object structures and context and potentially improving segmentation accuracy and generalization. 
Within SegCaps, a distinct collection of capsules is established for every pixel $(x, y)$ present in either the image or intermediate layers. The projected segmentation label for a given location $(x, y)$ corresponds to the identifier of the capsule from the final layer that exhibits the highest activation. SegCaps architecture is inspired by the U-Net architecture, therefore it comprises an encoder-decoder structure and skips connections. In the encoder part, convolutional capsule layers with kernel size $5 \times 5$ are stacked on top of a standard $5 \times 5$ convolutional layer to reduce the size of the input. SegCaps employ a novel dynamic routing algorithm, called locally-connected routing, where (i) children are only routed to parents within a defined spatially-local window and (ii) transformation matrices are shared for each member of the grid within a capsule type. In the decoder part, SegCaps introduces the concept of deconvolutional capsules (with kernel size $4 \times 4$), which operate using transposed convolutions, routed by the locally-connected routing. The network outputs both the segmentation mask of the object and the masked reconstruction of the positive input class. In our implementation, we removed the latter because we focus only on the lung nodule and not on the lung itself.

For a fair comparison between U-Net and SegCaps regarding the number of trainable parameters, we also tested smaller versions of U-Net with four and three layers in the encoder part. Note that the original U-Net has five layers.

\section{Results}
We report in Table~\ref{tab:lung-results} the results on the LIDC-IDRI test set for U-Net and SegCaps with different numbers of layers and percentages of training samples. We can see that the highest dice score is obtained by the original U-Net model with five layers when trained with 100\% of the training data. As mentioned before, since the number of trainable parameters of this network is very high, we choose to compare SegCaps with shallower U-Net models. We can see that in these cases, SegCaps performs better than U-Net models with the same number of layers or lower, when trained with the full training set and even on subsets containing 10\% or 50\% of the training data. 
\label{sec:lung_experiments}
\begin{table}[h]
\centering
\footnotesize
\begin{tabular}{l|c|c|c|c}
\hline
\textbf{Model}                & \multicolumn{1}{l|}{\textbf{Layers}} & \textbf{Training samples (\%)} & \textbf{Dice score}       & \multicolumn{1}{l}{\textbf{Parameters (M)}} \\ \hline
\multicolumn{1}{l|}{U-Net}    & \multicolumn{1}{c|}{\multirow{3}{*}{5}}       & \multicolumn{1}{c|}{100}                & \multicolumn{1}{c|}{0.81} & \multirow{3}{*}{31}                         \\
\multicolumn{1}{l|}{U-Net}    & \multicolumn{1}{c|}{}                         & \multicolumn{1}{c|}{50}                 & \multicolumn{1}{c|}{0.79} &                                             \\ 
\multicolumn{1}{l|}{U-Net}    & \multicolumn{1}{c|}{}                         & \multicolumn{1}{c|}{10}                 & \multicolumn{1}{c|}{0.78} &                                             \\ \hline
\multicolumn{1}{l|}{U-Net}    & \multicolumn{1}{c|}{\multirow{3}{*}{4}}       & \multicolumn{1}{c|}{100}                & \multicolumn{1}{c|}{0.74} & \multirow{3}{*}{7.6}                        \\ 
\multicolumn{1}{l|}{U-Net}    & \multicolumn{1}{c|}{}                         & \multicolumn{1}{c|}{50}                 & \multicolumn{1}{c|}{0.71} &                                             \\
\multicolumn{1}{l|}{U-Net}    & \multicolumn{1}{c|}{}                         & \multicolumn{1}{c|}{10}                 & \multicolumn{1}{c|}{0.68} &                                             \\ \hline
\multicolumn{1}{l|}{U-Net}    & \multicolumn{1}{c|}{\multirow{3}{*}{3}}       & \multicolumn{1}{c|}{100}                & \multicolumn{1}{c|}{0.72} & \multirow{3}{*}{1.8}                        \\ 
\multicolumn{1}{l|}{U-Net}    & \multicolumn{1}{c|}{}                         & \multicolumn{1}{c|}{50}                 & \multicolumn{1}{c|}{0.69} &                                             \\ 
\multicolumn{1}{l|}{U-Net}    & \multicolumn{1}{c|}{}                         & \multicolumn{1}{c|}{10}                 & \multicolumn{1}{c|}{0.66} &                                             \\ \hline
\multicolumn{1}{l|}{SegCaps}  & \multicolumn{1}{c|}{\multirow{3}{*}{4}}       & \multicolumn{1}{c|}{100}                & \multicolumn{1}{c|}{0.79} & \multirow{3}{*}{1.4}                        \\ 
\multicolumn{1}{l|}{SegCaps}  & \multicolumn{1}{c|}{}                         & \multicolumn{1}{c|}{50}                 & \multicolumn{1}{c|}{0.77} &                                             \\ 
\multicolumn{1}{l|}{SegCaps} & \multicolumn{1}{c|}{}                         & \multicolumn{1}{c|}{10}                 & \multicolumn{1}{c|}{0.76} &                                             \\ \hline
\end{tabular}
\caption{Dice score results for U-Net and SegCaps with different numbers of layers and percentages of training samples (LIDC-IDRI test set).}
\label{tab:lung-results}
\end{table}

\section{Summary}
This chapter briefly investigated whether capsule-based networks can achieve better results for the lung nodules segmentation task. We showed that, despite the model with the highest performance being the original U-Net model with 31M of parameters, a SegCaps model with only 1.4M parameters shows promising results compared to smaller U-Nets with 7.6M and 1.8M parameters. These results are also confirmed in low training data scenarios. This suggests that capsule-based networks have the potential to better segment lung nodules with fewer parameters and training samples. In the future, we will also test other 2D and 3D capsule-based networks, such as 3D-UCaps and OnlyCaps.
\label{sec:lung_summary}
\cleardoublepage
\chapter{Conclusion}\label{ch:conclusion}
%************************************************

This thesis has delved into the intriguing theory and applications of capsule networks, aiming to unlock their full potential in computer vision tasks. 
This thesis has been organized into three parts: Part~\ref{part:capsnet-fundamentals} provided an overview of capsule network fundamentals, Part~\ref{part:capsnet-components} addressed some of capsule networks limitations with novel proposed methods, and Part~\ref{part:capsnets-applications} demonstrated capsule network applications in real-world scenarios.
We began with Chapter~\ref{ch:limitconvnets} by highlighting the limitations of traditional convolutional networks, which often lose spatial relationships due to max pooling layers and lack of an explicit entity representation. Chapter~\ref{ch:capsnetsbackground} offered an overview of capsule networks, deep learning models emerged as a promising solution, offering hierarchical object and part encoding, robustness to viewpoint changes, and improved feature preservation.

Despite their promise, capsule networks have not seen widespread adoption compared to traditional convolutional networks or vision transformers, possibly due to the lack of standardized architectures and pre-trained models. However, this work underscores the importance of continued research and advancements in capsule networks to fully harness their potential and address their limitations. 

Throughout the thesis, we focused on three key questions to enhance our understanding and utilization of capsule networks:

\begin{enumerate}
\item \textbf{The routing algorithm:} In Chapter~\ref{ch:icann_routing_annealing}, we investigated the effectiveness of the routing algorithm and its necessity, especially in small-sized networks. Our experiments confirmed the importance of the routing algorithm. We introduced a novel method for annealing the number of routing iterations during training, yielding improved performance in architectures with fewer parameters.

\item \textbf{Learning effective part descriptions:} In Chapter~\ref{ch:icip_effcapsnets}, to enhance primary capsule representations, we explored the use of pruned backbones, promoting sparsity to improve computational efficiency and reduce memory and training time requirements. This approach demonstrated high generalization ability with resource savings.

\item \textbf{Part-whole relationships learning:} Understanding how capsules capture part-whole relationships is crucial. In Chapter~\ref{ch:rem}, we revealed that capsules with low entropy can extract more concise and discriminative part-relationships, shedding light on the mechanisms of part-aware representations and paving the way for more interpretable and effective capsule networks.
\end{enumerate}

Additionally, we showcased the practical applications of capsule networks in autonomous localization (Chapter~\ref{ch:lsvl}), quaternion-based rotation prediction in synthetic datasets (Chapter~\ref{ch:quaternions_capsnets}), and lung nodule segmentation in medical images ((Chapter~\ref{ch:lungs_nodules_capsnets}), highlighting their ability to capture spatial relationships in various domains.

 Therefore, this work contributes to advancing our understanding of capsule networks and their potential for revolutionizing computer vision tasks across diverse domains.

In closing, \citeauthor{genclassifiers}~\cite{genclassifiers} recently showed intriguing emergent properties of generative classifiers such as human-like shape bias in object recognition. They also understand certain perceptual illusions like those presented in Figure~\ref{fig:rabbit_duck}. It will be interesting to study how, and to what degree,
diffusion and large language vision models exhibit object-centric hierarchical representations in zero-shot recognition.

% ********************************************************************
% Backmatter
%*******************************************************
%\include{Parts/05_Appendix/P05}
%********************************************************************
% Other Stuff in the Back
%*******************************************************
\cleardoublepage%*******************************************************
% Images Attribution
%*******************************************************
\refstepcounter{dummy}
\pdfbookmark[0]{Images Attribution}{attributions}
\chapter*{Images Attribution}
\thispagestyle{empty}

\begin{itemize}
    \item Portrait of woman in d’hermine pass (Olga), 1923, Pablo Picasso
    \item Rabbit–duck illusion, unattributed drawing from the 23 October 1892 issue of Fliegende Blätter
    \item The Birth of Venus, mid 1480s, Sandro Botticelli
    \item Mona Lisa, 1503-1517, Leonardo da Vinci
    \item Self-Portrait with Thorn Necklace and Hummingbird, 1940, Frida Kahlo

\end{itemize}

\bigskip

\cleardoublepage%********************************************************************
% Bibliography
%*******************************************************
% work-around to have small caps also here in the headline
\manualmark
\markboth{\spacedlowsmallcaps{\bibname}}{\spacedlowsmallcaps{\bibname}} % work-around to have small caps also
%\phantomsection 
\refstepcounter{dummy}
\addtocontents{toc}{\protect\vspace{\beforebibskip}} % to have the bib a bit from the rest in the toc
\addcontentsline{toc}{chapter}{\tocEntry{\bibname}}
\label{app:bibliography}
\printbibliography

%\cleardoublepage\include{FrontBackmatter/Declaration}
%\cleardoublepage\include{FrontBackmatter/Colophon}
% ********************************************************************
% Game Over: Restore, Restart, or Quit?
%*******************************************************
\end{document}